\documentclass{article}

\usepackage[main, final]{neurips_2026}

\usepackage[utf8]{inputenc}
\usepackage[T1]{fontenc}
\usepackage[hidelinks]{hyperref}
\usepackage{url}
\usepackage{booktabs}
\usepackage{amsfonts}
\usepackage{nicefrac}
\usepackage{microtype}
\usepackage[table]{xcolor}

\usepackage{graphicx}
\usepackage{doi}
\usepackage{adjustbox}
\usepackage{enumitem}

\usepackage{amsmath,amssymb,bm,mathtools,bbm,amsthm}
\usepackage{multirow}
\usepackage{algorithm}
\usepackage{algpseudocode}
\usepackage{subcaption}
\usepackage{tcolorbox}
\usepackage{tikz}
\usepackage{tabularx}
\usepackage{float}
\usepackage{placeins}
\usepackage{needspace}

\title{DynamicRad: Content-Adaptive Sparse Attention for Long Video Diffusion}

\author{%
Yongji Long$^{1,2}$ \quad Shijun Liang$^{3}$ \quad Jintao Li$^{1,2,*}$ \quad Yun Li$^{1,2,*}$\\[0.6em]
{\normalfont\small $^{1}$ University of Electronic Science and Technology of China}\\
{\normalfont\small $^{2}$ Shenzhen Institute for Advanced Study, UESTC}\\
{\normalfont\small $^{3}$ Computational Mathematics, Science, \& Engineering at Michigan State University (MSU)}\\[0.6em]
{\normalfont\small $^{*}$ Corresponding authors: \texttt{j.t.li@i4ai.org} \quad \texttt{Yun.Li@ieee.org}}%
}

\begin{document}

\maketitle

\begin{abstract}
  Leveraging the natural spatiotemporal energy decay in video diffusion offers a path to efficiency, yet relying solely on rigid static masks risks losing critical long-range information in complex dynamics. To address this issue, we propose \textbf{DynamicRad}, a sparse-attention framework that constrains adaptive selection using a radial locality prior. DynamicRad introduces a \textbf{dual-mode} strategy: \textit{static-ratio} for speed-optimized execution and \textit{dynamic-threshold} for quality-first filtering. To avoid online configuration search, we integrate an offline Bayesian Optimization (BO) pipeline with a \textbf{semantic motion router}. The offline-trained router maps prompt embeddings to BO-selected sparsity regimes with a single projection module. Unlike online profiling methods, our offline BO optimizes attention reconstruction error (MSE) on a proxy task and reuses the selected configurations during inference. Dynamic selection still scores current Q/K features and constructs sparse indices. Experiments on HunyuanVideo and Wan2.1-14B demonstrate that DynamicRad achieves a strong efficiency--quality trade-off among block-sparse-attention baselines, obtaining \textbf{1.7$\times$--2.5$\times$ inference speedups} at \textbf{74.3\%--86.8\% effective block sparsity}. In some long-sequence settings, the dynamic mode matches or slightly improves upon a checkpoint-matched dense baseline on individual automatic quality metrics, while optional Mask-Aware LoRA further improves long-horizon coherence. Code is available at \url{https://github.com/Adamlong3/DynamicRad}.
\end{abstract}

\section{Introduction}
\label{sec:intro}

Recent diffusion-based video generators achieve strong visual fidelity and temporal coherence, yet scaling them to long horizons and high resolution remains difficult because self-attention grows quadratically with the spatiotemporal token count \citep{ho2020ddpm,song2021sde,rombach2022ldm,peebles2023dit,singer2022imagenvideo,hong2022cogvideo,bartal2024lumiere, brooks2024sora,vaswani2017attention}. Even with optimized kernels, attention-side compute and memory bandwidth remain major bottlenecks \citep{dao2022flashattention,dao2023flashattention2, bolya2023tome, ma2024deepcache}. 

A natural remedy is sparsity. Prior efficient attention methods mainly fall into two categories: static sparse patterns, which are hardware-friendly but rigid, and dynamic sparsification, which is adaptive but often introduces extra overhead or instability under heterogeneous and long-video settings \citep{child2019sparsetransformer,beltagy2020longformer,zaheer2020bigbird,kitaev2020reformer,wang2020linformer,choromanski2021performer,roy2021routing,tay2020efficient}. Recent evidence further suggests a spatiotemporal locality structure in video diffusion attention \citep{li2025radialattention,zhang2025sta, ding2025efficientvdit}, where post-softmax attention mass decays with spatial and temporal distance, motivating sparse masks that preserve near-field interactions while pruning long-range pairs \mbox{\citep{li2025radialattention,zhang2025sta}}. However, purely static sparsification remains brittle under complex motion, whereas fully dynamic schemes may be costly or unstable at scale.

In this work, we propose \textbf{DynamicRad}, a sparse-attention framework for long video diffusion that combines a structured radial candidate set with content-dependent selection. DynamicRad uses a shared block-sparse candidate set and supports two deployment modes: \textit{static-ratio}, which prioritizes throughput by using fixed retention ratios, and \textit{dynamic-threshold}, which uses attention-logit scores to retain candidate pairs. The mode and sparsity configuration are selected from an offline BO-compiled lookup table using a semantic motion router. An optional Mask-Aware LoRA is evaluated separately to compensate for sparsity-induced temporal artifacts. Unlike Radial Attention's content-independent final mask~\citep{li2025radialattention}, Dynamic mode selects links from current Q/K features within radial support. Recent methods such as AdaSpa~\citep{xia2025adaspa} also explore training-free dynamic sparsification for long video generation, but rely on online precise search over sparse indices; DynamicRad instead shifts regime-level configuration search offline while Dynamic mode still incurs scoring and indexing costs.

\paragraph{Contributions}
We summarize our contributions as follows:
(1) \textbf{Core Framework:} We propose DynamicRad, a unified sparse-attention framework that combines a shared kernel-friendly candidate set with two selection modes, bridging rigid-prior methods such as Radial Attention \citep{li2025radialattention} and dynamic methods such as SVG \citep{xi2025svg}. We further introduce a multi-dimensional parameterization, including a continuous long-range factor $\lambda$, that unifies static efficiency and dynamic flexibility without an online mask-prediction network inside the denoising loop.
(2) \textbf{Robust Deployment:} We introduce an offline BO-driven configuration pipeline coupled with a lightweight \textbf{Semantic Motion Router}. By optimizing \textbf{Attention MSE on a physics-based proxy task}, the search process can be completed in minutes, while the router maps prompt embeddings to BO-selected regimes in $<2$ms without fragile manual tuning.
(3) \textbf{Empirical Gains:} Integrating DynamicRad into HunyuanVideo-241f and Wan2.1-14B \citep{xue2024openvid,kong2024hunyuanvideo,wan2025wan,huang2023vbench}, we achieve up to \textbf{2.54$\times$ speedup} under high sparsity. We report the \textbf{backbone-frozen sparse-attention policy} separately from the optional \textbf{Mask-Aware LoRA} adaptation, which further improves temporal coherence under aggressive sparsity.

\section{Related Work}
\label{sec:related_work}

\paragraph{Evolution of Video Generation Backbones.}
Video generation has shifted from U-Net backbones \citep{singer2022imagenvideo} to DiT-style architectures \citep{peebles2023dit} due to superior scalability. Representative DiT models include Latte \citep{ma2024latte} and CogVideoX \citep{yang2024cogvideox}; autoregressive approaches include VideoPoet \citep{kondratyuk2023videopoet} and the long-video system StreamingT2V \citep{henschel2024streamingt2v}.

\paragraph{Sparse Attention \& Efficiency Paradigms.}
Scaling diffusion models to long horizons is hindered by the quadratic $\mathcal{O}(N^2)$ cost of attention \citep{vaswani2017attention,dao2022flashattention}. 
Prior works typically bifurcate into \textit{static patterns} that impose fixed locality priors \citep{child2019sparsetransformer,li2025radialattention,zhang2025sta} and \textit{dynamic routing} that adapts to content \citep{roy2021routing,xi2025svg}.
While Radial Attention \citep{li2025radialattention} is hardware-friendly, it lacks adaptivity. Conversely, SVG \citep{xi2025svg} enables dynamic routing but adds online overhead.
MOD-DiT~\citep{liu2026moddit} predicts later masks from early denoising and historical sparsity, whereas our regime router uses the prompt before denoising. BLASST~\citep{yuan2025blasst} performs threshold pruning for LLM inference rather than video spatiotemporal attention. We treat MA-LoRA as an optional robustness mechanism, not an algorithmic novelty claim.

To address these gaps, recent works have introduced novel structural efficiencies. Dynamic Diffusion Transformer (DyDiT) \citep{zhao2024dydit} dynamically adjusts computation in diffusion transformers along timestep and spatial dimensions, while sparse-attention methods such as \textbf{PowerAttention (PA)} \citep{chen2025powerattention} (originally proposed for long-context transformers), \textbf{SVG} \citep{xi2025svg}, and \textbf{Radial Attention} \citep{li2025radialattention} improve efficiency from different design perspectives.

\paragraph{Learning-based Efficiency.}
Recent approaches like VORTA \citep{sun2025vorta} and VSA \citep{zhang2025vsa} propose learning sparsity policies directly via gating networks. While effective, they require expensive retraining or per-step routing inference. 
In contrast, DynamicRad achieves adaptivity by leveraging physical motion priors and reusing pre-computed text embeddings, avoiding the need for invasive architectural changes. Its router is trained offline and frozen; optional Mask-Aware LoRA is a separate adaptation path.
\paragraph{Deployment Optimization Strategies.}
Our work integrates several orthogonal efficiency techniques. 
TeaCache \citep{liu2024teacache} and DeepCache \citep{ma2024deepcache} save compute by reusing cached model outputs or features across denoising steps; we reduce the intra-layer attention cost. 
In diffusion-based image restoration, Li et al. decouple data-consistency reconstruction from diffusion purification to improve inference efficiency \citep{li2026dcdp}.
To ensure temporal coherence under aggressive pruning, we employ Parameter-Efficient Fine-Tuning (PEFT) akin to LoRA \citep{hu2022lora}, but specifically target projection layers sensitive to sparsity. 
Finally, we utilize Bayesian Optimization \citep{snoek2012bo} in an offline pipeline to pre-compile a lightweight lookup table, enabling instantaneous adaptation at runtime.
More broadly, constrained optimization of competing objectives also appears in policy learning, where comparison-based updates have been used to improve feasible Pareto fronts \citep{li2026copro}.
\section{Method}
\label{sec:method}

\begin{figure*}[t]
  \centering
  \includegraphics[width=0.90\linewidth]{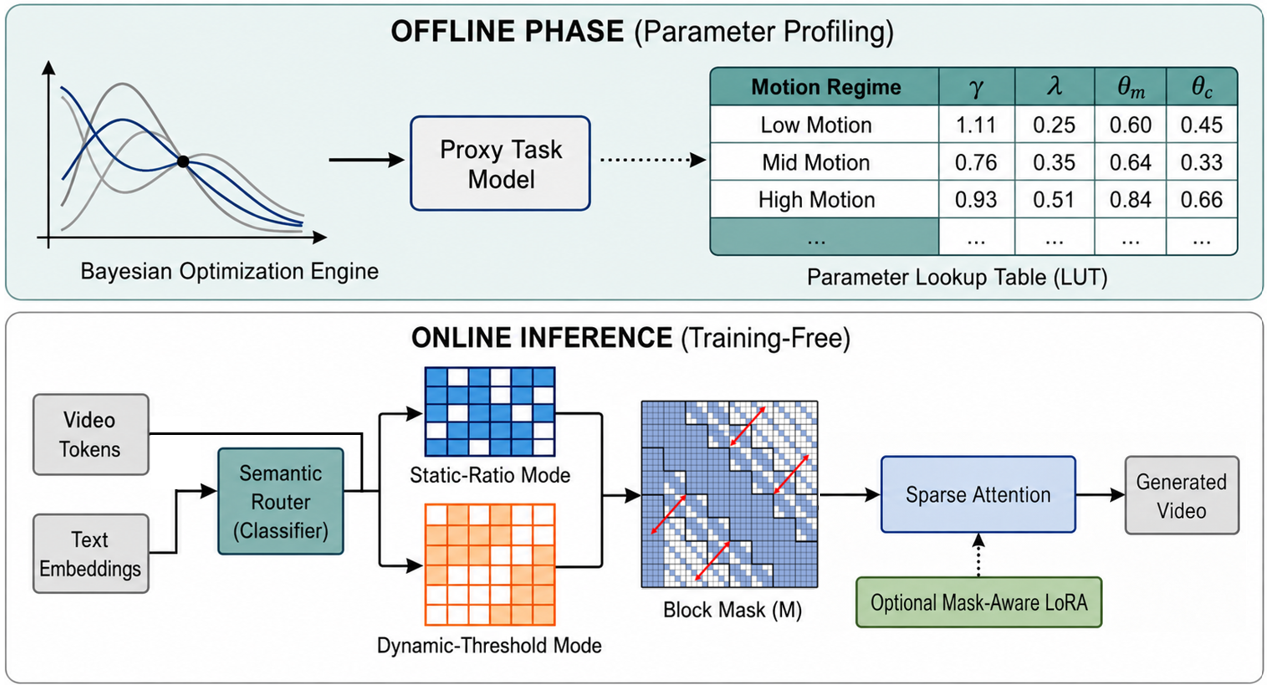}
  \caption{\textbf{Overview of DynamicRad.} Offline BO builds a motion-regime lookup table; its displayed values are schematic, while optimized configurations appear in Table~\ref{tab:bo_lookup}. At inference, the trained and frozen router selects a configuration from the prompt embedding. Video tokens bypass the router and join the selected Static-ratio or Dynamic-threshold path; only Dynamic scores video Q/K features. The diffusion backbone remains frozen, with optional separate Mask-Aware LoRA.}
  \label{fig:framework}
\end{figure*}

Radial Attention reduces attention cost via a static energy-decay mask~\citep{li2025radialattention}, but its fixed sparsity pattern can be brittle under long horizons and complex motion where long-range dependencies become critical. 
To address these limitations, we propose \textbf{DynamicRad}, a sparse-attention framework summarized in Figure~\ref{fig:framework}. The framework combines offline BO-based configuration, prompt-conditioned regime routing, and a shared candidate set with two selection modes: \textit{static-ratio} for throughput-first deployment and \textit{dynamic-threshold} for quality-first deployment. Both modes share the same kernel-friendly block aggregation logic. We additionally introduce an optional Mask-Aware LoRA adaptation path, which is evaluated separately from the backbone-frozen sparse-attention policy.
\subsection{Dynamic Neighborhood Partitioning and Parameter Space}
\label{subsec:dynamic-partitioning}
We parameterize the attention connectivity used by the Semantic Router through the radial width scale $\gamma$, block size $B_s$, mask threshold $\theta_m$, column density threshold $\theta_c$, and long-range factor $\lambda$. The Offline BO engine tunes $\gamma$, $\lambda$, $\theta_m$, $\theta_c$, and the mode-specific retention ratios or score thresholds, while $B_s$ is fixed within each run. This formulation allows the radial window width and long-range allocation to adapt to varying motion regimes.

\paragraph{Notation.}
Let $N_f$ be the number of frames and $N_t$ the number of tokens per frame. The total token count is $S = N_f N_t$.
We use $B_s \in \{32,64,128\}$ to denote the block size, aligning with hardware granularity. The sequence is padded such that $S' = \lceil S/B_s \rceil \cdot B_s$.

\paragraph{Radial window width via logarithmic grouping.}
Our construction uses logarithmic distance groups to narrow the radial window as temporal distance increases. For a frame pair $(i,j)$ with temporal distance $t = |i-j|$, we define a logarithmic group index $\mathcal{B}(t)$ and a base token span $L_0$.
We rigorously define the grouping for temporal neighbors ($t \ge 1$):
\begin{equation}
\label{eq:decay_len_impl_group}
\begin{split}
    \mathcal{B}(t) &= \lfloor \log_2 t \rfloor + 1 \quad (\text{for } t \ge 1), \\
    L_0 &= 2^{\lfloor \log_2 N_t \rfloor + 1}.
\end{split}
\end{equation}
Here, $L_0$ represents the implementation-aligned base token span under the flattened latent-token layout. Note that for the self-attention case ($t=0$), we default to dense attention. The decay length $\mathcal{L}(t)$ is an inverse power-law scaled by $\gamma$, so larger $\gamma$ widens the candidate window:
\begin{equation}
\label{eq:decay_len_impl}
\mathcal{L}(t) = \frac{L_0}{2^{\mathcal{B}(t)}} \cdot \gamma \approx \frac{L_0}{t} \cdot \gamma.
\end{equation}
The window width $w(i,j)$ is rounded to an integer and lower-bounded by $B_s$ before token-to-block aggregation:
\begin{equation}
\label{eq:window_width}
w(i,j) =
\begin{cases}
N_t, & t \le 1 \text{ (intra \& adjacent)}\\
\max\!\big(B_s,\ \lfloor \mathcal{L}(t)\rceil \big), & t > 1.
\end{cases}
\end{equation}

\paragraph{Long-range split rule controlled by the long-range factor $\lambda$.}
To capture periodic long-range dependencies (e.g., repetitive motion), we introduce a secondary decay schedule for the ``split'' stride, controlled by $\lambda$:
\begin{equation}
\label{eq:split_decay}
\mathcal{L}_{\text{split}}(t) = \frac{L_0}{2^{\mathcal{B}(t)}} \cdot \lambda.
\end{equation}
The split stride factor $\mathrm{sf}(t)$ determines the sampling frequency of long-range frames:
\begin{equation}
\label{eq:split_factor}
\mathrm{sf}(t) = \max\left(1, \left\lfloor \frac{B_s}{\mathcal{L}_{\text{split}}(t) + \epsilon} \right\rfloor \right).
\end{equation}
A frame pair $(i,j)$ is retained only if
\begin{equation}
\label{eq:frame_retain_rule}
t \bmod \mathrm{sf}(t) = 0,
\end{equation}
where $t = |i-j|$ denotes the temporal distance. This rule implies that as $t$ increases, $\mathcal{L}_{\text{split}}(t)$ decreases and the stride $\mathrm{sf}(t)$ increases, thereby sparsifying long-range temporal connections.

Let $u, v \in [0, N_t - 1]$ denote local token indices under the flattened latent-token memory layout within frame $i$ and frame $j$, respectively. For retained frame pairs, we define the candidate token-pair set as
\begin{equation}
\label{eq:candidate_set}
\mathcal{P}(i,j)=\{(u,v): |u-v|\le w(i,j)\},
\end{equation}
and set $\mathcal{P}(i,j)=\varnothing$ otherwise. Here, $|u-v|$ denotes locality under the flattened latent-token layout before block aggregation. In this way, the radial window $w(i,j)$ controls local token connectivity under the kernel memory layout, while the split rule in Eq.~\ref{eq:frame_retain_rule} sparsifies distant frame pairs along the temporal dimension.
\paragraph{Density control within blocks.}
The parameters $\theta_c$ and $\theta_m$ control the aggregation of token-level links into a block-level mask. For each block, we first mark a column as active if its retained-token ratio exceeds $\theta_c$; the block is then activated if the fraction of active columns exceeds $\theta_m$. Thus, $\theta_c$ suppresses noisy columns, while $\theta_m$ controls the final block retention for kernel efficiency.

\subsection{Two-Mode Intra-Candidate Selection}
\label{subsec:dual-mode-sparsification}
Given the candidate set $\mathcal{P}(i,j)$ for each inter-frame pair ($t>0$), DynamicRad selects a subset using one of two mutually exclusive modes. For $t=0$ (intra-frame) we do not prune and keep dense attention.

\paragraph{Static-ratio mode (static\_ratio).}
Designed for throughput-first deployment, this mode avoids the overhead of content-dependent scoring. We choose a retention ratio based on the decay state:
\begin{equation}
\label{eq:static_ratio_selector}
\rho(i, j) =
\begin{cases}
1.0, & t \le 1,\\
\rho_1, & t > 1\ \text{and}\ \mathcal{L}(t) \geq B_s,\\
\rho_2, & t > 1\ \text{and}\ \mathcal{L}(t) < B_s,
\end{cases}
\end{equation}
and uniformly sample
$
k=\max\!\left(1,\left\lfloor |\mathcal{P}(i,j)|\cdot \rho(i,j)\right\rfloor\right)
$
pairs from $\mathcal{P}(i,j)$ using a deterministic seed.
The generator is initialized once and is not reset for each frame-pair candidate set, so successive sets receive distinct draws; the seed reproduces the complete mask.

\paragraph{Dynamic-threshold mode (dynamic\_threshold).}
Designed for quality-first deployment, this mode uses attention logits as lightweight importance proxies.
For each candidate pair $(u,v)\in\mathcal{P}(i,j)$, we compute a proxy score by averaging pre-softmax logits over a small set of fused heads $H_f$:
\begin{equation}
\label{eq:dynamic_score}
s(u,v)=\frac{1}{H_f}\sum_{h=1}^{H_f}\frac{\mathbf{Q}^{(h)}_u\cdot(\mathbf{K}^{(h)}_v)^\top}{\sqrt{d_k}}.
\end{equation}
\paragraph{Score normalization.}
To make global thresholds more transferable across layers and prompts, we normalize scores within each candidate set $\mathcal{P}(i,j)$:
\begin{equation}
\label{eq:dynamic_score_norm}
\tilde{s}(u,v)=\frac{s(u,v)-\mu_{ij}}{\sigma_{ij}+\epsilon},
\end{equation}
where $\mu_{ij}$ and $\sigma_{ij}$ are the mean and std of $\{s(u',v'):(u',v')\in\mathcal{P}(i,j)\}$.
We compute $\tilde{s}(u,v)$ at the token-pair level and then aggregate token links to a block mask via $(\theta_c,\theta_m)$ to preserve kernel-friendly block sparsity.

\paragraph{Global thresholding with safeguards.}
We keep pairs whose normalized score exceeds a decay-state-dependent threshold:
\begin{equation}
\label{eq:dynamic_threshold_selector}
\tau(i,j)=
\begin{cases}
-\infty, & t\le 1,\\
\tau_1, & t>1\ \text{and}\ \mathcal{L}(t)\ge B_s,\\
\tau_2, & t>1\ \text{and}\ \mathcal{L}(t)< B_s.
\end{cases}
\end{equation}
where $\tau_1$ and $\tau_2$ correspond to near\_frame\_threshold and far\_frame\_threshold, respectively.
In our evaluated settings, pre-norm and scaled dot-product attention yield an empirically stable score scale across layers and denoising steps, making shared thresholds practical; using a small fused-head set ($H_f=2$) further reduces overhead while preserving rank-order utility for filtering.
To avoid fragile per-layer calibration, we use global thresholds $(\tau_1,\tau_2)$ shared across blocks, together with two safeguards:
(i) per-candidate-set score normalization (Eq.~\ref{eq:dynamic_score_norm});
(ii) if no pair passes the threshold, we retain the highest-scoring pair as a fallback.

The overall selection procedure is summarized in Algorithm~\ref{alg:dual-mode-sparsification}, while the detailed definitions of $w(i,j)$, $\mathcal{P}(i,j)$, score normalization, and thresholding are given in Eqs.~\ref{eq:window_width}--\ref{eq:dynamic_threshold_selector}.
\begin{algorithm}[t!]
\caption{DynamicRad: Candidate Construction and Dual-Mode Selection}
\label{alg:dual-mode-sparsification}
\begin{algorithmic}[1]
\Require $\mathbf{Q}, \mathbf{K}, N_f, N_t, B_s$, mode, configuration $\mathbf{x}$
\Ensure Block-sparse mask $\mathbf{M}_b \in \{0,1\}^{S_b \times S_b}$
\State Initialize $\mathbf{M}_b \leftarrow \mathbf{0}$
\For{each frame pair $(i,j)$}
    \State $t \leftarrow |i-j|$
    \If{$t = 0$}
        \State Mark all blocks within the intra-frame region in $\mathbf{M}_b$ \Comment{dense intra-frame attention}
        \State \textbf{continue}
    \EndIf
    \State Compute $w(i,j)$ by Eq.~\ref{eq:window_width} and $\mathrm{sf}(t)$ by Eq.~\ref{eq:split_factor}
    \If{$t \bmod \mathrm{sf}(t) \neq 0$}
        \State \textbf{continue}
    \EndIf
    \State Build $\mathcal{P}(i,j)$ by Eq.~\ref{eq:candidate_set}
    \If{mode = \texttt{static\_ratio}}
        \State Sample $k=\max(1,\lfloor |\mathcal{P}(i,j)|\rho(i,j)\rfloor)$ pairs
    \Else
        \State Compute $\tilde{s}(u,v)$ and keep pairs with $\tilde{s}(u,v)\ge \tau(i,j)$
        \If{no pair is kept}
            \State Keep the highest-scoring pair
        \EndIf
    \EndIf
    \State Aggregate selected pairs into a local block mask using $(\theta_m,\theta_c)$
    \State If Dynamic kept pairs but local mask is empty, activate the top-scoring pair's block
    \State Merge the local block mask into $\mathbf{M}_b$
\EndFor
\State \Return $\mathbf{M}_b$
\end{algorithmic}
\end{algorithm}

\subsection{Parameter-Efficient Fine-Tuning for Mask-Aware Refinement}
\label{subsec:lora_finetune}
Similar to adaptive rank allocation strategies like AdaLoRA \citep{zhang2023adalora}, our Mask-Aware LoRA selectively refines projections that are most sensitive to the sparsity pattern. While sparse selection reduces computation, aggressive sparsity may impair temporal consistency for long-horizon generation. We adopt a parameter-efficient fine-tuning strategy that adapts only mask-related components while freezing the diffusion backbone.
\paragraph{Where to adapt.}
Let the attention logits be
\begin{equation}
\label{eq:attn_logits}
\mathbf{A} = \frac{\mathbf{Q}\mathbf{K}^\top}{\sqrt{d_k}} \in \mathbb{R}^{S'\times S'},
\end{equation}
and let $\mathbf{M}_b\in\{0,1\}^{S_b\times S_b}$ denote the final block-sparse mask. We expand $\mathbf{M}_b$ to a token-level mask $\widetilde{\mathbf{M}}\in\{0,1\}^{S'\times S'}$ by broadcasting each block to its corresponding token region. The masked attention is computed as
\begin{equation}
\label{eq:masked_attn}
\text{Attn}(\mathbf{Q},\mathbf{K},\mathbf{V}) = \text{Softmax}\!\left(\mathbf{A} + \log(\widetilde{\mathbf{M}})\right)\mathbf{V}
\end{equation}
where the logarithm is applied elementwise with $\log 0=-\infty$.
In DynamicRad, the dynamic threshold mode computes approximate scores (Eq.~\ref{eq:dynamic_score}) and applies adaptive thresholds (Eq.~\ref{eq:dynamic_threshold_selector}). We apply LoRA adapters to (i) the selector-related projections and (ii) the $Q/K$ projection layers that shape the score geometry under sparsity, while freezing the rest of the backbone.

\subsection{Automated Configuration: Semantic-Aware Motion Routing}
\label{subsec:bo_predictor}
Since text prompts inherently encode semantic motion priors (e.g., 'reading' vs. 'running') \citep{wang2024motionctrl, guo2024animatediff}, we leverage the prompt embedding to predict the global motion regime. Optimal sparsity relies heavily on the motion dynamics implied by the text prompt. Naive heuristic rules often fail to capture semantic nuances (e.g., ``a fast snail'' may imply low absolute motion despite the word ``fast''). To address this, we introduce a \textbf{Semantic Motion Router} that maps prompt embeddings directly to a BO-selected configuration regime.
\paragraph{Optimization Objective and Efficiency.}
Since running full video generation for parameter search is computationally prohibitive, we perform BO on the lightweight \textbf{Proxy Task} defined in \textbf{Appendix~\ref{app:bo_details}}. 
The objective is to minimize the \textbf{Mean Squared Error (MSE)} between the sparse approximation and the dense proxy attention, subject to a sparsity budget:
\begin{equation}
\label{eq:bo_objective}
\begin{split}
    \mathcal{L}(\mathbf{x}) &= \frac{\|\mathbf{A}_{\text{dense}}-\mathbf{A}_{\text{sparse}}(\mathbf{x})\|_F^2}{\|\mathbf{A}_{\text{dense}}\|_F^2} \\
    &\quad + \alpha \cdot \mathrm{ReLU}\!\big(\mathcal{T}-\mathcal{S}(\mathbf{x})\big)
\end{split}
\end{equation}
where $\mathcal{T}$ is the target sparsity (e.g., $80\%$) and $\mathcal{S}(\mathbf{x})$ is the achieved sparsity under configuration $\mathbf{x}$.
Thanks to the low-dimensional search space and the smoothness of the proxy landscape, the TPE optimizer typically \textbf{converges within 30 trials}, completing the profiling in $<15$ minutes. This ensures that DynamicRad can be rapidly adapted to new resolutions or aspect ratios.

\paragraph{Architecture and Inference.}
Since modern video diffusion models (e.g., HunyuanVideo) utilize heavy text encoders (e.g., LLM or CLIP) for conditioning, we can reuse the pre-computed text embeddings $\mathbf{e}_{\text{text}}$ to guide sparsity. We append a lightweight 2-layer MLP projection head $\phi(\cdot)$ to predict a motion intensity score:
\begin{equation}
    s_{\text{motion}} = \phi(\text{Pool}(\mathbf{e}_{\text{text}})) \in [0, 1].
\end{equation}
This score is discretized to select the motion regime $c \in \{\text{Low}, \text{Mid}, \text{High}\}$ (e.g., $s < 0.3 \to \text{Low}$), which indexes the offline BO lookup table.
This router introduces a small runtime overhead. Since $\mathbf{e}_{\text{text}}$ is pre-computed, the router requires a single MLP pass taking $<2$ms, which is small relative to the generation latency. At inference time, we encode the prompt, predict a motion regime, retrieve the corresponding sparsity configuration from the lookup table, and run generation with the selected mode.
\paragraph{Training with Optical Flow Proxy.}
To supervise $\phi(\cdot)$ without expensive human annotation, we employ a physics-grounded proxy task. We construct a dataset $\mathcal{D}_{\text{train}}$ from a subset of OpenVid-1M \citep{xue2024openvid}. For each video-text pair, we use RAFT \citep{teed2020raft} to estimate optical flow between adjacent frames and take its average magnitude as the motion supervision target $y_{\text{flow}}$. The projection head is trained to minimize the regression loss $\mathcal{L} = \| s_{\text{motion}} - \text{Norm}(y_{\text{flow}}) \|_2^2$. This aligns the semantic understanding of the router with the physical motion priors used in our offline BO optimization.

\section{Experiments}
\label{sec:experiments}
\subsection{Experimental Setup}
\label{subsec:exp_setups}

\paragraph{Implementation \& Hardware.}
We implement DynamicRad using \textbf{FlashInfer} kernels ($B_s=32$) on \textbf{NVIDIA H100 GPUs}.
Standard inference uses a single H100, while HunyuanVideo-241f ($>$200k tokens) employs an \textbf{8$\times$H100 cluster} (sequence parallelism) to prevent OOM.
The \textbf{Semantic Motion Router} is frozen during inference; its parameter search follows the offline BO pipeline (Sec.~\ref{subsec:bo_predictor}).
Unless otherwise specified, the main comparisons marked with $^\dagger$ report the \textbf{backbone-frozen DynamicRad sparse-attention policy} with an offline-trained, frozen router and without Mask-Aware LoRA.
Mask-Aware LoRA ($r=16$) is fine-tuned for 1k steps on a subset of OpenVid-1M using the standard diffusion denoising loss under the selected sparse mask, while keeping the backbone frozen, and is reported as an optional adaptation setting.
Detailed hyperparameters, overhead quantification, and proxy-task analysis are provided in the supplementary material.

\paragraph{Baselines \& Protocols.}
We compare DynamicRad against state-of-the-art sparse-attention baselines using a common block-sparse backend. Specifically, we evaluate \textbf{PowerAttention (PA)} \citep{chen2025powerattention}, \textbf{SVG} \citep{xi2025svg}, and \textbf{Radial Attention} \citep{li2025radialattention}, which are reported in Table~\ref{tab:main_results}, and our reproduction of \textbf{AdaSpa}~\citep{xia2025adaspa} in Appendix~\ref{app:rebuttal_baselines}.
\textit{Exclusion of other methods:} We discuss MOD-DiT~\citep{liu2026moddit} and BSA~\citep{zhan2025bsa},
but do not include them in the controlled timing table, which evaluates the implemented baselines with a common block-sparse backend.
We also exclude orthogonal kernel-level sparse-attention systems, such as 
LiteAttention~\citep{shmilovich2025liteattention} and 
BLASST~\citep{yuan2025blasst}, because their implementation assumptions and 
target deployment settings differ from our block-sparse radial attention framework.

Following \citep{li2025radialattention}, long-context sparse settings (241f/145f) use HunyuanVideo and Wan2.1-14B checkpoints fine-tuned on \textbf{OpenVid-1M} \citep{xue2024openvid}, while standard-length settings use original checkpoints; unless Mask-Aware LoRA is enabled, DynamicRad remains an inference-time sparse-attention policy. The Original-LX rows instead extrapolate native checkpoints without length adaptation. Full-FT Dense uses the same target-length-adapted checkpoint as the matched sparse variants and serves as their quality reference.
The Semantic Motion Router was trained on 20k OpenVid-1M video-text pairs using RAFT-derived optical-flow labels (taking $<1$h on a single H100).
For evaluation, we report \textbf{VBench} scores \citep{huang2023vbench} for subject consistency and \textbf{VisionReward} \citep{chen2024visionreward} for human-aligned aesthetic quality. VBench Total and Dynamic Degree appear in Appendix~\ref{app:rebuttal_quality}.

\subsection{Main Results}
\label{subsec:exp_default}
\paragraph{Efficiency--quality trade-off.}
Table~\ref{tab:main_results} shows that DynamicRad provides a strong efficiency--quality trade-off among the evaluated sparse-attention baselines. The \textit{static-ratio} mode serves as a throughput-first operating point. On HunyuanVideo-241f, it achieves a \textbf{2.54$\times$ speedup} (324s), surpassing Radial Attention (351s). Similarly, on Wan2.1-14B (145f), it achieves a \textbf{2.25$\times$ speedup} (vs.~Radial Attention's 2.02$\times$), reducing inference time compared to the dense baseline. The Routed operating point takes 342/2490s at 241f/145f versus Radial Attention's 351/2560s and has slightly higher reported quality, but the sparsity budgets differ; this is a trade-off comparison rather than matched-budget dominance.

\begin{table*}[t!]
\centering
\setlength{\tabcolsep}{2.0pt}
\renewcommand{\arraystretch}{0.98}
\caption{\textbf{Quality--Efficiency Comparison.}
All videos are 768p ($1280 \times 768$); HunyuanVideo-241f uses 8$\times$H100, and other settings use one H100.
Speedups use Original (standard length) or Original-LX (long length). Full-FT Dense is a checkpoint-matched quality reference; its latency and speedup are \textemdash{} because it is outside this efficiency comparison.
$^\dagger$: DynamicRad without Mask-Aware LoRA, with an offline-trained frozen router; $^\ddagger$: long-context SVG reproduction.
Sparsity is the fraction of padded dense-grid blocks absent from the final BSR execution index. S.C./A.Q./I.Q. denote VBench subject consistency, aesthetic quality, and imaging quality.
\textbf{Bold}: overall best; \underline{underline}: best baseline, or best DynamicRad when a baseline leads.}
\label{tab:main_results}

\resizebox{\linewidth}{!}{
\begin{tabular}{l c l c c c c c c c}
\toprule
\multirow{2}{*}{\textbf{Model}} &
\multirow{2}{*}{\textbf{\#Frames}} &
\multirow{2}{*}{\textbf{Method}} &
\multicolumn{4}{c}{\textbf{Quality} $\uparrow$} &
\multicolumn{3}{c}{\textbf{Efficiency}} \\
\cmidrule(lr){4-7}\cmidrule(lr){8-10}
& & &
\textbf{VisionReward} &
\textbf{S.C.} &
\textbf{A.Q.} &
\textbf{I.Q.} &
\textbf{Sparsity (\%)} &
\textbf{Latency (s)} $\downarrow$ &
\textbf{Speedup} \\
\midrule

\multirow{14}{*}{HunyuanVideo} & \multirow{6}{*}{120} &
Original & 0.141 & 0.962 & 0.645 & 0.675 & 0.0 & 1705 & 1.00 \\
& & PA & 0.138 & 0.956 & 0.637 & 0.662 & 70.5 & 1024 & 1.67 \\
& & SVG & \underline{0.144} & \textbf{0.965} & \underline{0.651} & \underline{0.679} & \underline{78.2} & \underline{872} & \underline{1.96} \\
& & Radial Attention & 0.140 & 0.961 & 0.646 & 0.673 & 76.6 & 884 & 1.93 \\
& & \cellcolor{gray!10}Ours (Static)$^\dagger$ & \cellcolor{gray!10}0.138 & \cellcolor{gray!10}0.959 & \cellcolor{gray!10}0.642 & \cellcolor{gray!10}0.668 & \cellcolor{gray!10}\textbf{84.8} & \cellcolor{gray!10}\textbf{861} & \cellcolor{gray!10}\textbf{1.98} \\
& & \cellcolor{gray!10}Ours (Dynamic)$^\dagger$ & \cellcolor{gray!10}\textbf{0.146} & \cellcolor{gray!10}\underline{0.961} & \cellcolor{gray!10}\textbf{0.653} & \cellcolor{gray!10}\textbf{0.681} & \cellcolor{gray!10}81.6 & \cellcolor{gray!10}917 & \cellcolor{gray!10}1.86 \\

\cmidrule{2-10}

& \multirow{8}{*}{241} &
Original-LX & 0.116 & 0.948 & 0.592 & 0.597 & 0.0 & 824 & 1.00 \\
& & Full-FT Dense & 0.122 & 0.949 & 0.619 & 0.644 & 0.0 & \textemdash & \textemdash \\
& & PA & 0.108 & 0.960 & 0.611 & 0.649 & 80.4 & \underline{346} & \underline{2.38} \\
& & SVG$^\ddagger$ & 0.121 & 0.957 & 0.614 & 0.651 & 80.2 & 408 & 2.02 \\
& & Radial Attention & \underline{0.125} & \underline{0.961} & \underline{0.626} & \underline{0.659} & \underline{80.6} & 351 & 2.35 \\
& & \cellcolor{gray!10}Ours (Static)$^\dagger$ & \cellcolor{gray!10}0.118 & \cellcolor{gray!10}0.951 & \cellcolor{gray!10}0.612 & \cellcolor{gray!10}0.668 & \cellcolor{gray!10}\textbf{86.8} & \cellcolor{gray!10}\textbf{324} & \cellcolor{gray!10}\textbf{2.54} \\
& & \cellcolor{gray!10}Ours (Dynamic)$^\dagger$ & \cellcolor{gray!10}\textbf{0.131} & \cellcolor{gray!10}\textbf{0.969} & \cellcolor{gray!10}\textbf{0.659} & \cellcolor{gray!10}\textbf{0.687} & \cellcolor{gray!10}83.6 & \cellcolor{gray!10}378 & \cellcolor{gray!10}2.18 \\
& & \cellcolor{gray!10}Ours (Routed)$^\dagger$ & \cellcolor{gray!10}0.127 & \cellcolor{gray!10}0.965 & \cellcolor{gray!10}0.642 & \cellcolor{gray!10}0.676 & \cellcolor{gray!10}84.7 & \cellcolor{gray!10}342 & \cellcolor{gray!10}2.41 \\

\midrule

\multirow{14}{*}{Wan2.1-14B} & \multirow{6}{*}{72} &
Original & \textbf{0.136} & \underline{0.958} & \underline{0.620} & \underline{0.650} & 0.0 & 1680 & 1.00 \\
& & PA & 0.126 & 0.944 & 0.594 & 0.620 & 66.3 & 1040 & 1.62 \\
& & SVG & 0.116 & 0.938 & 0.580 & 0.605 & 70.2 & 990 & 1.70 \\
& & Radial Attention & 0.129 & 0.952 & 0.610 & 0.641 & \underline{70.4} & \underline{945} & \underline{1.78} \\
& & \cellcolor{gray!10}Ours (Static)$^\dagger$ & \cellcolor{gray!10}0.127 & \cellcolor{gray!10}0.949 & \cellcolor{gray!10}0.606 & \cellcolor{gray!10}0.637 & \cellcolor{gray!10}\textbf{78.8} & \cellcolor{gray!10}\textbf{875} & \cellcolor{gray!10}\textbf{1.92} \\
& & \cellcolor{gray!10}Ours (Dynamic)$^\dagger$ & \cellcolor{gray!10}\underline{0.134} & \cellcolor{gray!10}\textbf{0.959} & \cellcolor{gray!10}\textbf{0.621} & \cellcolor{gray!10}\textbf{0.656} & \cellcolor{gray!10}74.3 & \cellcolor{gray!10}980 & \cellcolor{gray!10}1.71 \\

\cmidrule{2-10}

& \multirow{8}{*}{145} &
Original-LX & 0.098 & 0.921 & 0.573 & 0.585 & 0.0 & 5160 & 1.00 \\
& & Full-FT Dense & \underline{0.146} & 0.958 & 0.586 & \underline{0.673} & 0.0 & \textemdash & \textemdash \\
& & PA & 0.132 & 0.963 & 0.592 & 0.642 & 73.4 & \underline{2550} & \underline{2.02} \\
& & SVG$^\ddagger$ & 0.133 & 0.965 & 0.596 & 0.648 & \underline{73.8} & 3035 & 1.70 \\
& & Radial Attention & 0.142 & \underline{0.972} & \underline{0.602} & 0.661 & 73.3 & 2560 & 2.02 \\
& & \cellcolor{gray!10}Ours (Static)$^\dagger$ & \cellcolor{gray!10}0.137 & \cellcolor{gray!10}0.969 & \cellcolor{gray!10}0.606 & \cellcolor{gray!10}0.647 & \cellcolor{gray!10}\textbf{81.6} & \cellcolor{gray!10}\textbf{2289} & \cellcolor{gray!10}\textbf{2.25} \\
& & \cellcolor{gray!10}Ours (Dynamic)$^\dagger$ & \cellcolor{gray!10}\textbf{0.148} & \cellcolor{gray!10}\textbf{0.979} & \cellcolor{gray!10}\textbf{0.621} & \cellcolor{gray!10}\textbf{0.686} & \cellcolor{gray!10}77.0 & \cellcolor{gray!10}2850 & \cellcolor{gray!10}1.81 \\
& & \cellcolor{gray!10}Ours (Routed)$^\dagger$ & \cellcolor{gray!10}0.144 & \cellcolor{gray!10}0.974 & \cellcolor{gray!10}0.613 & \cellcolor{gray!10}0.670 & \cellcolor{gray!10}78.6 & \cellcolor{gray!10}2490 & \cellcolor{gray!10}2.07 \\

\bottomrule
\end{tabular}
}
\end{table*}

\paragraph{Quality preservation.}
The \textit{dynamic-threshold} mode preserves quality under high sparsity. On HunyuanVideo-241f, it improves over checkpoint-matched Full-FT Dense on VisionReward (0.131 vs.~0.122). We hypothesize that content-aware filtering removes low-utility long-range interactions in long sequences, although this effect may depend on the prompt and model setting. On Wan-145f, VisionReward is 0.148 versus 0.146 for matched dense. VBench Total differs by only +0.14/+0.17 points (Appendix~\ref{app:rebuttal_quality}), supporting quality preservation under the matched evaluation protocol.
\paragraph{Robustness Across Architectures.}
The gains are consistent across HunyuanVideo and Wan2.1-14B. On Wan2.1-14B (72f), static mode maintains a \textbf{1.92$\times$ speedup} with \textbf{78.8\% sparsity}, while dynamic mode matches dense temporal consistency (VBench S.C. 0.959 vs.~0.958), suggesting transfer across different tokenization strategies. An A800 run confirms the same acceleration trend (Appendix~\ref{app:rebuttal_timing}).
\subsection{Ablation Study}
\label{subsec:exp_ablation}
\begin{figure*}[t!]
    \centering
    \begin{minipage}{0.98\textwidth}
        \fcolorbox{gray!50}{gray!10}{
            \parbox{0.97\linewidth}{
                \textbf{\small Prompt (Low-Motion):} \small \textit{“A high-quality portrait of an elderly woman reading a book by a window, static camera, minimal movement, sunlight gently illuminating her face, cinematic lighting.”}
            }
        }
        \centering
        \begin{subfigure}[b]{0.22\linewidth}
            \includegraphics[width=\linewidth]{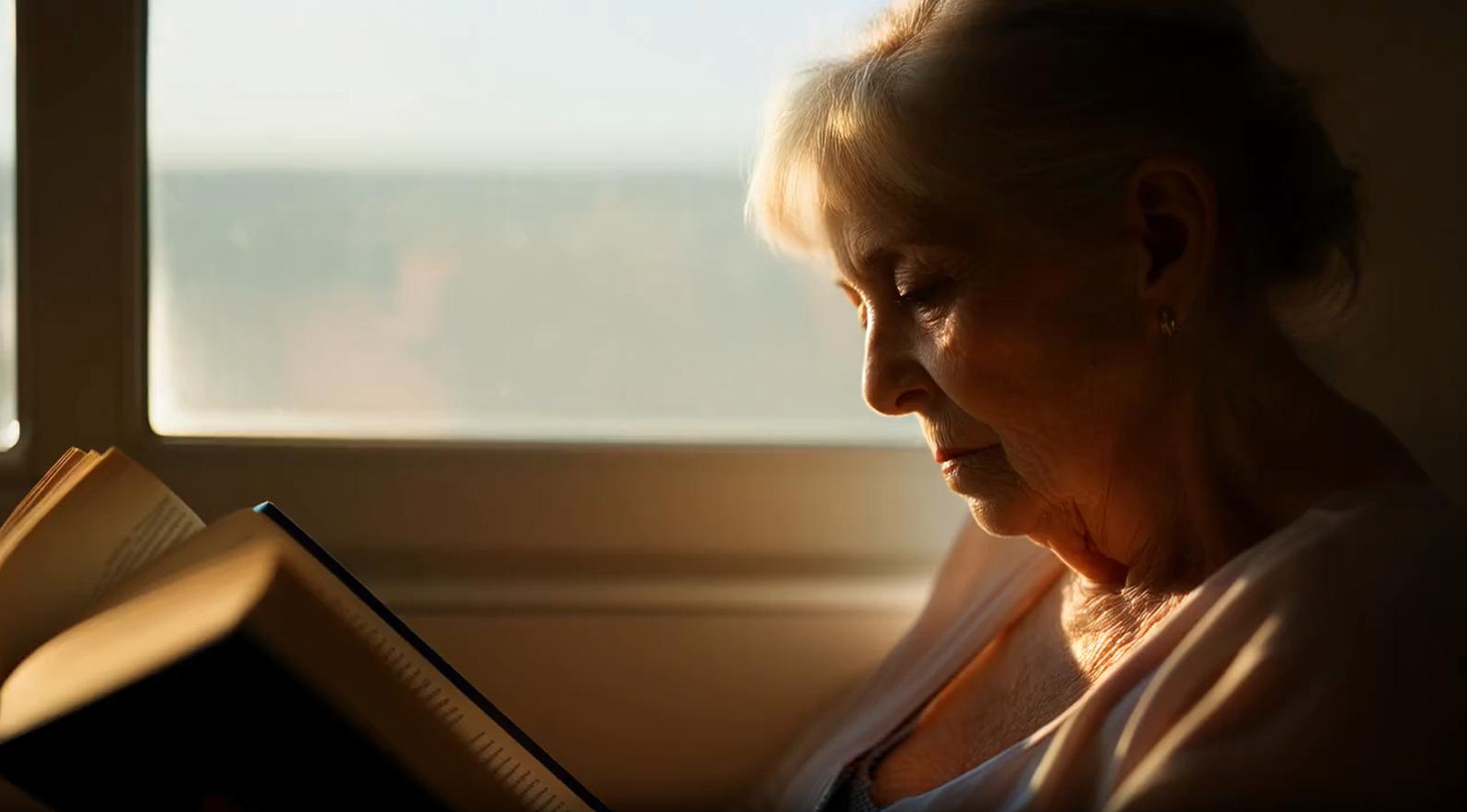}
            \caption*{\scriptsize Frame $t=0$}
        \end{subfigure}
        \hfill
        \begin{subfigure}[b]{0.22\linewidth}
            \includegraphics[width=\linewidth]{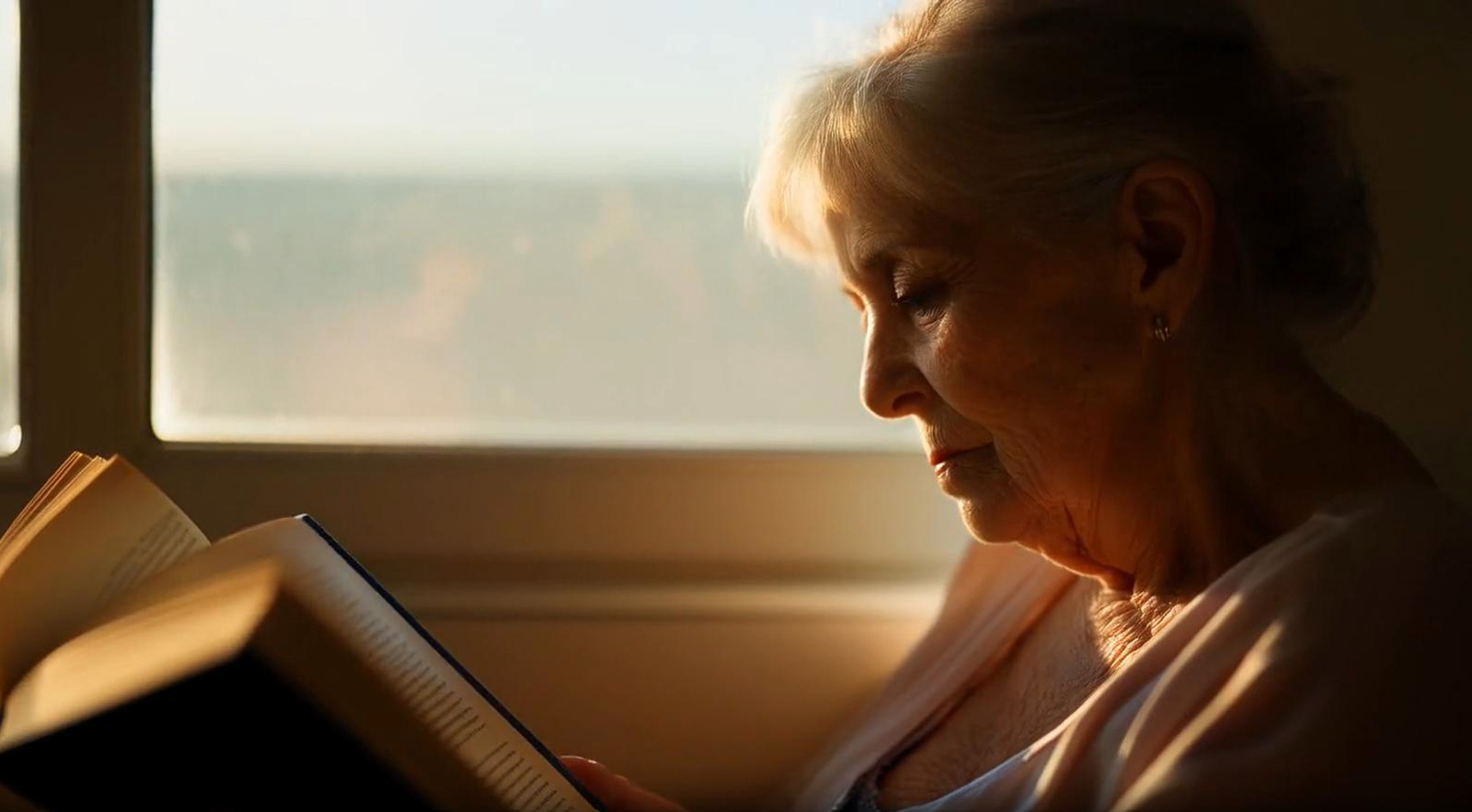}
            \caption*{\scriptsize Frame $t=60$}
        \end{subfigure}
        \hfill
        \begin{subfigure}[b]{0.22\linewidth}
            \includegraphics[width=\linewidth]{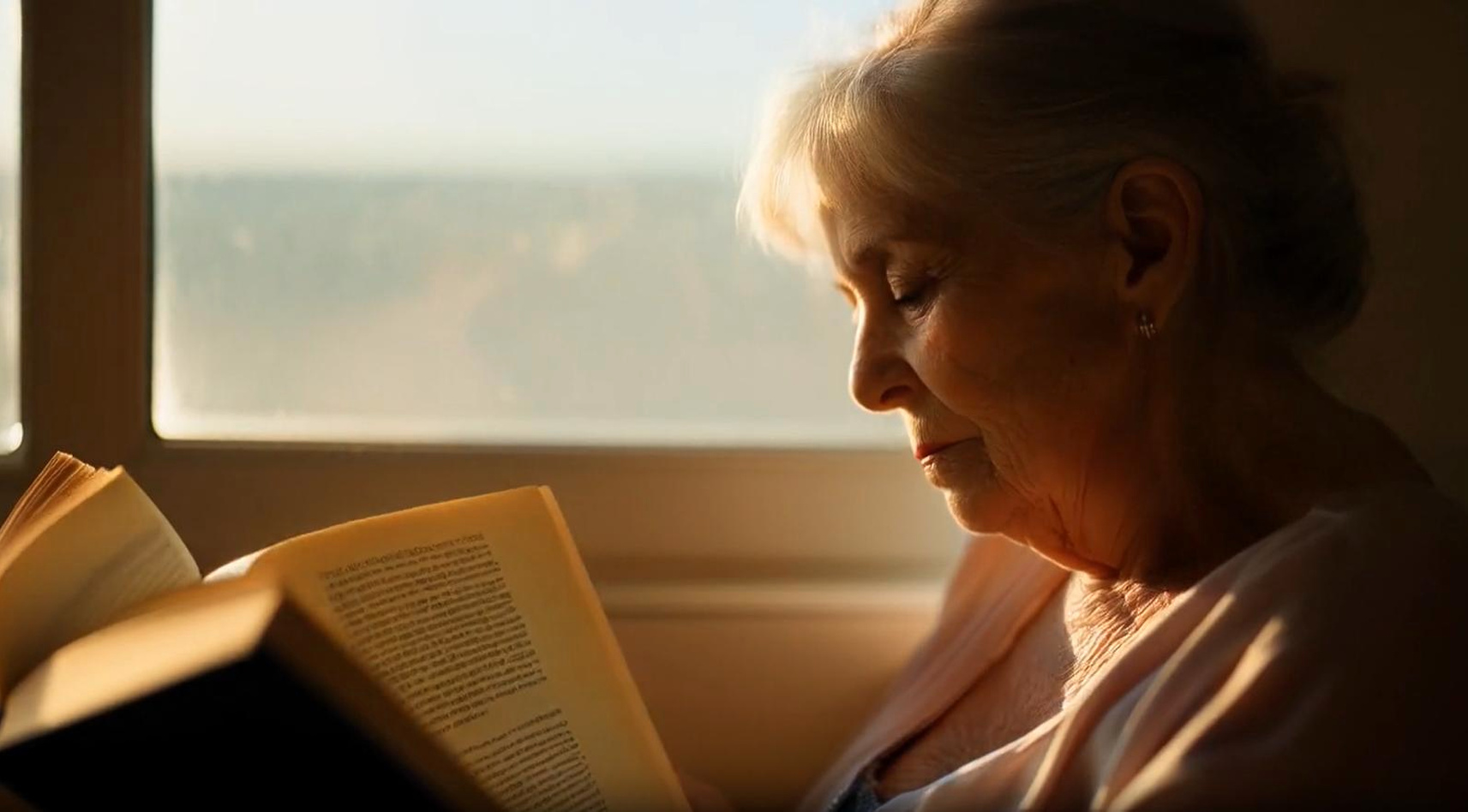}
            \caption*{\scriptsize Frame $t=120$}
        \end{subfigure}
        \hfill
        \begin{subfigure}[b]{0.02\linewidth}
            \centering  $\Rightarrow$
        \end{subfigure}
        \hfill
        \begin{subfigure}[b]{0.22\linewidth}
            \setlength{\fboxsep}{1pt} \setlength{\fboxrule}{0.8pt}
            \fcolorbox{black}{white}{\includegraphics[width=\linewidth]{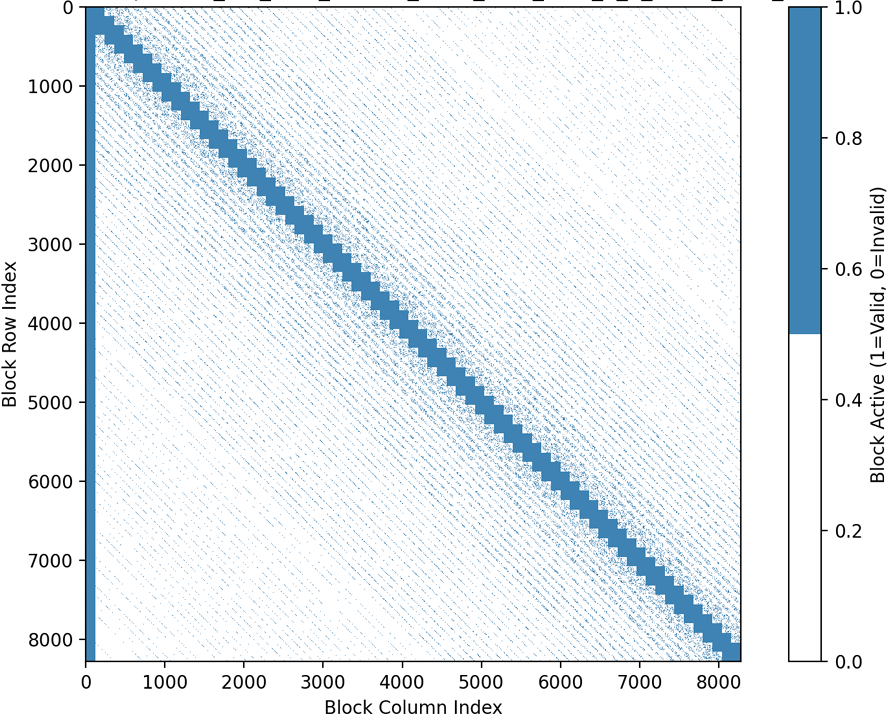}}
            \caption*{\textbf{\scriptsize Static Mask (86.6\%)} \\ \tiny (Local \& Diagonal Only)}
        \end{subfigure}
    \end{minipage}

    \begin{minipage}{0.98\textwidth}
        \fcolorbox{gray!50}{gray!10}{
            \parbox{0.97\linewidth}{
                \textbf{\small Prompt (High-Motion):} \small \textit{“FPV drone shot flying extremely fast through a narrow cyberpunk canyon, chasing a futuristic racing car, neon lights, motion blur, rapid camera movement.”}
            }
        }
        \centering
        \begin{subfigure}[b]{0.22\linewidth}
            \includegraphics[width=\linewidth]{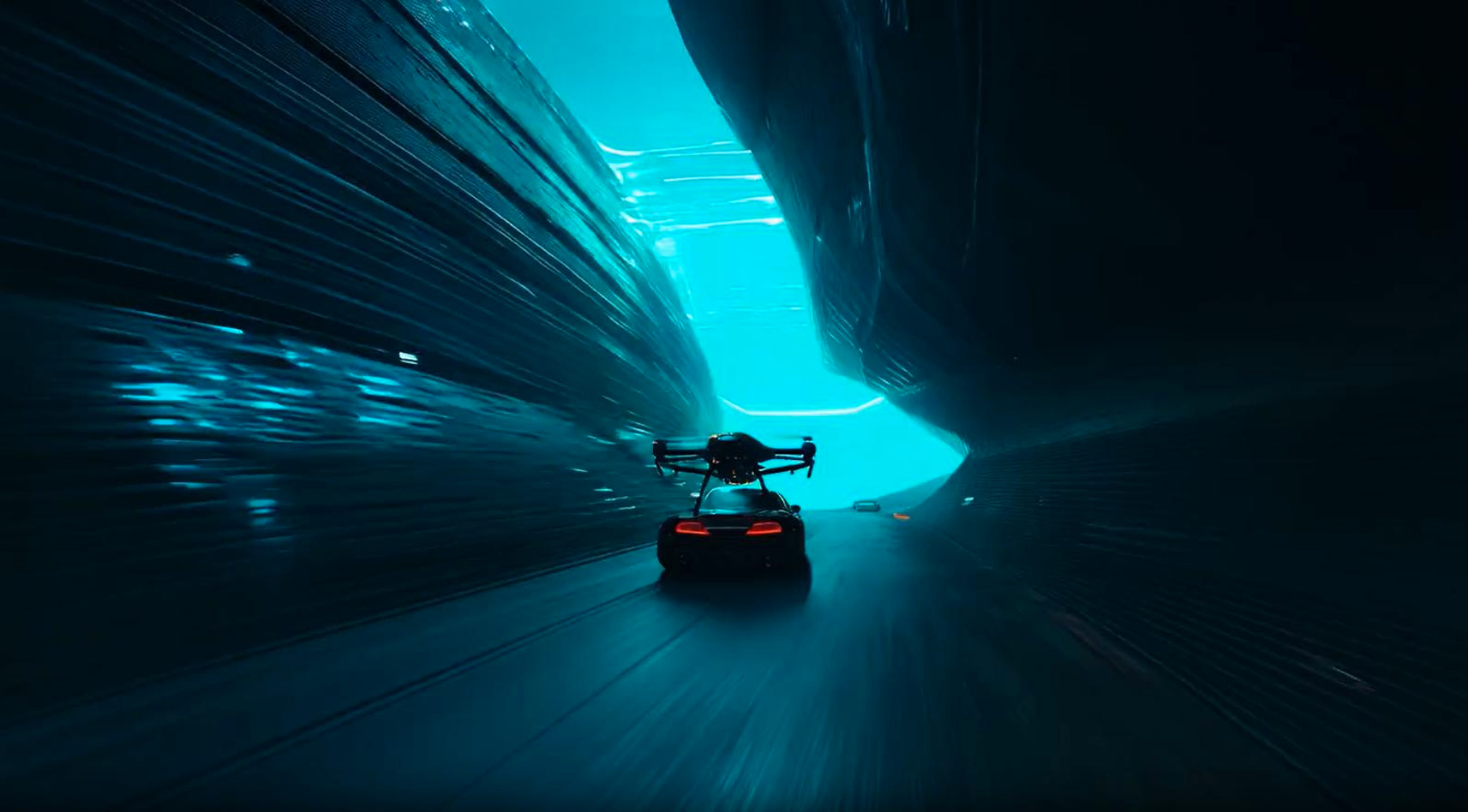}
            \caption*{\scriptsize Frame $t=0$}
        \end{subfigure}
        \hfill
        \begin{subfigure}[b]{0.22\linewidth}
            \includegraphics[width=\linewidth]{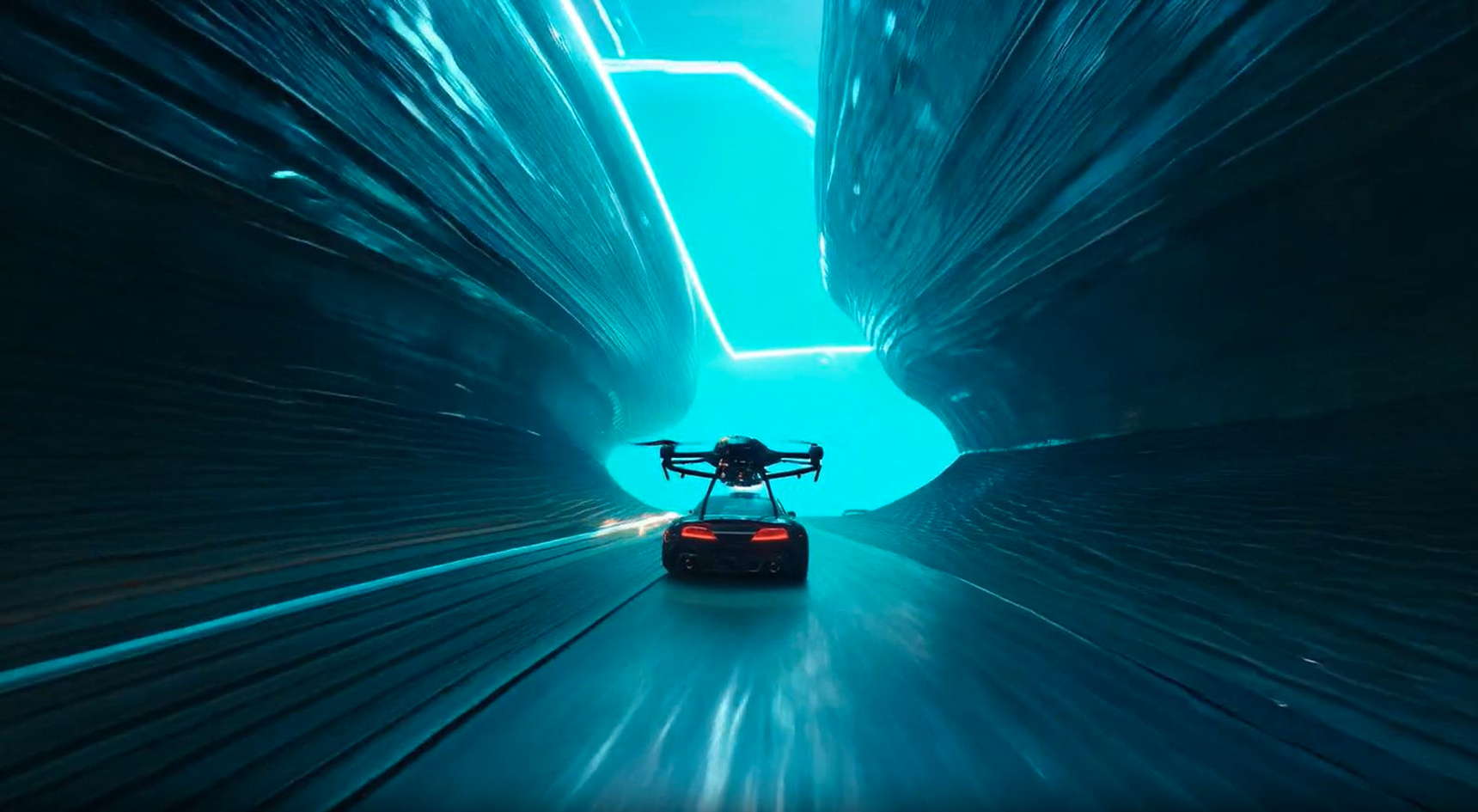}
            \caption*{\scriptsize Frame $t=60$}
        \end{subfigure}
        \hfill
        \begin{subfigure}[b]{0.22\linewidth}
            \includegraphics[width=\linewidth]{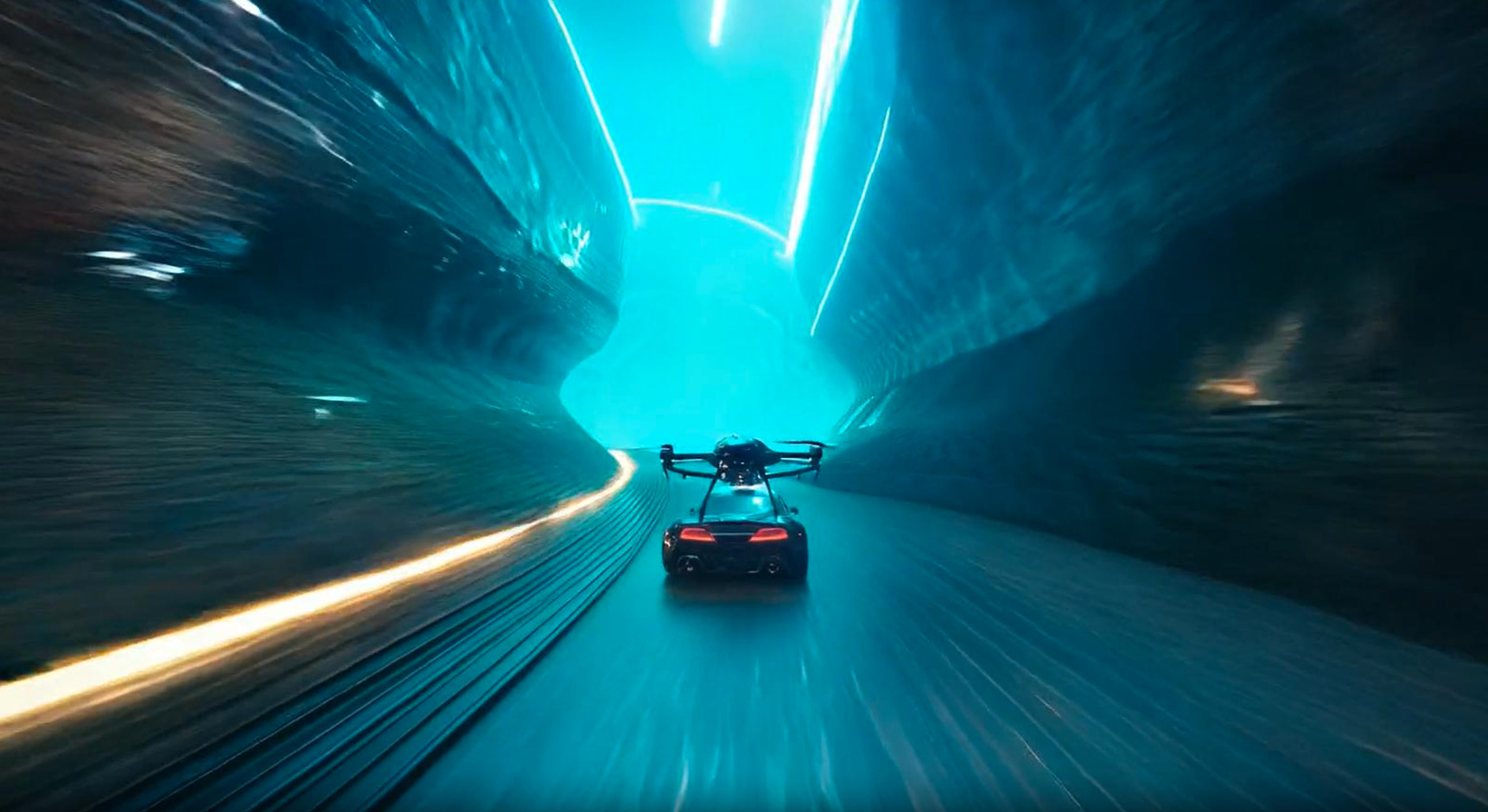}
            \caption*{\scriptsize Frame $t=120$}
        \end{subfigure}
        \hfill
        \begin{subfigure}[b]{0.02\linewidth}
            \centering  $\Rightarrow$
        \end{subfigure}
        \hfill
        \begin{subfigure}[b]{0.22\linewidth}
            \setlength{\fboxsep}{1pt} \setlength{\fboxrule}{0.8pt}
            \fcolorbox{red}{white}{\includegraphics[width=\linewidth]{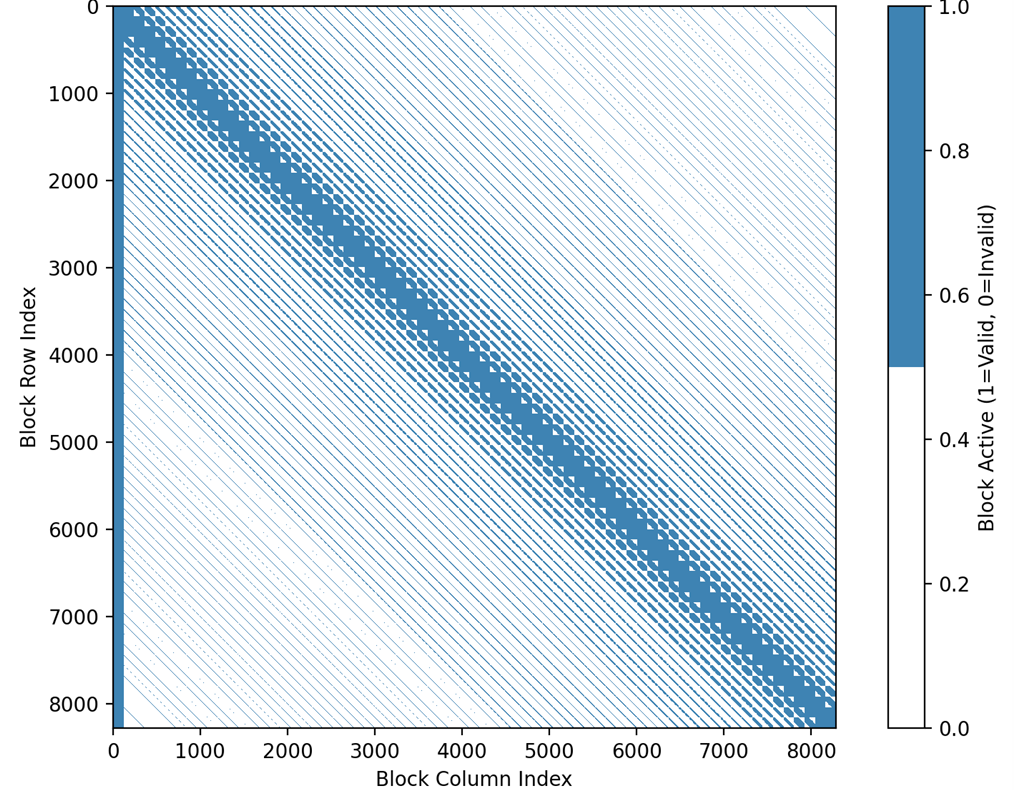}}
            \caption*{\textbf{\scriptsize Dynamic Mask (77.9\%)} \\ \tiny (Captures Long-Range)}
        \end{subfigure}
    \end{minipage}

    \caption{\textbf{Regime-aware adaptivity.} For low-motion prompts, the router selects \textit{static-ratio}, yielding a highly sparse near-diagonal mask. For high-motion prompts, it switches to \textit{dynamic-threshold}, preserving off-diagonal long-range connections useful for fast motion.}
    \label{fig:qualitative_vis}
\end{figure*}

\paragraph{Configurable Parameters \& Dual-Mode Selection.}
The introduction of configurable knobs ($\lambda, B_s, \theta$) is crucial for stability across resolutions. On top of this, the selection mode serves as the primary trade-off lever:
\textbf{Static-ratio mode} acts as an efficiency-first option. Compared to a naive configurable baseline, our static optimization (Dual-mode pruning) reduces latency to \textbf{338s}.
\textbf{Dynamic-threshold mode} utilizes attention scores (Eq.~\ref{eq:dynamic_score}) to filter tokens. Although it introduces a slight overhead (385s in this ablation), it recovers fine-grained details (VisionReward stabilizes around 0.119) relative to the selector-only Static row, which is not the deployed Static operating point.
\paragraph{Regime-Aware Adaptivity (Offline Predictor).}
The \textbf{Offline BO Predictor} (Sec.~\ref{subsec:bo_predictor}) is an important component of DynamicRad. As shown in \textbf{Table~\ref{tab:synthetic_ablation_wide}}, the predictor identifies BO-selected sparsity regimes for the specific motion context, effectively reducing latency further (Static mode: $338s \to \mathbf{324s}$) while improving quality (VisionReward $0.117 \to 0.122$).
\textbf{Table~\ref{tab:bo_lookup}} and \textbf{Figure~\ref{fig:qualitative_vis}} illustrate this trend: the system assigns aggressive static masking ($\gamma=2.0$, \textbf{86.6\%} sparsity) to low-motion prompts (e.g., reading) while automatically switching to dynamic selection with denser long-range connections for high-motion inputs.

\paragraph{Mask-Aware LoRA.}
Finally, integrating \textbf{Mask-Aware LoRA} (Sec.~\ref{subsec:lora_finetune}) further improves temporal consistency (VBench S.C. reaches \textbf{0.977}). By refining the logit distribution to accommodate the sparse structure, MA-LoRA mitigates aggressive pruning artifacts with a modest latency increase from 378s to 389s.
\Needspace{6\baselineskip}
\paragraph{Micro-Benchmark of Routing Overhead.}
On H100/Wan-72f, the deployed Dynamic path measures \textbf{1.71 ms} per invocation for QK scoring, thresholding, block aggregation, index construction, and planning; the router runs once per video. Complete generation takes $980\pm9$s versus $1680\pm15$s for dense attention (1.71$\times$ speedup), including all generation and host overheads. Appendix~\ref{app:rebuttal_timing} gives the component protocol and A800 result.

\begin{table*}[t!]
\centering
\caption{\textbf{BO-Optimized Configurations.} Low motion permits 86.6\% sparsity; high motion uses larger $\lambda$ for denser long-range links.
Near/far values are retention ratios $\rho$ (Static) or thresholds $\tau$ (Dynamic). Green and red shading mark speed- and quality-oriented configurations.}
\label{tab:bo_lookup}
\setlength{\tabcolsep}{3.0pt}
\renewcommand{\arraystretch}{0.90}
\resizebox{\linewidth}{!}{
\begin{tabular}{l l c c c c c c c l}
\toprule
\multirow{2}{*}{\textbf{Motion}} & \multirow{2}{*}{\textbf{Mode}} & \textbf{Width} & \textbf{Long} & \textbf{Mask} & \textbf{Column} & \textbf{Near} & \textbf{Far} & \textbf{Best} & \multirow{2}{*}{\textbf{Rationale}} \\
& & ($\gamma$) & ($\lambda$) & ($\theta_m$) & ($\theta_c$) & ($\rho_1 / \tau_1$) & ($\rho_2 / \tau_2$) & \textbf{Sparsity} & \\
\midrule
\multirow{2}{*}{\textbf{Low}} 
& Static & 2.0 & 0.3 & 0.75 & 0.20 & 0.25 & 0.55 & \textbf{86.6\%} & \cellcolor{green!10}Low retention; sparse long-range. \\
& Dynamic & 1.3 & 0.4 & 0.70 & 0.45 & -1.2 & 2.4 & 82.7\% & Balanced profile. \\
\midrule
\multirow{2}{*}{\textbf{Mid}} 
& Static & 1.3 & 0.7 & 0.60 & 0.15 & 0.75 & 0.55 & 84.0\% & Target satisfied. \\
& Dynamic & 1.4 & 0.7 & 0.70 & 0.45 & -1.5 & 2.0 & 80.6\% & Target satisfied. \\
\midrule
\multirow{2}{*}{\textbf{High}} 
& Static & 1.6 & 0.8 & 0.65 & 0.15 & 0.55 & 0.50 & 81.7\% & \cellcolor{red!10}Denser local budget. \\
& Dynamic & 1.6 & 0.9 & 0.55 & 0.40 & -2.0 & 2.4 & \textbf{77.9\%} & \cellcolor{red!10}Higher $\lambda$; loose thresholds. \\
\bottomrule
\end{tabular}
}
\end{table*}

\begin{table*}[t!]
\centering
\caption{\textbf{Component ablations on HunyuanVideo-241f at 768p (8$\times$H100).}
The backbone-frozen sparse policy and optional Mask-Aware LoRA are evaluated on a fixed prompt--seed subset; quality scores may therefore differ slightly from Table~\ref{tab:main_results}.}
\label{tab:synthetic_ablation_wide}
\setlength{\tabcolsep}{4pt}
\small
\begin{adjustbox}{width=\linewidth,center}
\begin{tabular}{l|c|c|c|c}
\toprule
Variant (HunyuanVideo, 241f, 768p, \textbf{8$\times$H100}) & Sparsity$\uparrow$ & Latency(s)$\downarrow$ & VisionReward$\uparrow$ & VBench S.C.$\uparrow$ \\
\midrule
Radial Attention + Configurable params & 81.3\% & 351 & 0.123 & 0.956 \\
\midrule
+ Dual-mode pruning (Ours, Static mode) & 85.5\% & 338 & 0.117 & 0.946 \\
+ Dual-mode pruning (Ours, Dynamic mode) & 82.3\% & 385 & 0.119 & 0.948 \\
\midrule
+ Offline predictor (auto config, Static mode) & \textbf{86.8}\% & \textbf{324} & 0.122 & 0.956 \\
+ Offline predictor (auto config, Dynamic mode) & 83.6\% & 378 & 0.127 & 0.963 \\
\midrule
+ MA-LoRA (Mask-Aware LoRA, applied on Dynamic mode) & 83.6\% & 389 & \textbf{0.143} & \textbf{0.977} \\
\bottomrule
\end{tabular}
\end{adjustbox}
\end{table*}

\FloatBarrier
\section{Conclusion}
\label{sec:conclusion}

We presented \textbf{DynamicRad}, a sparse-attention framework that combines a kernel-friendly radial candidate set with content-adaptive selection for long video diffusion. By implementing a dual-mode strategy via an offline BO-driven router, DynamicRad achieves \textbf{1.7$\times$--2.5$\times$ speedups} with \textbf{74.3\%--86.8\% effective block sparsity} on HunyuanVideo and Wan2.1-14B. As an optional adaptation path, lightweight Mask-Aware LoRA further preserves temporal coherence under aggressive pruning.

\paragraph{Limitations and Future Work.}
DynamicRad still has several boundary conditions. 
(1) \textbf{Extreme Fast Motion:} Under highly dynamic prompts (e.g., FPV drone racing), minor artifacts may still appear because localized block-sparse patterns cannot fully capture highly dispersed attention. 
(2) \textbf{Ambiguous or Multi-Phase Prompts:} The Semantic Motion Router predicts a global motion regime from text; for ambiguous prompts or videos with temporal phase shifts, a single fixed regime may be suboptimal, although dynamic fallbacks mitigate this issue. 
(3) \textbf{Hardware Generalization:} Because the method relies on specific block sizes ($B_s$), non-server GPUs may require local tuning and LUT recalibration. 
Future work will study frame-level dynamic routing and extend the regime-aware design to image-to-video generation.

\begin{ack}
The project was supported by the National Natural Science Foundation of China (Grant No. W2431044).

The authors have no competing interests to disclose.
\end{ack}

\clearpage
\setlength{\bibsep}{4pt}
\bibliographystyle{plainnat}
\bibliography{references}
\clearpage
\appendix
\onecolumn

\section{Illustrative Complexity Scaling}
\label{app:complexity}

This section compares dense attention with an idealized radial-decay model to illustrate the scaling of the locality prior. Let $N_f$ denote the number of frames, $N_t$ the number of tokens per frame, and $D$ the feature dimension. The total sequence length is $S = N_f N_t$.

\subsection{Standard Dense Attention}
The standard self-attention mechanism computes the similarity between all pairs of tokens. The computational cost consists of two parts: the matrix multiplication $\mathbf{Q}\mathbf{K}^\top$ and the attention-value aggregation $\mathbf{A}\mathbf{V}$.
\begin{equation}
    \mathcal{C}_{\text{dense}} = \mathcal{O}(S^2 D) = \mathcal{O}(N_f^2 N_t^2 D).
\end{equation}
The memory complexity for storing the attention map is $\mathcal{O}(S^2)$, which is quadratic with respect to the frame count $N_f$.

\subsection{Idealized Radial-Decay Model}
For this illustration, let the inter-frame window be $\widetilde{w}(t)=L_0\gamma/t$ for $t\geq 1$, before applying the minimum width $B_s$, the long-range split rule, either mode's intra-candidate selection, or block aggregation. There are at most two frames at each temporal distance from frame $i$, so for $N_f\geq 2$ the window-sum proxy satisfies
\begin{equation}
    \widetilde{\Omega}_i
    = \sum_{\substack{j=0\\j\neq i}}^{N_f-1}\widetilde{w}(|i-j|)
    \leq 2L_0\gamma\sum_{t=1}^{N_f-1}\frac{1}{t}
    \leq 2L_0\gamma\bigl(1+\ln(N_f-1)\bigr).
\end{equation}
Assuming attention arithmetic is proportional to this idealized link count, and treating $\gamma$ as constant, $L_0=\Theta(N_t)$ gives
\begin{equation}
    \widetilde{\mathcal{C}}_{\text{radial}}
    = \mathcal{O}(S\widetilde{\Omega}_{\mathrm{avg}}D)
    = \mathcal{O}(N_fN_t^2\ln N_f D).
\end{equation}
Relative to dense attention, this idealized arithmetic estimate scales as $\mathcal{O}(\ln N_f/N_f)$ in the frame count. It is an illustration of the radial prior, not a complexity bound for the deployed DynamicRad algorithm: the minimum block width, split retention, mode-specific selection, block aggregation, and mask-construction overhead determine its actual cost.

\section{Implementation Details: Efficiency and Robustness}
\label{app:selector_efficiency}

A common concern for dynamic sparse attention is the trade-off between implementation overhead and stability. In this section, we detail how DynamicRad minimizes computational costs via proxy scoring and ensures robustness via fallback mechanisms.

\paragraph{Head Fusion Proxy (Method).} 
Instead of computing scores across all heads (e.g., $H=32$), we hypothesize that attention structures are shared. We compute filtering logits using only the first $H_f=2$ heads:
\begin{equation}
    S_{ij}^{\text{proxy}} = \frac{1}{H_f} \sum_{h=1}^{H_f} \frac{Q_{i,h} K_{j,h}^\top}{\sqrt{d}}.
\end{equation}

\paragraph{Theoretical Overhead Analysis.}
The following summarizes the theoretical cost reductions; a measured H100/Wan-72f latency breakdown is reported in Appendix~\ref{app:micro_benchmark}:
\begin{itemize}
    \item \textbf{Selection Cost Reduction:} By using $H_f=2$ out of $H=32$ heads, the logit computation cost is reduced by a factor of \textbf{16$\times$}. The subsequent sorting and thresholding are performed on this reduced subset, ensuring the selection overhead remains a small fraction of the full $\mathcal{O}(N^2)$ attention.
    \item \textbf{Router Efficiency:} The Semantic Motion Router uses a lightweight MLP on pre-computed embeddings. Its complexity is $\mathcal{O}(1)$ relative to the video length $N_f$, incurring negligible cost compared to the quadratic attention mechanism.
\end{itemize}

\paragraph{Robustness and Fallback Mechanisms.}
For ambiguous prompts where the Semantic Router predicts low confidence (or classifies incorrectly), DynamicRad employs two safeguards to prevent quality collapse:
\begin{enumerate}
    \item \textbf{Score-Based Fallback:} In the \textit{dynamic-threshold} mode, if the threshold rejects every pair in a nonempty frame-pair candidate set, we retain its highest-scoring pair. If selected pairs are nonempty but no block survives aggregation, we activate the block containing the highest-scoring candidate pair. This prevents the final block mask for the retained frame pair from being empty.
    \item \textbf{Conservative Default:} If the router output is out-of-distribution (e.g., gibberish prompts), the system defaults to the \textit{Static-ratio (Mid-Motion)} profile, providing a safe baseline performance similar to standard sparse attention.
\end{enumerate}


\section{Offline Bayesian Optimization Details}
\label{app:bo_details}

To ensure DynamicRad adapts robustly to varying video dynamics without incurring online tuning costs, we utilize a rigorous offline Bayesian Optimization (BO) strategy. This section details the proxy task formulation, clarifies its scope, and analyzes the optimization convergence.

\subsection{Detailed Hyperparameter Configuration}
\label{app:hyperparams}

To support reproducibility, Table~\ref{tab:hyperparams} summarizes the fixed system parameters and the search ranges explored by the Bayesian optimizer. Structural parameters are fixed to match hardware-aware kernel constraints, whereas sparsity-related filtering parameters are optimized offline within the ranges listed below.

\begin{table}[H]
\centering
\caption{\textbf{Hyperparameter Settings and Optimization Bounds.} 
System parameters are fixed within each run rather than optimized by BO. For mode-specific parameters, we report the \textbf{Search Range} $[min, max]$ explored by the Bayesian Optimizer.}
\label{tab:hyperparams}
\vspace{2mm}

\resizebox{\linewidth}{!}{
\begin{tabular}{l c c l}
\toprule
\multicolumn{4}{c}{\textbf{I. Fixed System Parameters}} \\
\midrule
\textbf{Parameter} & \textbf{Value} & \multicolumn{2}{c}{\textbf{Description}} \\
\midrule
Block Size ($B_s$) & 128/64/32 & \multicolumn{2}{l}{Fixed granularity for FlashInfer/Sage kernels.} \\
Fused Heads ($H_f$) & 2 & \multicolumn{2}{l}{Fixed proxy for fast dynamic scoring.} \\
\midrule

\multicolumn{4}{c}{\textbf{II. Adaptive Mode Parameters (Search Ranges)}} \\
\midrule
\textbf{Mode} & \textbf{Parameter} & \textbf{Optimization Range (via BO)} & \textbf{Description} \\
\midrule
\multirow{4}{*}{\textit{Common}} 
& Width Scale ($\gamma$) & $[1.0, 3.0]$ & Controls radial window width. \\
& Long Factor ($\lambda$) & $[0.1, 1.0]$ & Controls long-range split density. \\
& Col Density Thresh ($\theta_c$) & $[0.1, 1.0]$ & Minimal density to consider a column ``active''. \\
& Mask Threshold ($\theta_m$) & $[0.1, 1.0]$ & Range for block activation threshold. \\
\midrule
\multirow{2}{*}{\textit{Static-ratio}} 
& topk\_ratio1 & $[0.1, 1.0]$ & For near-field blocks. \\
& topk\_ratio2 & $[0.1, 0.8]$ & For far-field blocks. \\
\midrule
\multirow{2}{*}{\textit{Dynamic-threshold}} 
& near\_frame\_thresh & $[-10.0, 5.0]$ & Threshold on $\tilde{s}$ for $t>1$, $\mathcal{L}(t)\ge B_s$. \\
& far\_frame\_thresh & $[-5.0, 8.0]$ & Threshold on $\tilde{s}$ for $t>1$, $\mathcal{L}(t)<B_s$. \\
\bottomrule
\end{tabular}
}
\end{table}

\subsection{Mathematical Formulation of the Proxy Task}
\label{subsec:proxy_formulation}
Running full diffusion generation loops for hyperparameter search is computationally prohibitive. Instead, we construct a lightweight \textbf{proxy task} that simulates the feature drift inherent in video attention. 

Let $\mathbf{h}_t \in \mathbb{R}^{D}$ denote the hidden state (Key/Query feature) of a token at frame $t$. We model the temporal evolution of features as an auto-regressive process governed by a drift rate $\delta \in [0, 1]$:
\begin{equation}
    \mathbf{h}_{t} = \sqrt{1 - \delta^2} \cdot \mathbf{h}_{t-1} + \delta \cdot \boldsymbol{\epsilon}_t, \quad \boldsymbol{\epsilon}_t \sim \mathcal{N}(\mathbf{0}, \mathbf{I}),
\end{equation}
where $\boldsymbol{\epsilon}_t$ represents new semantic information entering the frame.
\begin{itemize}
    \item \textbf{Low Motion ($\delta=0.02$):} Features remain highly correlated ($\mathbf{h}_t \approx \mathbf{h}_{t-1}$), mimicking static backgrounds.
    \item \textbf{Mid Motion ($\delta=0.15$):} Features exhibit moderate drift, representing standard character movements or slow camera pans.
    \item \textbf{High Motion ($\delta=0.40$):} Features drift rapidly, simulating fast-moving objects or rapid scene changes.
\end{itemize}

\paragraph{Alignment with Semantic Router.}
RAFT-derived optical-flow magnitude defines the real-video motion regimes that supervise the router (Section 3.4). The simulator parameter $\delta$ is a dimensionless feature-drift magnitude, not a conversion from pixel-space optical flow. Values 0.02, 0.15, and 0.40 are representative low-, mid-, and high-drift stress levels ordered consistently with those regimes.

The proxy attention matrix $\mathbf{A}_{\text{proxy}}$ is then computed as $\mathbf{A}_{ij} \propto \exp(\mathbf{h}_i^\top \mathbf{h}_j / \sqrt{D})$.

\subsection{Scope of the Proxy-Model Analysis}
\label{subsec:error_bound}
The AR proxy is a controlled test of how sparsity settings respond to feature drift, not a formal model of every video-diffusion attention pattern. A locality prior can discard useful long-range links. Dynamic-threshold mode scores only pairs inside the radial candidate set $\mathcal{P}(i,j)$, so it cannot recover links excluded by that set; within it, score-based selection can retain high-scoring pairs that static-ratio sampling may miss. We do not claim a deterministic error advantage for Dynamic mode. 

\subsection{Optimization Objective}
We search a mode-specific parameter vector, with the block size $B_s$ fixed within each profiling run:
\begin{equation}
\begin{aligned}
\mathbf{x}_{\mathrm{static}} &= [\gamma,\lambda,\theta_m,\theta_c,\rho_1,\rho_2],\\
\mathbf{x}_{\mathrm{dynamic}} &= [\gamma,\lambda,\theta_m,\theta_c,\tau_1,\tau_2].
\end{aligned}
\end{equation}
For either mode, $\mathbf{x}$ denotes its corresponding vector. The objective balances reconstruction fidelity (approximating the dense attention matrix) and computational efficiency (sparsity target):
\begin{equation}
\label{eq:bo_objective_app}
    \mathcal{L}(\mathbf{x}) = \underbrace{\frac{||\mathbf{A}_{\text{dense}} - \mathbf{A}_{\text{sparse}}(\mathbf{x})||_F^2}{||\mathbf{A}_{\text{dense}}||_F^2}}_{\text{Normalized MSE}} + \alpha \cdot \underbrace{\text{ReLU}(\mathcal{T} - \mathcal{S}(\mathbf{x}))}_{\text{Sparsity Penalty}}
\end{equation}
where $\mathcal{S}(\mathbf{x})$ is the sparsity ratio, $\mathcal{T}=0.80$ is the target sparsity, and $\alpha=10.0$ is a penalty coefficient.

\subsection{Optimization Algorithm and Efficiency}
We employ the Tree-structured Parzen Estimator (TPE) algorithm~\citep{bergstra2011tpe} to explore the non-differentiable search space efficiently. The complete offline profiling procedure is summarized in Algorithm~\ref{alg:bo_process}.

\begin{algorithm}[t]
\caption{Offline Parameter Profiling via TPE}
\label{alg:bo_process}
\begin{algorithmic}[1]
\Require Proxy Simulator $\mathcal{P}(\delta)$, Search Space $\Omega$, Trials $N=30$.
\Ensure Optimal configs $\{\mathbf{x}^*_{\text{low}}, \mathbf{x}^*_{\text{mid}}, \mathbf{x}^*_{\text{high}}\}$.
\For{regime $r \in \{\text{Low}, \text{Mid}, \text{High}\}$ defined in Sec.~\ref{subsec:bo_predictor}}
    \State Set drift rate $\delta_r$ (0.02, 0.15, or 0.40).
    \State Initialize trial history $\mathcal{H} \leftarrow \emptyset$.
    \For{$i = 1$ to $N$}
        \State Sample $\mathbf{x}_i$ from $\Omega$ using TPE acquisition function based on $\mathcal{H}$.
        \State Generate features $\mathbf{Q}, \mathbf{K}$ using $\mathcal{P}(\delta_r)$.
        \State Compute Sparse Mask $\mathbf{M}(\mathbf{x}_i)$.
        \State Calculate Loss $\ell_i = \mathcal{L}(\mathbf{x}_i)$ via Eq.~\eqref{eq:bo_objective_app}.
        \State Update history $\mathcal{H} \leftarrow \mathcal{H} \cup \{(\mathbf{x}_i, \ell_i)\}$.
    \EndFor
    \State $\mathbf{x}^*_r \leftarrow \arg\min_{\mathbf{x} \in \mathcal{H}} \mathcal{L}(\mathbf{x})$
\EndFor
\State \Return Lookup Table $\mathcal{T}$ \Comment{Indexed by Semantic Router outputs}
\end{algorithmic}
\end{algorithm}

\vspace{10pt}

\noindent\textbf{Efficiency Analysis.} 
While our proxy task requires computing dense attention baselines for loss calculation, it remains significantly faster than standard hyperparameter tuning on video diffusion models. 
Replacing proxy evaluations with full video generations would require repeated generation runs for each candidate configuration and motion regime. Our proxy-based optimization completes the reported calibration in approximately \textbf{15 minutes}, allowing adaptation to new resolutions or aspect ratios without iterative video sampling.

\begin{figure*}[t]
    \centering
    \includegraphics[width=0.95\linewidth]{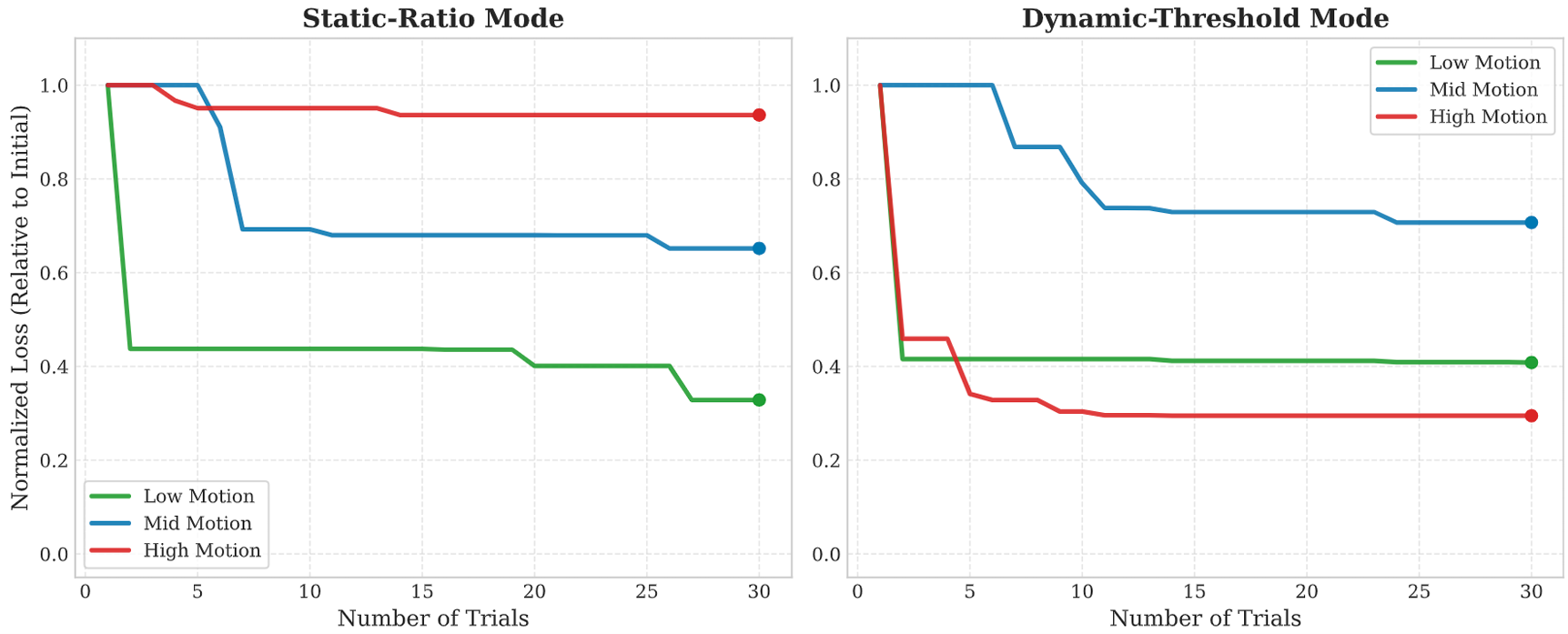}
    \caption{\textbf{BO Convergence Analysis.} The optimization process (simulated on proxy tasks) typically converges within 30 trials. \textbf{Left:} Static mode struggles with High Motion (red line stays high), necessitating a mode switch. \textbf{Right:} Dynamic mode successfully minimizes reconstruction error for High Motion scenes, validating its capability to capture complex dynamics.}
    \label{fig:bo_convergence}
\vspace{-0.1 cm}
\end{figure*}

\subsection{Convergence Analysis and Qualitative Validation}
\label{subsec:bo_analysis}

\paragraph{Convergence Behaviors (Figure~\ref{fig:bo_convergence}).}
We first analyze the optimization trajectories to validate the necessity of our dual-mode design. As shown in Figure~\ref{fig:bo_convergence}:
\begin{itemize}
    \item \textbf{Static Mode Limitation:} For High Motion scenes (red line, Left), the loss stagnates near 1.0 over the tested parameter range. This suggests that the tested static masks do not capture this proxy's rapid feature drift as well as Dynamic mode.
    \item \textbf{Dynamic Mode Recovery:} In contrast, the Dynamic mode (red line, Right) drives the loss down to $\sim$0.30. By retaining high-scoring off-diagonal pairs within radial support, it better captures the long-range dependencies represented in this proxy.
\end{itemize}

\begin{figure*}[t]
    \centering
    \setlength{\tabcolsep}{1pt}
    \renewcommand{\arraystretch}{0.1}
    
    \fcolorbox{gray!50}{gray!10}{
        \parbox{0.97\linewidth}{
            \vspace{2pt}
            \textbf{\small Test Prompt (High Motion):} \small \textit{“FPV drone shot flying through a futuristic sci-fi tunnel at high speed. Repeating neon ring lights, metallic walls, motion blur, symmetrical composition, cinematic lighting, 4k.”}
            \vspace{2pt}
        }
    }
    \vspace{4pt}
    
    \begin{subfigure}{\textwidth}
        \textbf{\scriptsize (a) Untuned (Static Mode): $\gamma{=}3.0, \lambda{=}0.1, \theta_m{=}0.75, \mathbf{r{=}[0.2, 0.1]}$} \\
        \includegraphics[width=0.195\textwidth]{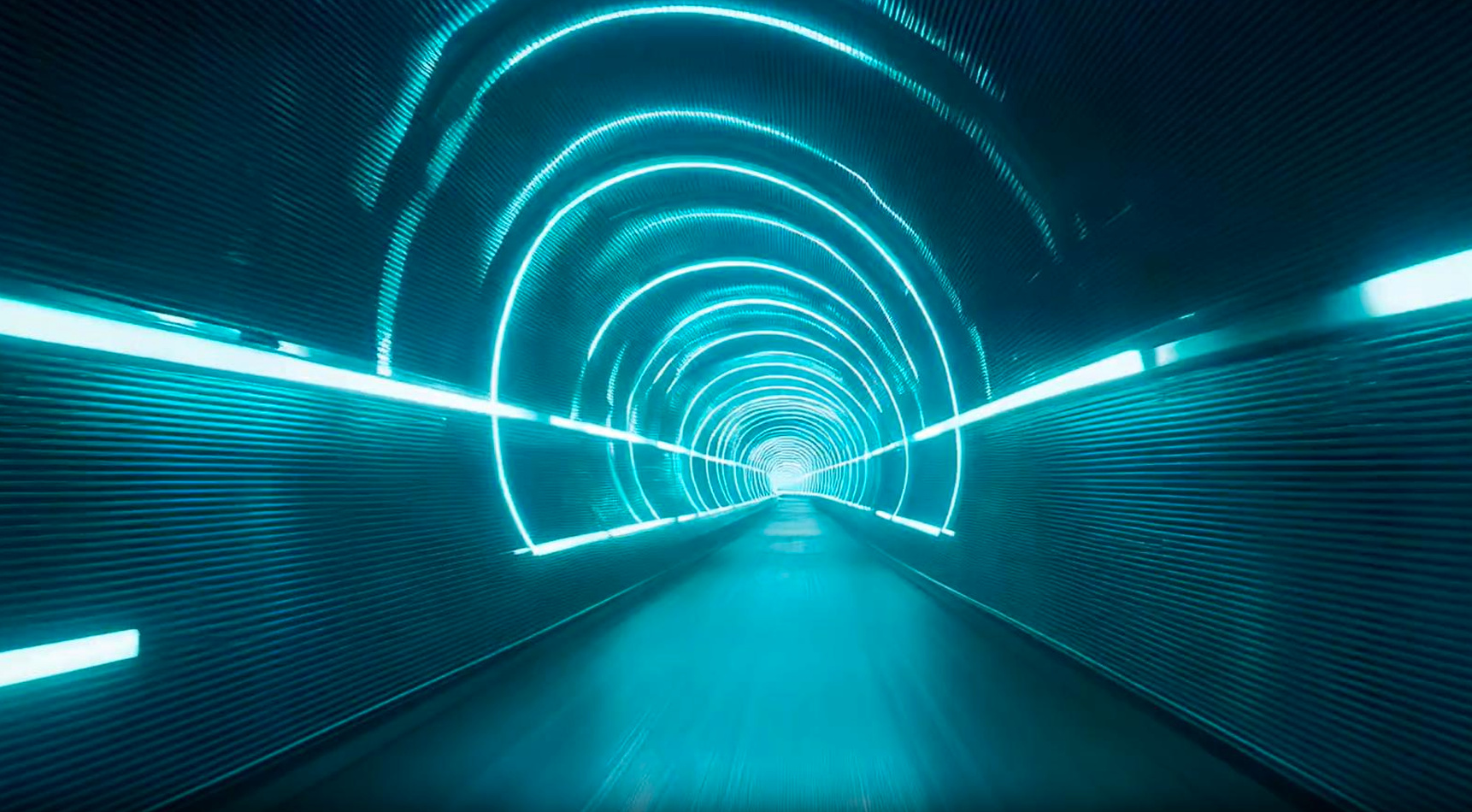} \hfill
        \includegraphics[width=0.195\textwidth]{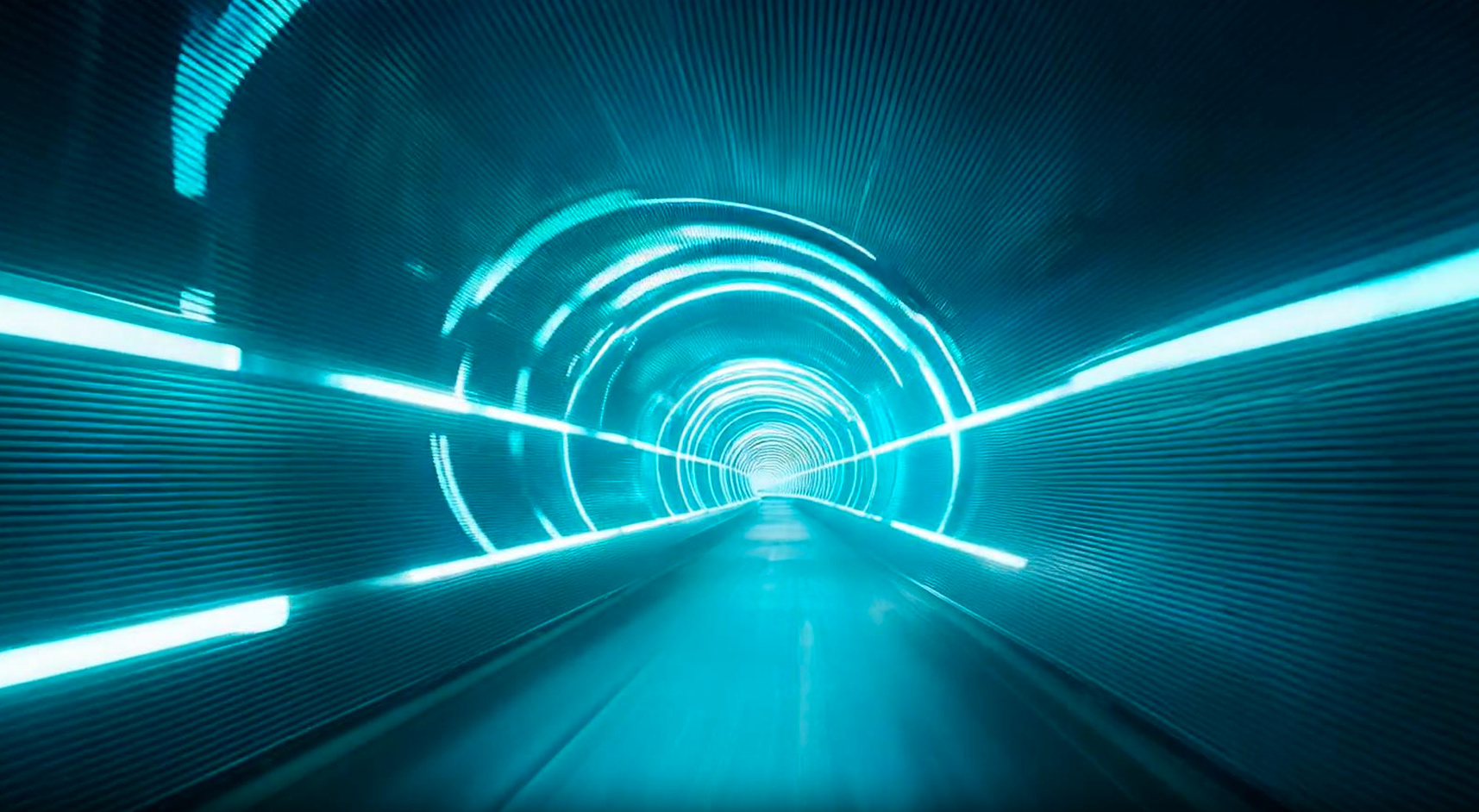} \hfill
        \includegraphics[width=0.195\textwidth]{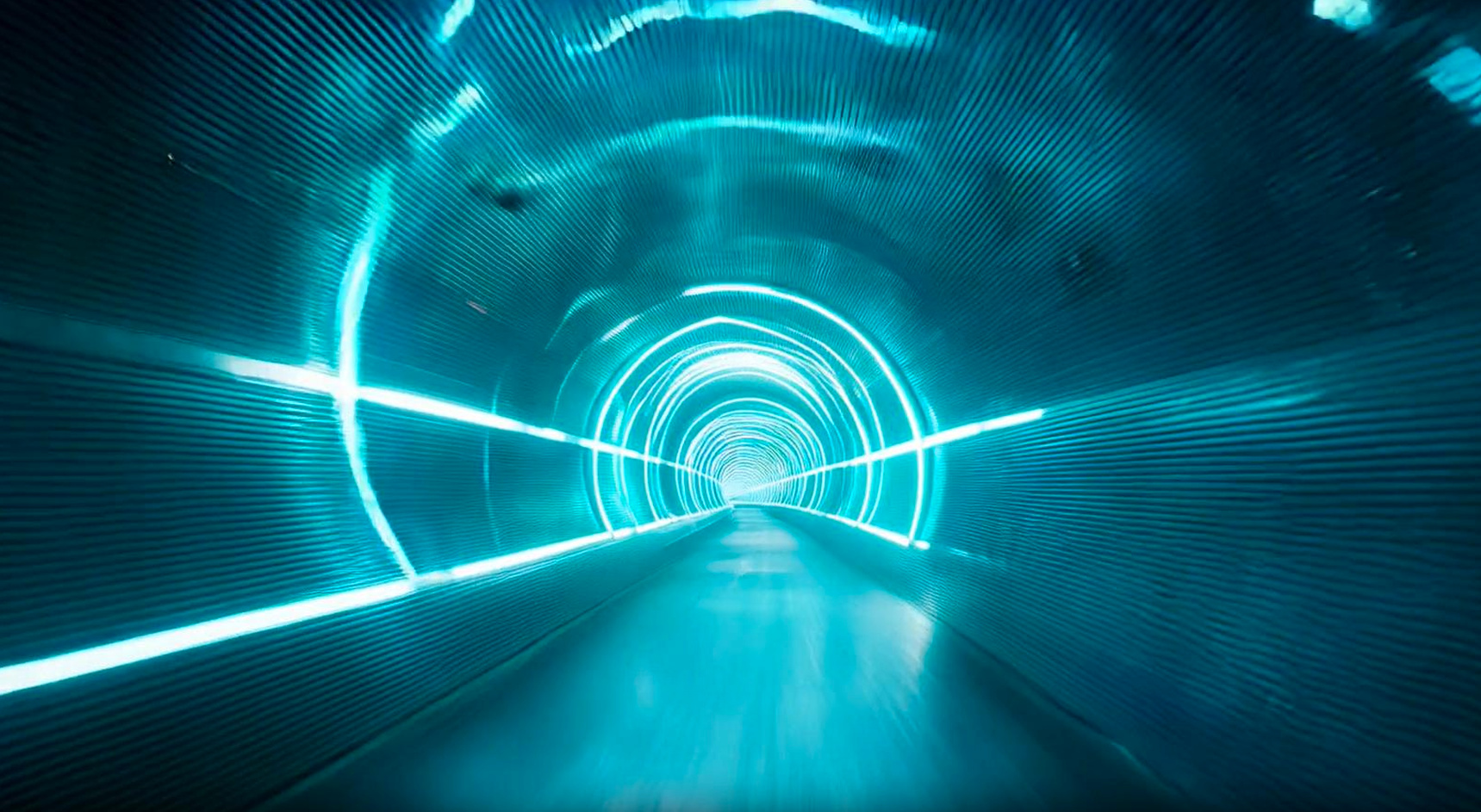} \hfill
        \includegraphics[width=0.195\textwidth]{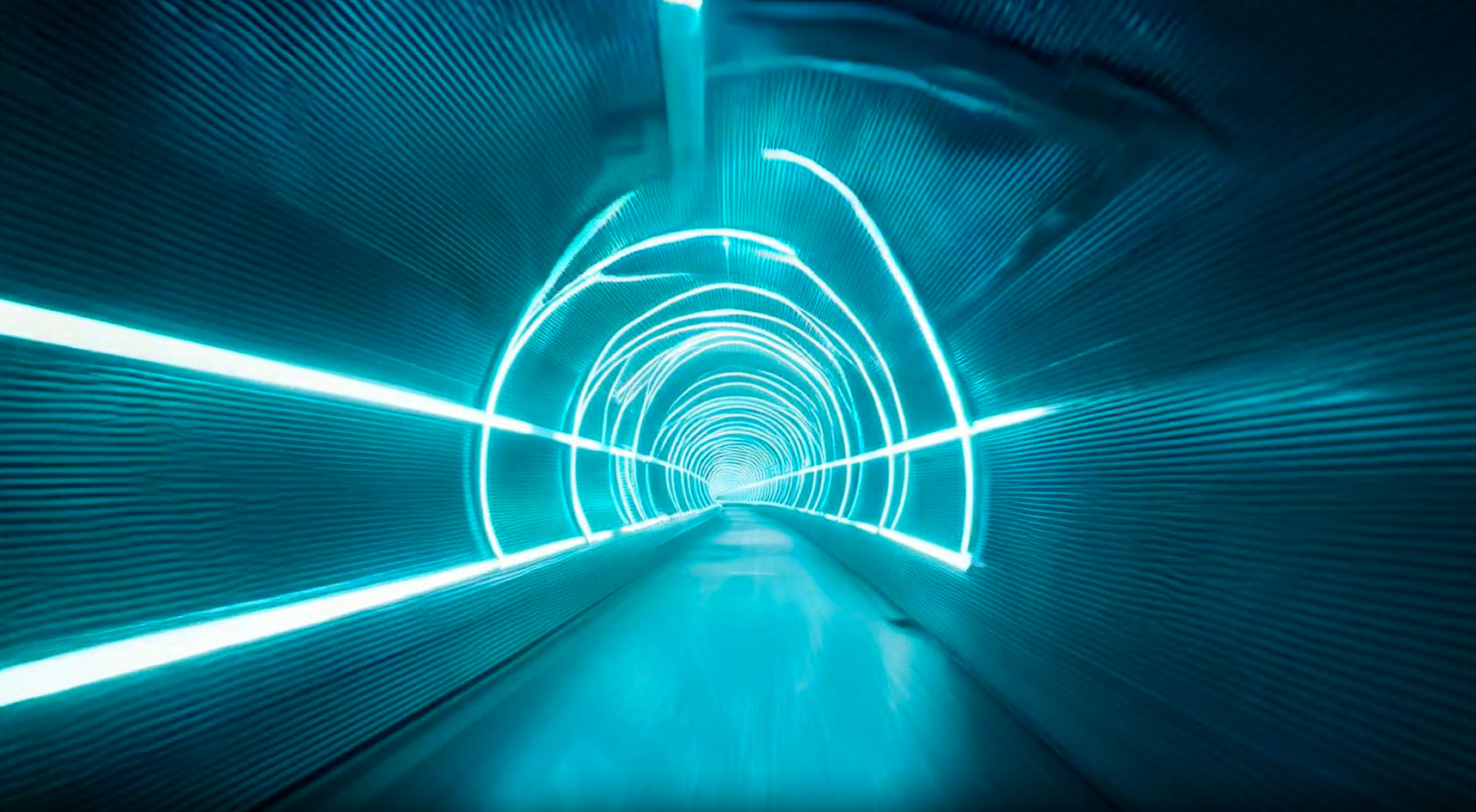} \hfill
        \includegraphics[width=0.195\textwidth]{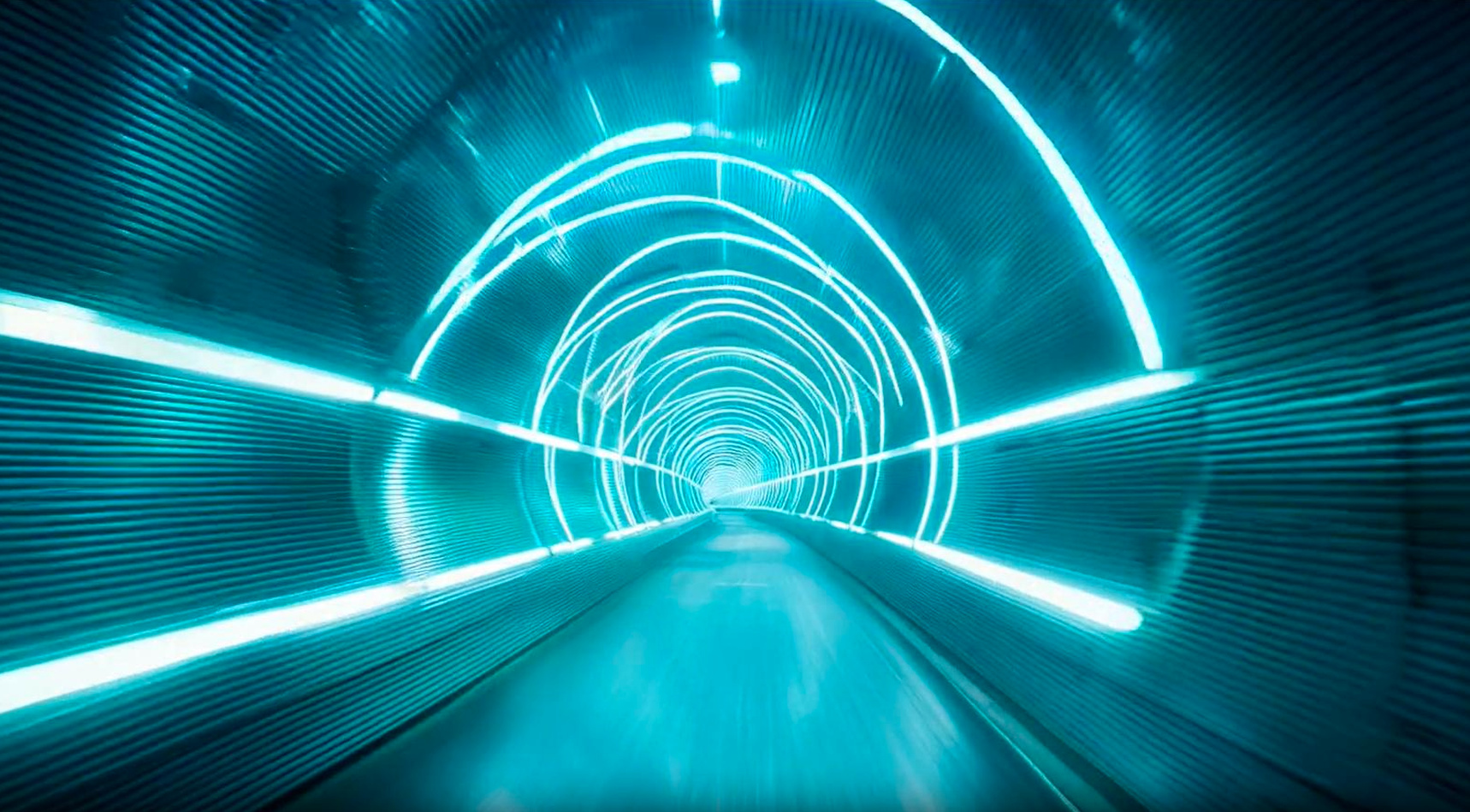} 
    \end{subfigure}
    \vspace{4pt} 
    
    \begin{subfigure}{\textwidth}
        \textbf{\scriptsize (b) BO-Optimized (Dynamic Mode): $\gamma{=}1.6, \lambda{=}0.9, \theta_m{=}0.55, \boldsymbol{\tau}{=}[-2.0, 2.4]$} \\
        \includegraphics[width=0.195\textwidth]{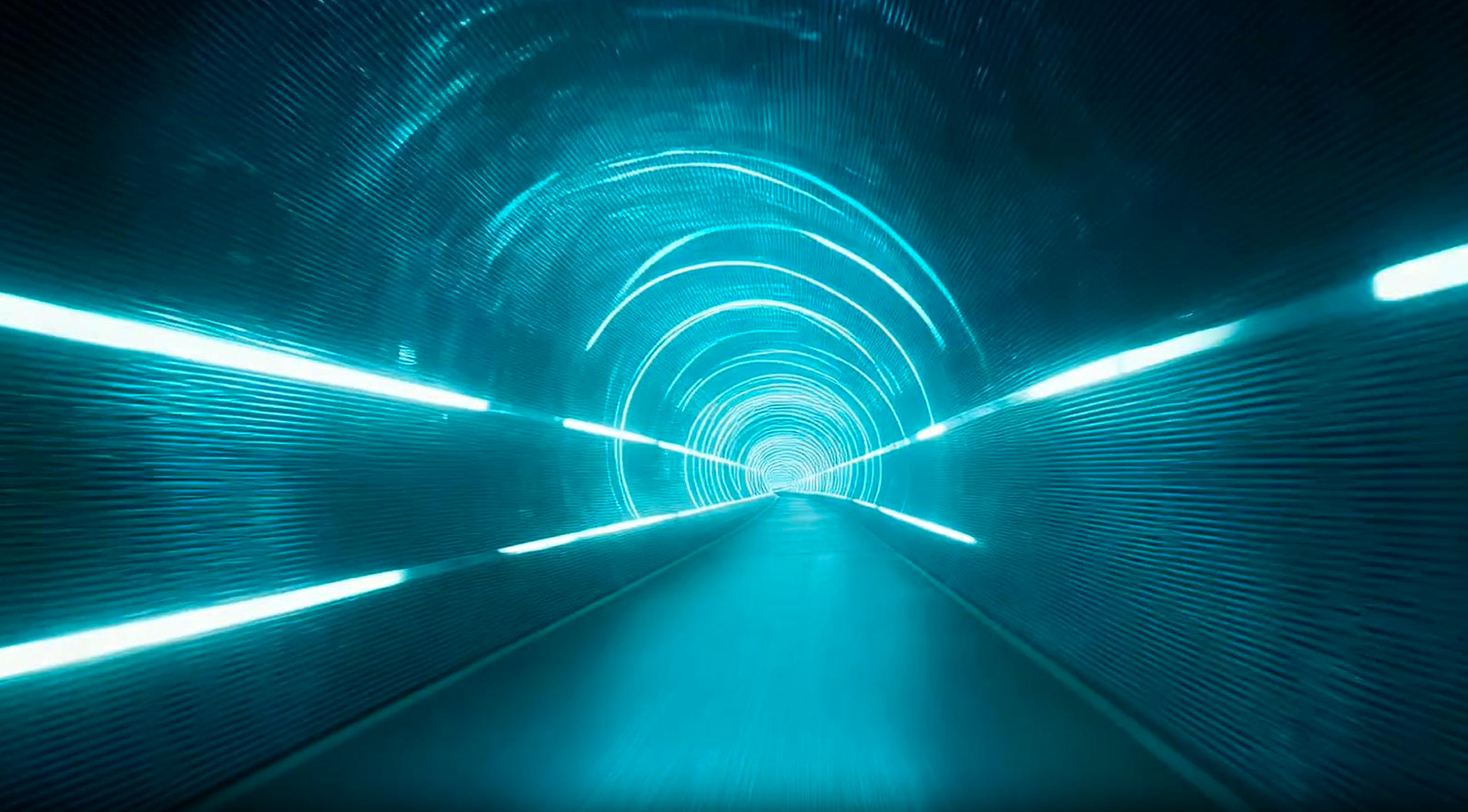} \hfill
        \includegraphics[width=0.195\textwidth]{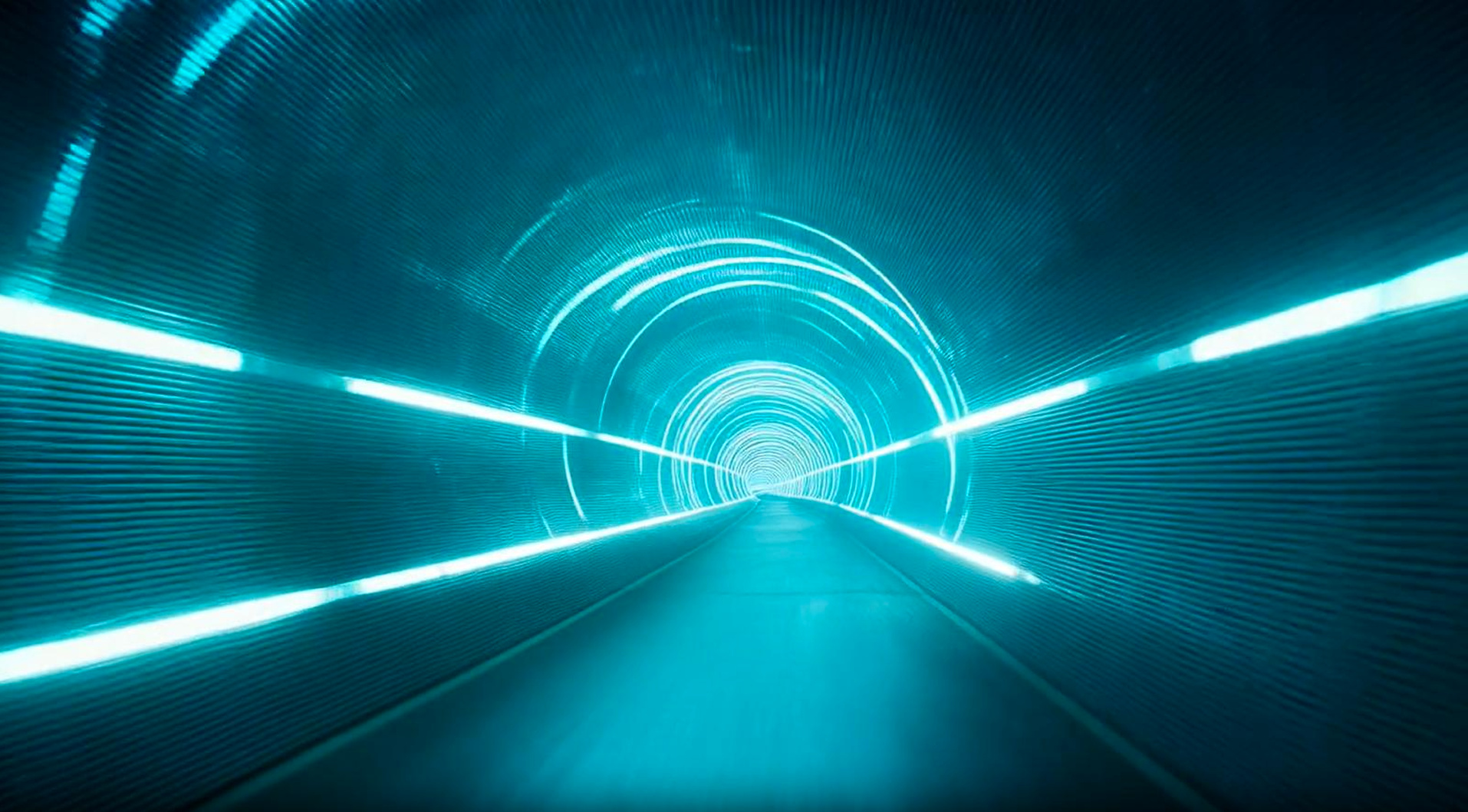} \hfill
        \includegraphics[width=0.195\textwidth]{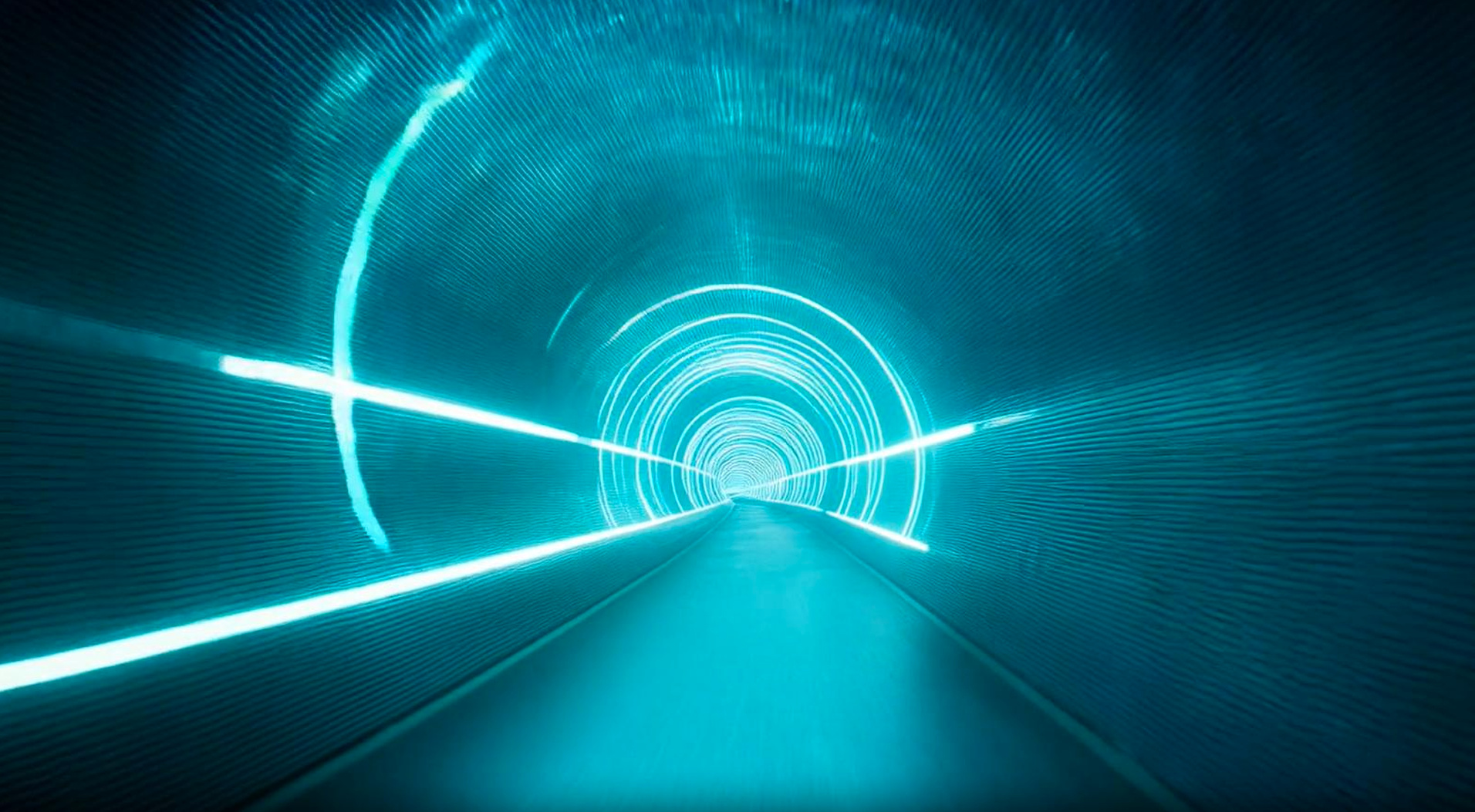} \hfill
        \includegraphics[width=0.195\textwidth]{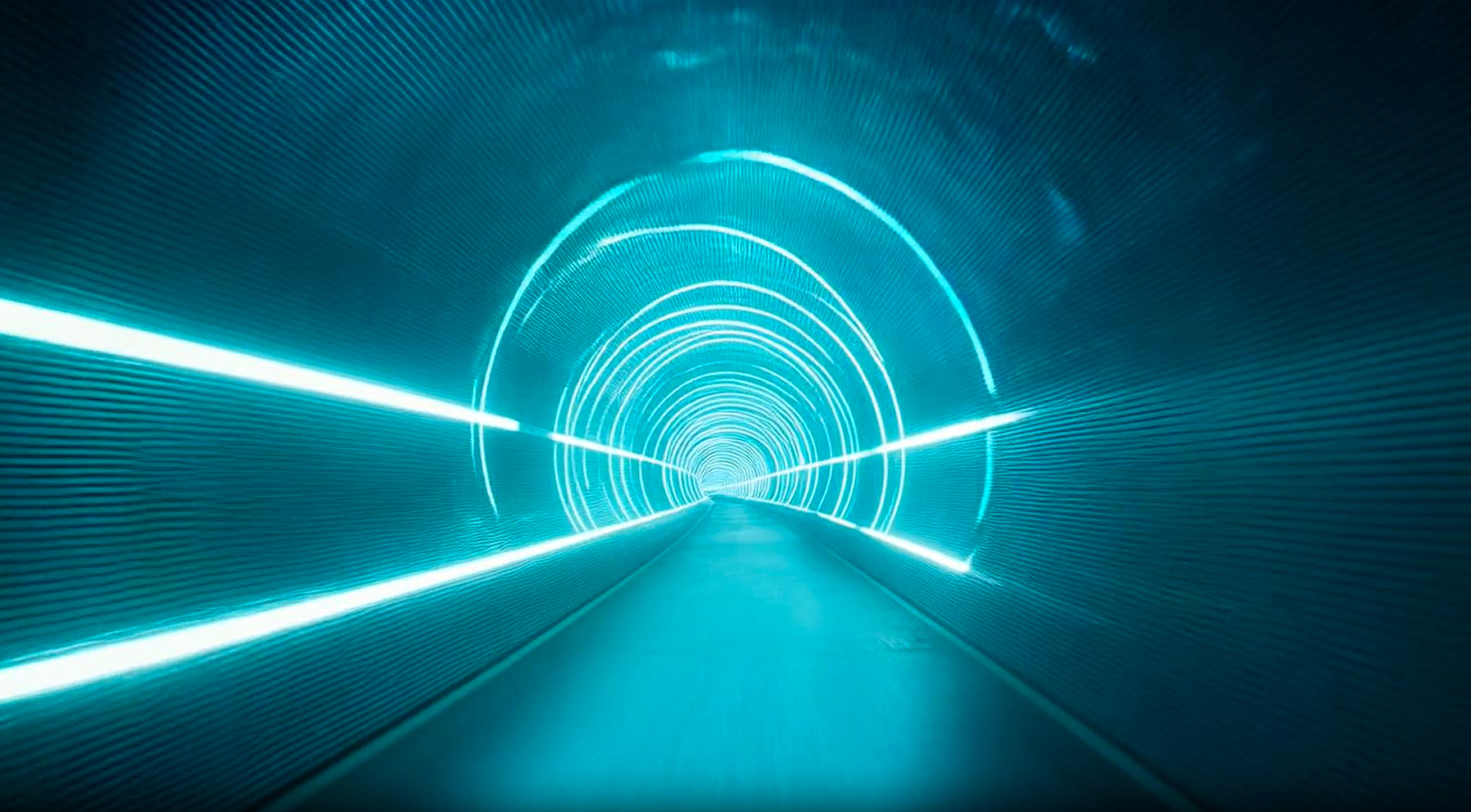} \hfill
        \includegraphics[width=0.195\textwidth]{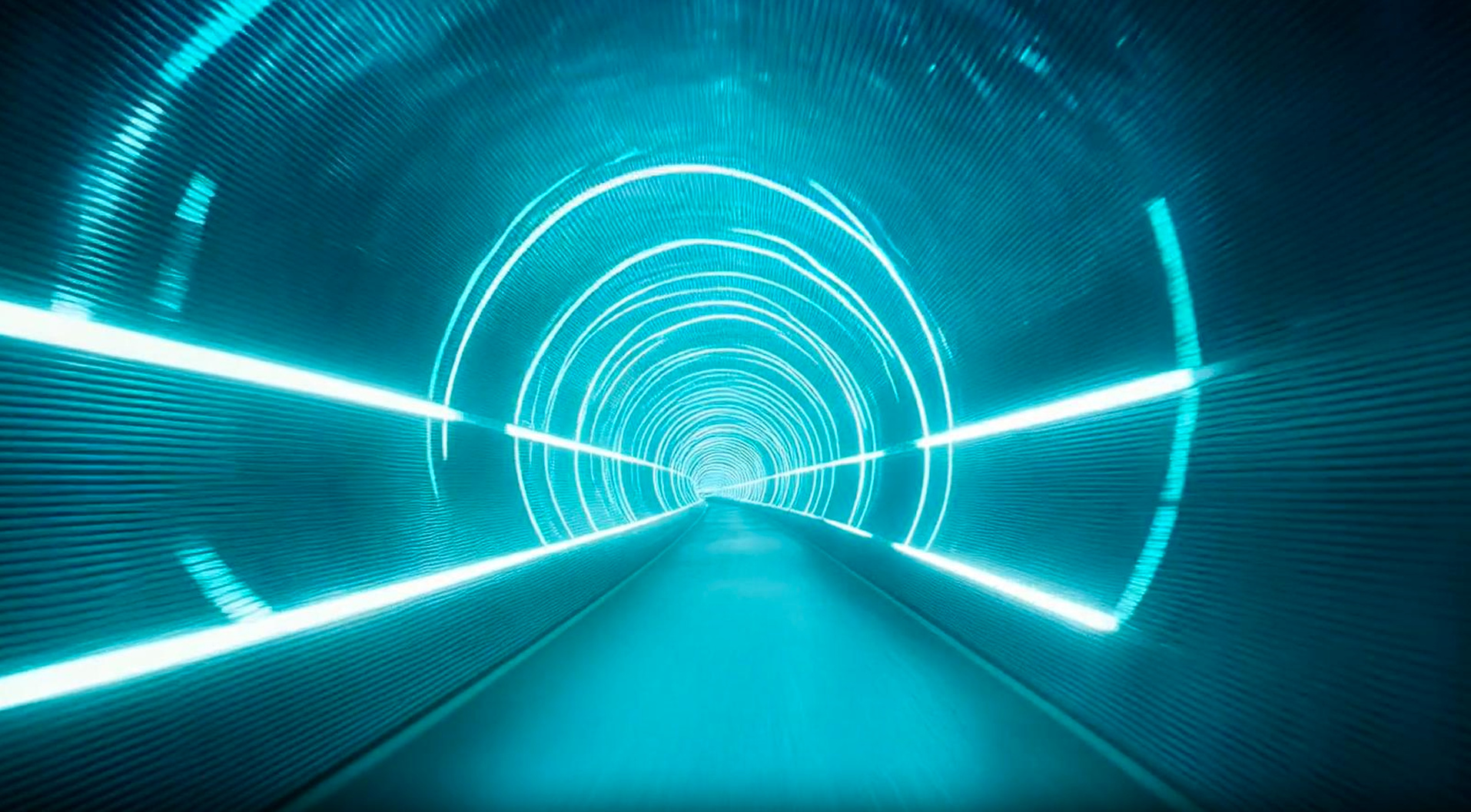}
    \end{subfigure}

    \caption{\textbf{Visual Ablation of Parameter Tuning and Mode Selection.} 
    \textbf{(a)} The \textit{Static-ratio} mode with naive heuristic parameters ($r_{\text{near}}=0.2$, $\gamma=3.0$, which widens the radial window) imposes a rigid sparsity pattern. In this high-speed FPV scene, its low retention of candidate links fails to preserve enough useful temporal connections, causing the \textbf{repeating neon rings to flicker and the tunnel geometry to distort} across frames.
    \textbf{(b)} The \textit{Dynamic-threshold} mode configured by our pipeline. The \textbf{Semantic Router} classifies this as a high-motion regime, triggering the retrieval of BO-optimized parameters (relaxed $\tau_{\text{near}}=-2.0$, denser $\lambda=0.9$) that better preserve the \textbf{symmetrical structure and smooth motion blur} in this example while balancing reconstruction fidelity and sparsity.
    (Notation: $\mathbf{r}=[r_{\text{near}}, r_{\text{far}}]$ denotes retention ratios; $\boldsymbol{\tau}=[\tau_{\text{near}}, \tau_{\text{far}}]$ denotes attention score thresholds).}
    \label{fig:bo_ablation}
\end{figure*}

\paragraph{Validation of Proxy Alignment (Figure~\ref{fig:bo_ablation}).}
The visual ablation illustrates the association between lower proxy MSE and better generations; a matched-sparsity quantitative analysis is reported in Appendix~\ref{app:rebuttal_proxy}.
\begin{itemize}
    \item \textbf{High MSE $\rightarrow$ Artifacts:} As shown in Figure~\ref{fig:bo_ablation}(a), the high reconstruction error in static mode is associated with visual artifacts (flickering neon rings) and geometric distortion.
    \item \textbf{Low MSE $\rightarrow$ Fidelity:} Conversely, Figure~\ref{fig:bo_ablation}(b) demonstrates that when the BO pipeline minimizes the Proxy MSE, the visual fidelity is improved.
\end{itemize}
Together, the qualitative and quantitative results support proxy MSE as a configuration-ranking signal under the tested prompts and sparsity range, rather than a universal predictor of video quality.

\section{Measured Dynamic-Mode Overhead}
\label{app:micro_benchmark}
\label{app:rebuttal_timing}

We instrument the deployed Wan2.1-14B-72f pipeline on one H100 under the Table~\ref{tab:main_results} configuration. After 20 warm-up iterations, component latencies are measured over 200 iterations using synchronized CUDA events for GPU stages and synchronized wall-clock timing for host-side index construction and planning. Each Dynamic invocation builds its mask, CSR/BSR indices, and execution plan from current Q/K activations; no content-dependent result is reused across layers or denoising steps. The router runs once per video from its prompt embedding (1.73 ms per video).

\begin{table}[H]
\centering
\caption{Measured deployed Dynamic-control overhead per attention invocation on H100/Wan-72f. The total includes QK scoring; no projected kernel-fusion benefit is counted.}
\label{tab:micro_benchmark}
\begin{tabular}{lr}
\toprule
Component & Latency (ms) \\
\midrule
Dynamic QK proxy scoring & 0.39 \\
Thresholding and token-to-block aggregation & 0.74 \\
CSR/BSR index construction & 0.31 \\
FlashInfer sparse planning & 0.27 \\
\midrule
Mask construction, indexing, and planning & 1.32 \\
Total Dynamic-control overhead & \textbf{1.71} \\
\bottomrule
\end{tabular}
\end{table}

Complete-generation timing independently uses synchronized host wall-clock measurements on five fixed prompts, three runs per prompt, with matched prompt-wise seeds, checkpoint, precision, resolution, 50 sampling steps, and CFG 5.0. Dense takes $1680\pm15$s and DynamicRad $980\pm9$s (1.71$\times$). This includes text encoding, routing, sparse control and execution, all denoising steps, host orchestration, and VAE decoding. On A800, the corresponding Wan-72f pipeline takes $2480\pm30$s dense and $1530\pm20$s Dynamic (1.62$\times$). H100/FlashInfer is the primary controlled comparison; A800 establishes transfer within the tested NVIDIA CUDA stack.

\section{Additional Long-Context Baselines}
\label{app:rebuttal_baselines}

Table~\ref{tab:main_results} includes the long-context PA and SVG comparisons requested in review. SVG is our reproduction of its dense-warm-up profiling and per-head spatial/temporal pattern assignment, using the same adapted checkpoint, prompts, seeds, resolution, 50 steps, CFG, precision, hardware, and block-sparse backend as the matched long-context runs. Its latency includes warm-up, profiling, assignment, mask construction, and sparse execution. PA uses its deterministic power-of-two connectivity under the same protocol and a sparsity budget matched to Radial Attention. These results show that both methods can run at long context.

We also reproduced AdaSpa under the same H100 timing protocol and block-sparse backend, with block size 64, approximately 80\% sparsity, preserved text-related sink interactions, row-wise balanced selection, head-adaptive hierarchical allocation, ten dense warm-up steps, search at step 10, and an LSE-cached update at step 30. Timing includes warm-up, online search, LSE computation, updates, mask construction, and sparse attention. The 241f result extends the published procedure and is not an official AdaSpa result.

\begin{table}[H]
\centering
\caption{AdaSpa (our reproduction) on HunyuanVideo under the matched protocol.}
\label{tab:adaspa_reproduction}
\begin{tabular}{lccccc}
\toprule
Frames & VR & S.C. & A.Q. & I.Q. & Latency (s) \\
\midrule
120 & 0.141 & 0.963 & 0.648 & 0.675 & 958 \\
241 & 0.126 & 0.963 & 0.643 & 0.668 & 420 \\
\bottomrule
\end{tabular}
\end{table}
AdaSpa reaches 80.0\%/80.1\% sparsity and 1.78$\times$/1.96$\times$ speedup at 120f/241f. At 241f, DynamicRad Dynamic is 42s faster with higher quality metrics; Static is the lower-latency endpoint.

\section{Proxy and Router Validation}
\label{app:rebuttal_proxy}

\paragraph{Proxy MSE versus video quality.}
We evaluate 48 Dynamic-threshold configurations, 24 each for low and high motion, at 78\%--82\% final executed block sparsity. Each uses three fixed prompts from its regime with matched prompt-wise seeds, checkpoint, CFG, sampling steps, precision, and resolution. Sparsity is computed from the final BSR index after token-to-block aggregation, dense intra-frame insertion, and fallback. Table~\ref{tab:proxy_correlation} correlates mean attention-reconstruction MSE with video metrics across configurations.

\begin{table}[H]
\centering
\caption{Correlation between proxy MSE and video quality at matched sparsity.}
\label{tab:proxy_correlation}
\begin{tabular}{llrr}
\toprule
Regime & Metric & Pearson $r$ & Spearman $\rho$ \\
\midrule
Low & VisionReward & $-0.49$ & $-0.46$ \\
Low & A.Q. & $-0.52$ & $-0.49$ \\
Low & I.Q. & $-0.58$ & $-0.54$ \\
Low & M.S. & $-0.45$ & $-0.42$ \\
High & VisionReward & $-0.62$ & $-0.58$ \\
High & A.Q. & $-0.64$ & $-0.60$ \\
High & I.Q. & $-0.71$ & $-0.66$ \\
High & M.S. & $-0.68$ & $-0.63$ \\
\bottomrule
\end{tabular}
\end{table}
Within this range, lower proxy MSE is associated with better scores; final sparsity has absolute Spearman correlation at most 0.19 with these metrics. Across all 48 Dynamic-threshold configurations, partial Spearman correlations controlling for executed sparsity are $-0.39$ to $-0.51$ for low motion and $-0.54$ to $-0.64$ for high motion across VisionReward, A.Q., I.Q., and M.S. We use proxy MSE as a ranking signal in this tested setting, not as a universal quality predictor.

\paragraph{Radial support and routing.}
At the same 83.6\% final block sparsity on Hunyuan-241f, Table~\ref{tab:matched_masks} separates the radial prior from the selector. Global top-$k$ scores the full grid; DynamicRad scores only radial candidates.
\begin{table}[H]
\centering
\caption{Mask ablation at matched 83.6\% sparsity on Hunyuan-241f.}
\label{tab:matched_masks}
\begin{tabular}{lrrr}
\toprule
Mask & Latency (s) & VR & S.C. \\
\midrule
Random mask & 331 & 0.108 & 0.935 \\
Local temporal window & 334 & 0.116 & 0.947 \\
Global top-$k$ (no radial) & 472 & 0.129 & 0.966 \\
DynamicRad Dynamic & 378 & 0.131 & 0.969 \\
\bottomrule
\end{tabular}
\end{table}

The frozen router is evaluated on 2,000 held-out OpenVid-1M video--text pairs with RAFT-derived motion labels. Its row-normalized confusion matrix is shown in Table~\ref{tab:router_confusion}; balanced accuracy is 88.8\%, and most errors are between neighboring regimes.
\begin{table}[H]
\centering
\caption{Router confusion matrix (true regime by row, predicted regime by column).}
\label{tab:router_confusion}
\begin{tabular}{lrrr}
\toprule
 & Low & Mid & High \\
\midrule
Low & 91.2\% & 8.1\% & 0.7\% \\
Mid & 8.4\% & 84.8\% & 6.8\% \\
High & 1.0\% & 8.6\% & 90.4\% \\
\bottomrule
\end{tabular}
\end{table}

On Hunyuan-241f, Random/Predicted/Oracle routing gives VR 0.120/0.127/0.128 and S.C. 0.952/0.965/0.967, at comparable 84.8\%/84.7\%/84.9\% sparsity and 346/342/350s latency. Random uniformly selects among the same three BO policies over 100 assignments; Oracle uses the RAFT-derived label. Predicted routing approaches the non-deployable oracle and exceeds random assignment at similar cost.

Using a fixed pool of 43 candidate configurations in a separate proxy-drift experiment at ten drift levels, three and five contiguous motion buckets each select three unique configurations and yield mean normalized MSE 0.45626. Ten buckets select four configurations and yield 0.45613, a 0.029\% decrease. We retain three regimes: finer bucketing contributes little in this tested proxy setting. The BO configurations jointly couple mode, decay, long-range, threshold, and retention parameters; arbitrary interpolation of these vectors need not preserve the calibrated sparsity budget.

\section{Additional Video-Quality Evaluation}
\label{app:rebuttal_quality}

\begin{table}[H]
\centering
\caption{VBench Total and Dynamic Degree (DD) at long context. High-motion DD uses a RAFT-defined subset. Full-FT Dense is the checkpoint- and protocol-matched dense reference.}
\label{tab:vbench_total_dd}
\begin{tabular}{llrrr}
\toprule
Setting & Method & Total (\%) & DD (\%) & High-motion DD (\%) \\
\midrule
\multirow{5}{*}{Hunyuan-241f} & Full-FT Dense & 80.10 & 66.2 & 78.2 \\
& Radial & 80.03 & 65.8 & 77.6 \\
& Static & 79.91 & 64.9 & 76.6 \\
& Dynamic & 80.24 & 67.0 & 79.0 \\
& Routed & 80.18 & 66.7 & 78.6 \\
\midrule
\multirow{5}{*}{Wan-145f} & Full-FT Dense & 79.78 & 64.8 & 76.4 \\
& Radial & 79.70 & 64.4 & 75.9 \\
& Static & 79.55 & 63.7 & 75.0 \\
& Dynamic & 79.95 & 65.6 & 77.3 \\
& Routed & 79.88 & 65.2 & 76.9 \\
\bottomrule
\end{tabular}
\end{table}
Static incurs a small dynamicity trade-off, while Dynamic and Routed preserve matched-dense-level motion. Under our local pipeline, the native checkpoints score 81.82 Total at Hunyuan-120f and 82.26 at Wan-72f, versus 80.10/79.78 for Full-FT Dense at 241f/145f. The 1.72/2.48-point differences reflect the longer horizon and finite target-length adaptation, and public native-length scores use different evaluation settings. All matched long-context methods share checkpoint, prompts and seeds, frame count, generation settings, precision, and metric pipeline.

\section{Additional Qualitative Results}
\label{app:more_visuals}

In this section, we provide a comprehensive gallery of generated samples to demonstrate the robustness of DynamicRad across two state-of-the-art video diffusion backbones: Wan2.1-14B and HunyuanVideo.

For a controlled visual comparison, we compare a dense reference against DynamicRad. The gallery label ``Original (Dense)'' denotes the native dense checkpoint at standard length and checkpoint-matched Full-FT Dense at long length, not the Original-LX extrapolation row of Table~\ref{tab:main_results}. Both methods use identical random seeds, prompts, sampling steps, CFG scale, and backbone checkpoints within each matched setting.

Unless otherwise specified, DynamicRad results in this appendix use the \textbf{Dynamic mode with Mask-Aware LoRA (MA-LoRA)}, which is the sparsity-aware refinement introduced in Sec.~3.3. The gallery illustrates visual fidelity and temporal coherence; efficiency results for the backbone-frozen policy and the separate MA-LoRA ablation are reported in Tables~\ref{tab:main_results} and~\ref{tab:synthetic_ablation_wide}, respectively.

\textbf{Generation Configuration.} 
To validate performance stability across varying temporal horizons, we generate samples at \textbf{High Definition ($1280 \times 720$)} resolution using \textbf{50 sampling steps} and a CFG scale of 5.0.
Specifically, we evaluate both models under standard and long-context settings:
\begin{itemize}
    \item \textbf{HunyuanVideo:} Evaluated at both \textbf{120 frames} (Standard) and \textbf{241 frames} (Long-Context, $>$200k tokens).
    \item \textbf{Wan2.1-14B:} Evaluated at both \textbf{72 frames} (Standard) and \textbf{145 frames} (Long-Context).
\end{itemize}

Each strip below contains \textbf{5 frames} sampled uniformly from the generated video to visualize the temporal progression.
\clearpage

\subsection{Results on Wan2.1-14B}

\begin{figure}[H]
    \centering
    \setlength{\tabcolsep}{1pt} 
    \renewcommand{\arraystretch}{0.1}
    
    \begin{minipage}{\textwidth}
        \fcolorbox{gray!50}{gray!10}{
            \parbox{0.97\linewidth}{
                \vspace{2pt}
                \textbf{\small Case 1 (Fluid Dynamics):} \small \textit{“Close-up of a raging campfire at night, sparks flying upwards, dynamic flames flickering, realistic texture, cinematic lighting, 4k, high frame rate.”}
                \vspace{2pt}
            }
        }
        \vspace{3pt}
        \begin{subfigure}{\textwidth}
        \vspace{3pt}
            \textbf{\scriptsize Original (Dense):} \\
            \includegraphics[width=0.195\textwidth]{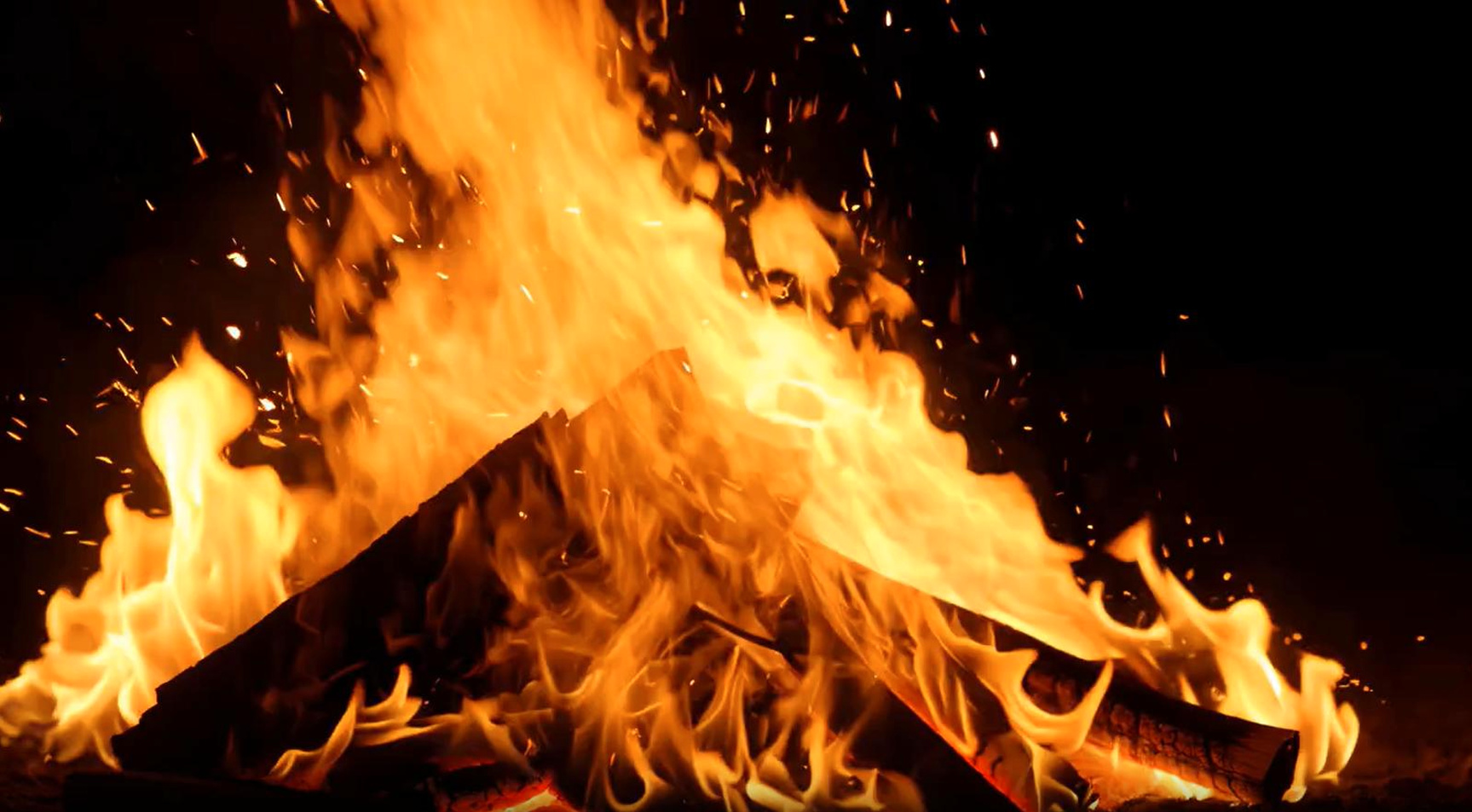} \hfill
            \includegraphics[width=0.195\textwidth]{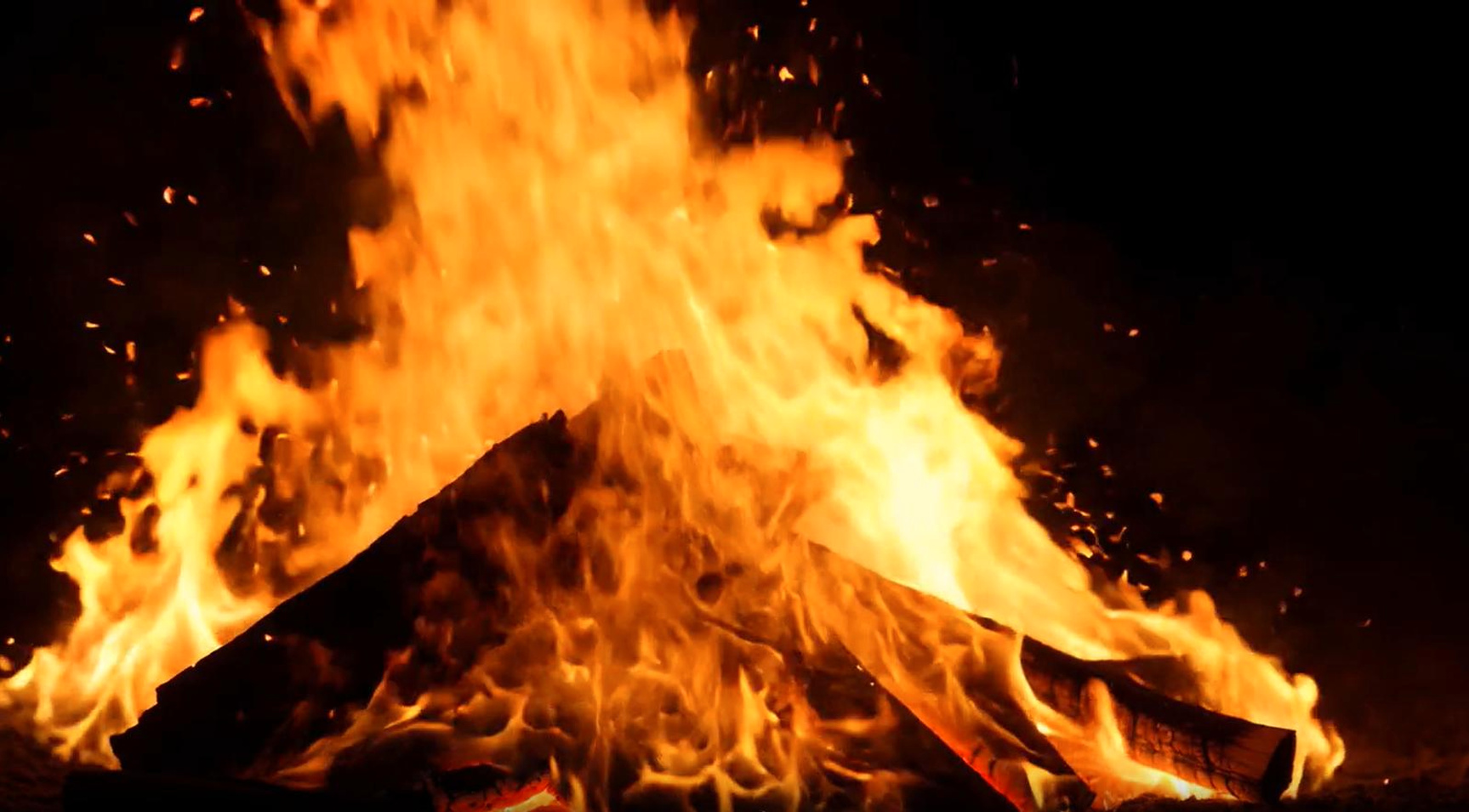} \hfill
            \includegraphics[width=0.195\textwidth]{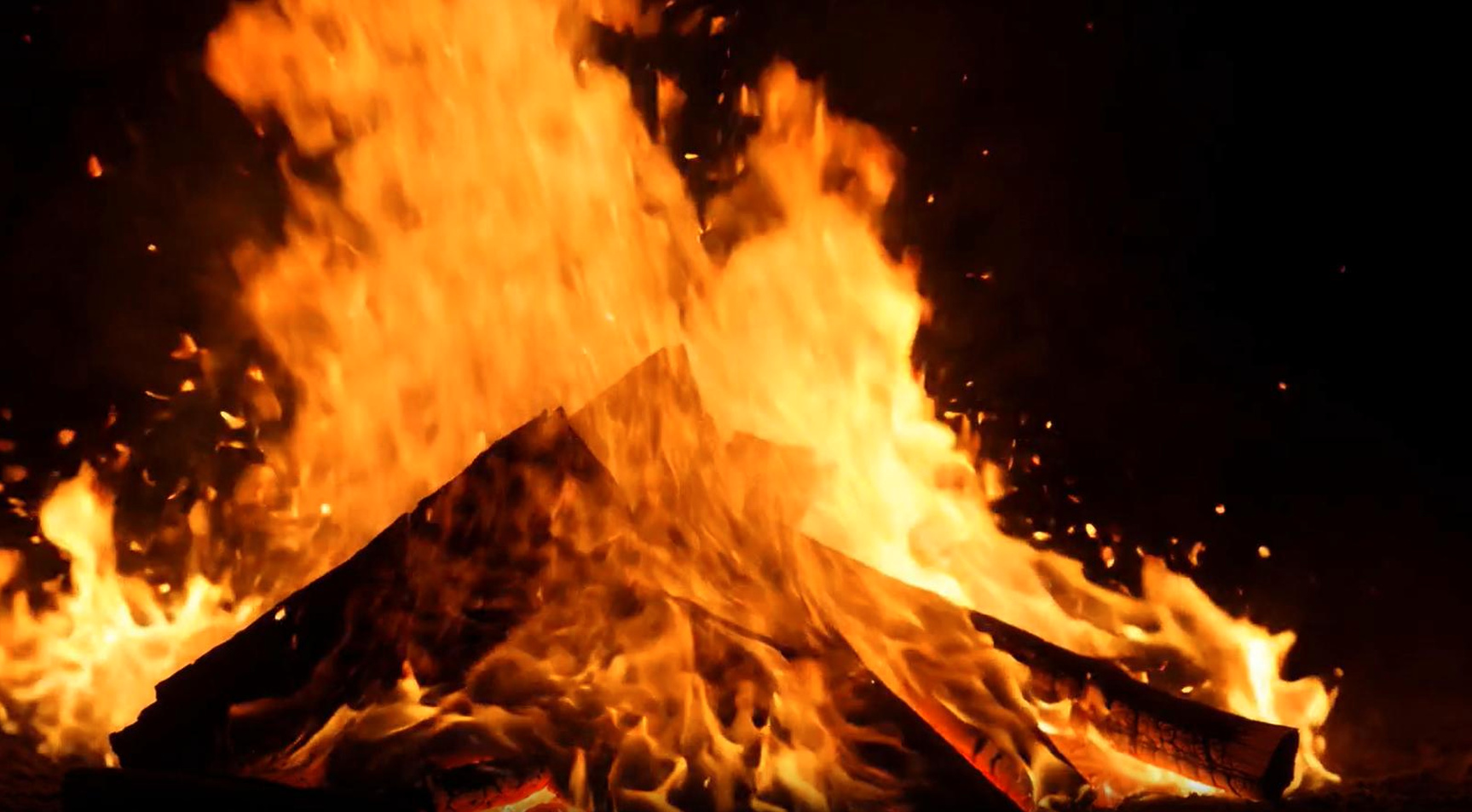} \hfill
            \includegraphics[width=0.195\textwidth]{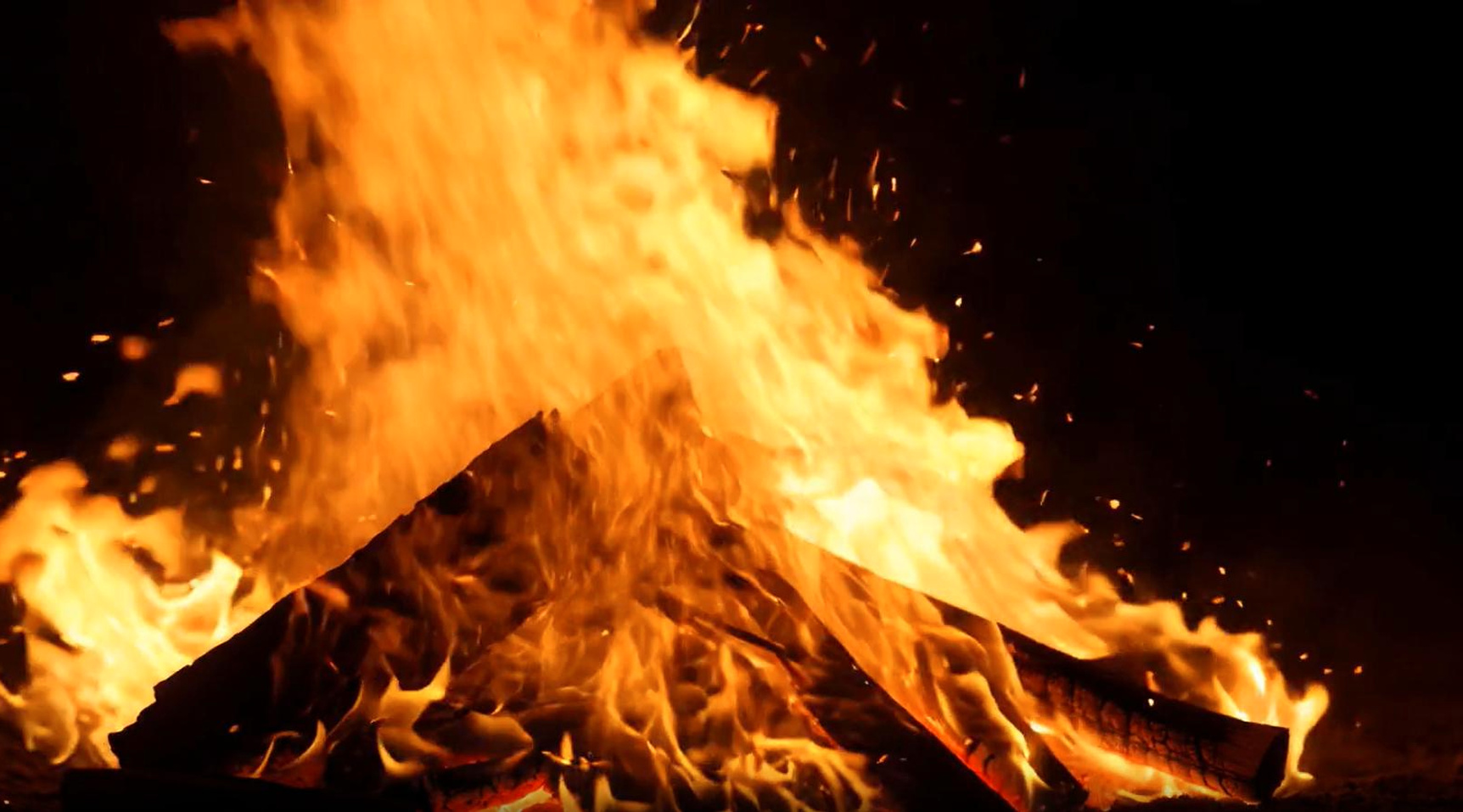} \hfill
            \includegraphics[width=0.195\textwidth]{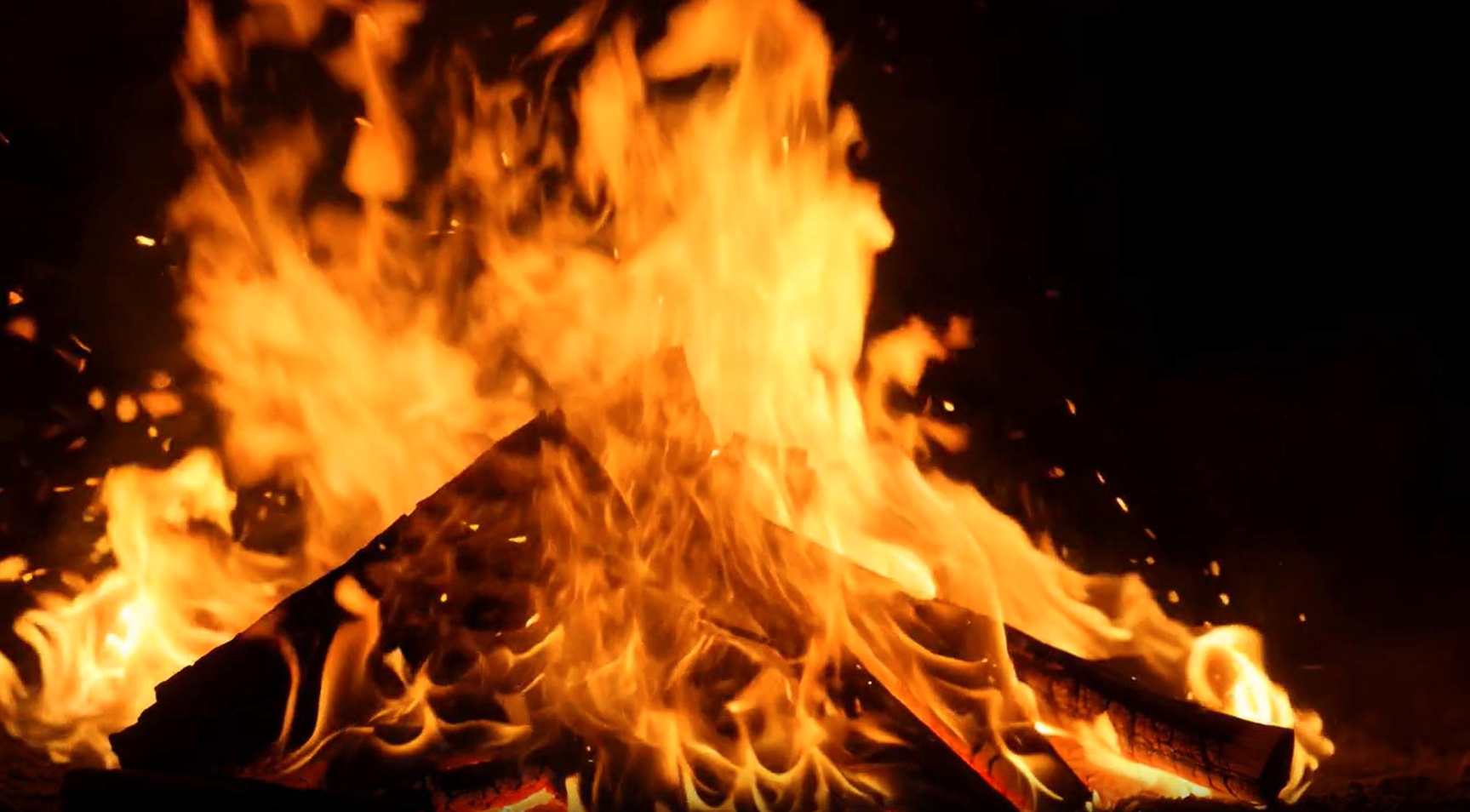}
        \end{subfigure}
        \vspace{1pt}
        \begin{subfigure}{\textwidth}
            \textbf{\scriptsize DynamicRad (Ours):} \\
            \includegraphics[width=0.195\textwidth]{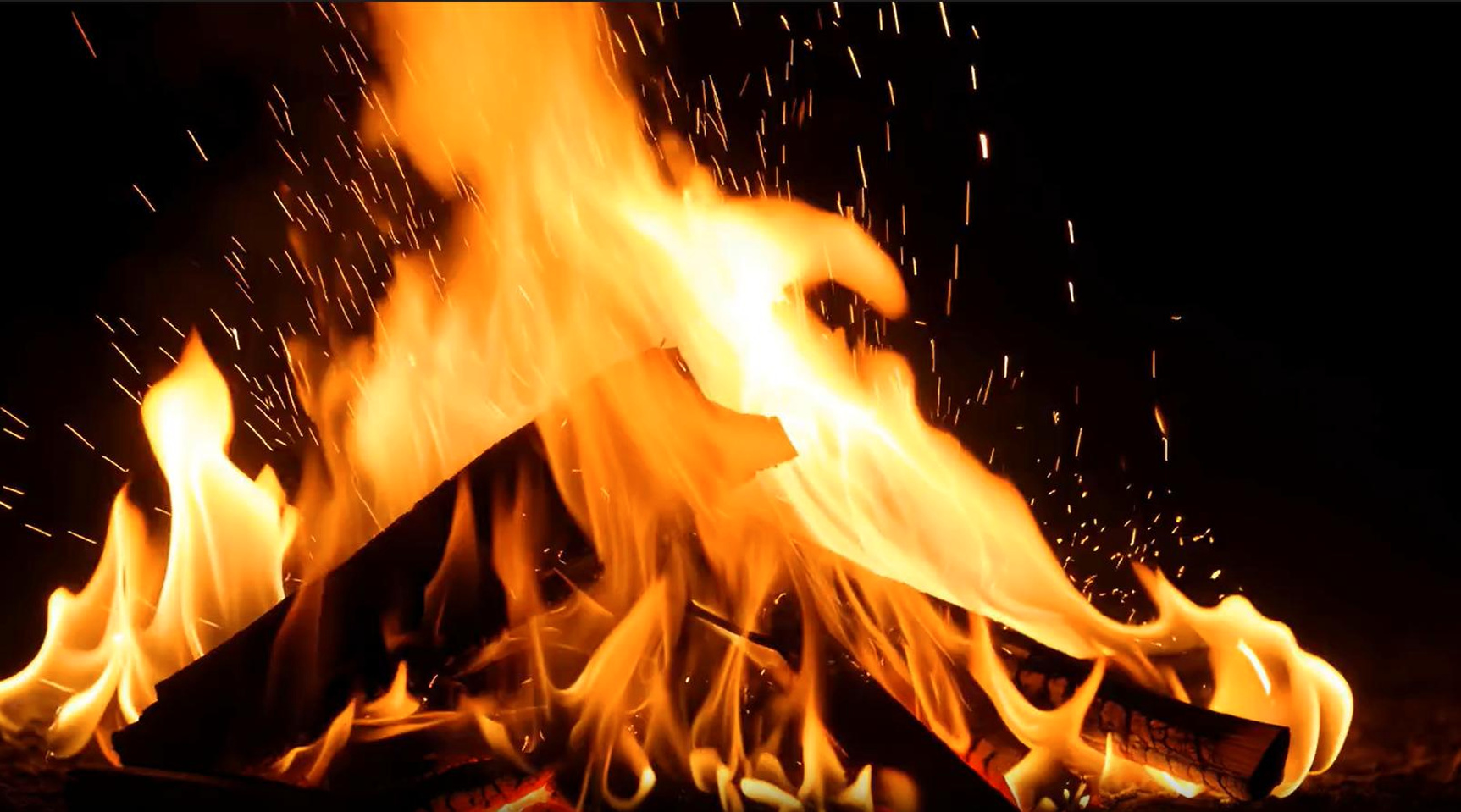} \hfill
            \includegraphics[width=0.195\textwidth]{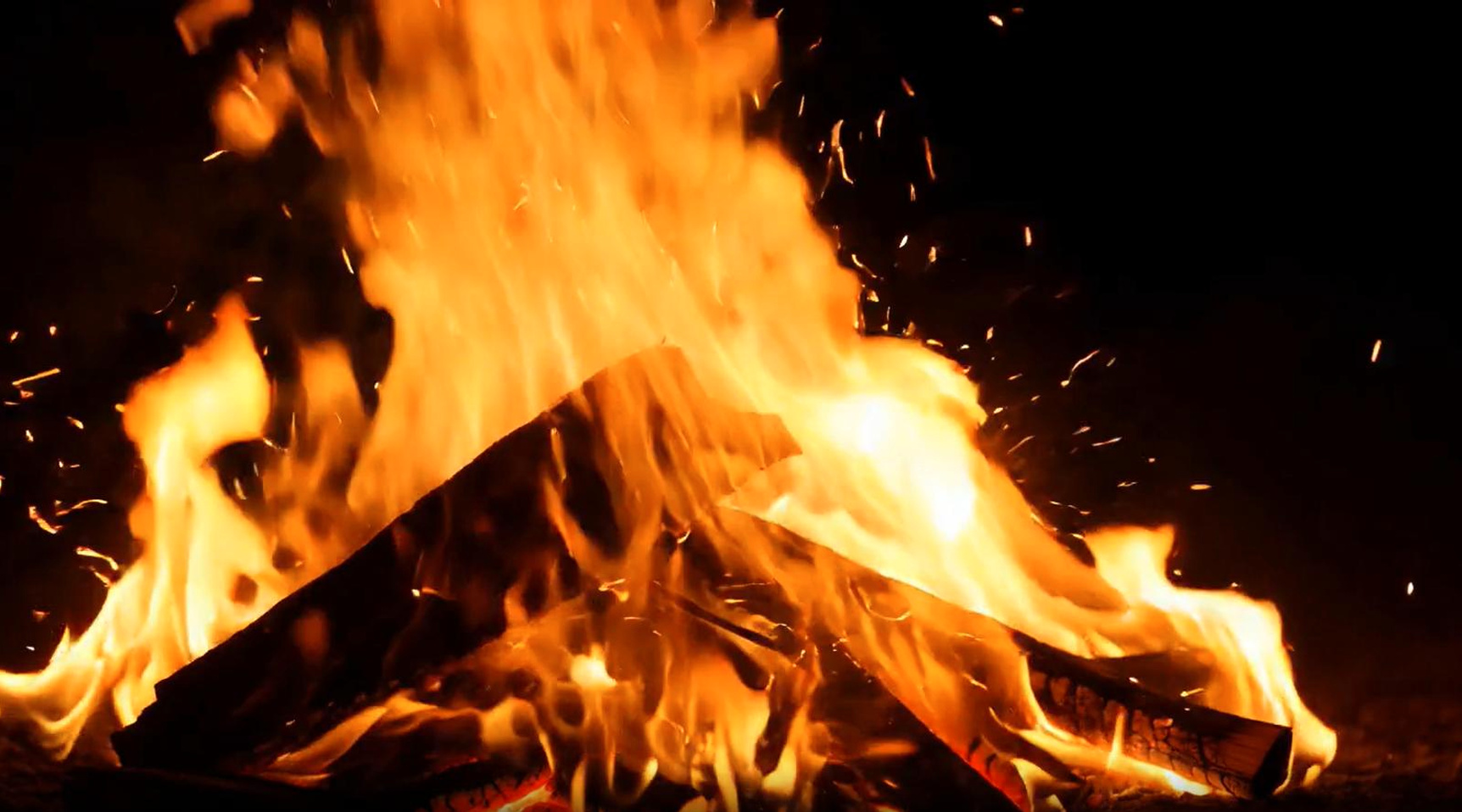} \hfill
            \includegraphics[width=0.195\textwidth]{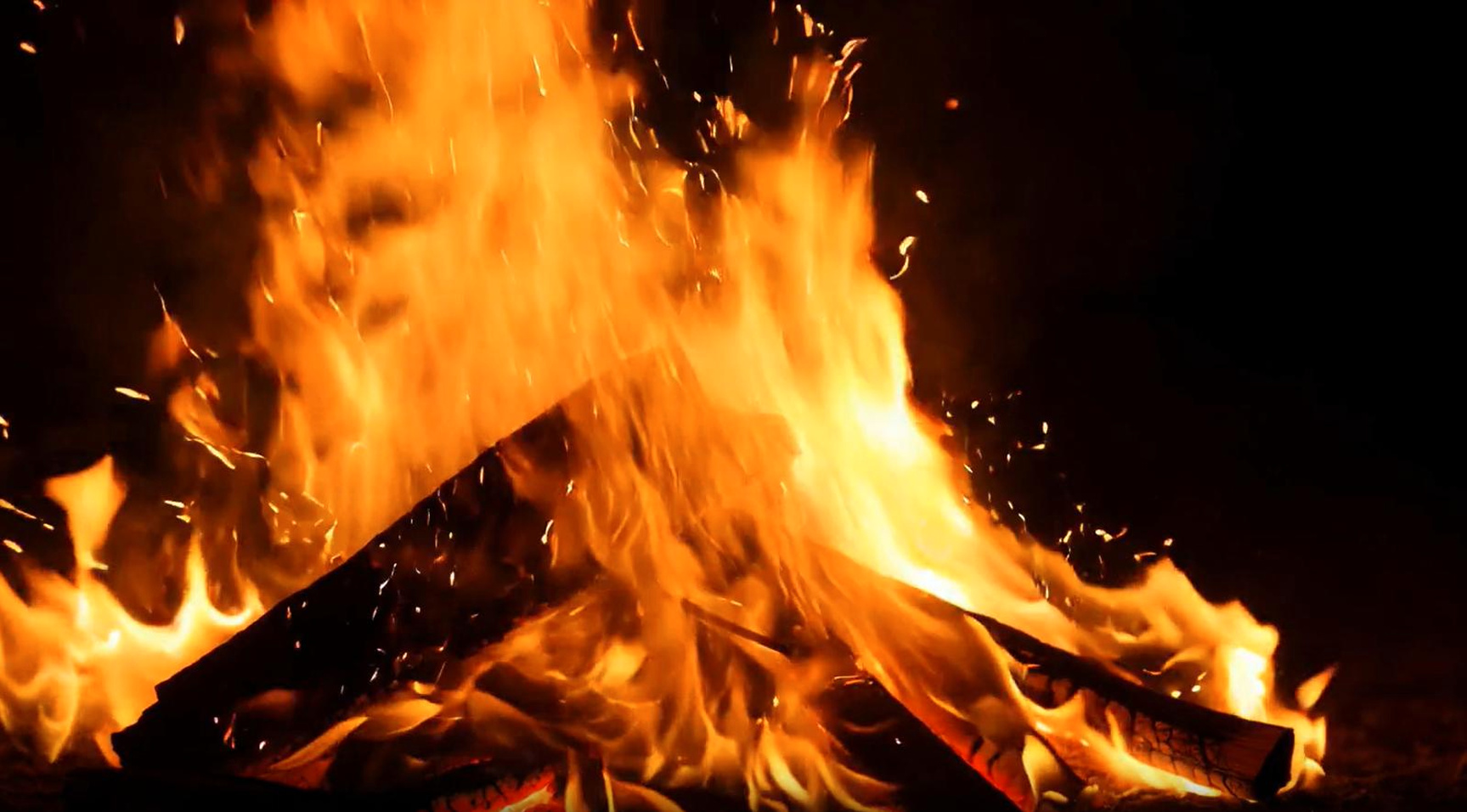} \hfill
            \includegraphics[width=0.195\textwidth]{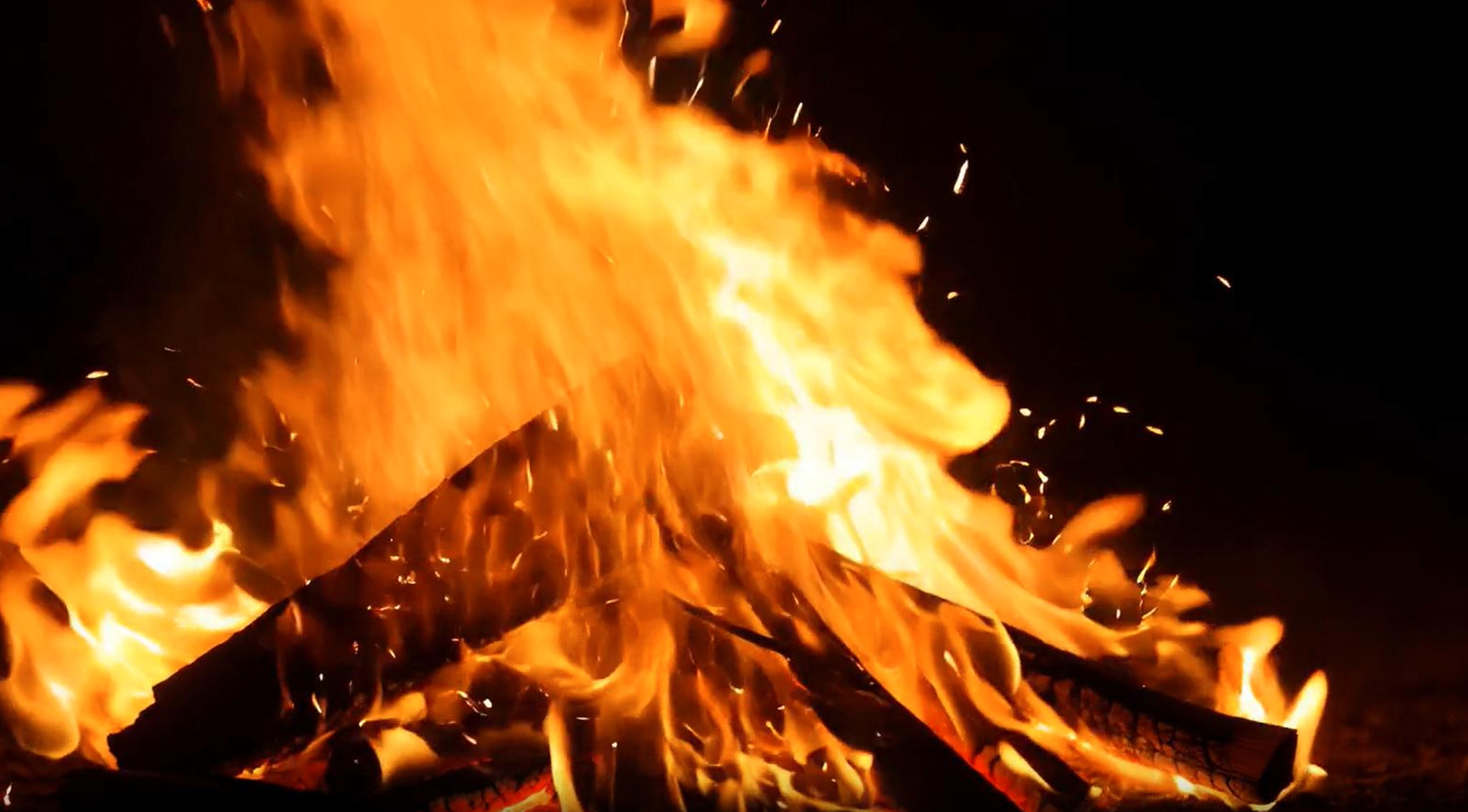} \hfill
            \includegraphics[width=0.195\textwidth]{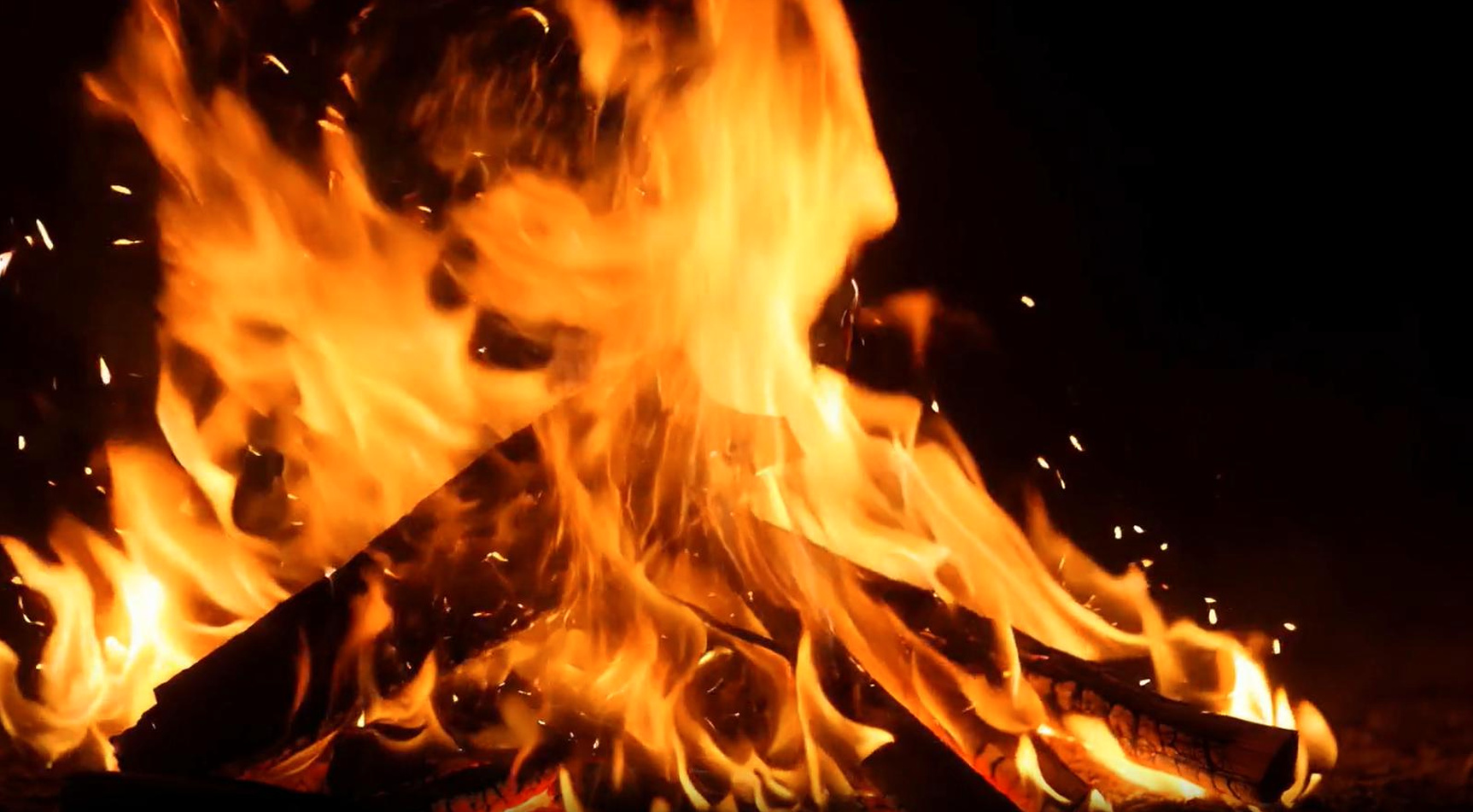}
        \end{subfigure}
    \end{minipage}
    \vspace{20pt}

    \begin{minipage}{\textwidth}
        \fcolorbox{gray!50}{gray!10}{
            \parbox{0.97\linewidth}{
                \vspace{2pt}
                \textbf{\small Case 2 (Fast Motion):} \small \textit{“A majestic white horse galloping through a field of golden wheat, wind blowing through its mane, dynamic camera angle tracking the horse, sunlight breaking through clouds, 4k, slow motion.”}
                \vspace{2pt}
            }
        }
        \vspace{3pt}
        \begin{subfigure}{\textwidth}
        \vspace{3pt}
            \textbf{\scriptsize Original (Dense):} \\
            \includegraphics[width=0.195\textwidth]{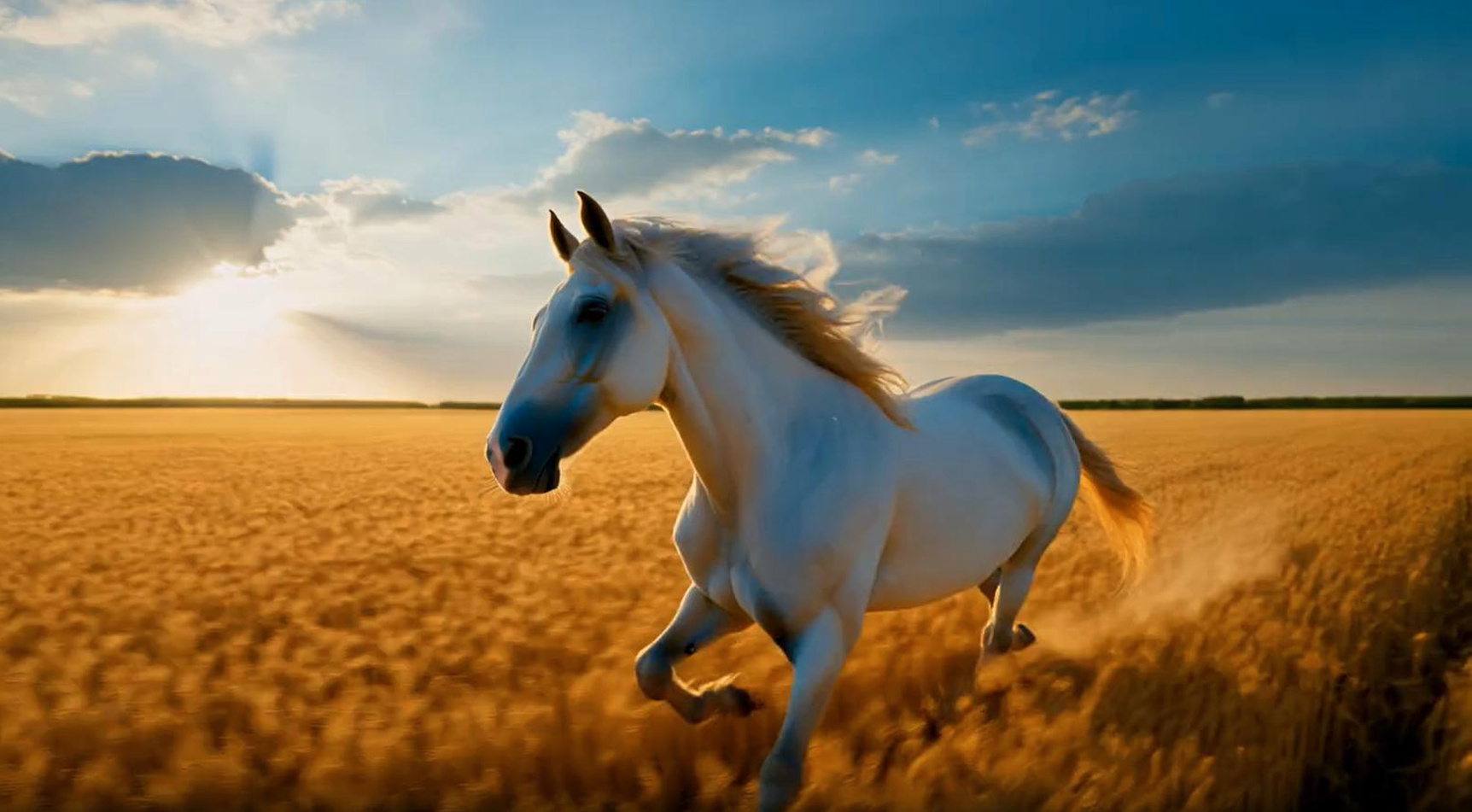} \hfill
            \includegraphics[width=0.195\textwidth]{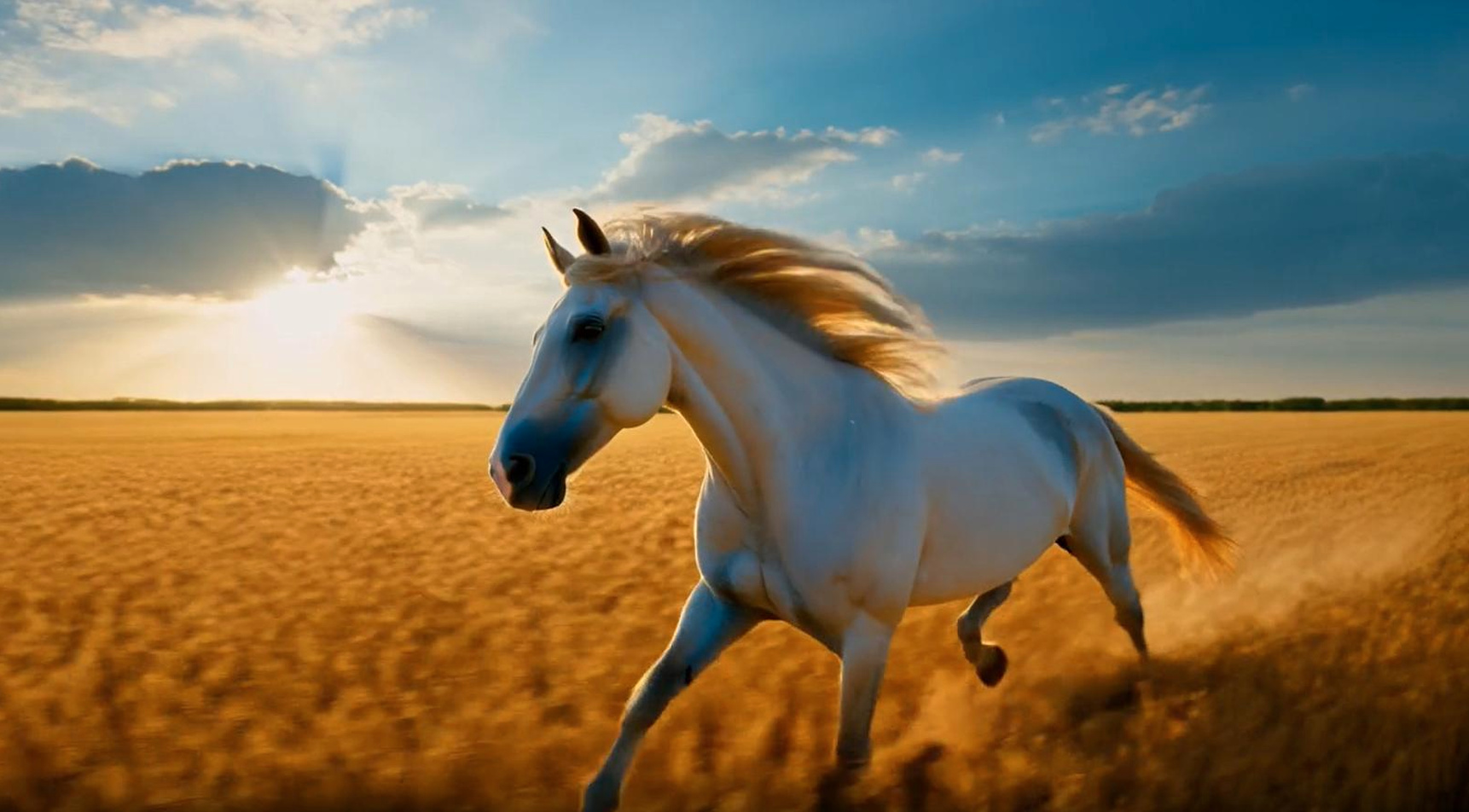} \hfill
            \includegraphics[width=0.195\textwidth]{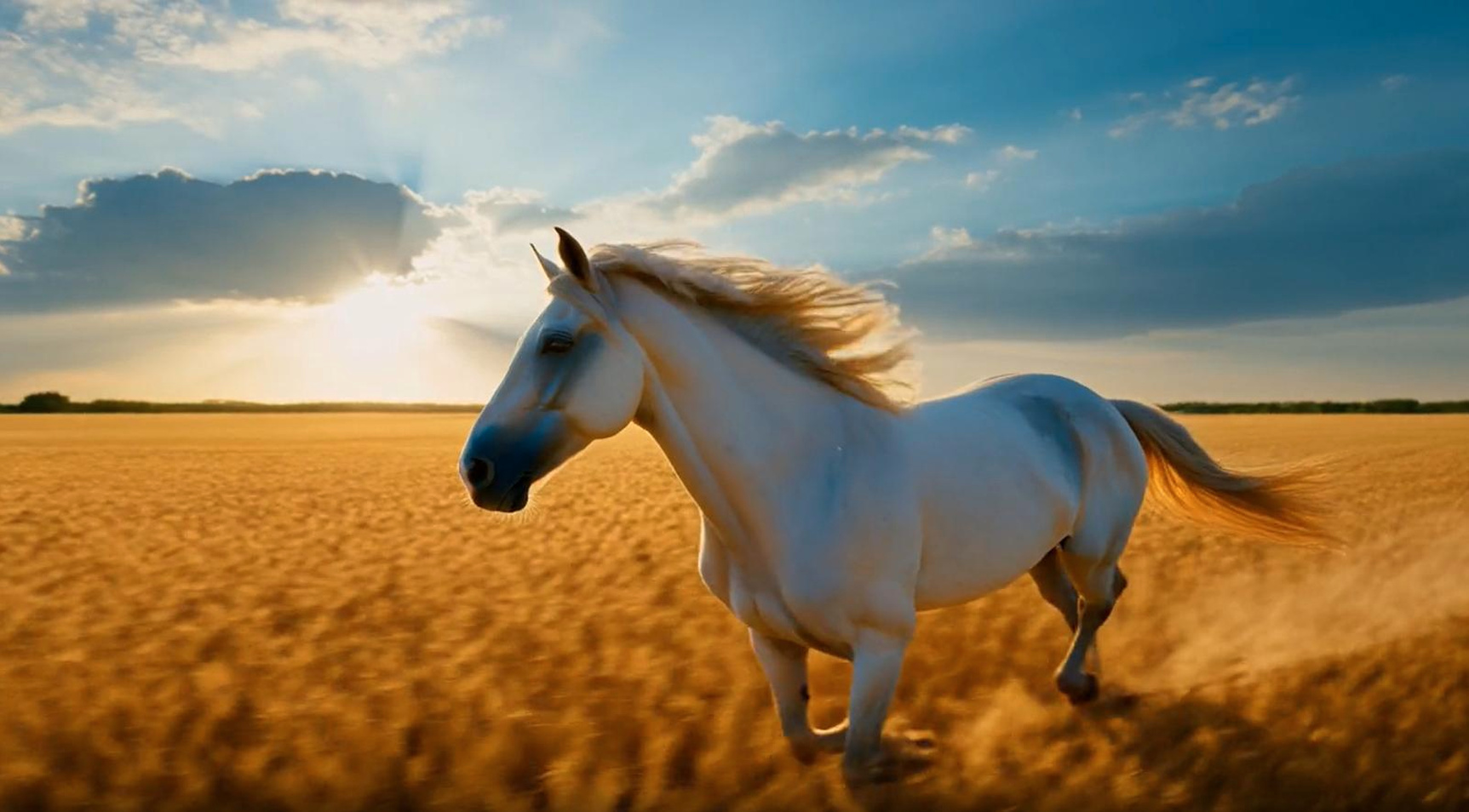} \hfill
            \includegraphics[width=0.195\textwidth]{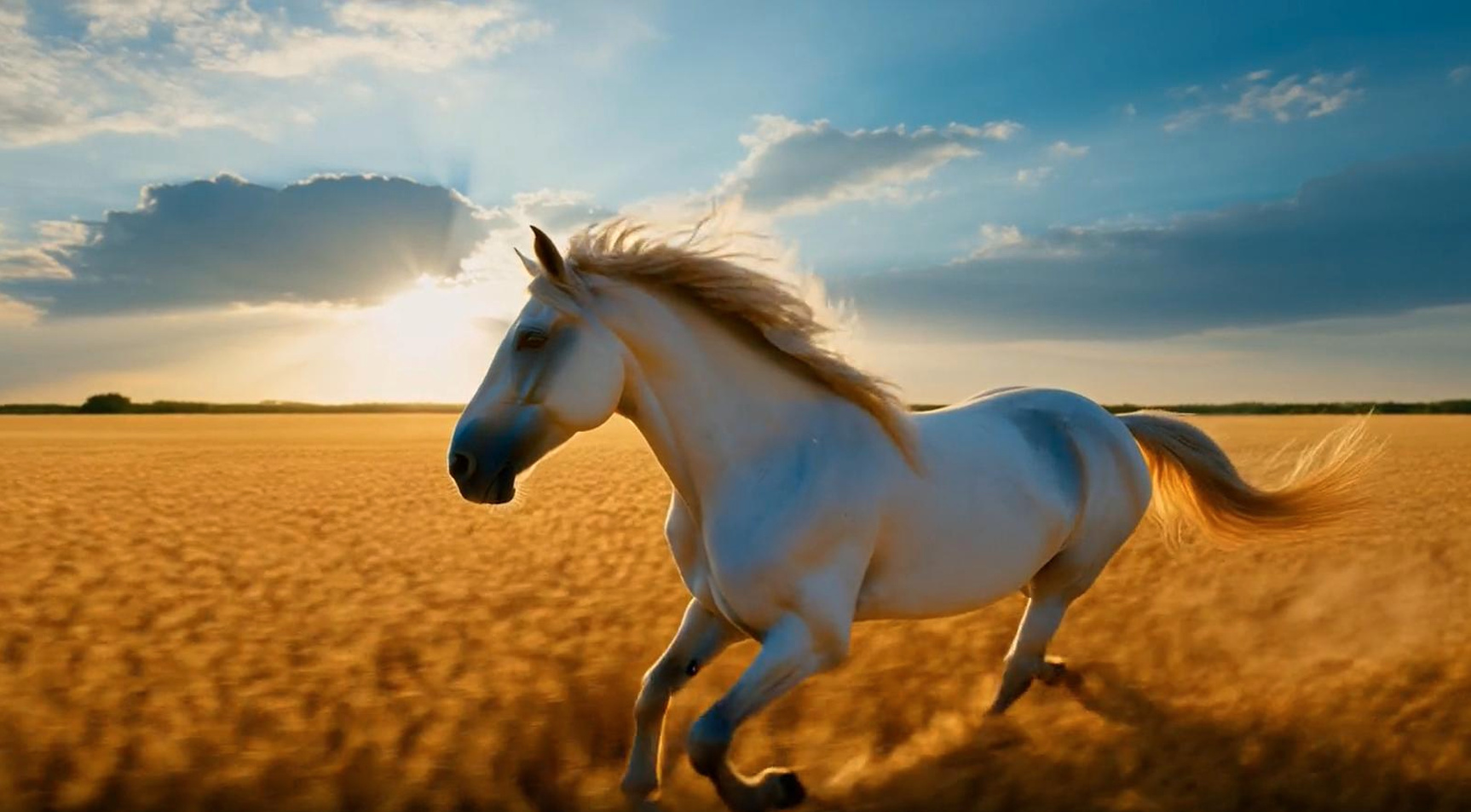} \hfill
            \includegraphics[width=0.195\textwidth]{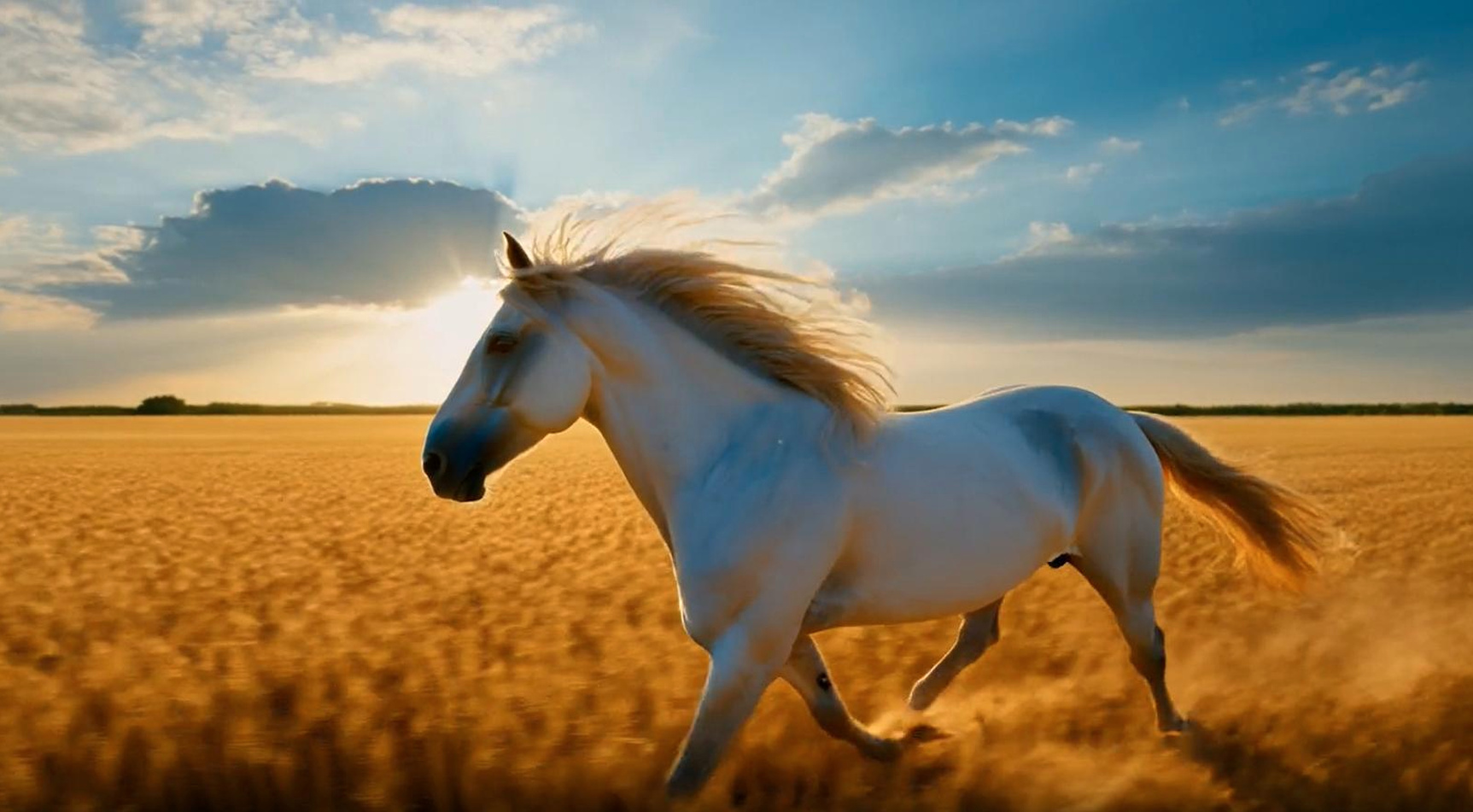}
        \end{subfigure}
        \vspace{1pt}
        \begin{subfigure}{\textwidth}
            \textbf{\scriptsize DynamicRad (Ours):} \\
            \includegraphics[width=0.195\textwidth]{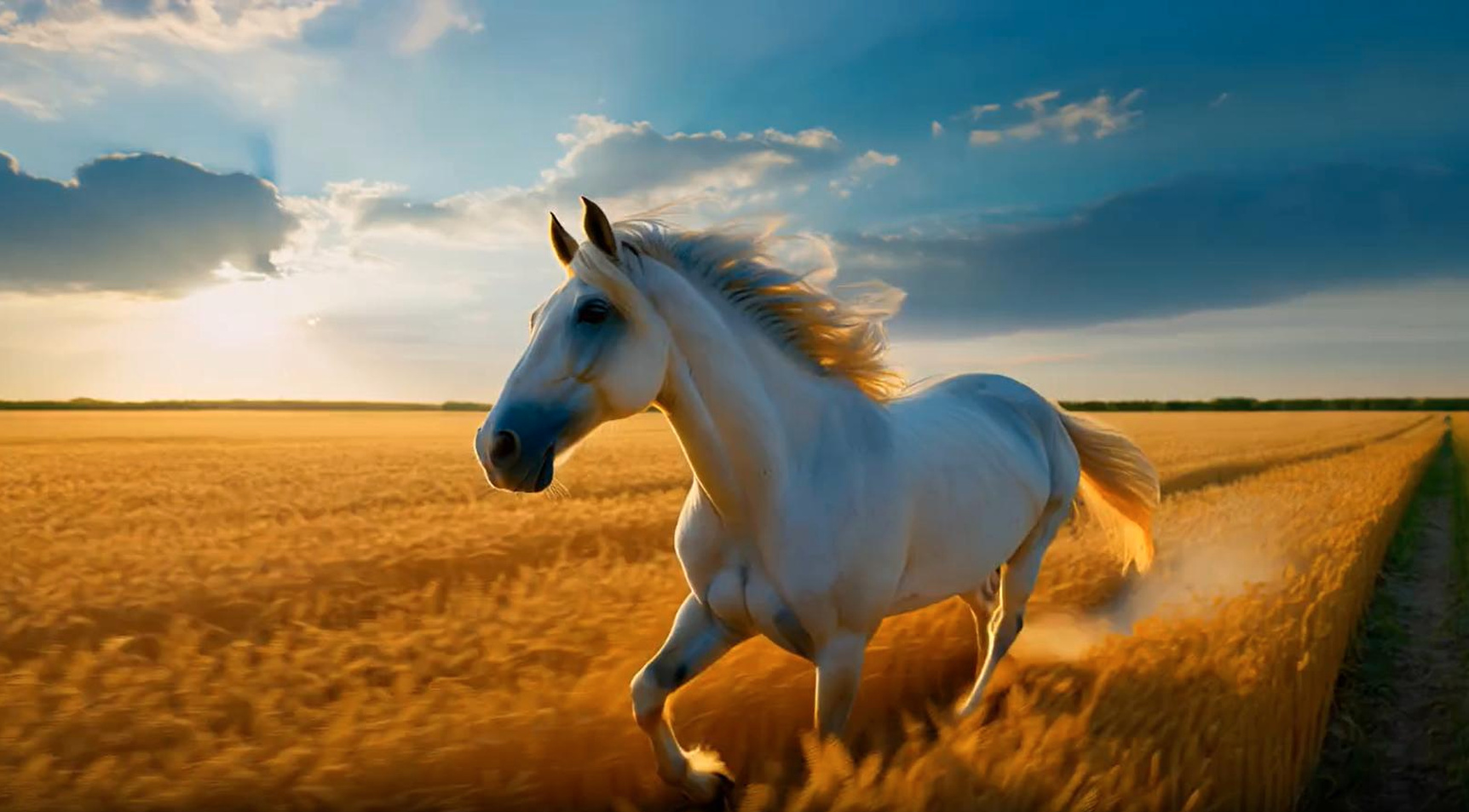} \hfill
            \includegraphics[width=0.195\textwidth]{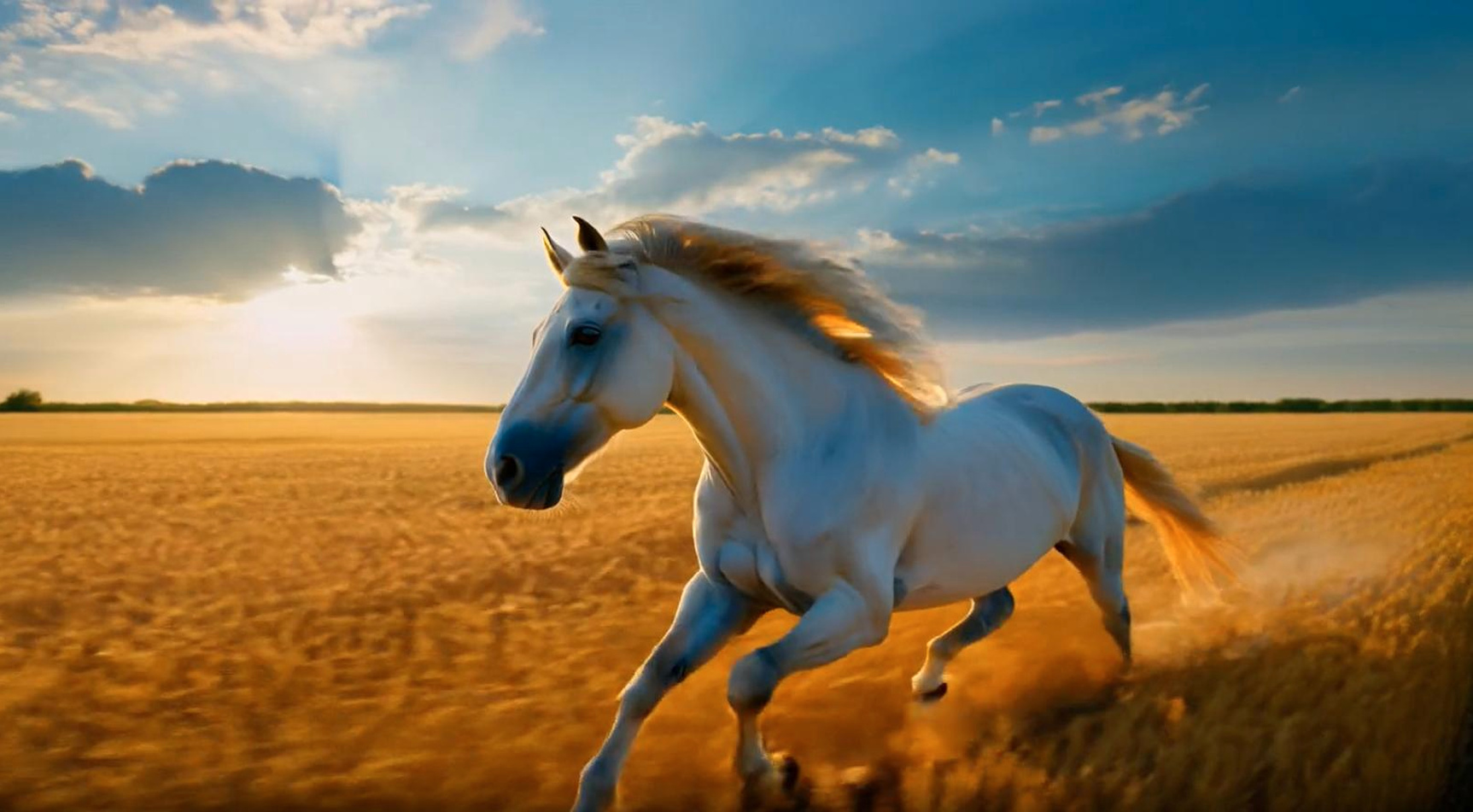} \hfill
            \includegraphics[width=0.195\textwidth]{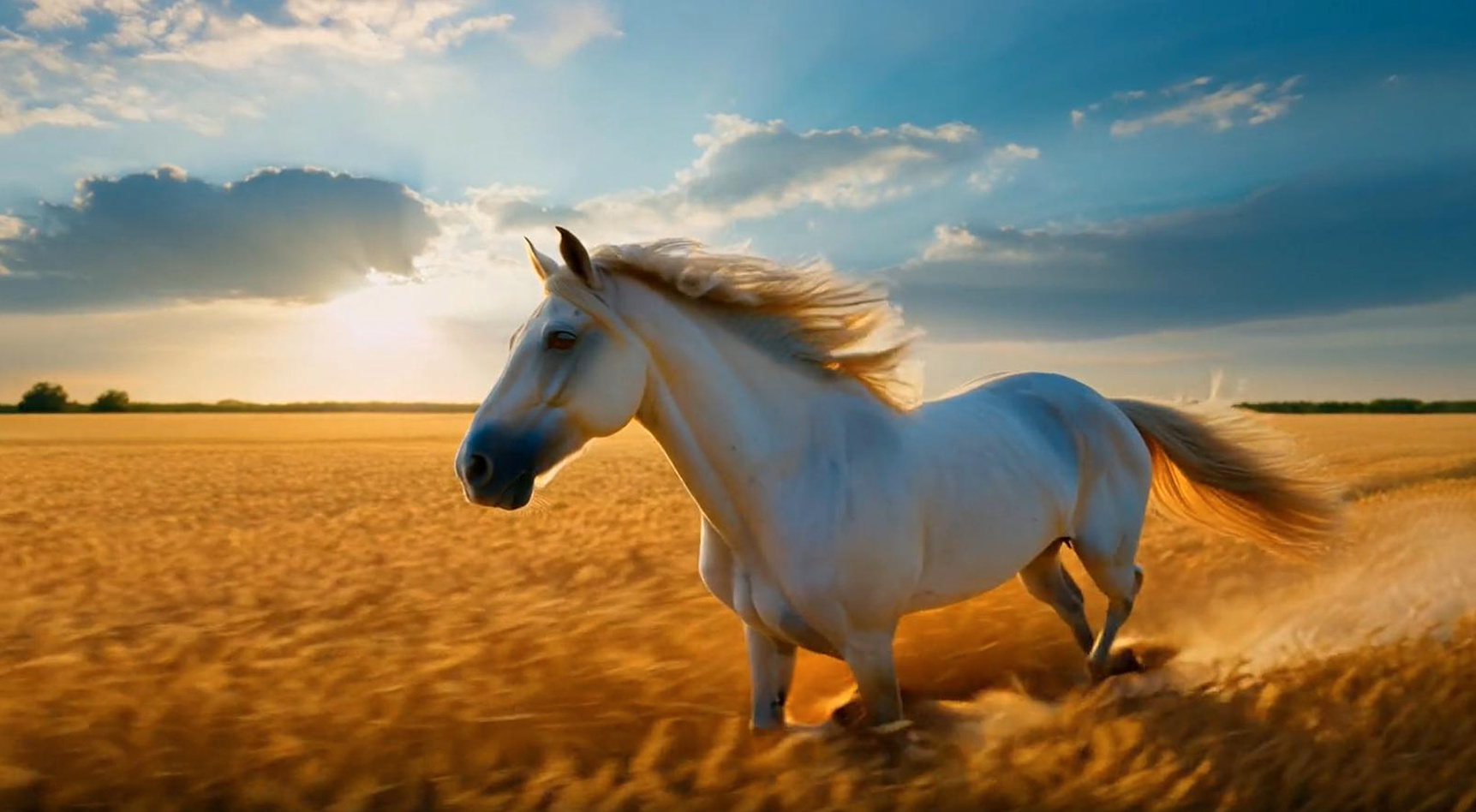} \hfill
            \includegraphics[width=0.195\textwidth]{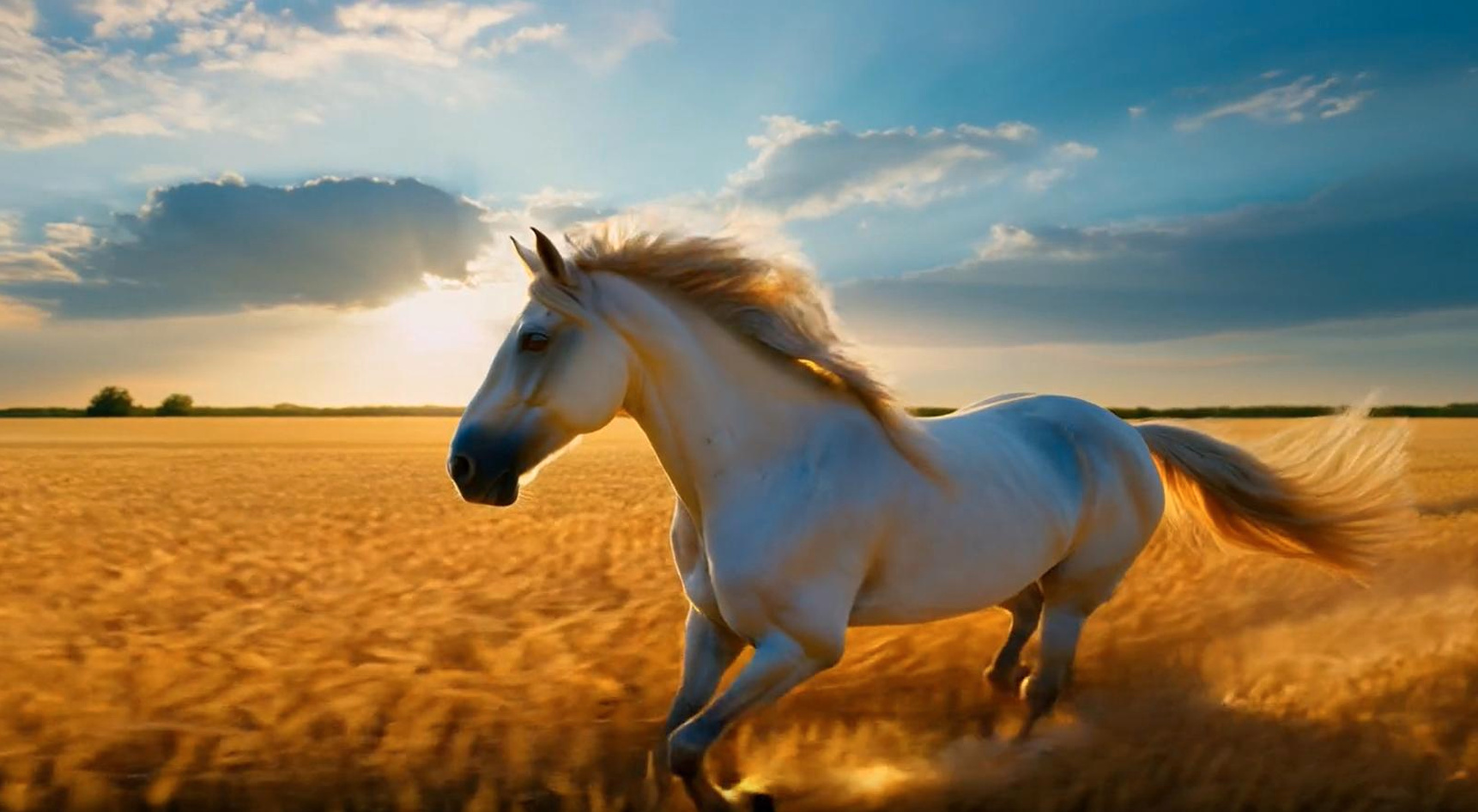} \hfill
            \includegraphics[width=0.195\textwidth]{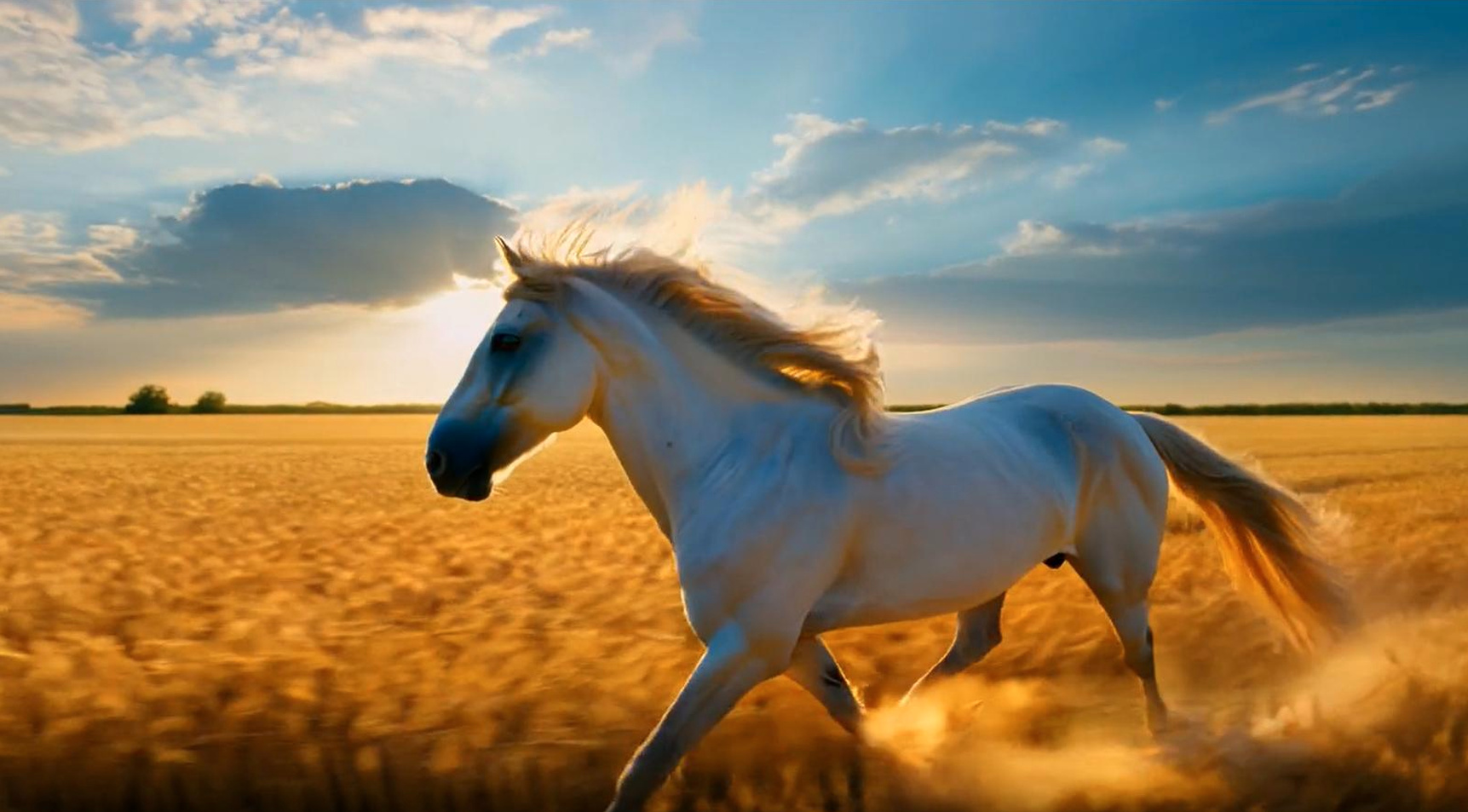}
        \end{subfigure}
    \end{minipage}
    \vspace{20pt}
    
    \begin{minipage}{\textwidth}
        \fcolorbox{gray!50}{gray!10}{
            \parbox{0.97\linewidth}{
                \vspace{2pt}
                \textbf{\small Case 3 (Stylized/Anime):} \small \textit{“Anime style, Makoto Shinkai style, a young girl standing on a train station platform, cherry blossoms falling, vibrant blue sky with fluffy clouds, high quality animation.”}
                \vspace{2pt}
            }
        }
        \vspace{3pt}
        \begin{subfigure}{\textwidth}
        \vspace{3pt}
            \textbf{\scriptsize Original (Dense):} \\
            \includegraphics[width=0.195\textwidth]{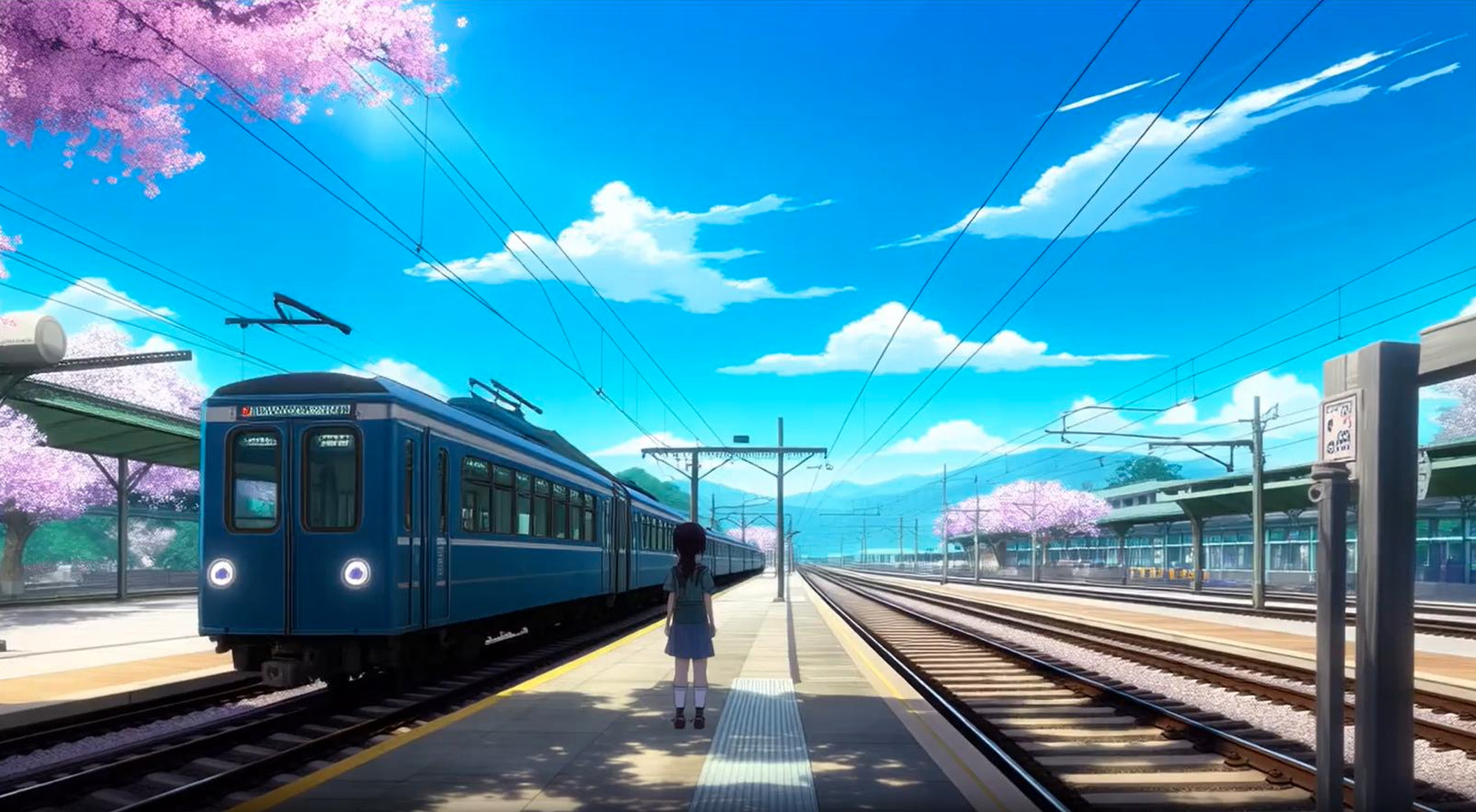} \hfill
            \includegraphics[width=0.195\textwidth]{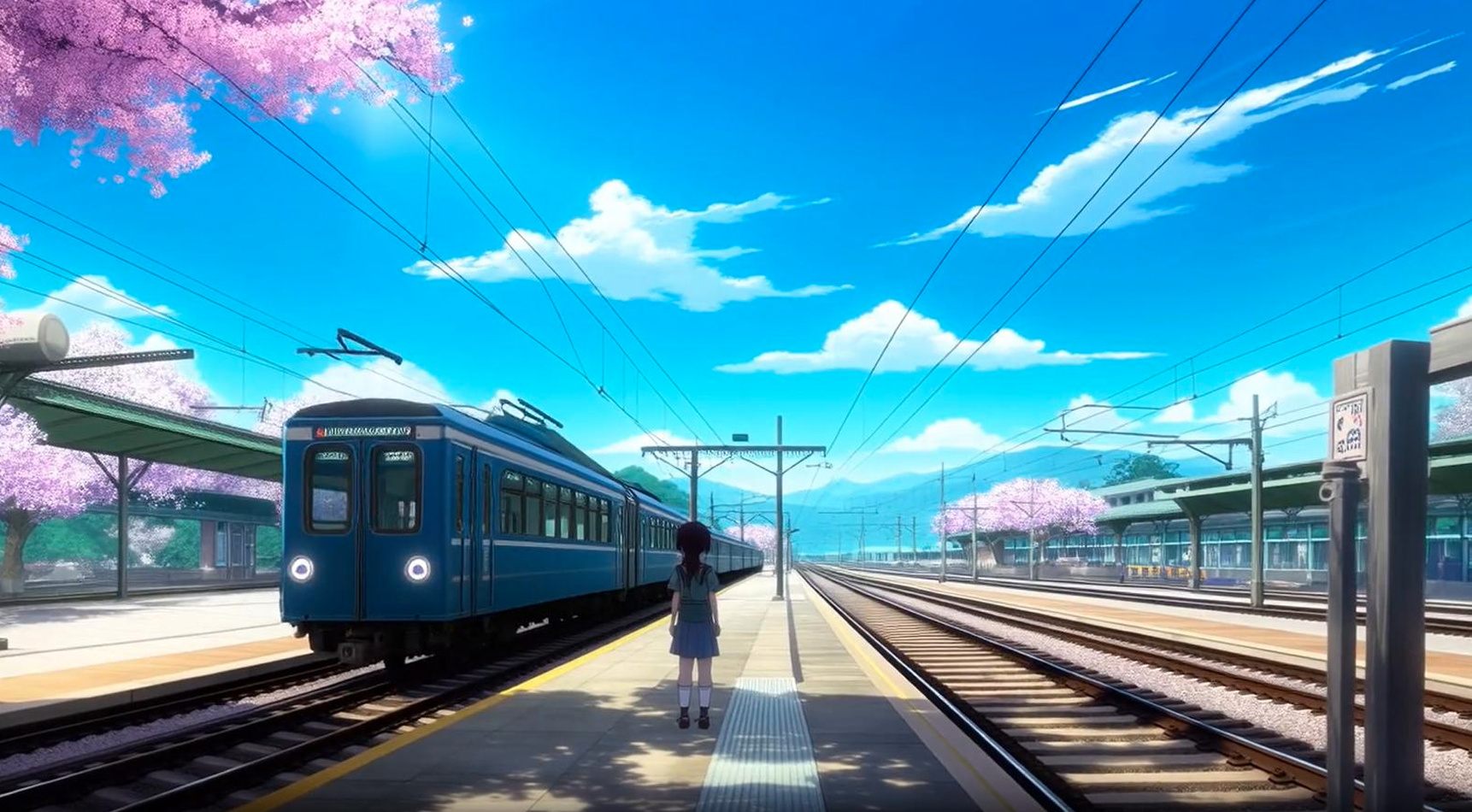} \hfill
            \includegraphics[width=0.195\textwidth]{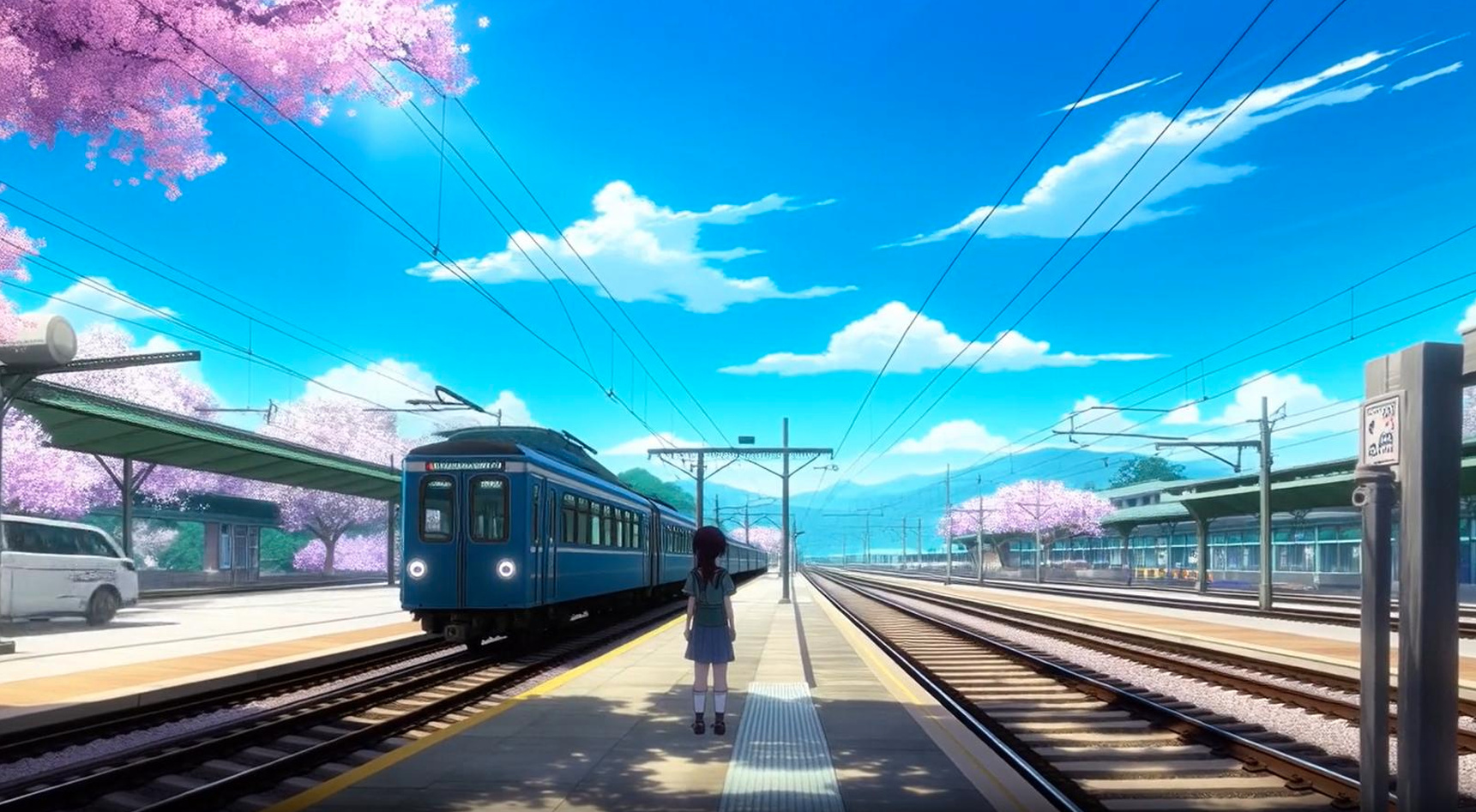} \hfill
            \includegraphics[width=0.195\textwidth]{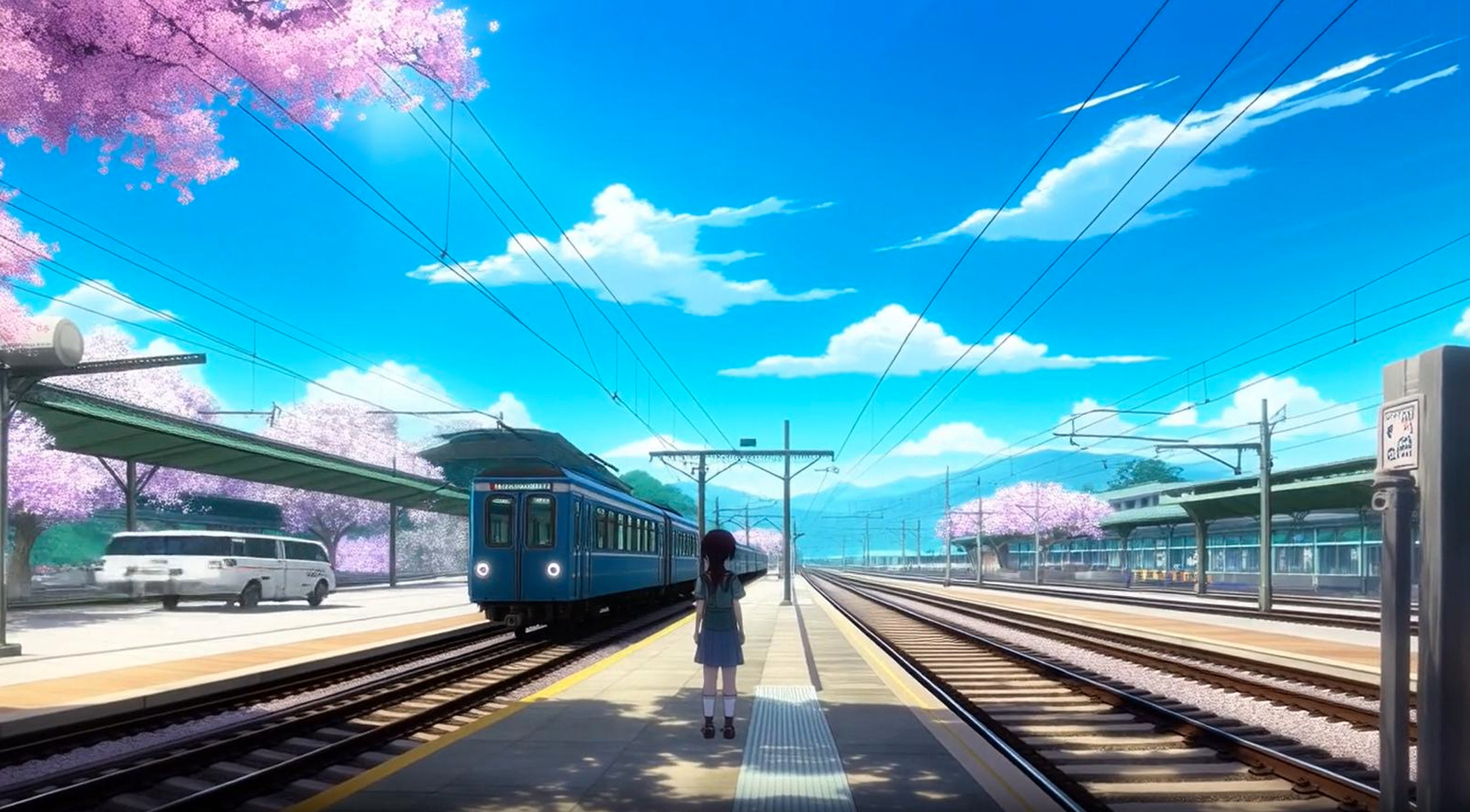} \hfill
            \includegraphics[width=0.195\textwidth]{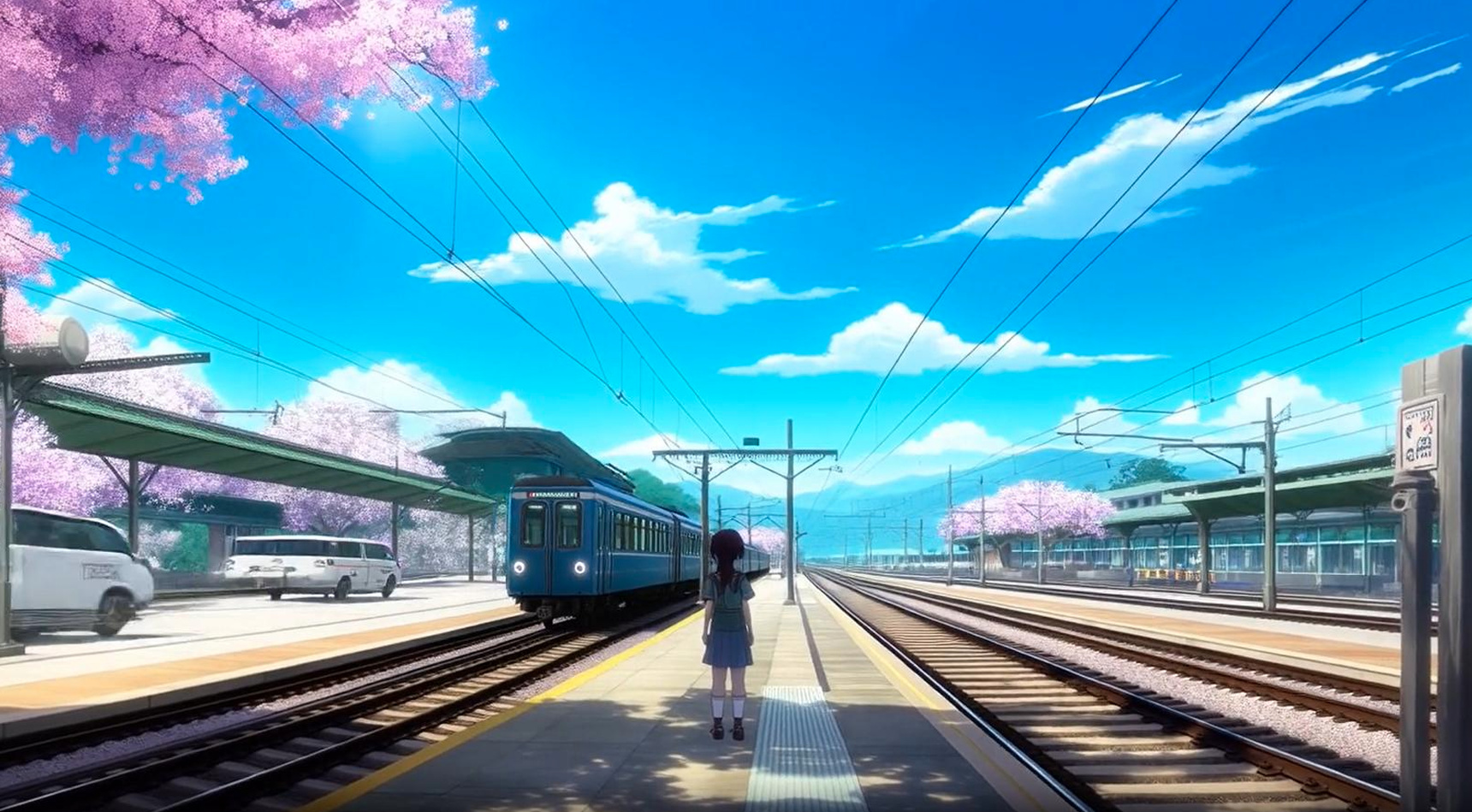}
        \end{subfigure}
        \vspace{1pt}
        \begin{subfigure}{\textwidth}
            \textbf{\scriptsize DynamicRad (Ours):} \\
            \includegraphics[width=0.195\textwidth]{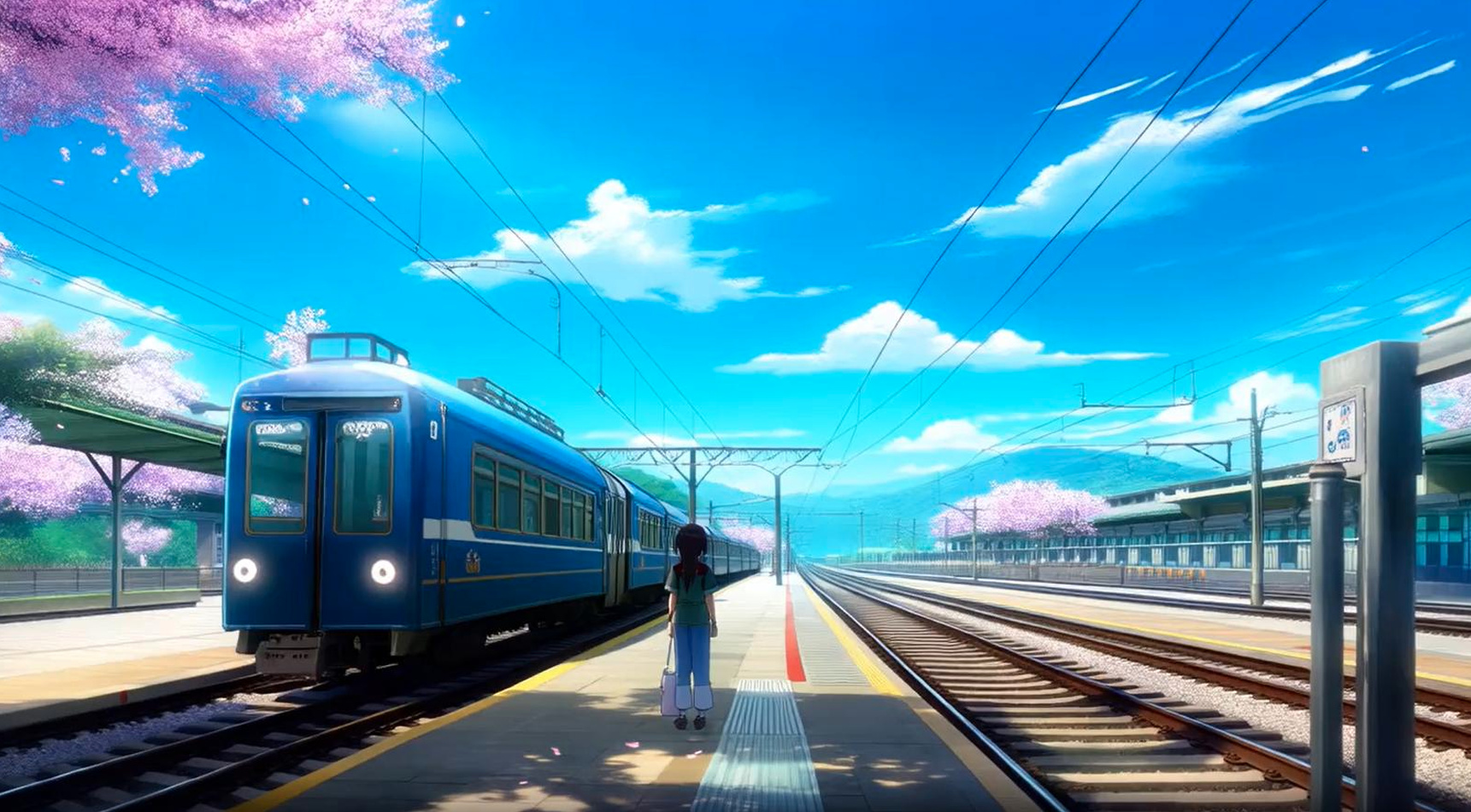} \hfill
            \includegraphics[width=0.195\textwidth]{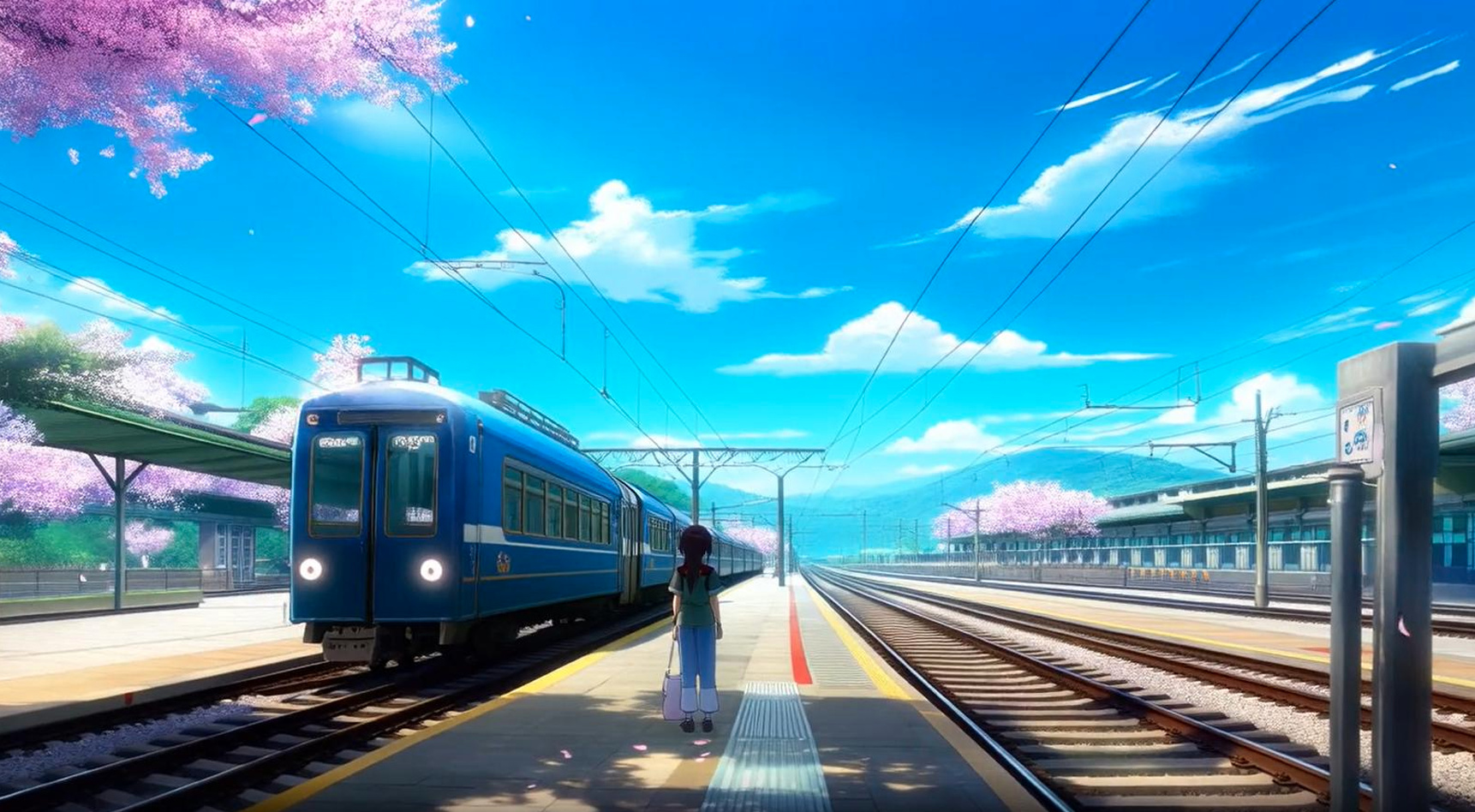} \hfill
            \includegraphics[width=0.195\textwidth]{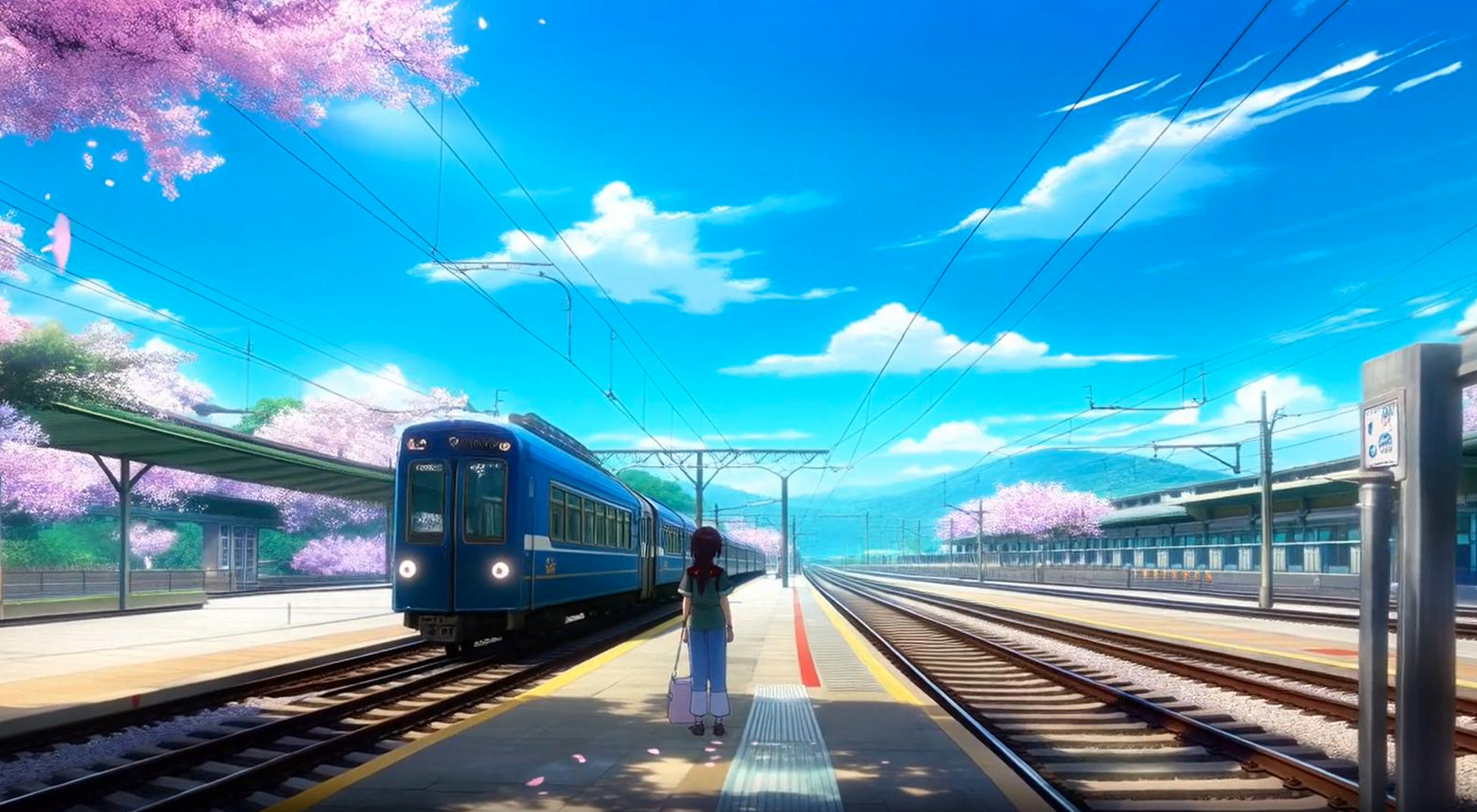} \hfill
            \includegraphics[width=0.195\textwidth]{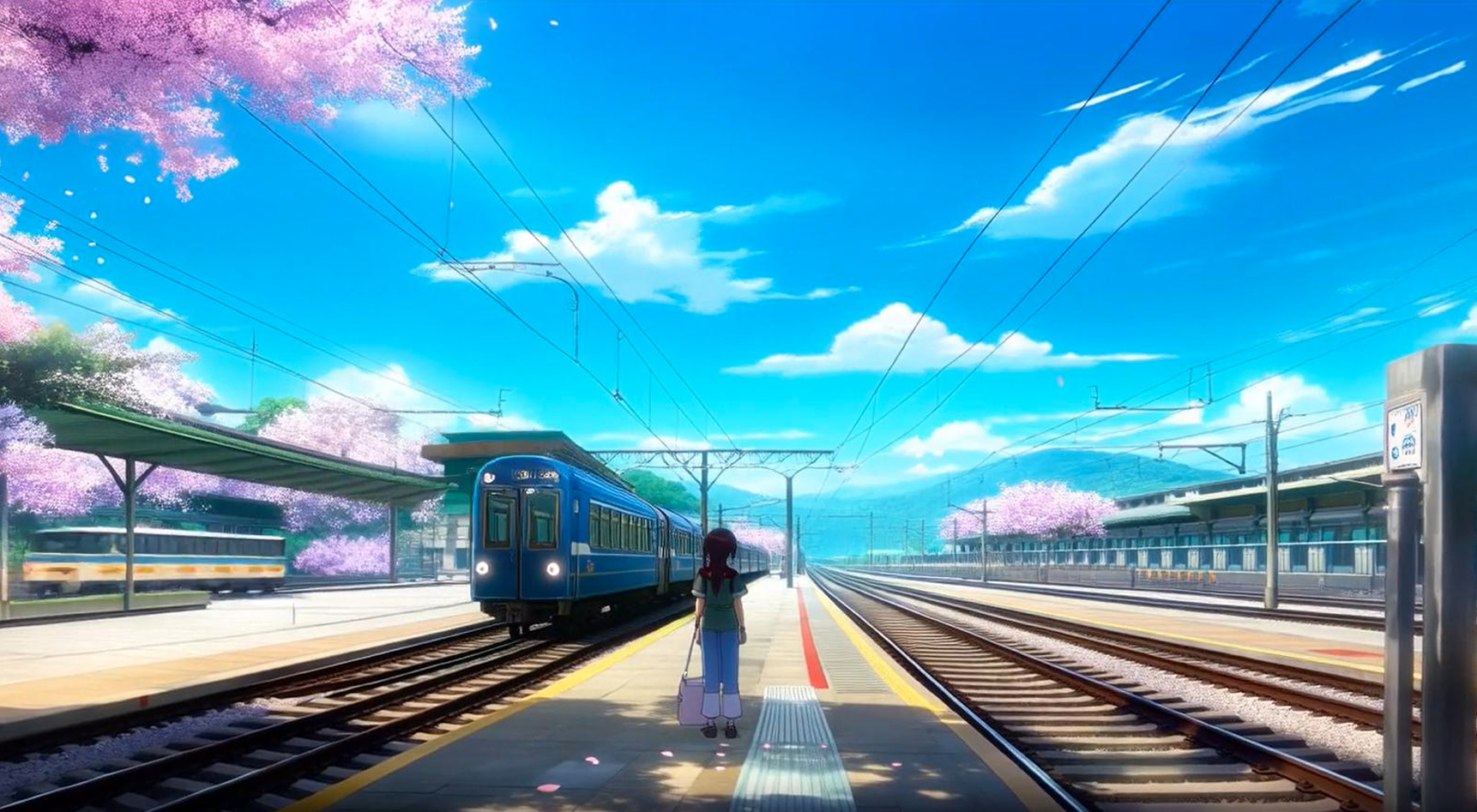} \hfill
            \includegraphics[width=0.195\textwidth]{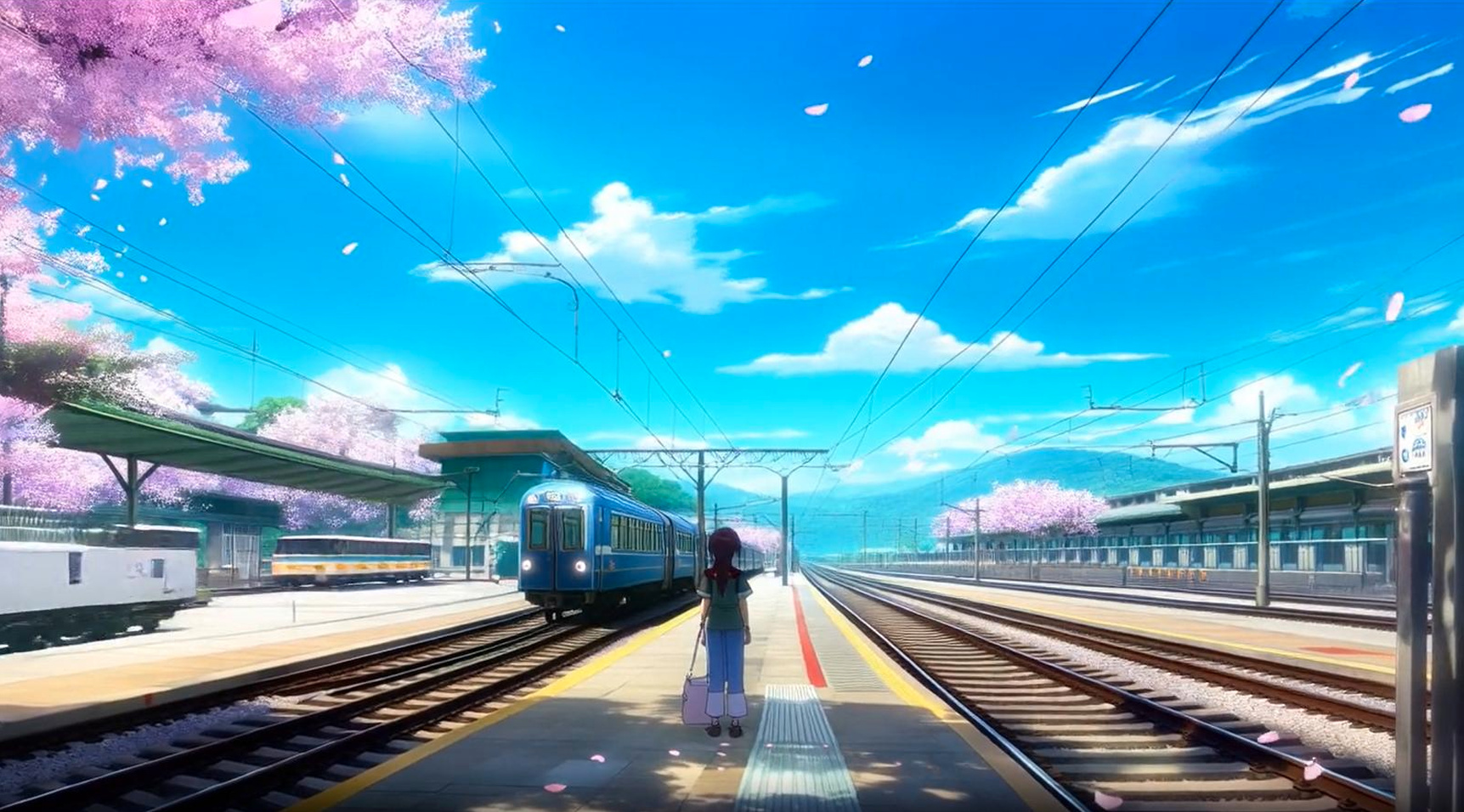}
        \end{subfigure}
    \end{minipage}

    \label{fig:wan_gallery_1}
\end{figure}

\clearpage

\begin{figure}[H]
    \centering
    \setlength{\tabcolsep}{1pt} \renewcommand{\arraystretch}{0.1}
    \begin{minipage}{\textwidth}
        \fcolorbox{gray!50}{gray!10}{
            \parbox{0.97\linewidth}{
                \vspace{2pt}
                \textbf{\small Case 4 (Complex Scene):} \small \textit{“Cyberpunk city street at night, neon lights reflecting on wet pavement, a mysterious figure walking through steam and rain, cinematic lighting, volumetric fog, highly detailed, 8k.”}
                \vspace{2pt}
            }
        }
        \vspace{3pt}
        \begin{subfigure}{\textwidth}
        \vspace{3pt}
            \textbf{\scriptsize Original (Dense):} \\
            \includegraphics[width=0.195\textwidth]{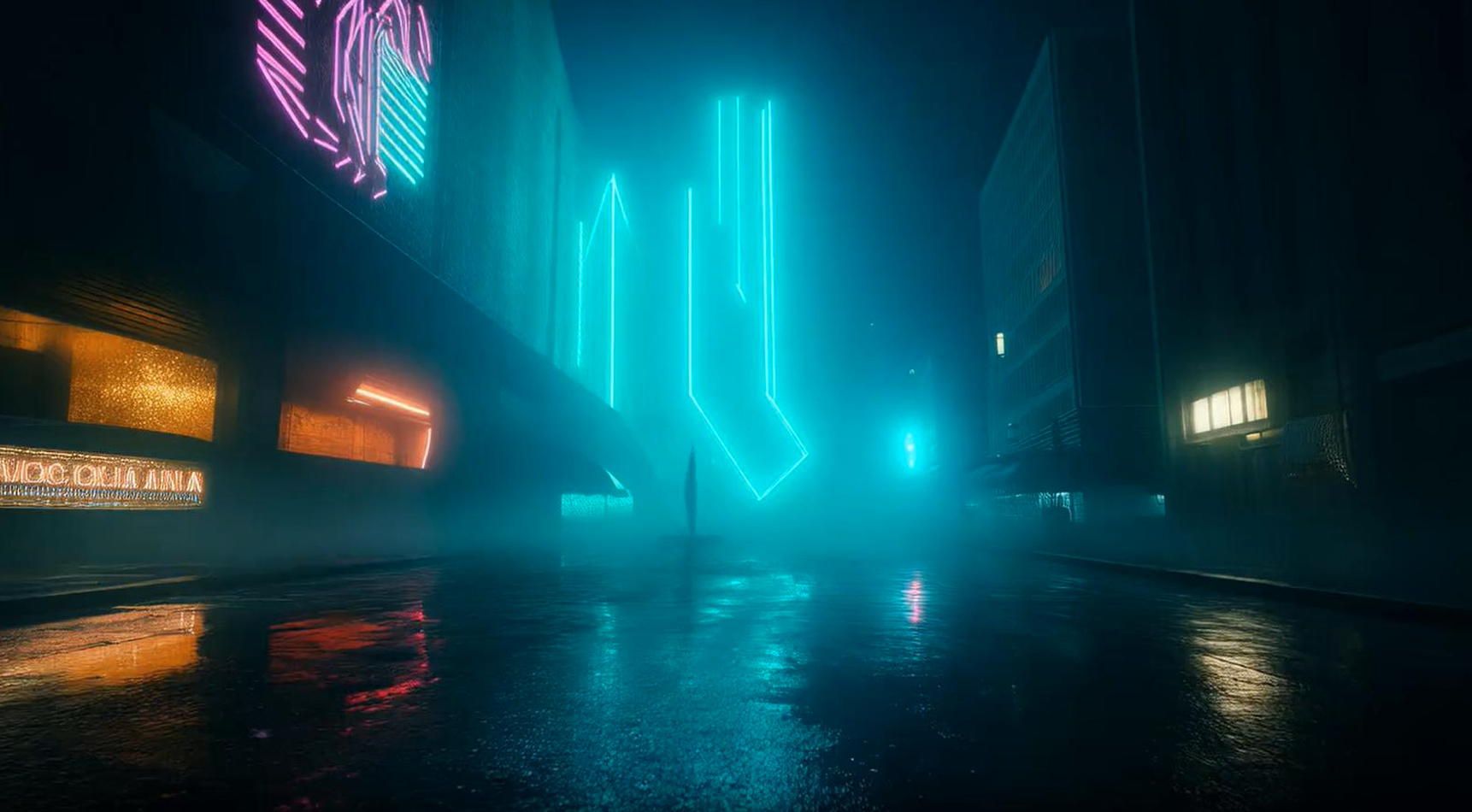} \hfill
            \includegraphics[width=0.195\textwidth]{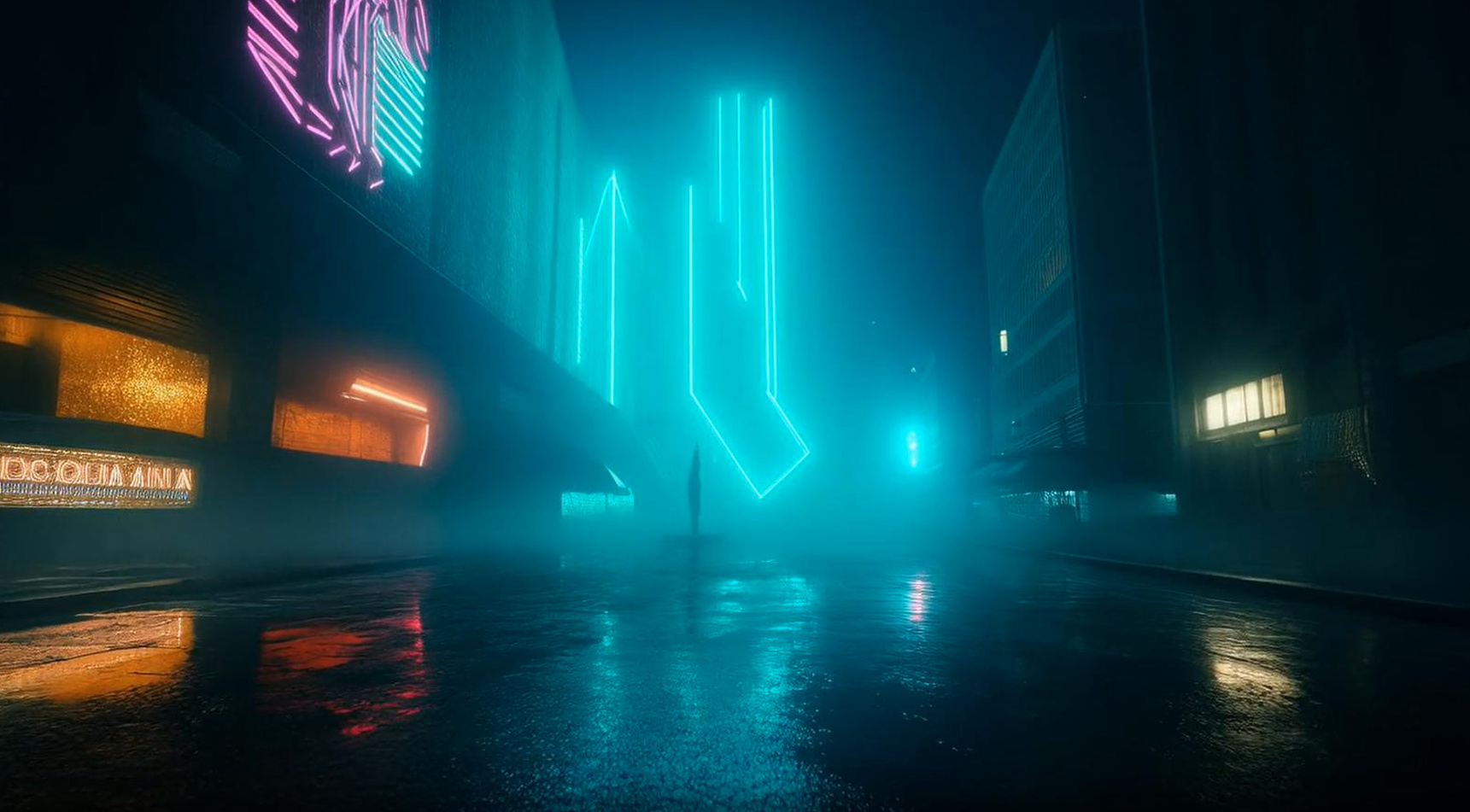} \hfill
            \includegraphics[width=0.195\textwidth]{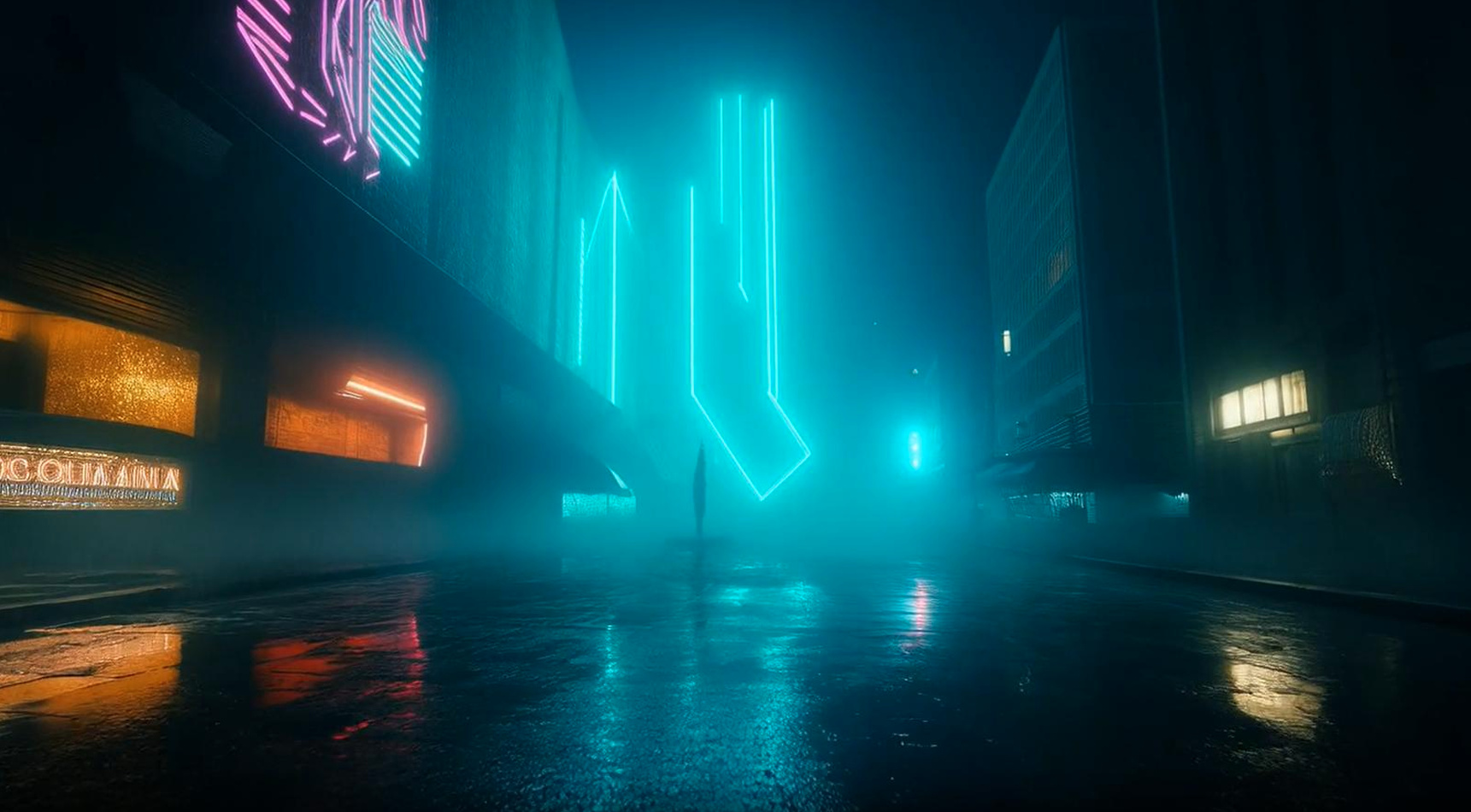} \hfill
            \includegraphics[width=0.195\textwidth]{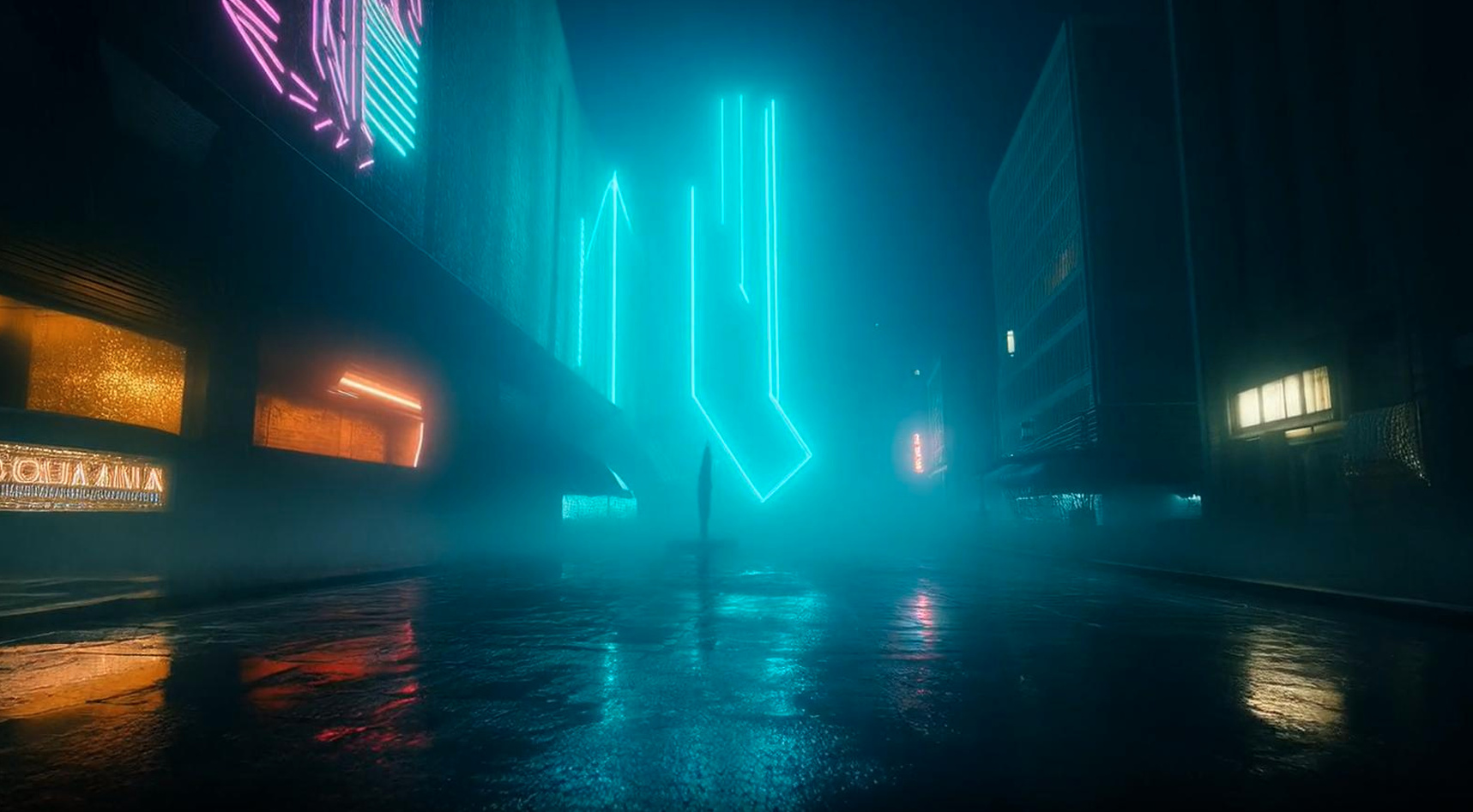} \hfill
            \includegraphics[width=0.195\textwidth]{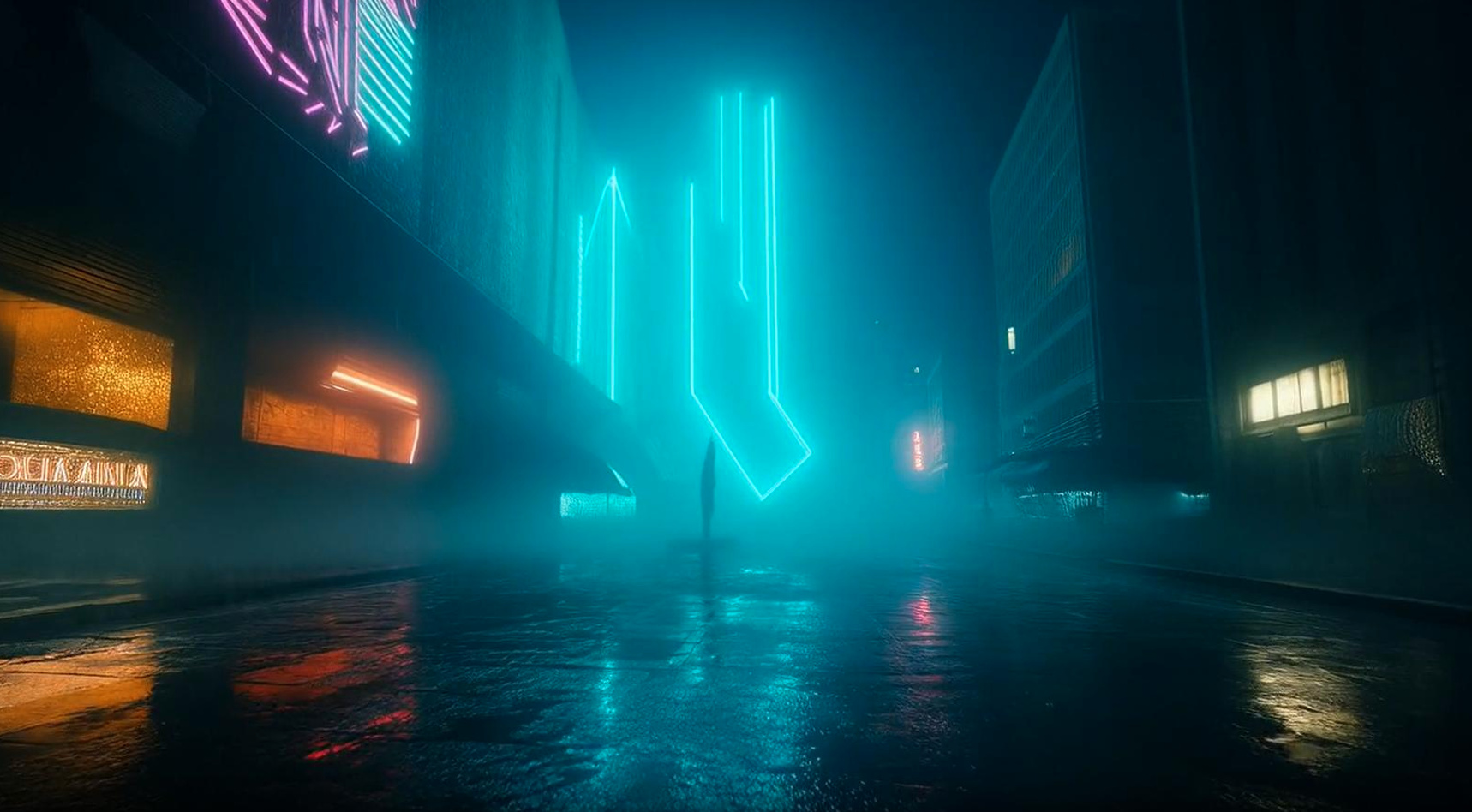}
        \end{subfigure}
        \vspace{1pt}
        \begin{subfigure}{\textwidth}
            \textbf{\scriptsize DynamicRad (Ours):} \\
            \includegraphics[width=0.195\textwidth]{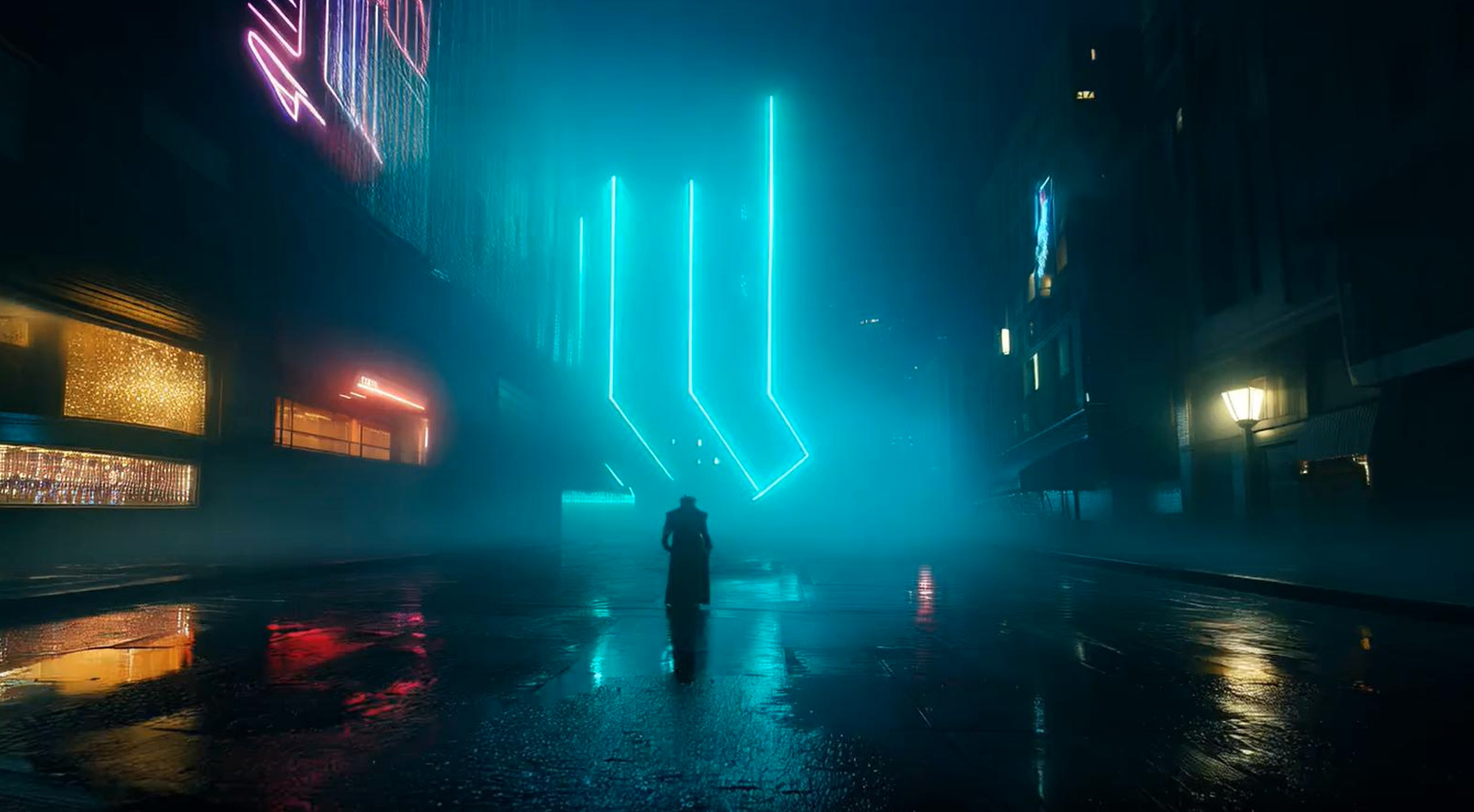} \hfill
            \includegraphics[width=0.195\textwidth]{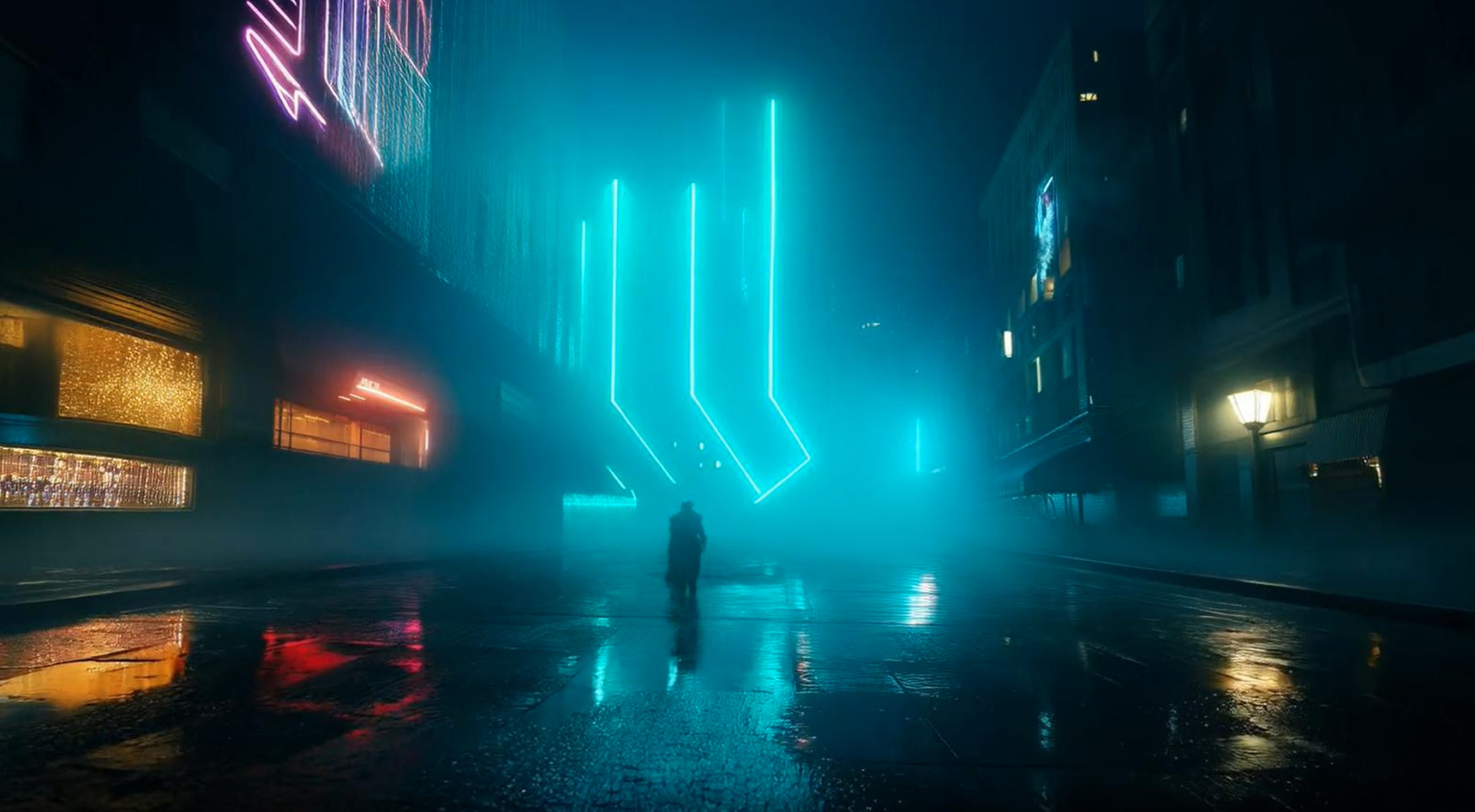} \hfill
            \includegraphics[width=0.195\textwidth]{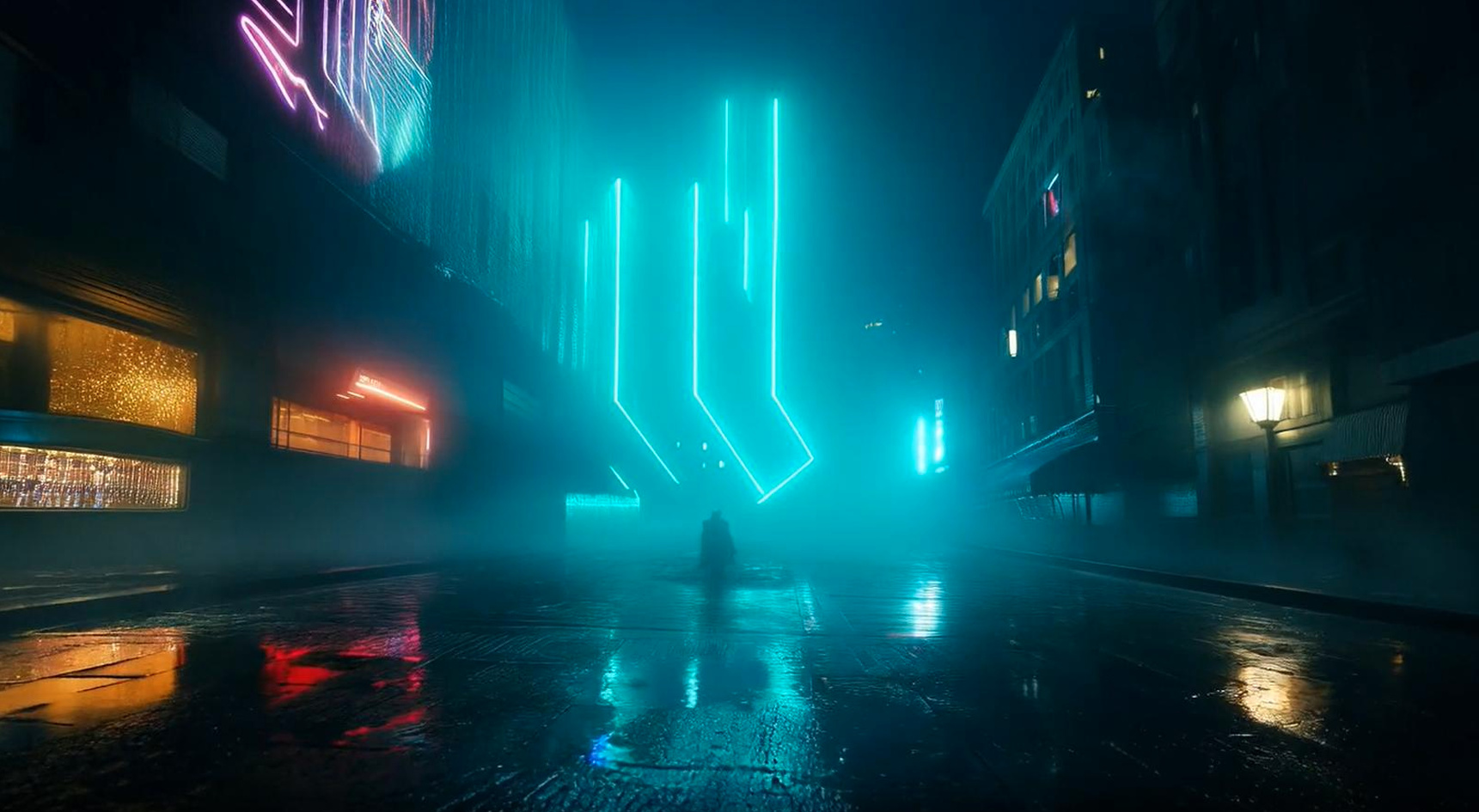} \hfill
            \includegraphics[width=0.195\textwidth]{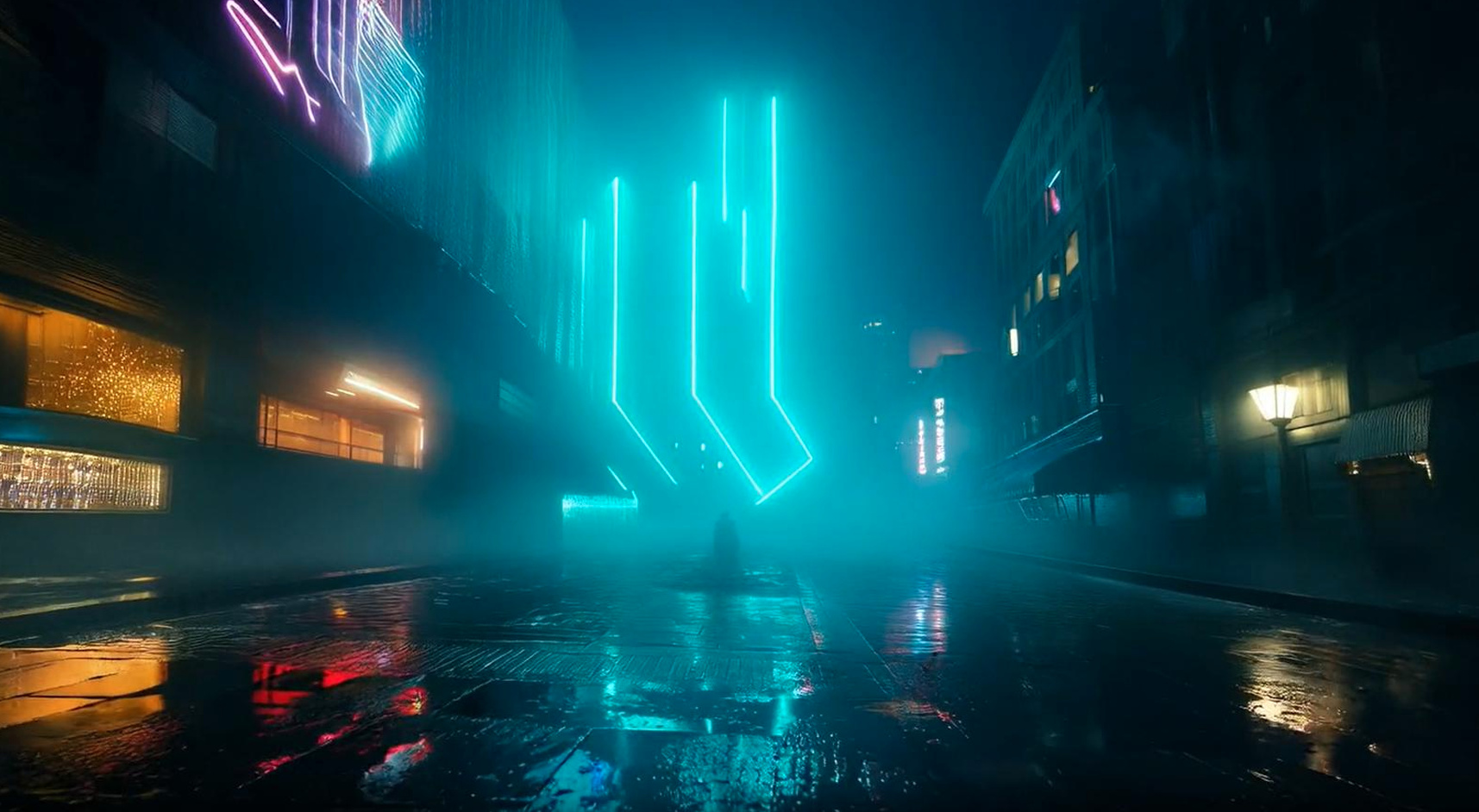} \hfill
            \includegraphics[width=0.195\textwidth]{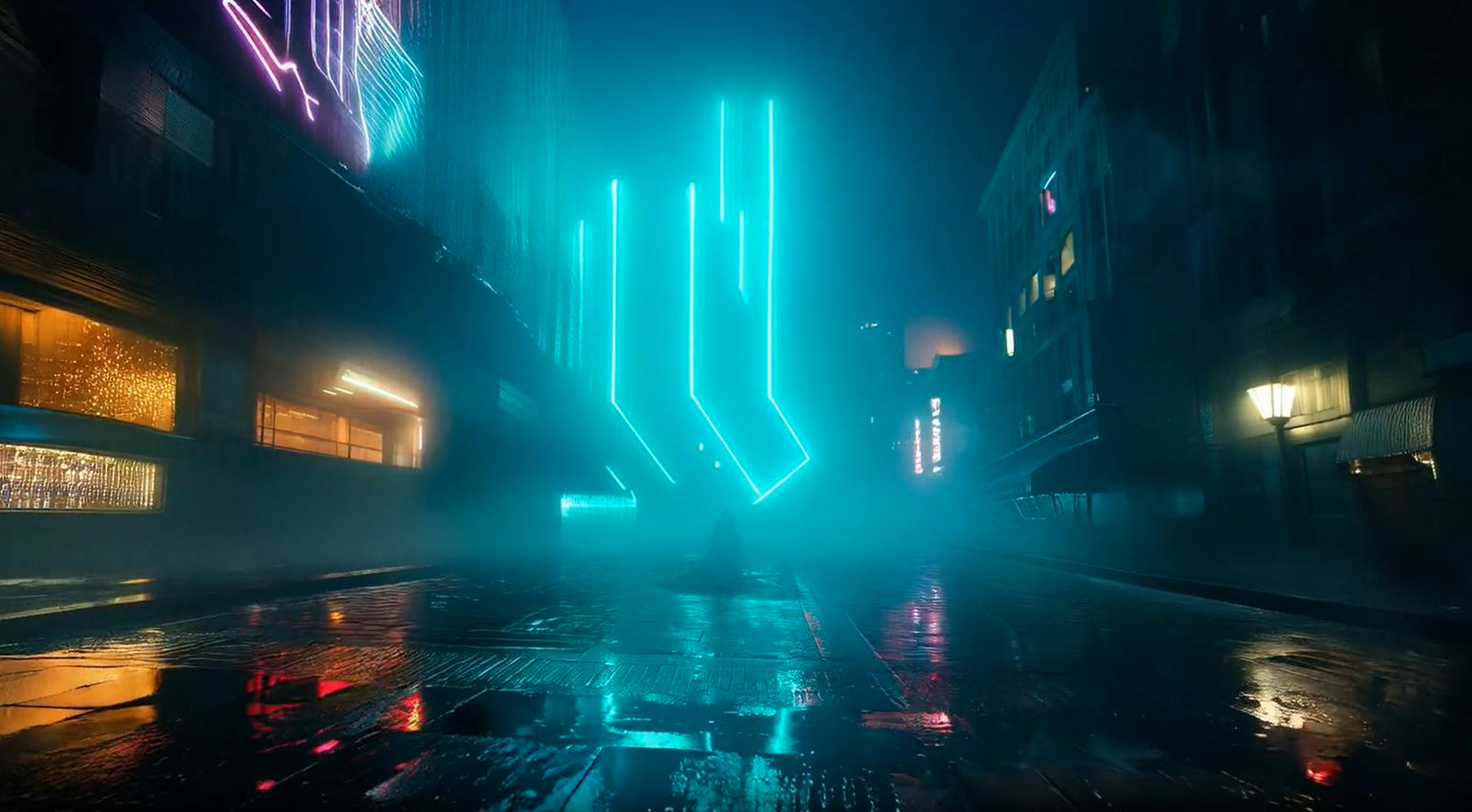}
        \end{subfigure}
    \end{minipage}
    \label{fig:wan_gallery_2}
\end{figure}

\vspace{30pt}

\subsection{Results on HunyuanVideo}

\begin{figure}[H]
    \centering
    \setlength{\tabcolsep}{1pt} \renewcommand{\arraystretch}{0.1}
    
    \begin{minipage}{\textwidth}
        \fcolorbox{gray!50}{gray!10}{
            \parbox{0.97\linewidth}{
                \vspace{2pt}
                \textbf{\small Case 5 (Portrait/Low Motion):} \small \textit{“A cinematic close-up of an elderly craftsman carving wood, detailed wrinkles, focused expression, warm workshop lighting, 4k, slow camera movement.”}
                \vspace{2pt}
            }
        }
        \vspace{3pt}
        \begin{subfigure}{\textwidth}
        \vspace{3pt}
            \textbf{\scriptsize Original (Dense):} \\
            \includegraphics[width=0.195\textwidth]{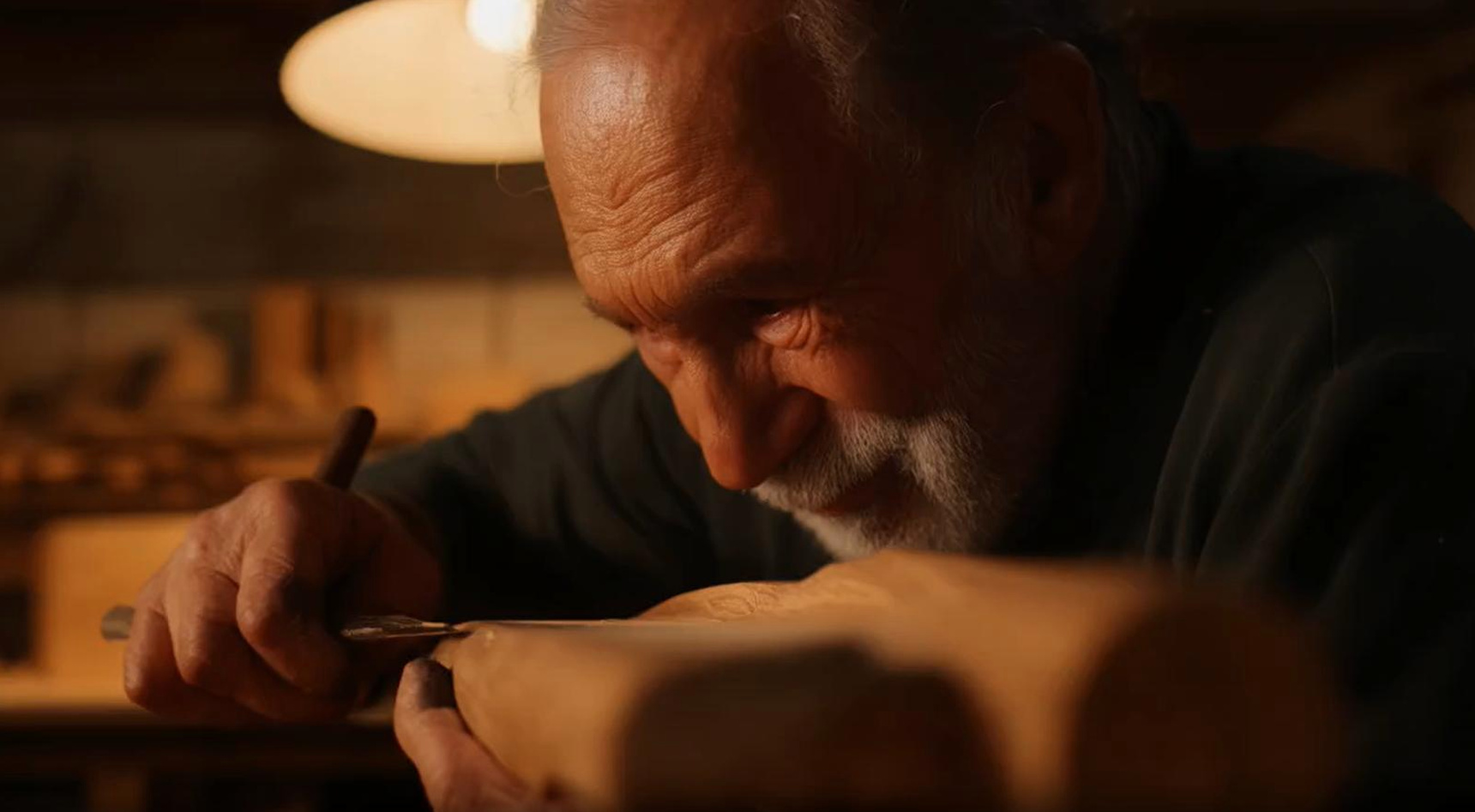} \hfill
            \includegraphics[width=0.195\textwidth]{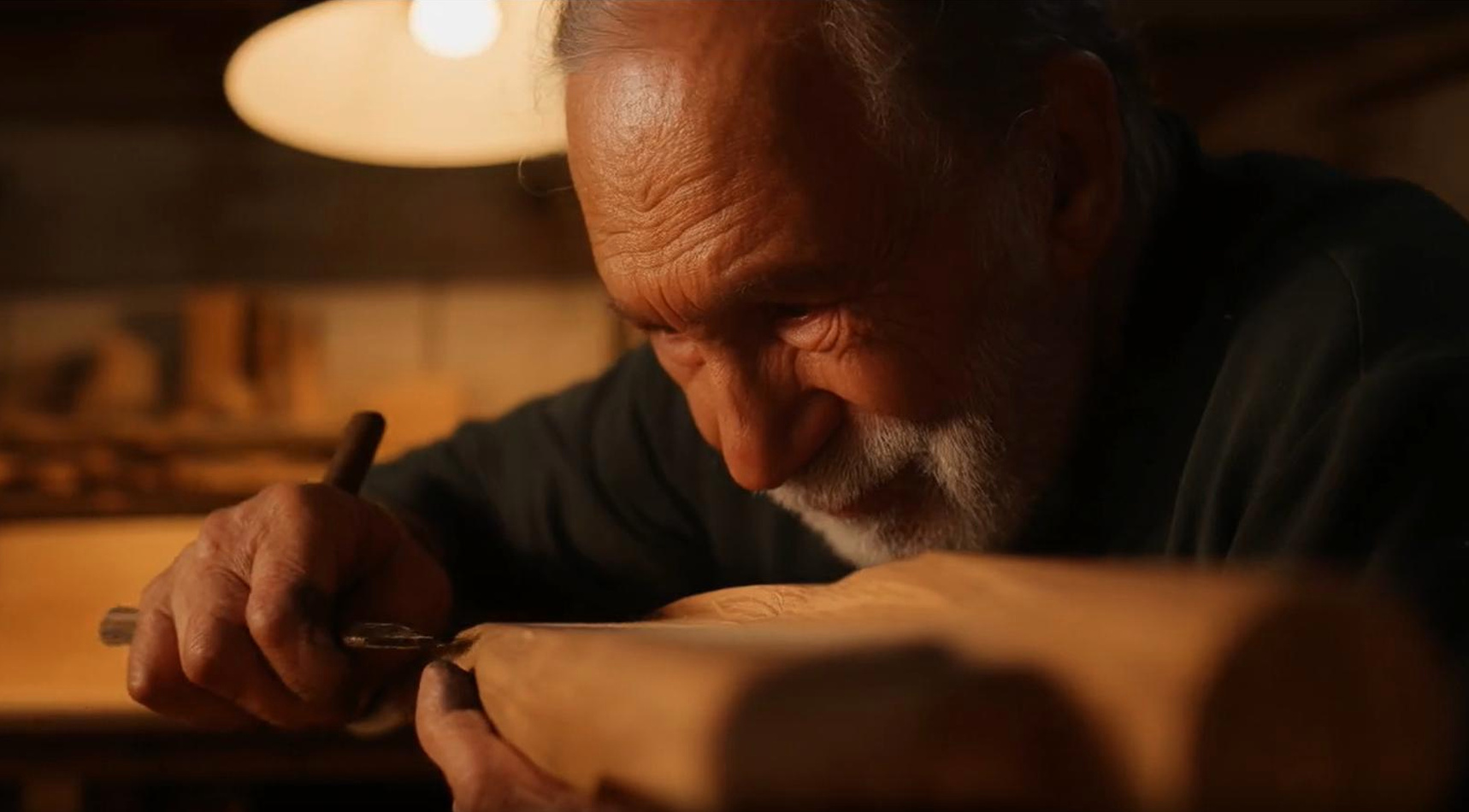} \hfill
            \includegraphics[width=0.195\textwidth]{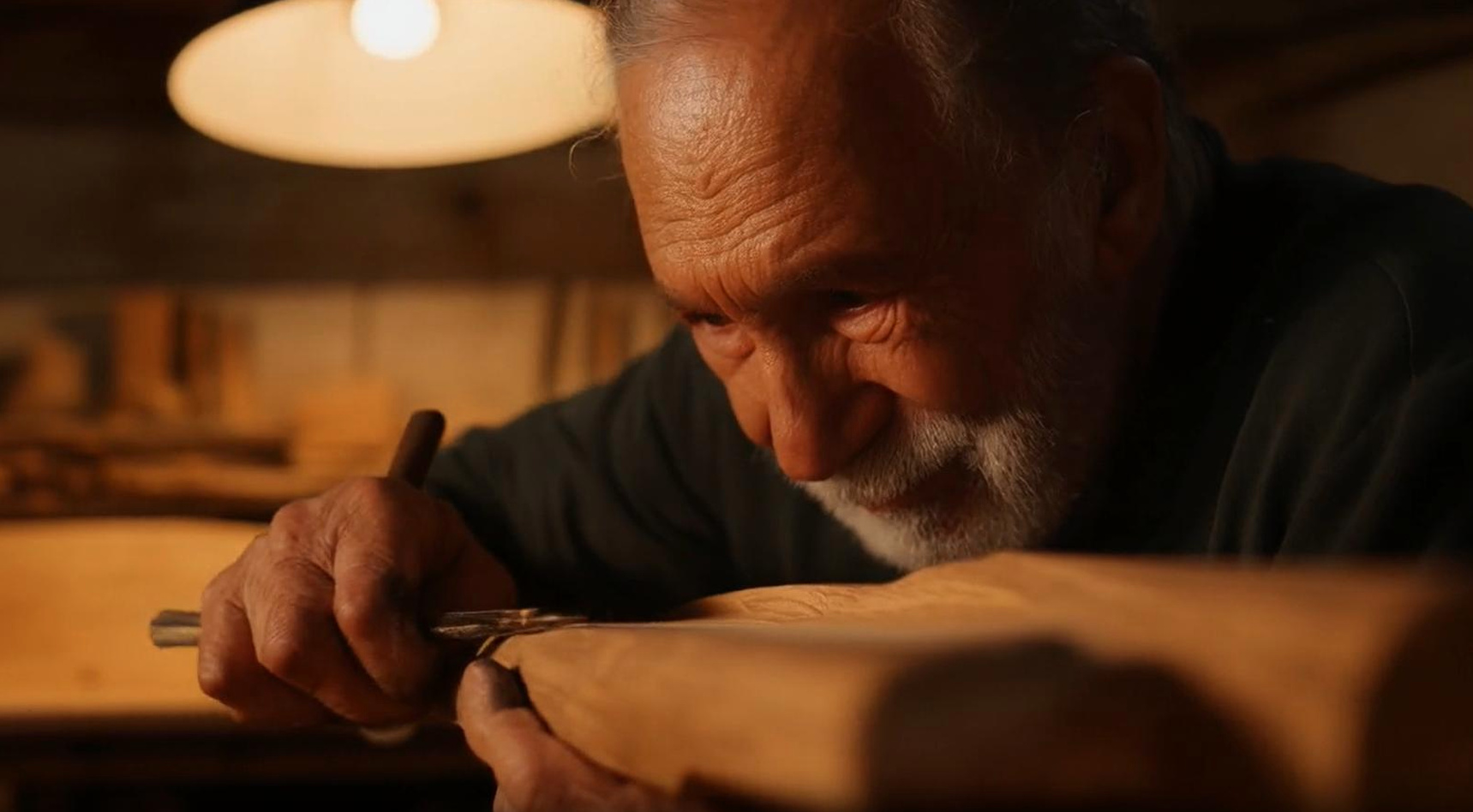} \hfill
            \includegraphics[width=0.195\textwidth]{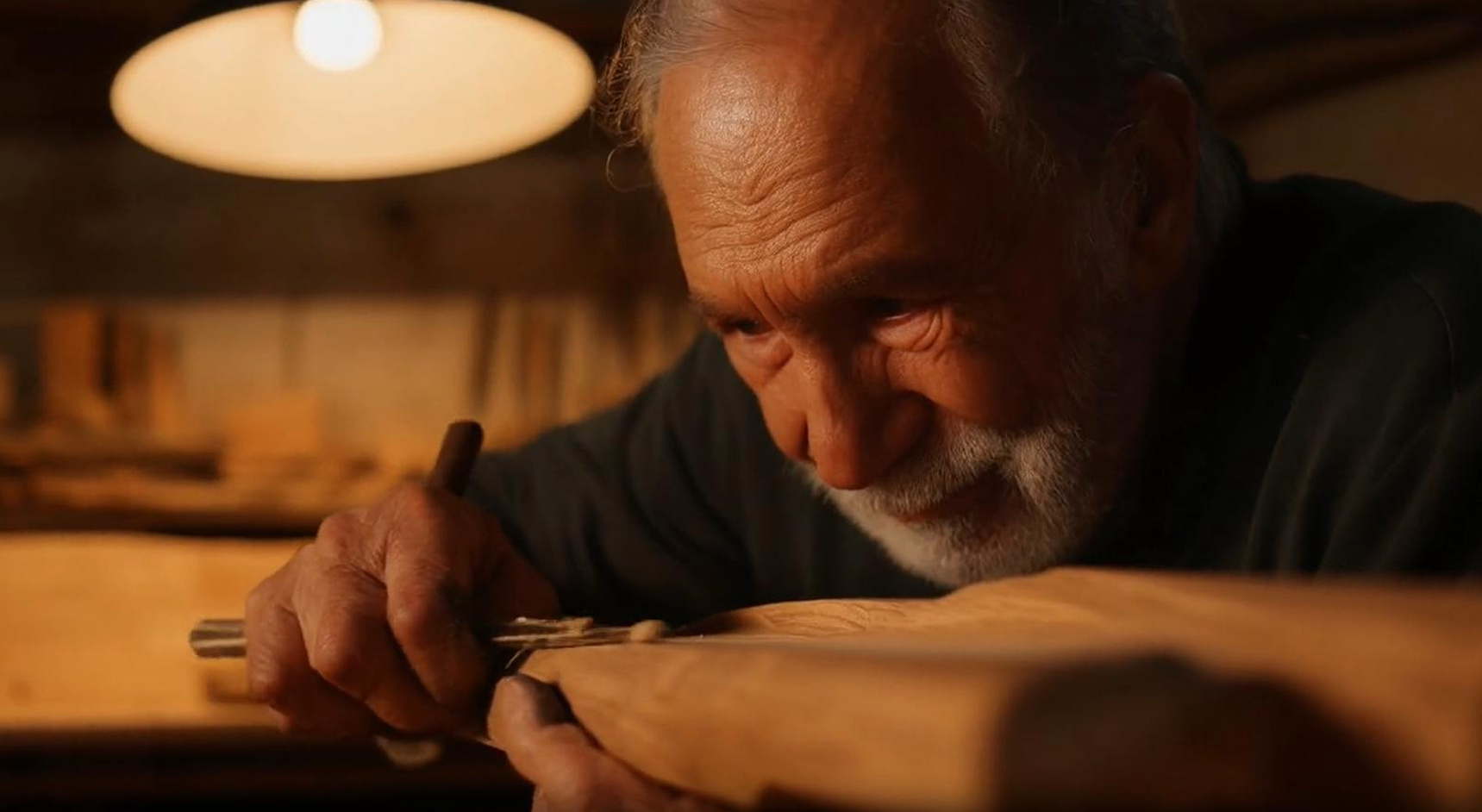} \hfill
            \includegraphics[width=0.195\textwidth]{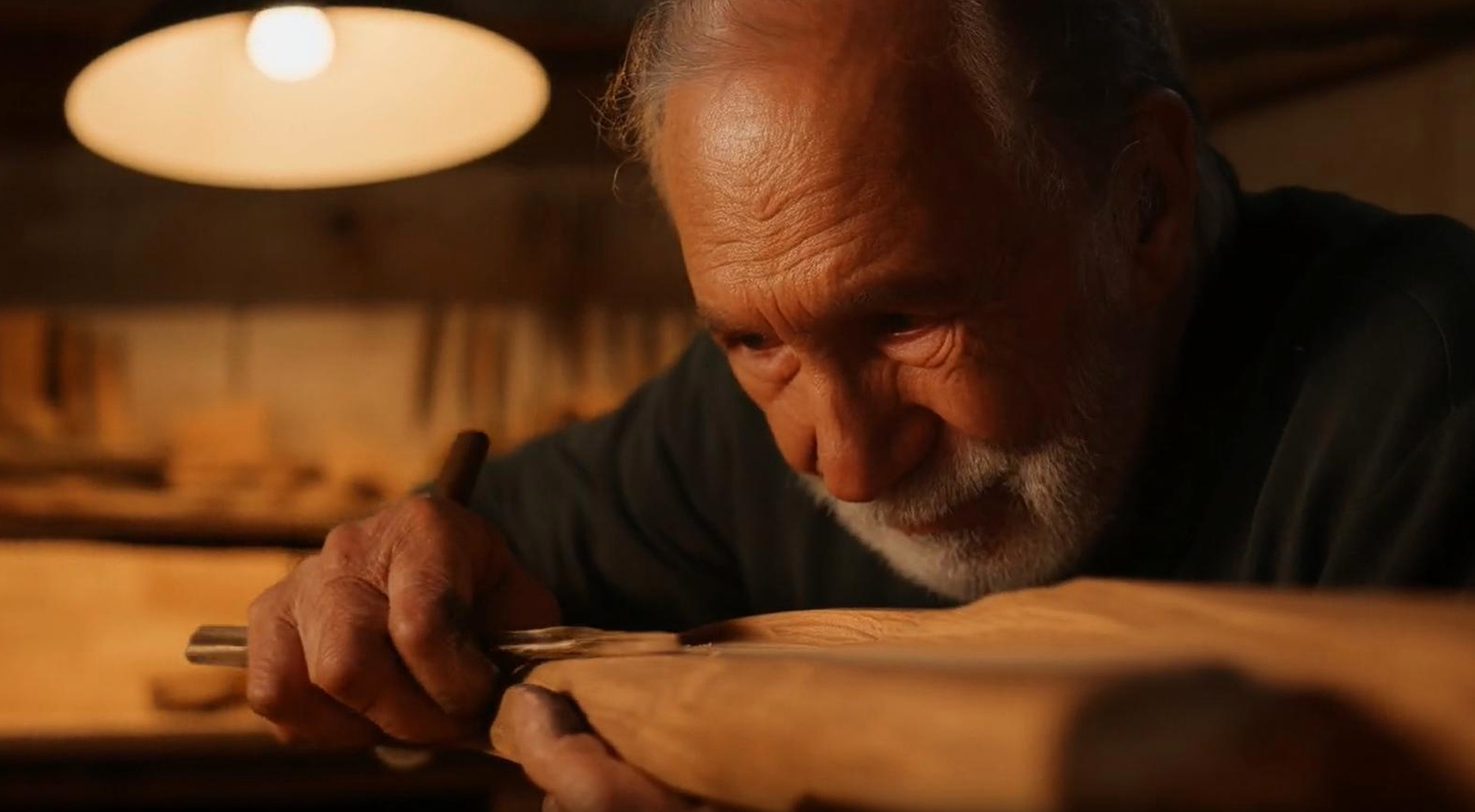}
        \end{subfigure}
        \vspace{1pt}
        \begin{subfigure}{\textwidth}
            \textbf{\scriptsize DynamicRad (Ours):} \\
            \includegraphics[width=0.195\textwidth]{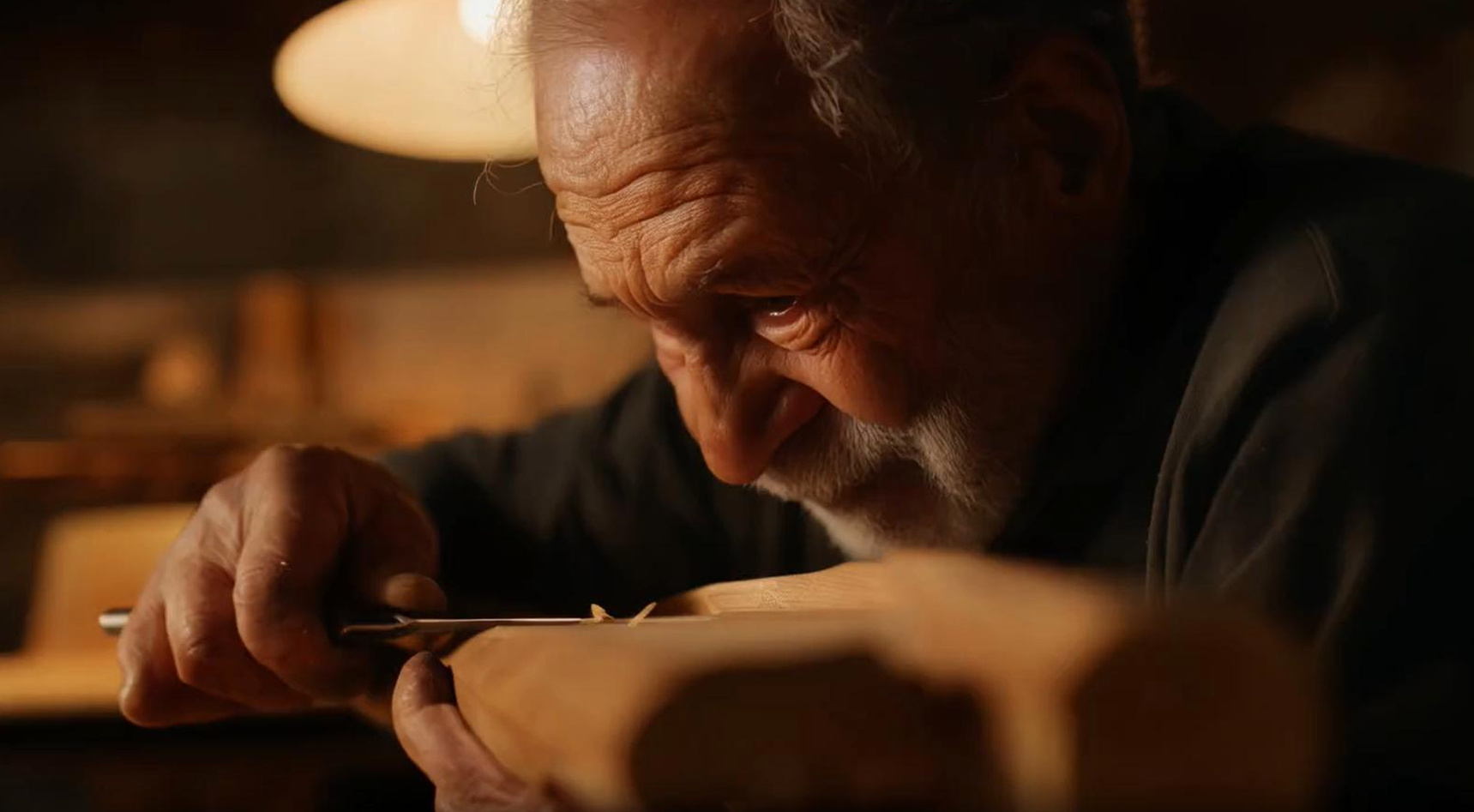} \hfill
            \includegraphics[width=0.195\textwidth]{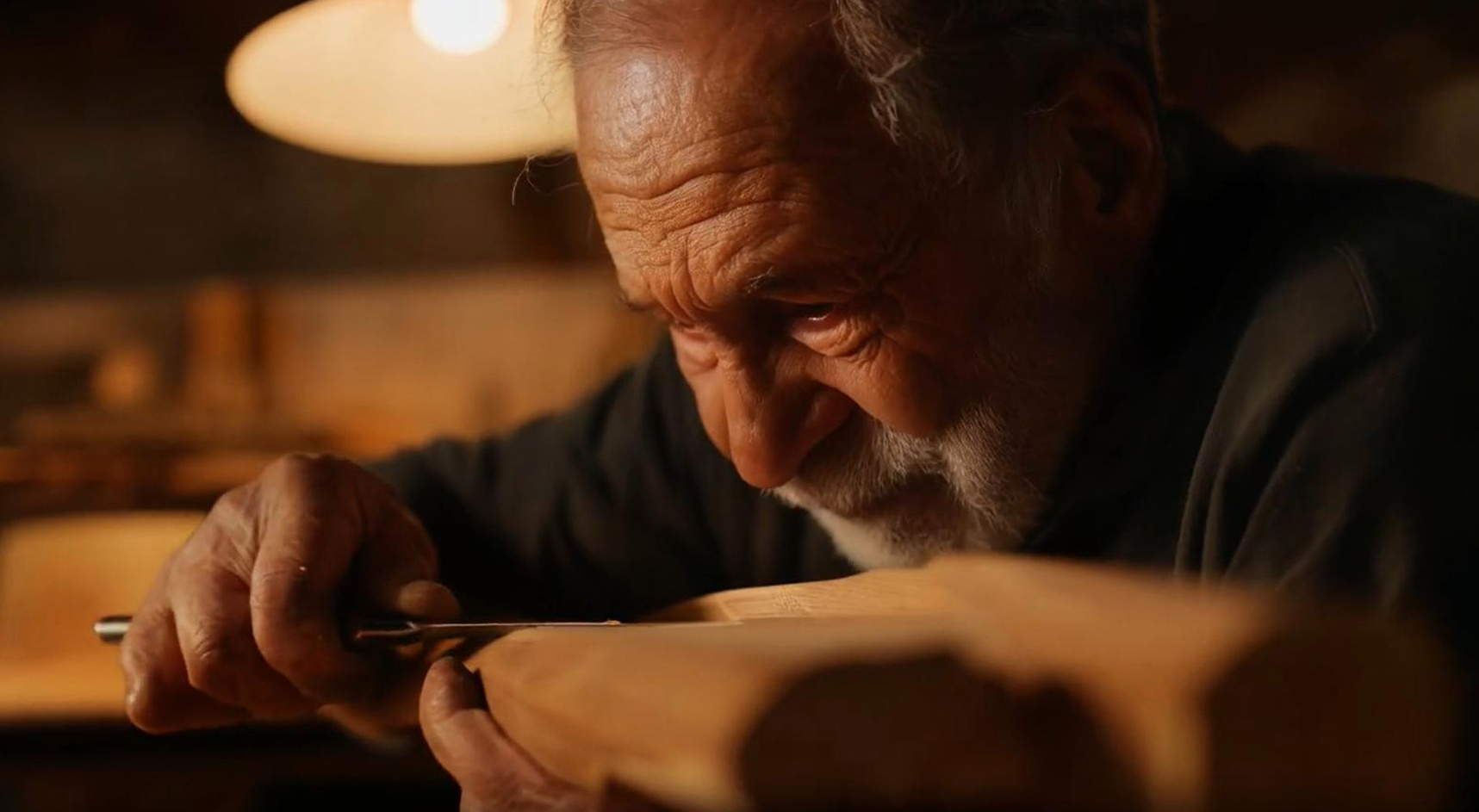} \hfill
            \includegraphics[width=0.195\textwidth]{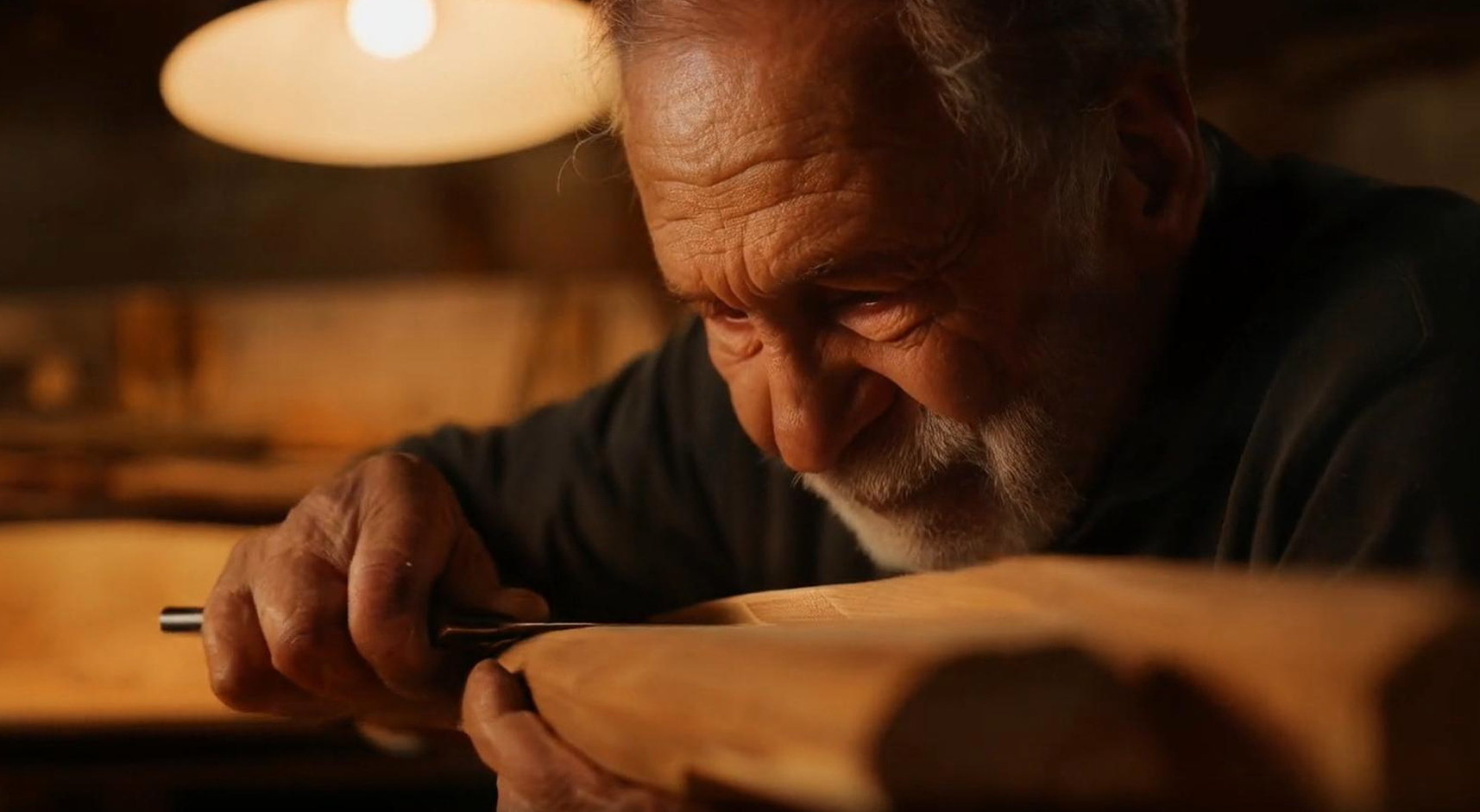} \hfill
            \includegraphics[width=0.195\textwidth]{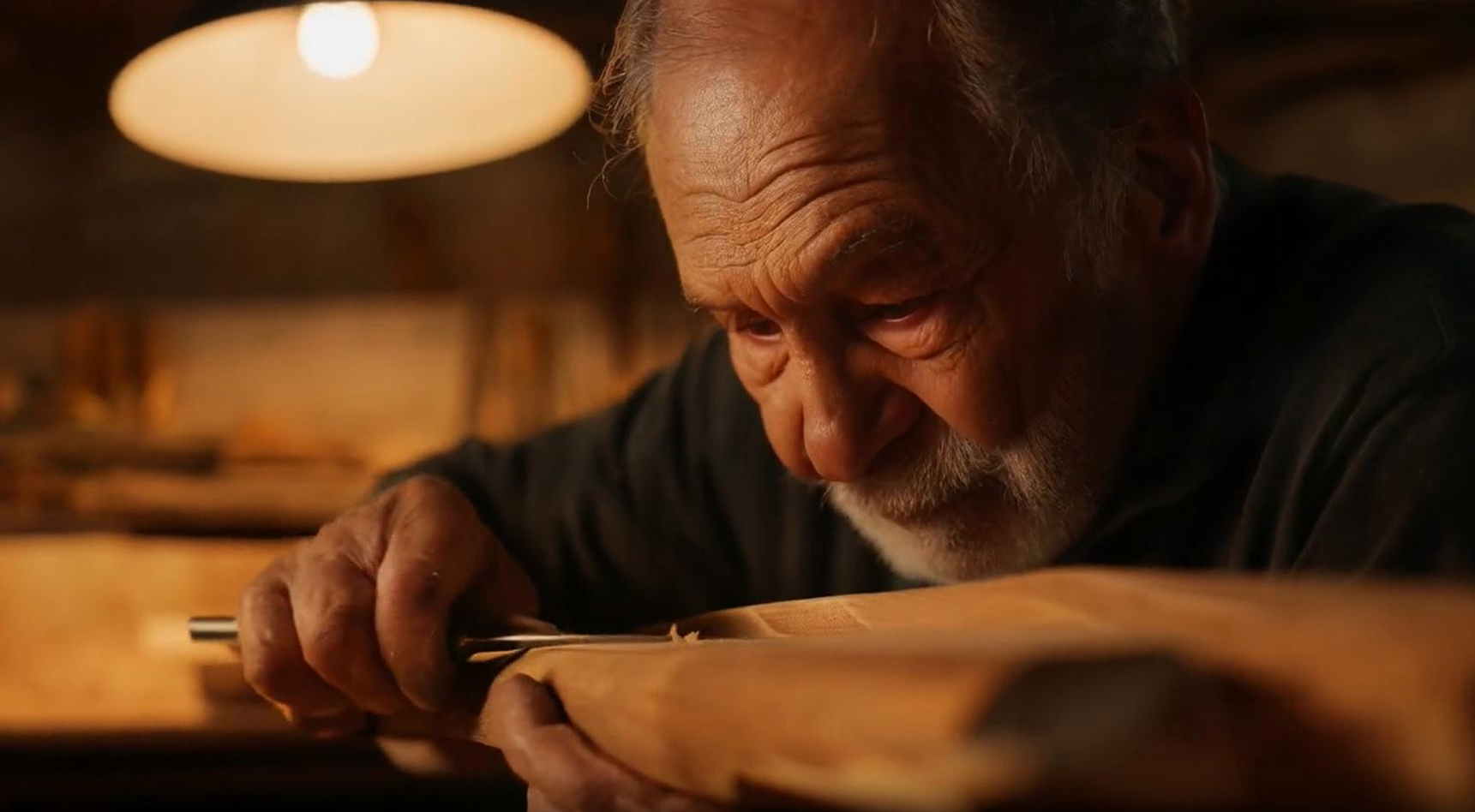} \hfill
            \includegraphics[width=0.195\textwidth]{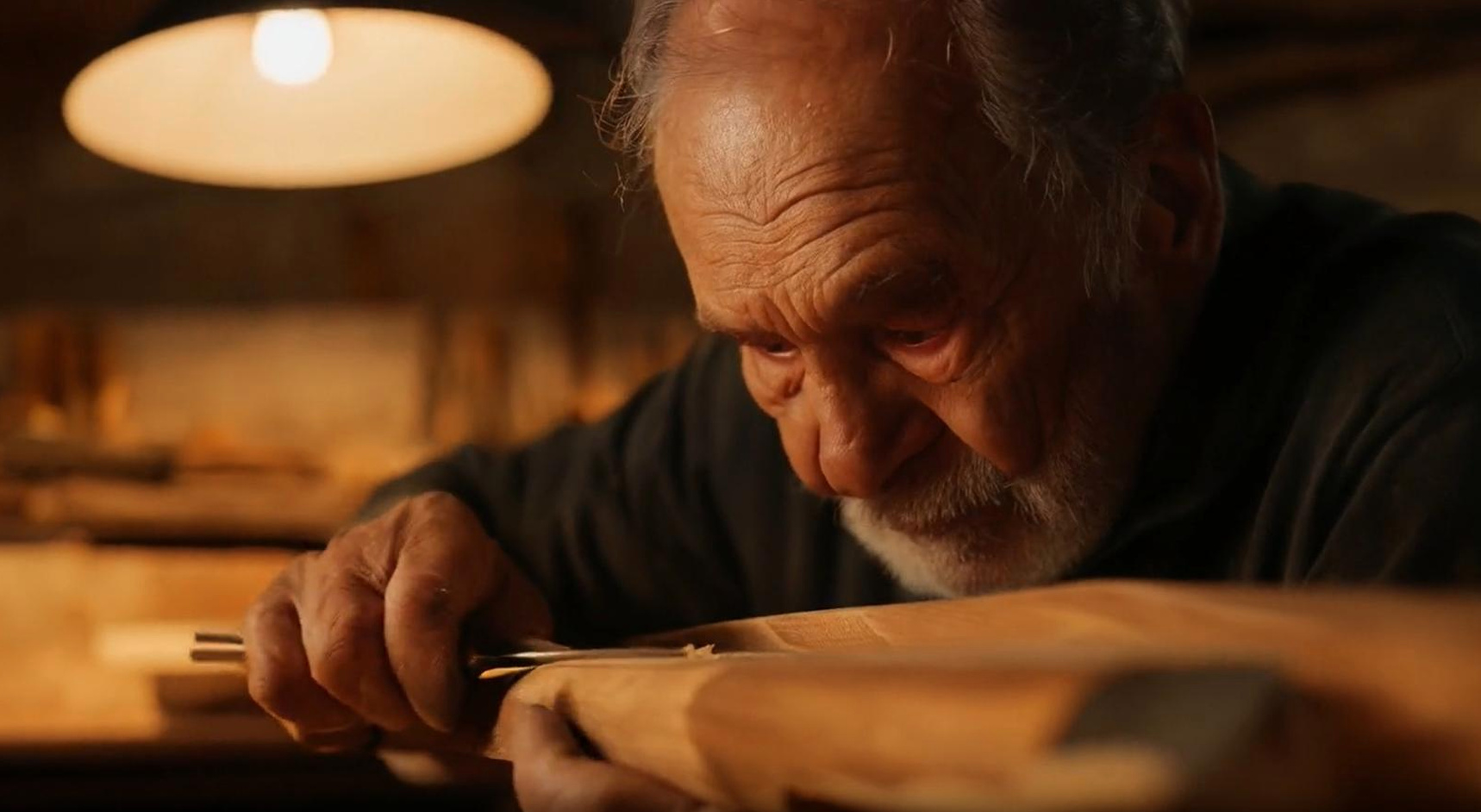}
        \end{subfigure}
    \end{minipage}
    \vspace{20pt}

    \begin{minipage}{\textwidth}
        \fcolorbox{gray!50}{gray!10}{
            \parbox{0.97\linewidth}{
                \vspace{2pt}
                \textbf{\small Case 6 (High Motion/Water):} \small \textit{“A professional surfer riding a massive blue wave, water spray, dynamic camera tracking, high speed, sunlight reflection, 4k.”}
                \vspace{2pt}
            }
        }
        \vspace{3pt}
        \begin{subfigure}{\textwidth}
        \vspace{3pt}
            \textbf{\scriptsize Original (Dense):} \\
            \includegraphics[width=0.195\textwidth]{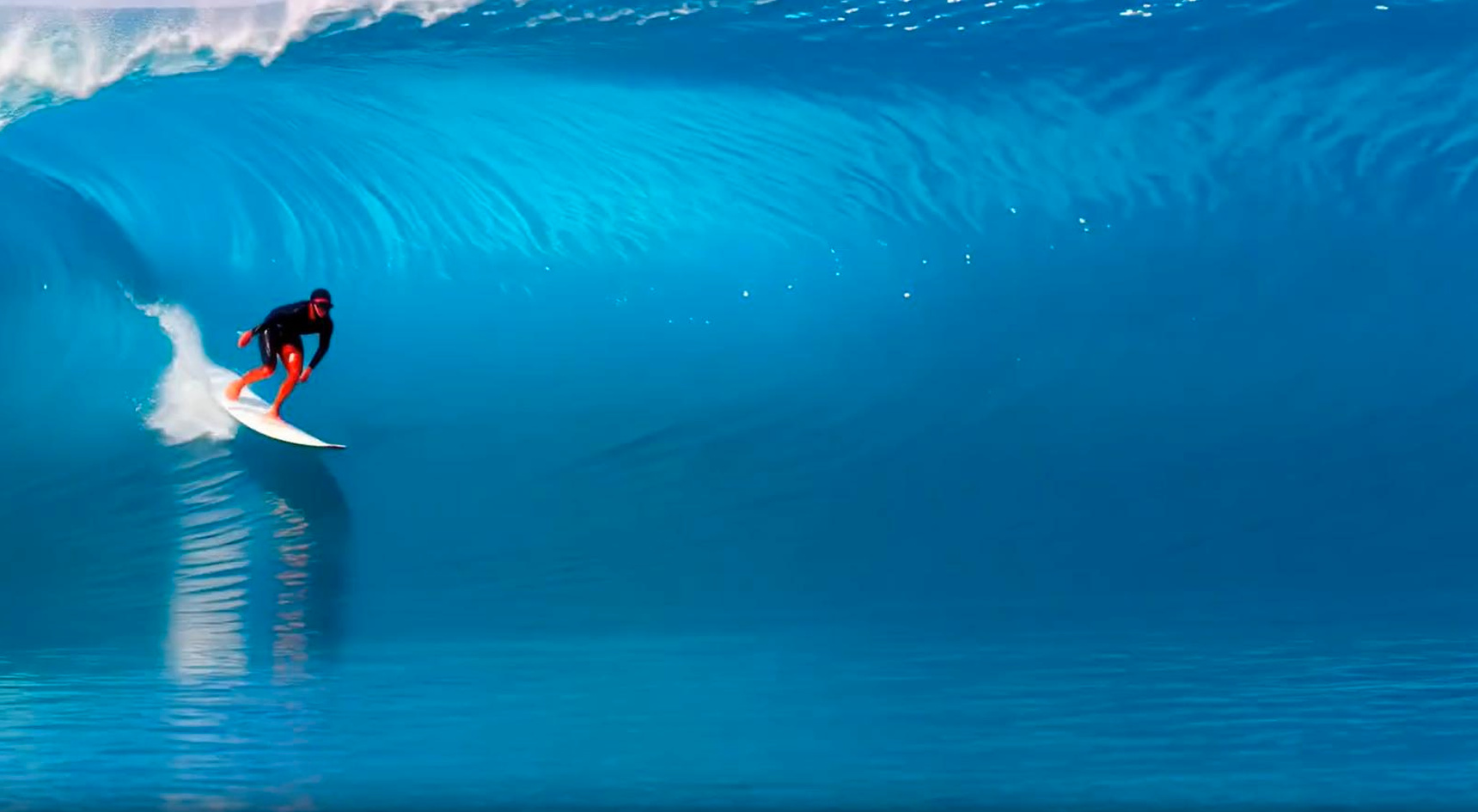} \hfill
            \includegraphics[width=0.195\textwidth]{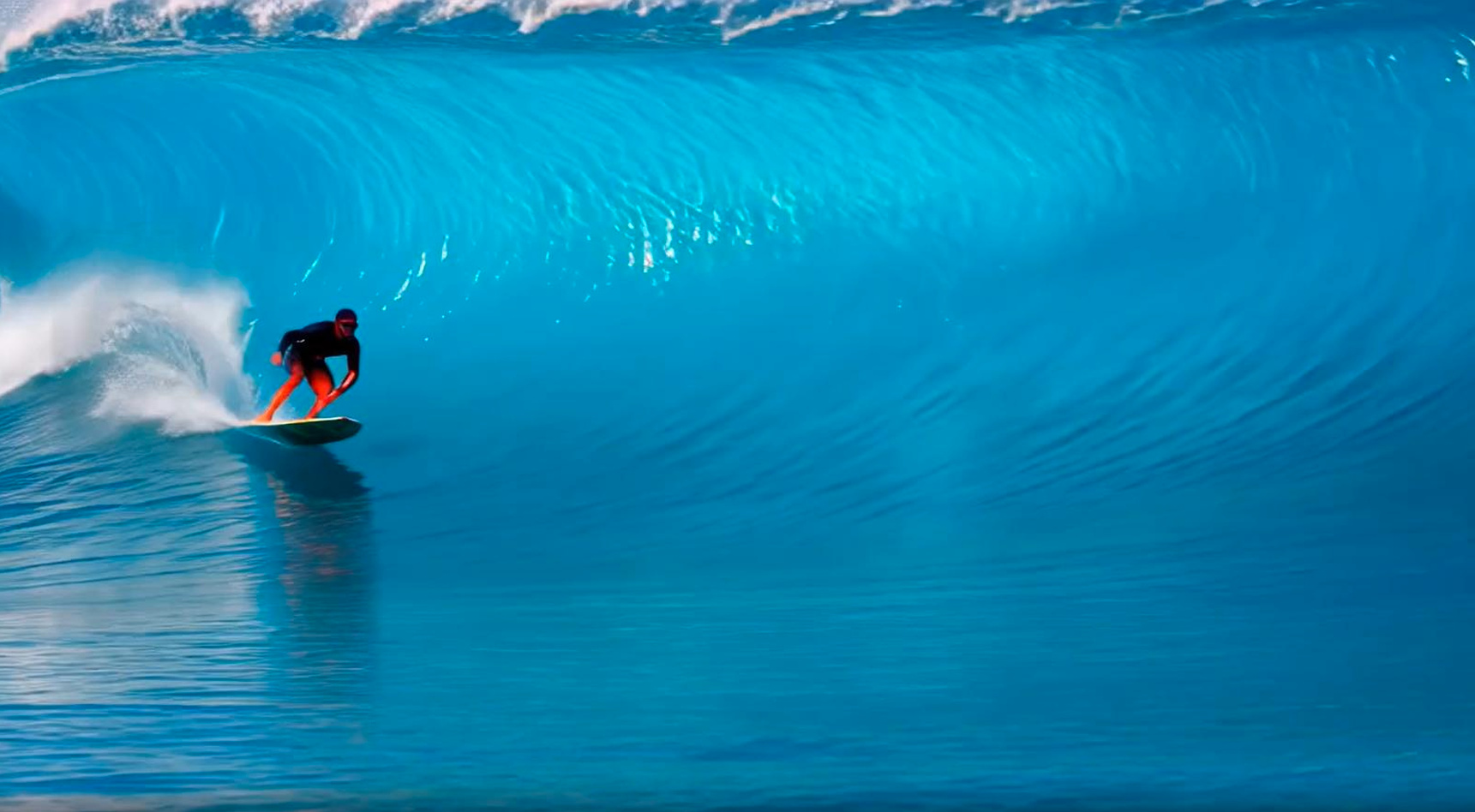} \hfill
            \includegraphics[width=0.195\textwidth]{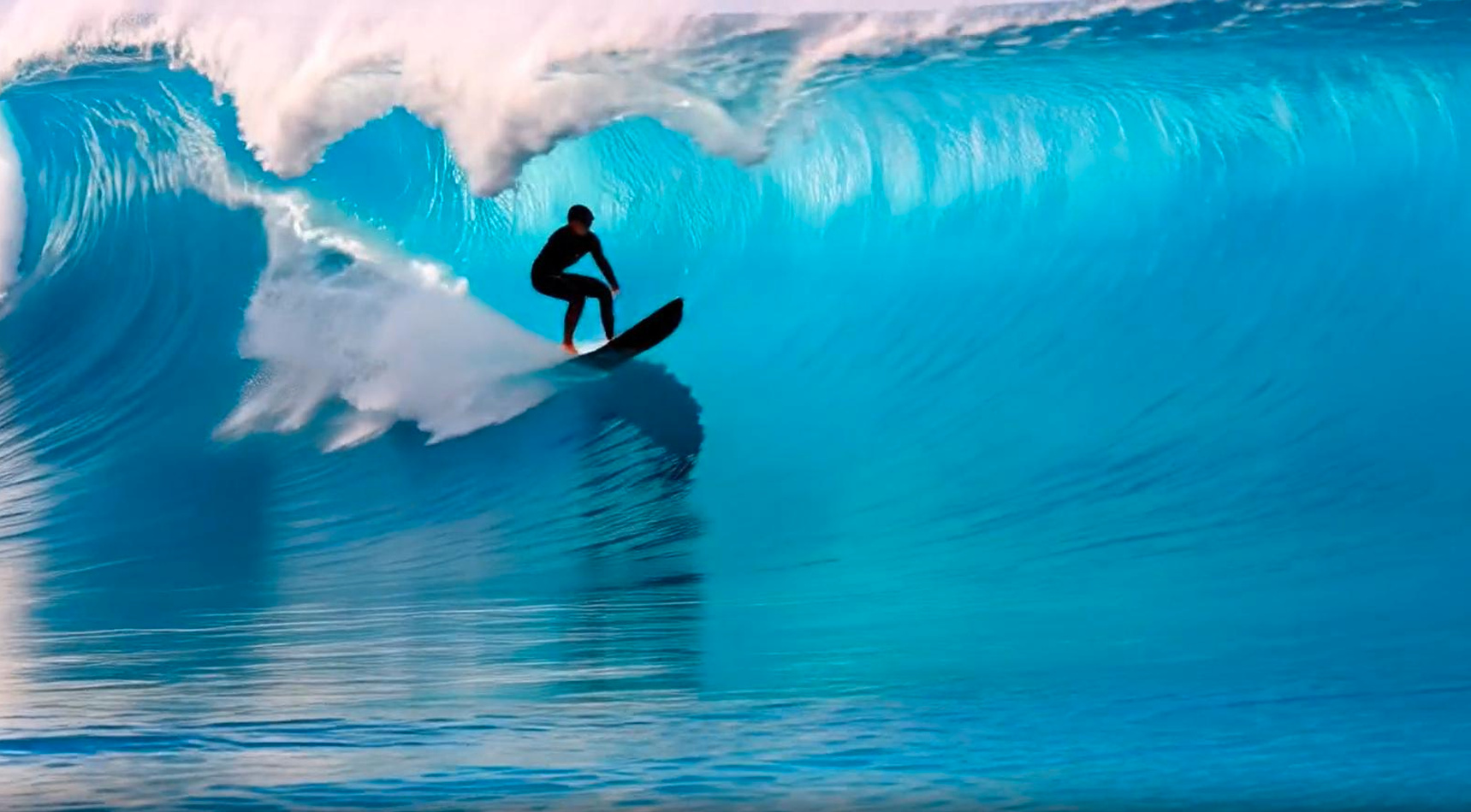} \hfill
            \includegraphics[width=0.195\textwidth]{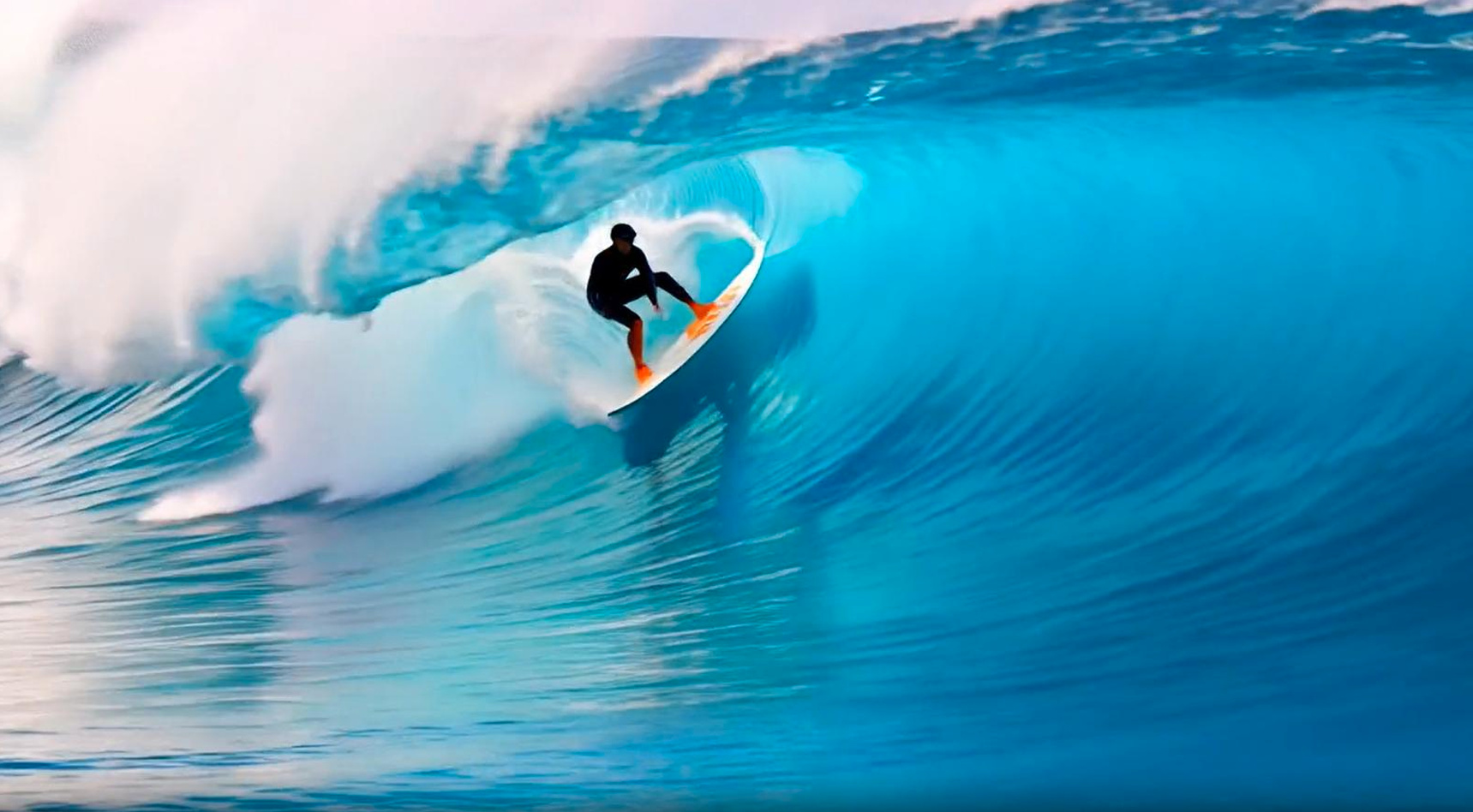} \hfill
            \includegraphics[width=0.195\textwidth]{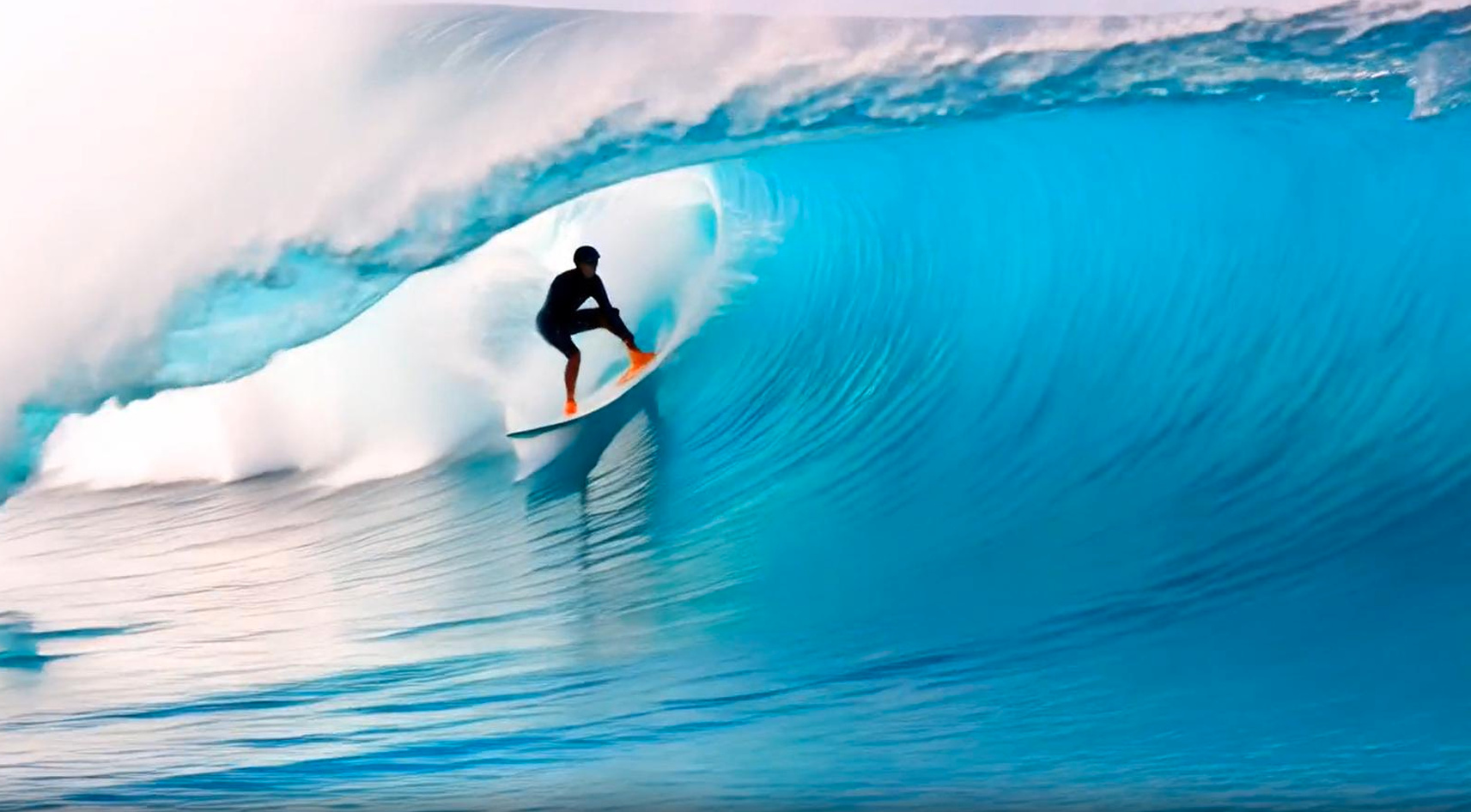}
        \end{subfigure}
        \vspace{1pt}
        \begin{subfigure}{\textwidth}
            \textbf{\scriptsize DynamicRad (Ours):} \\
            \includegraphics[width=0.195\textwidth]{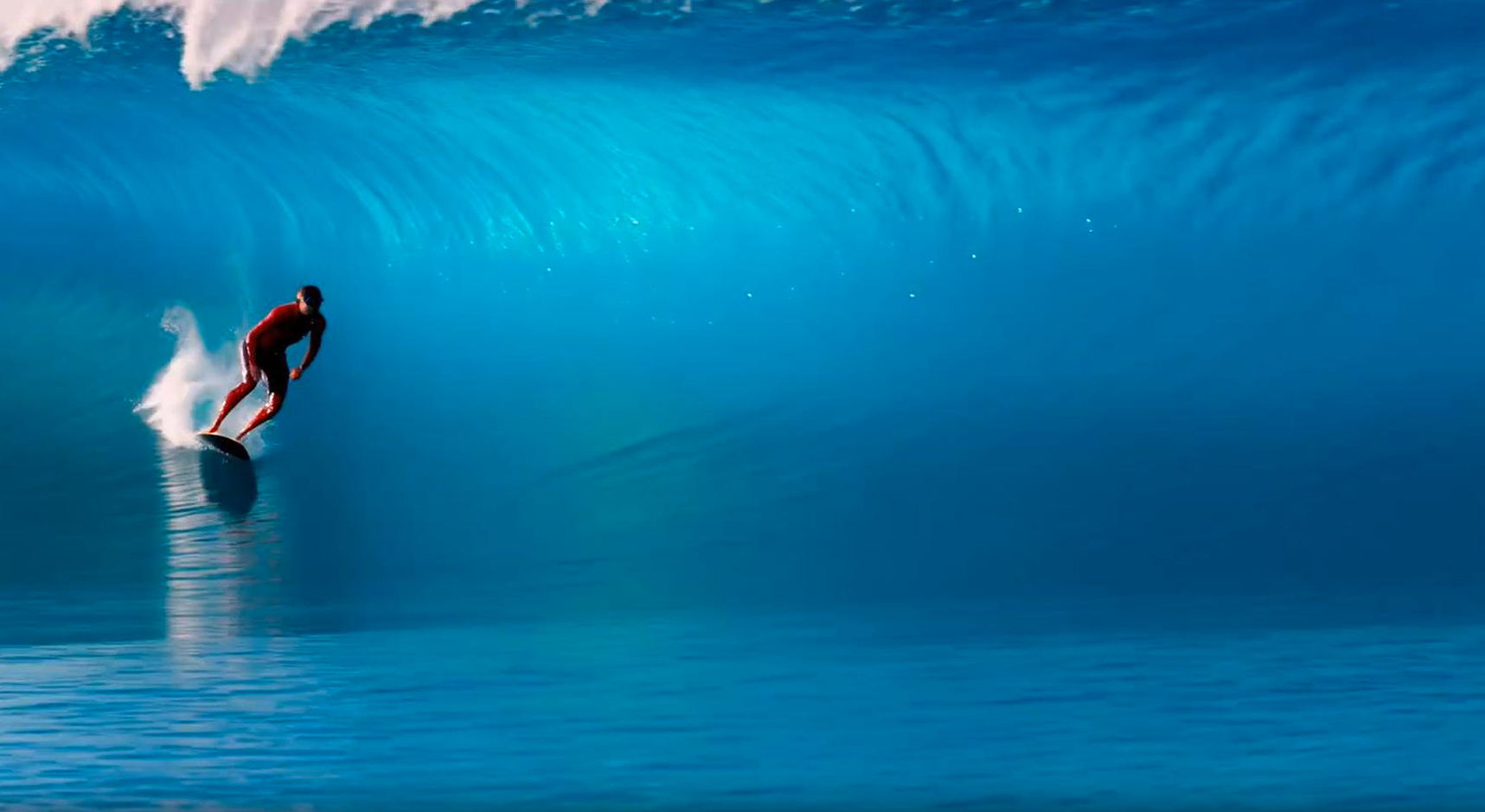} \hfill
            \includegraphics[width=0.195\textwidth]{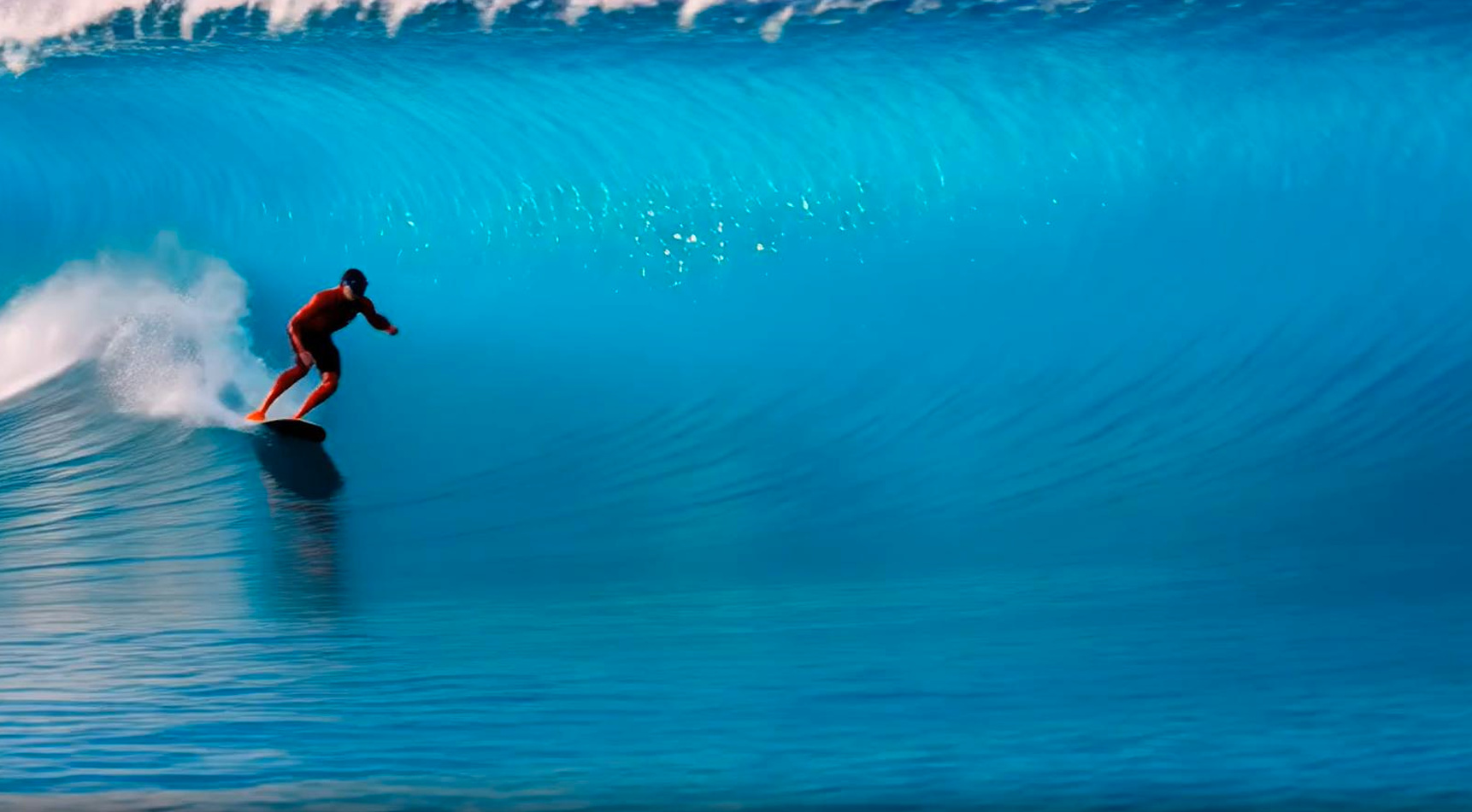} \hfill
            \includegraphics[width=0.195\textwidth]{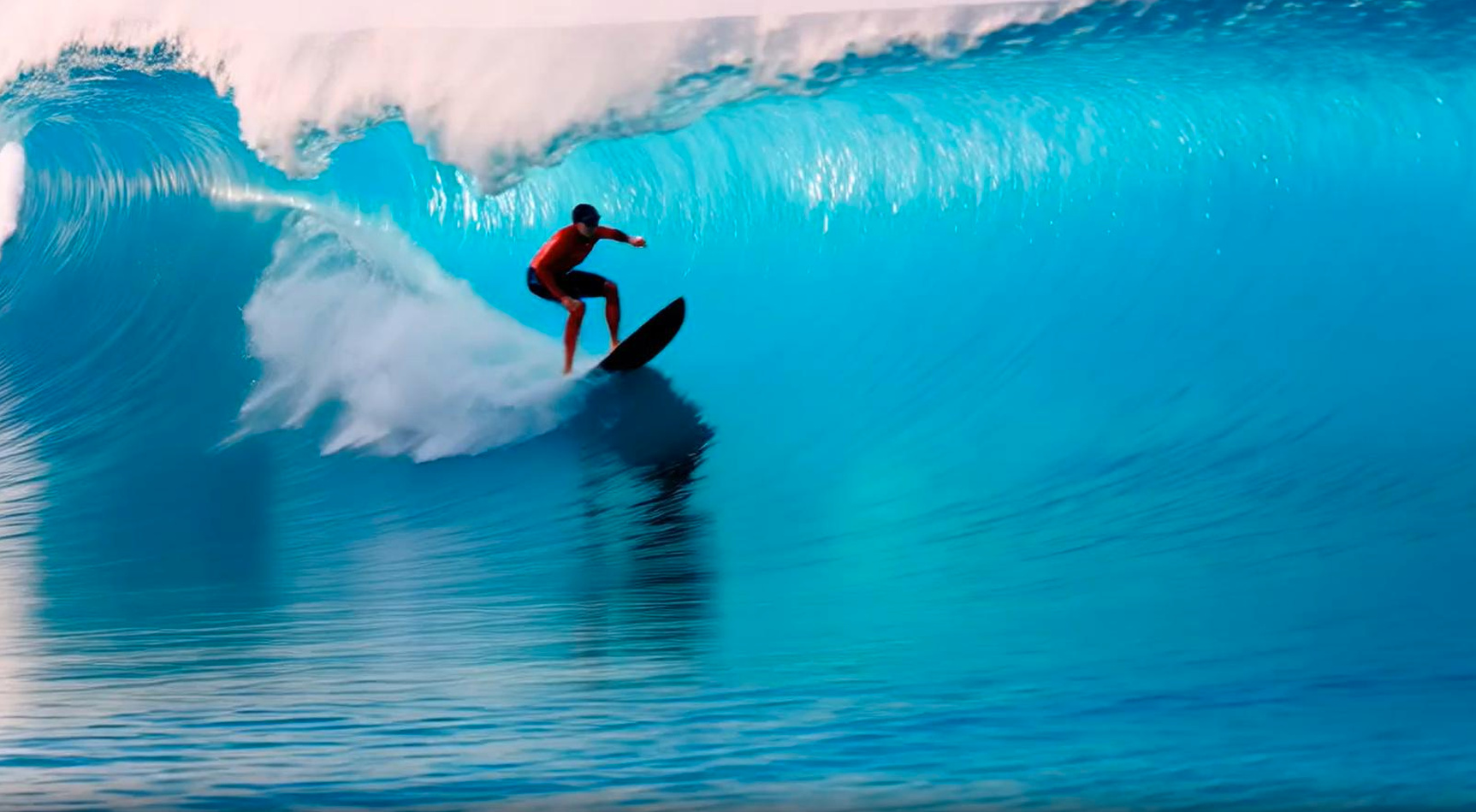} \hfill
            \includegraphics[width=0.195\textwidth]{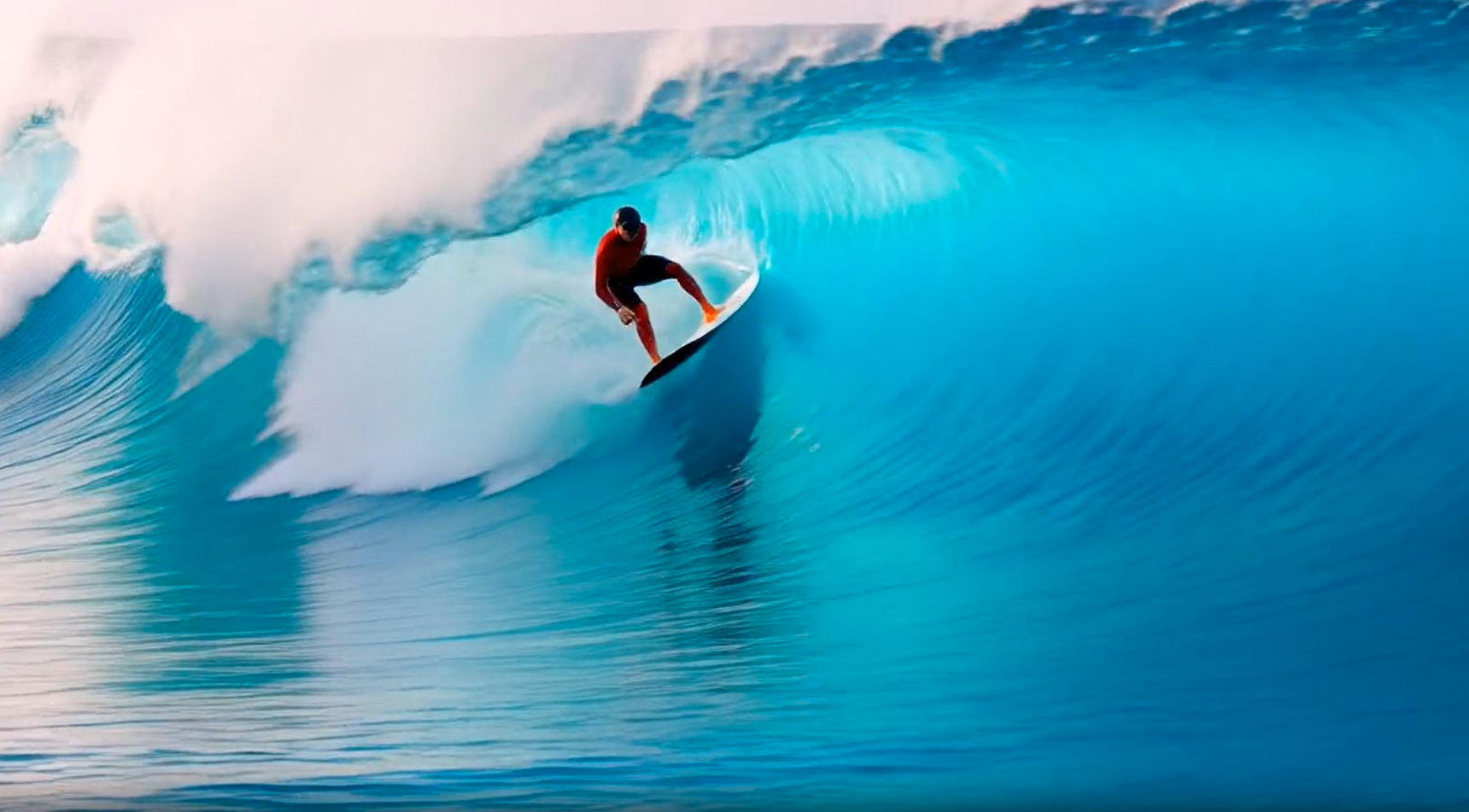} \hfill
            \includegraphics[width=0.195\textwidth]{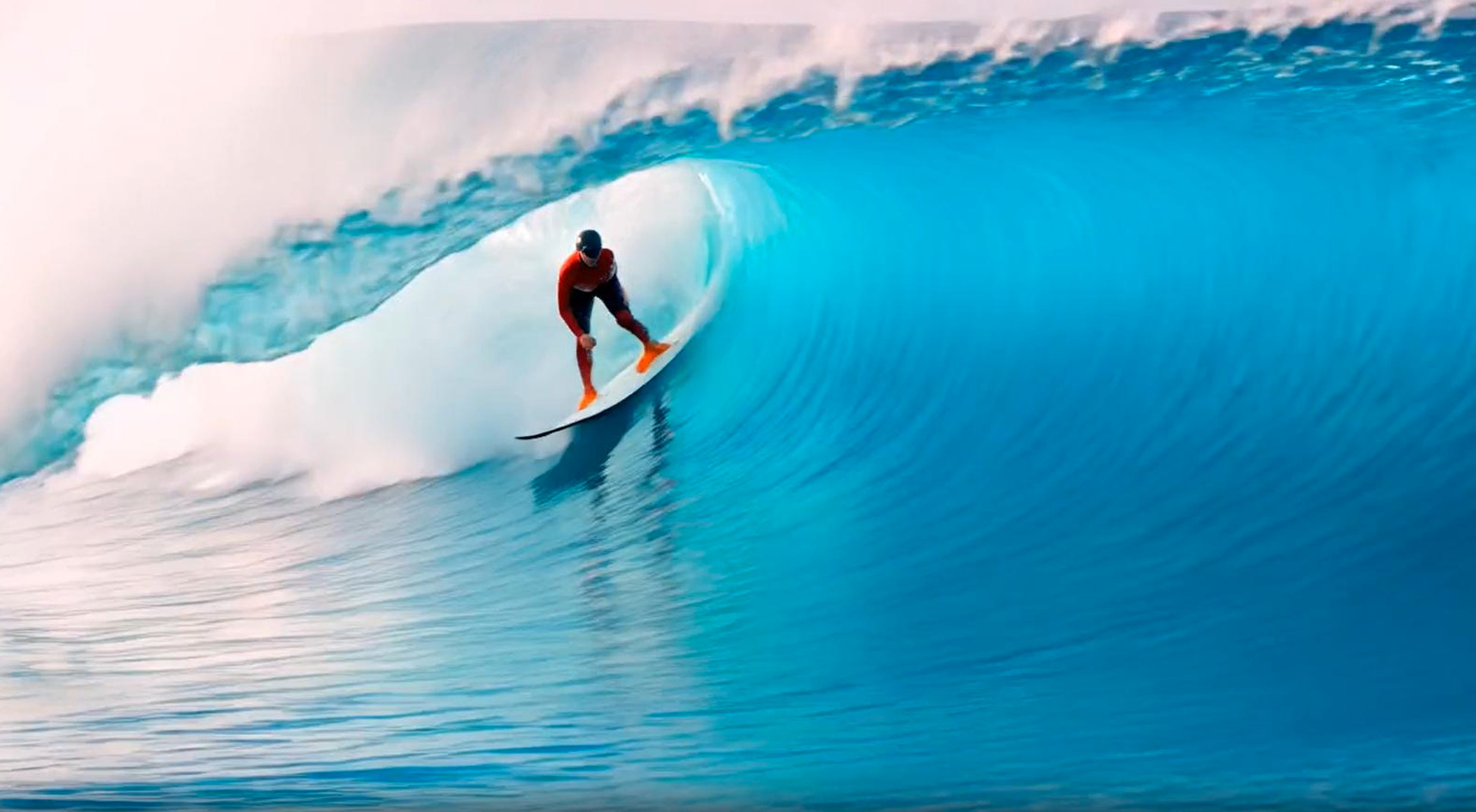}
        \end{subfigure}
    \end{minipage}

    \label{fig:hunyuan_gallery_1}
\end{figure}

\clearpage

\begin{figure}[H]
    \centering
    \setlength{\tabcolsep}{1pt} \renewcommand{\arraystretch}{0.1}

    \begin{minipage}{\textwidth}
        \fcolorbox{gray!50}{gray!10}{
            \parbox{0.97\linewidth}{
                \vspace{2pt}
                \textbf{\small Case 7 (Sci-Fi/Texture):} \small \textit{“A spaceship fleet engaging in a battle near a nebula, lasers firing, explosions, colorful gas clouds, high detail, sci-fi movie style.”}
                \vspace{2pt}
            }
        }
        \vspace{3pt}
        \begin{subfigure}{\textwidth}
        \vspace{3pt}
            \textbf{\scriptsize Original (Dense):} \\
            \includegraphics[width=0.195\textwidth]{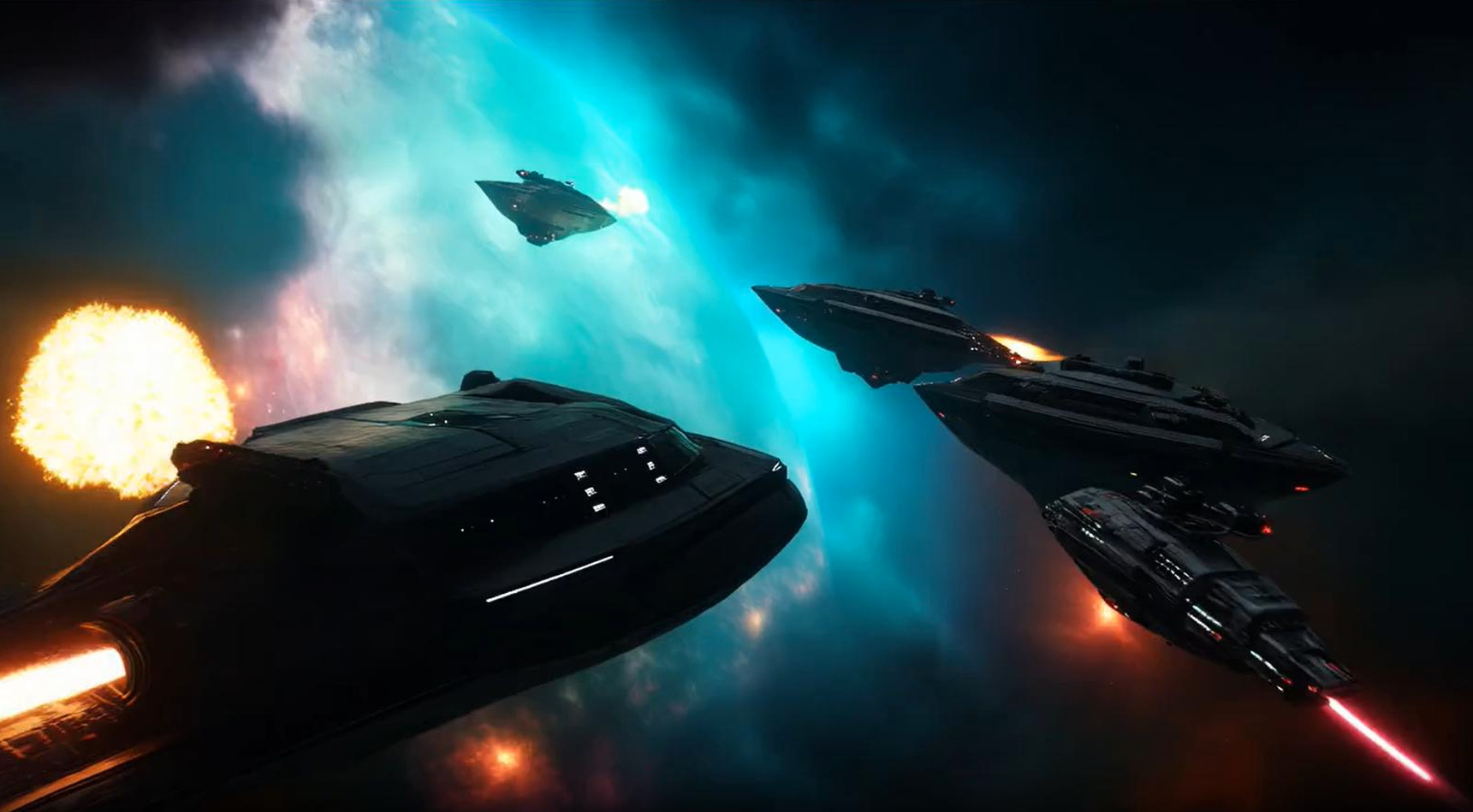} \hfill
            \includegraphics[width=0.195\textwidth]{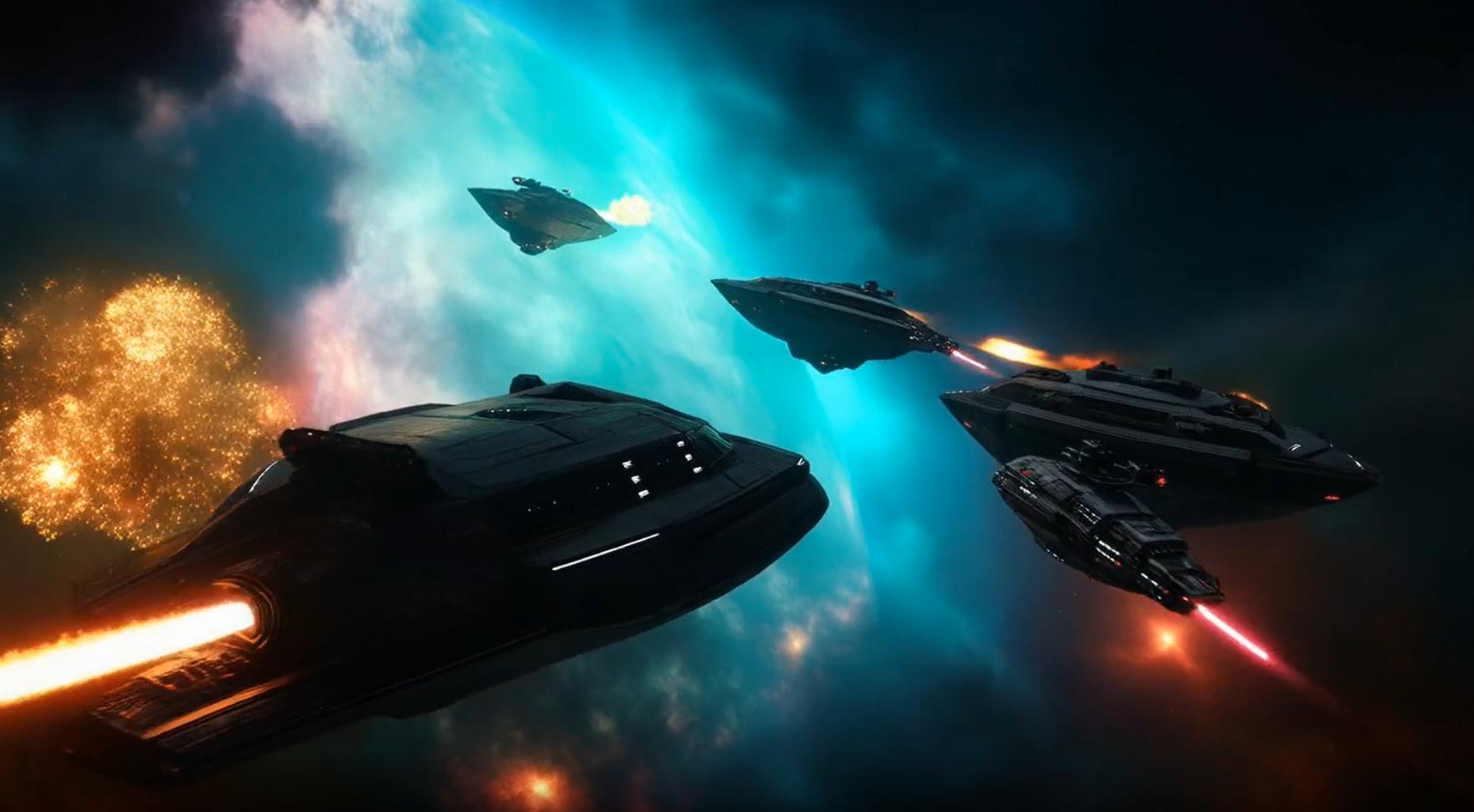} \hfill
            \includegraphics[width=0.195\textwidth]{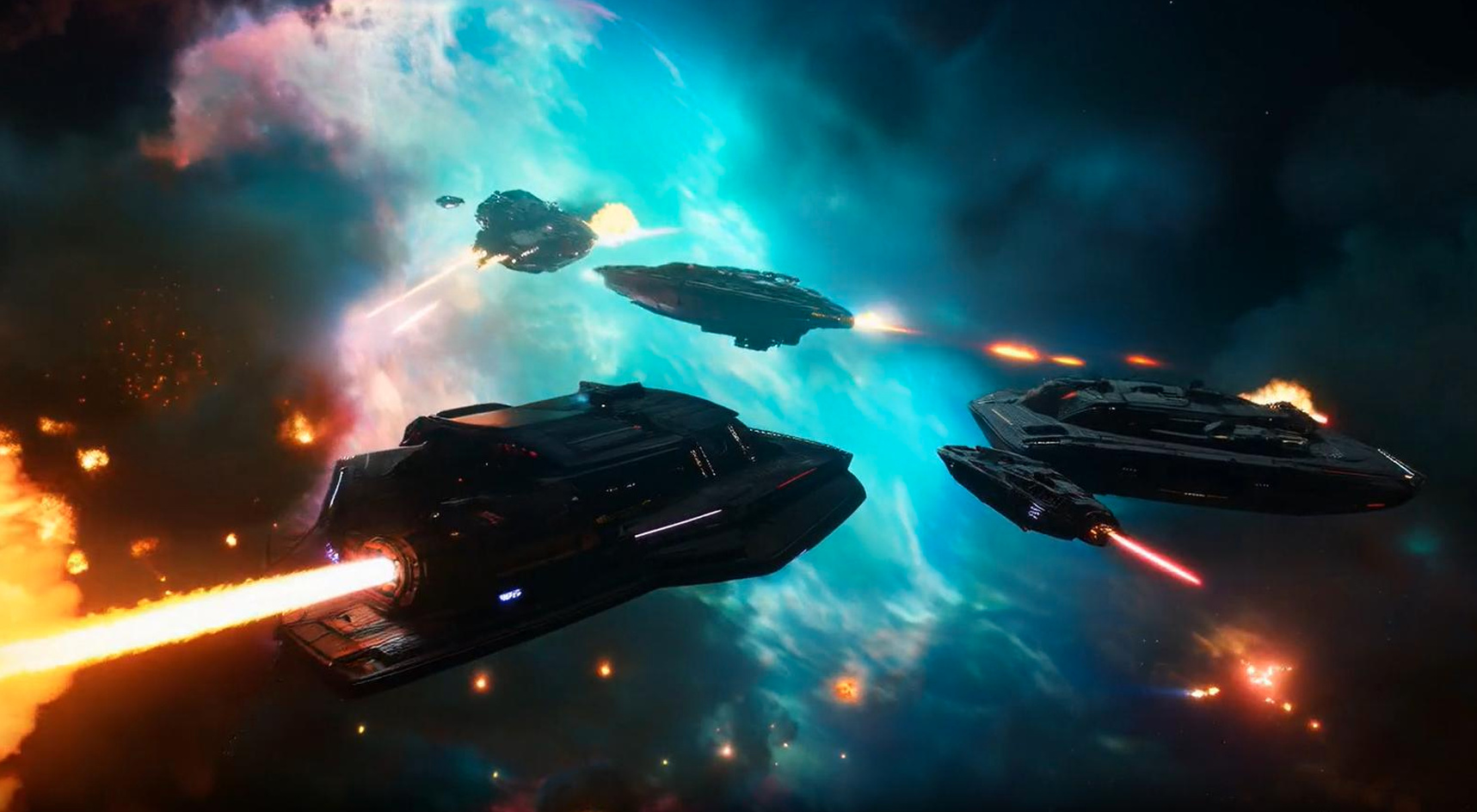} \hfill
            \includegraphics[width=0.195\textwidth]{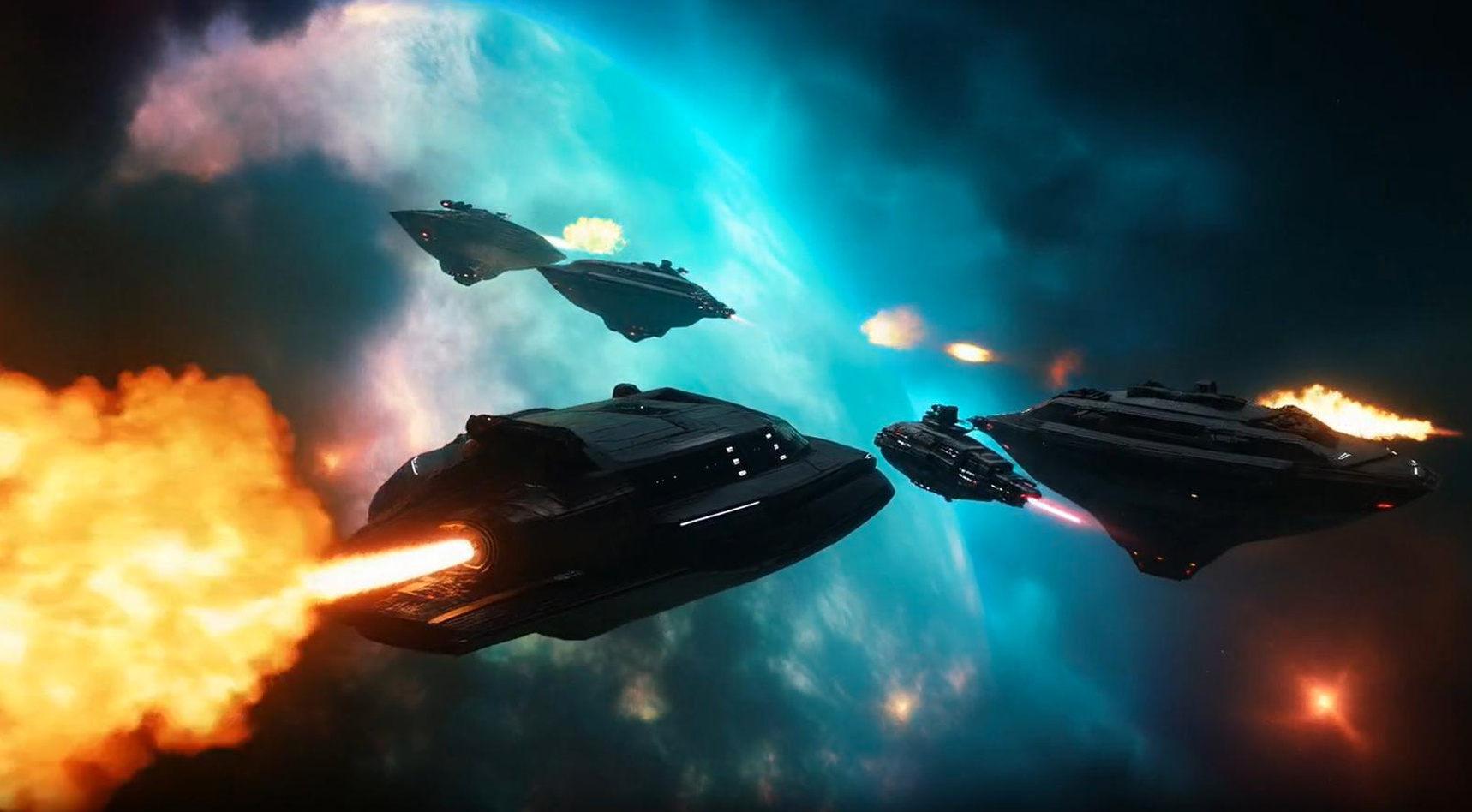} \hfill
            \includegraphics[width=0.195\textwidth]{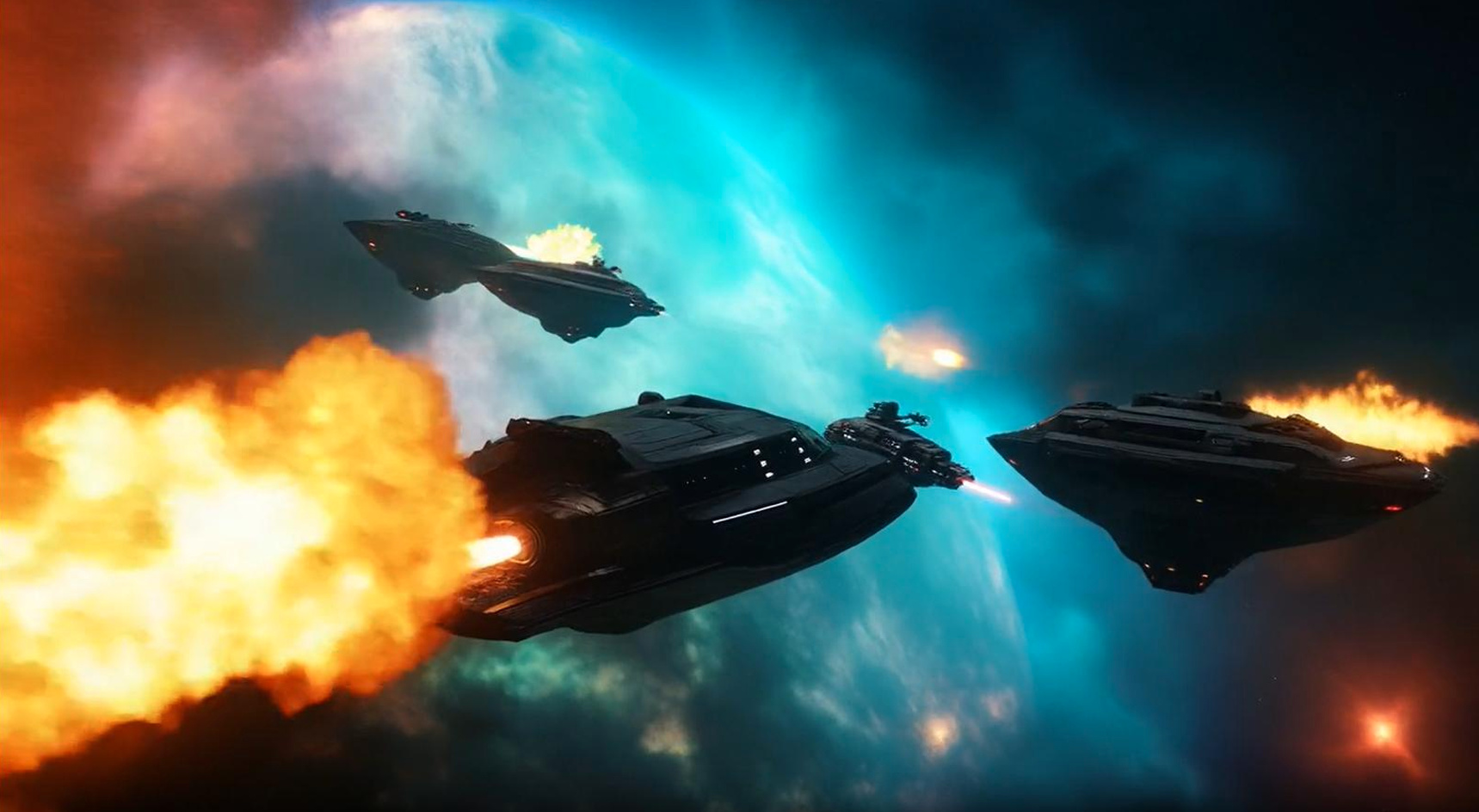}
        \end{subfigure}
        \vspace{1pt}
        \begin{subfigure}{\textwidth}
            \textbf{\scriptsize DynamicRad (Ours):} \\
            \includegraphics[width=0.195\textwidth]{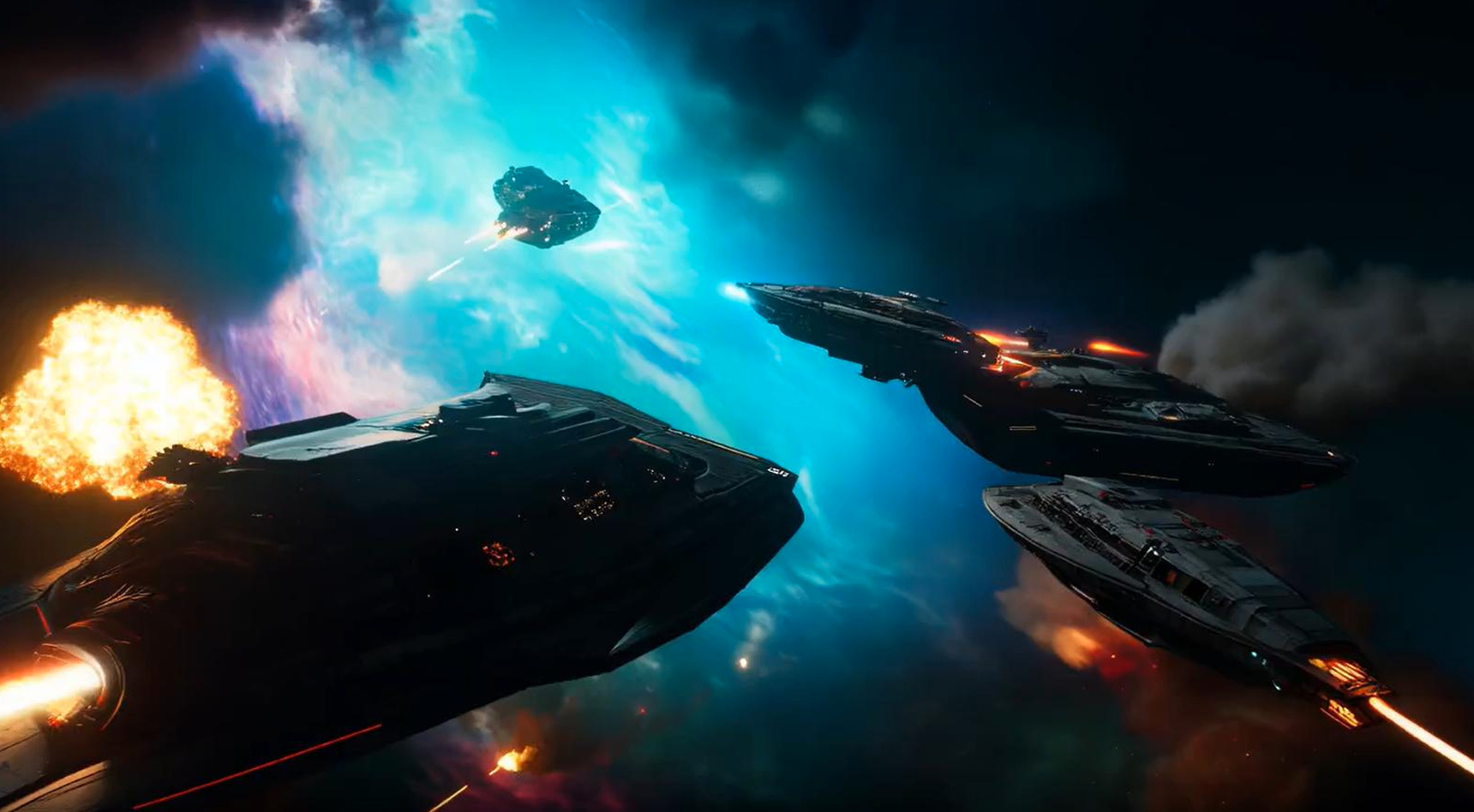} \hfill
            \includegraphics[width=0.195\textwidth]{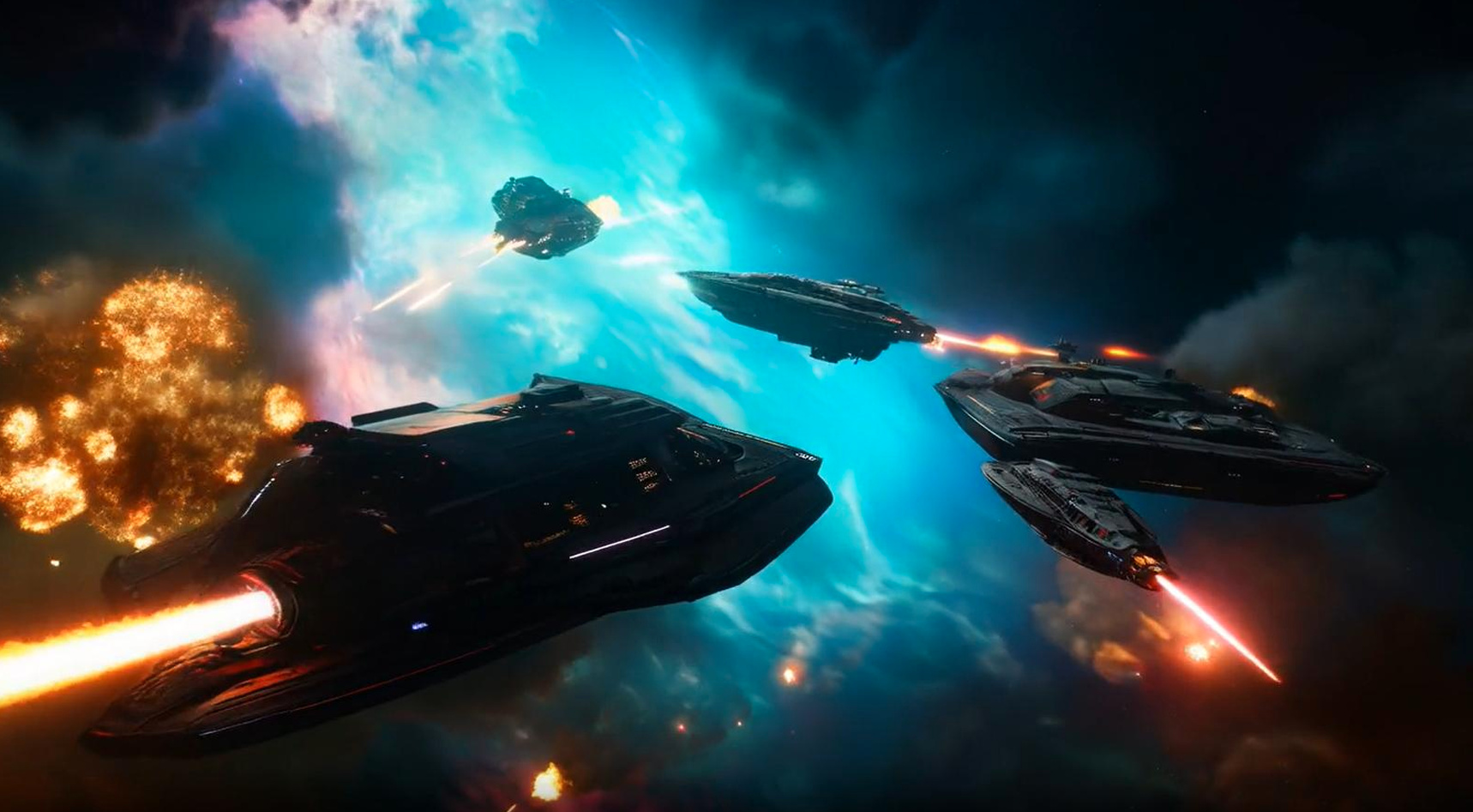} \hfill
            \includegraphics[width=0.195\textwidth]{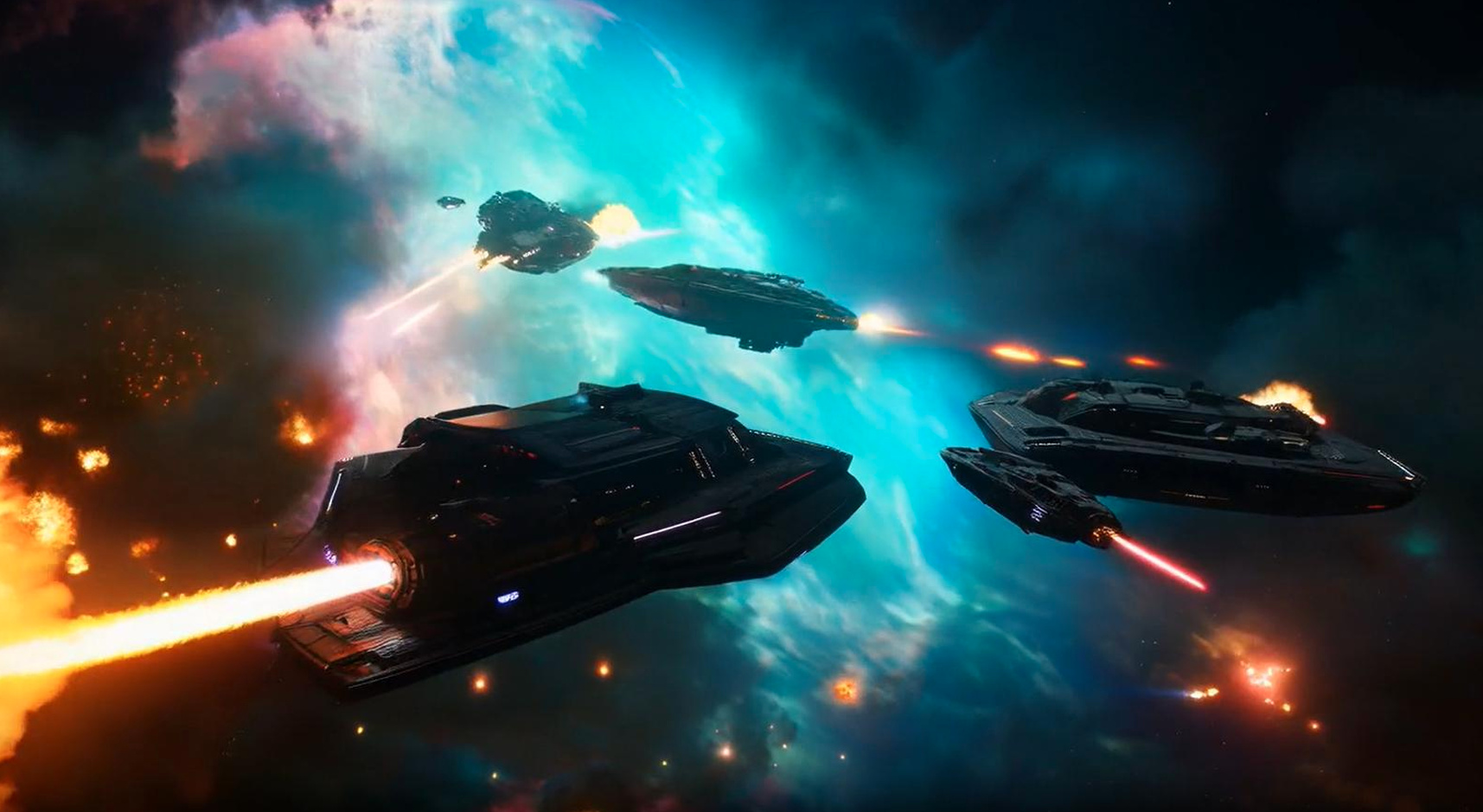} \hfill
            \includegraphics[width=0.195\textwidth]{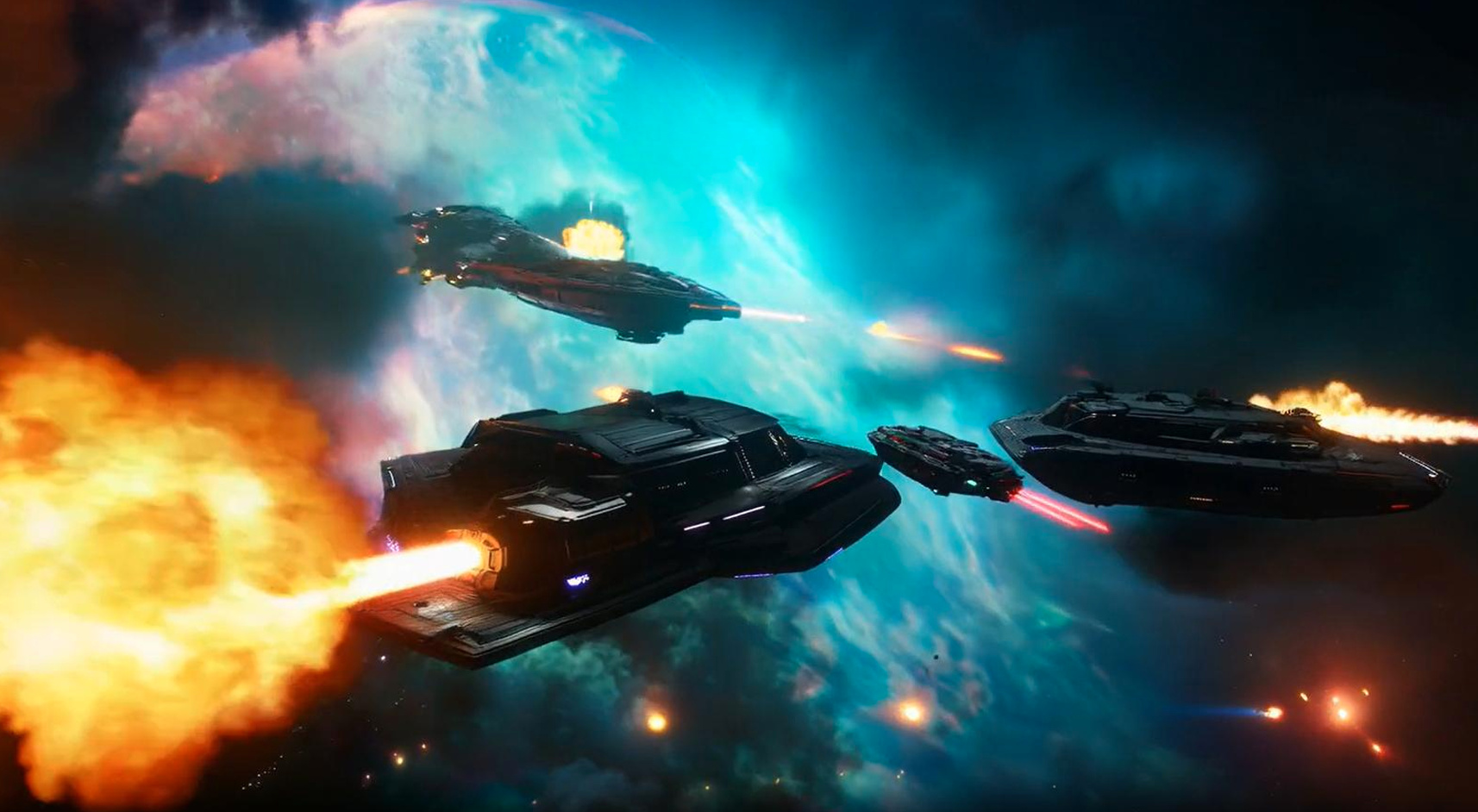} \hfill
            \includegraphics[width=0.195\textwidth]{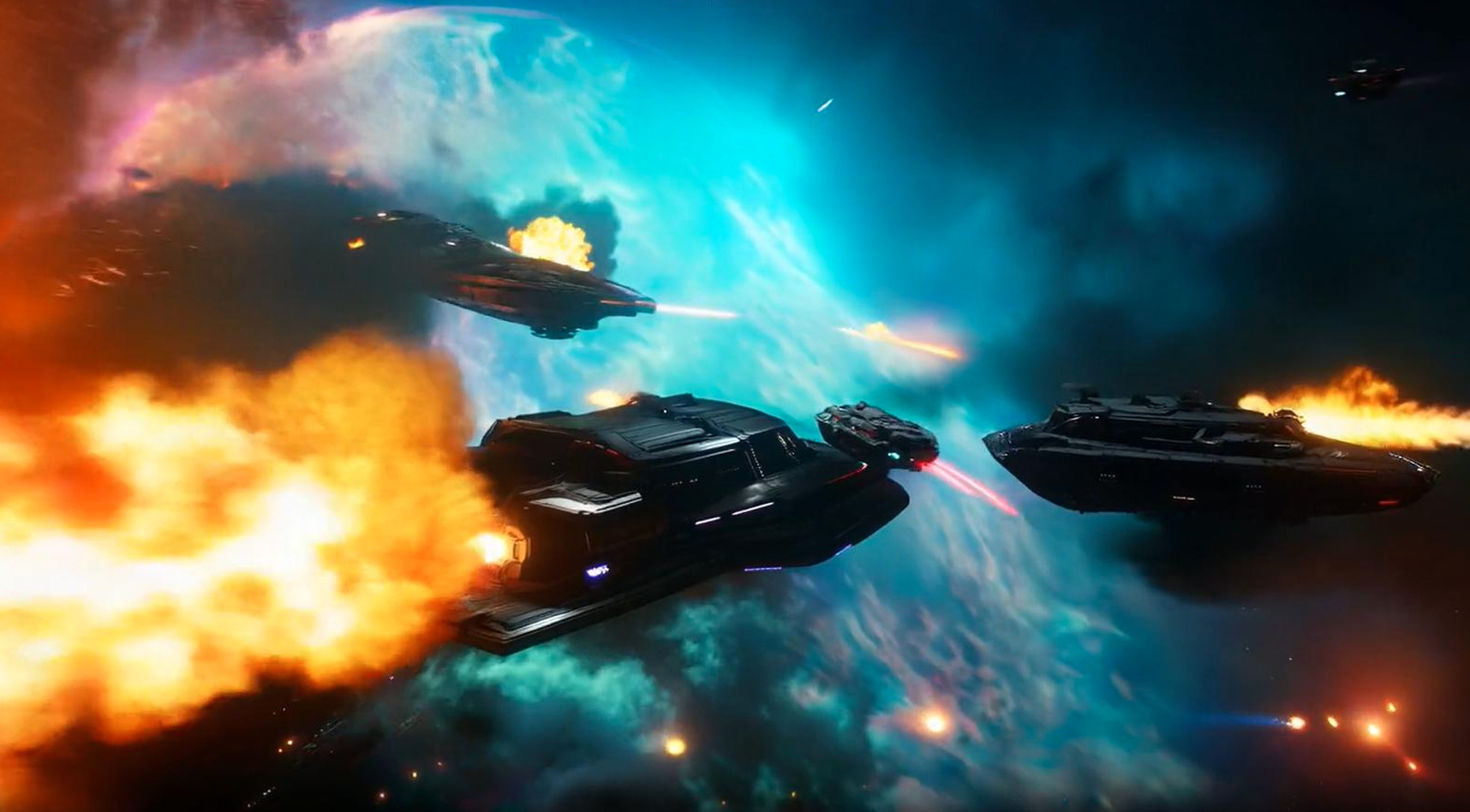}
        \end{subfigure}
    \end{minipage}
    \vspace{20pt}

    \begin{minipage}{\textwidth}
        \fcolorbox{gray!50}{gray!10}{
            \parbox{0.97\linewidth}{
                \vspace{2pt}
                \textbf{\small Case 8 (Urban/Timelapse):} \small \textit{“Time-lapse of a busy City intersection at night, car trails, crowds moving fast, neon signs flashing, 4k, high quality.”}
                \vspace{2pt}
            }
        }
        \vspace{3pt}
        \begin{subfigure}{\textwidth}
        \vspace{3pt}
            \textbf{\scriptsize Original (Dense):} \\
            \includegraphics[width=0.195\textwidth]{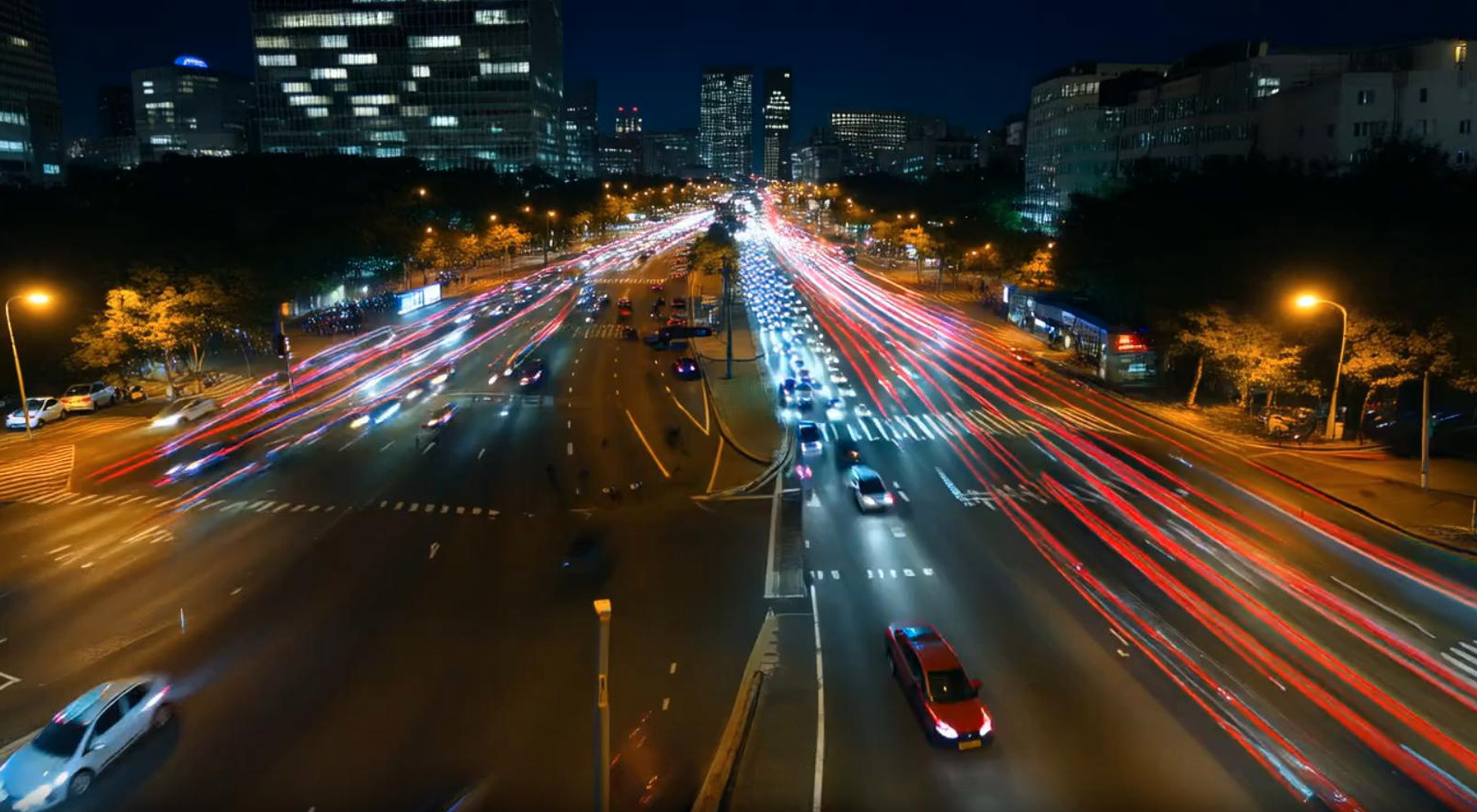} \hfill
            \includegraphics[width=0.195\textwidth]{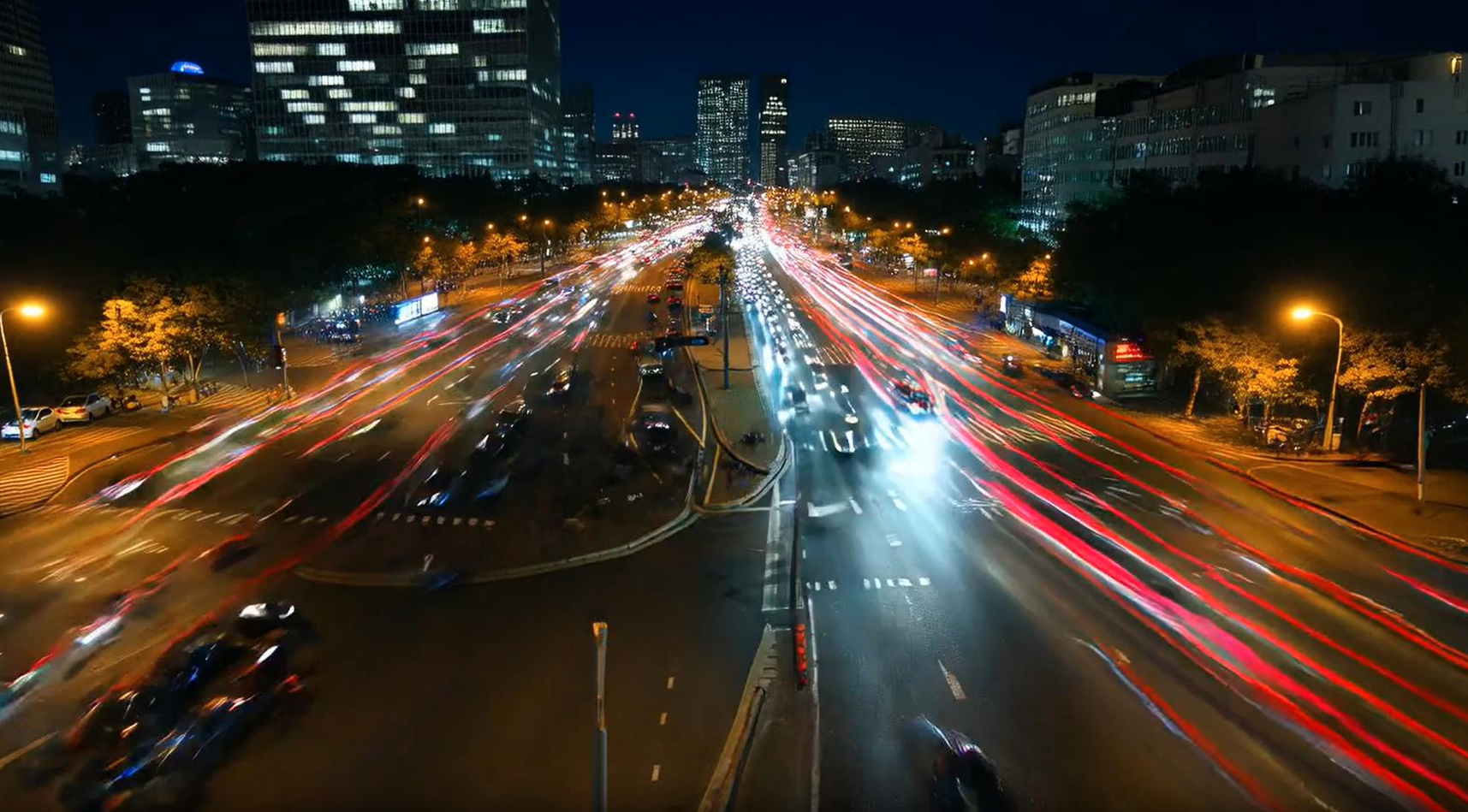} \hfill
            \includegraphics[width=0.195\textwidth]{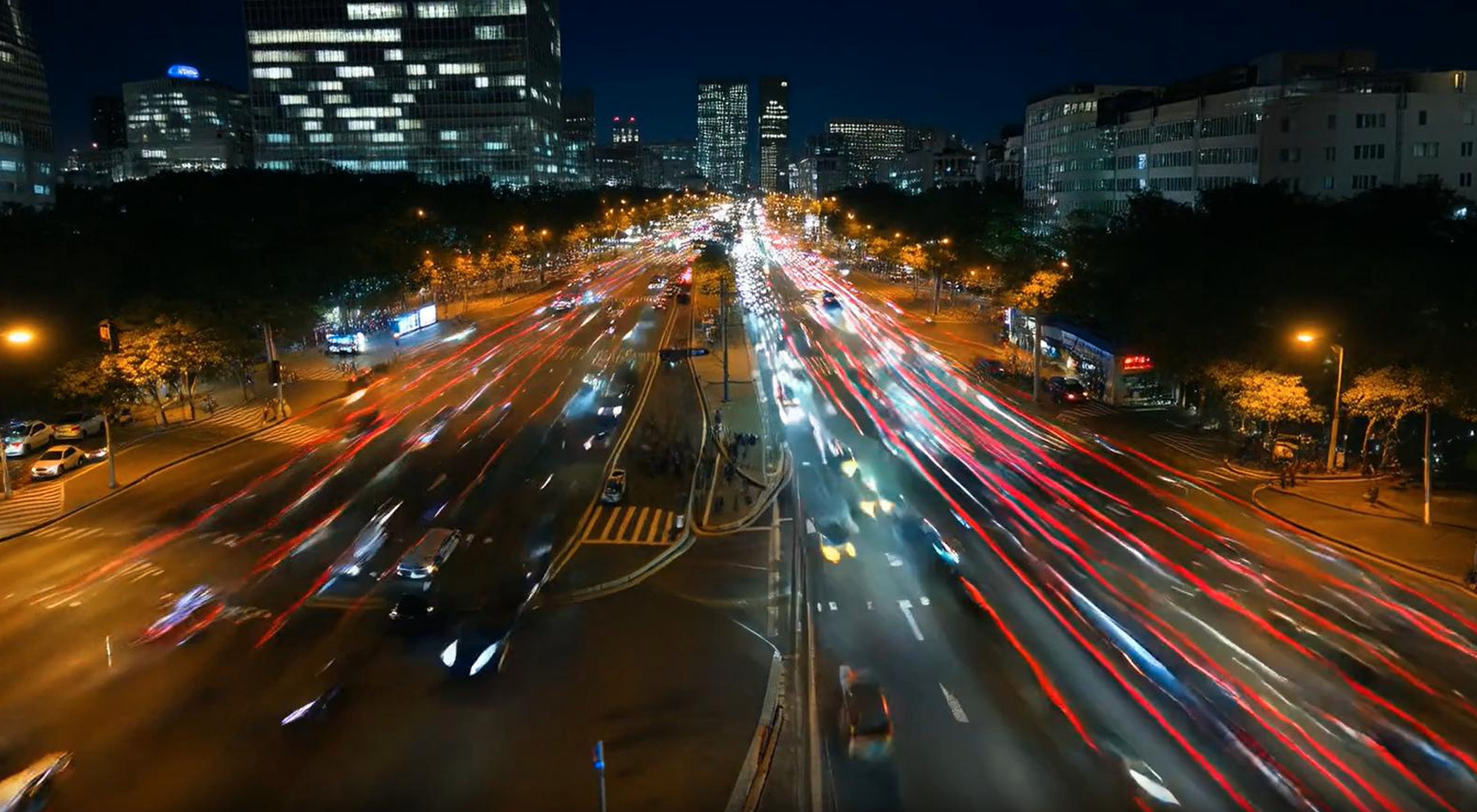} \hfill
            \includegraphics[width=0.195\textwidth]{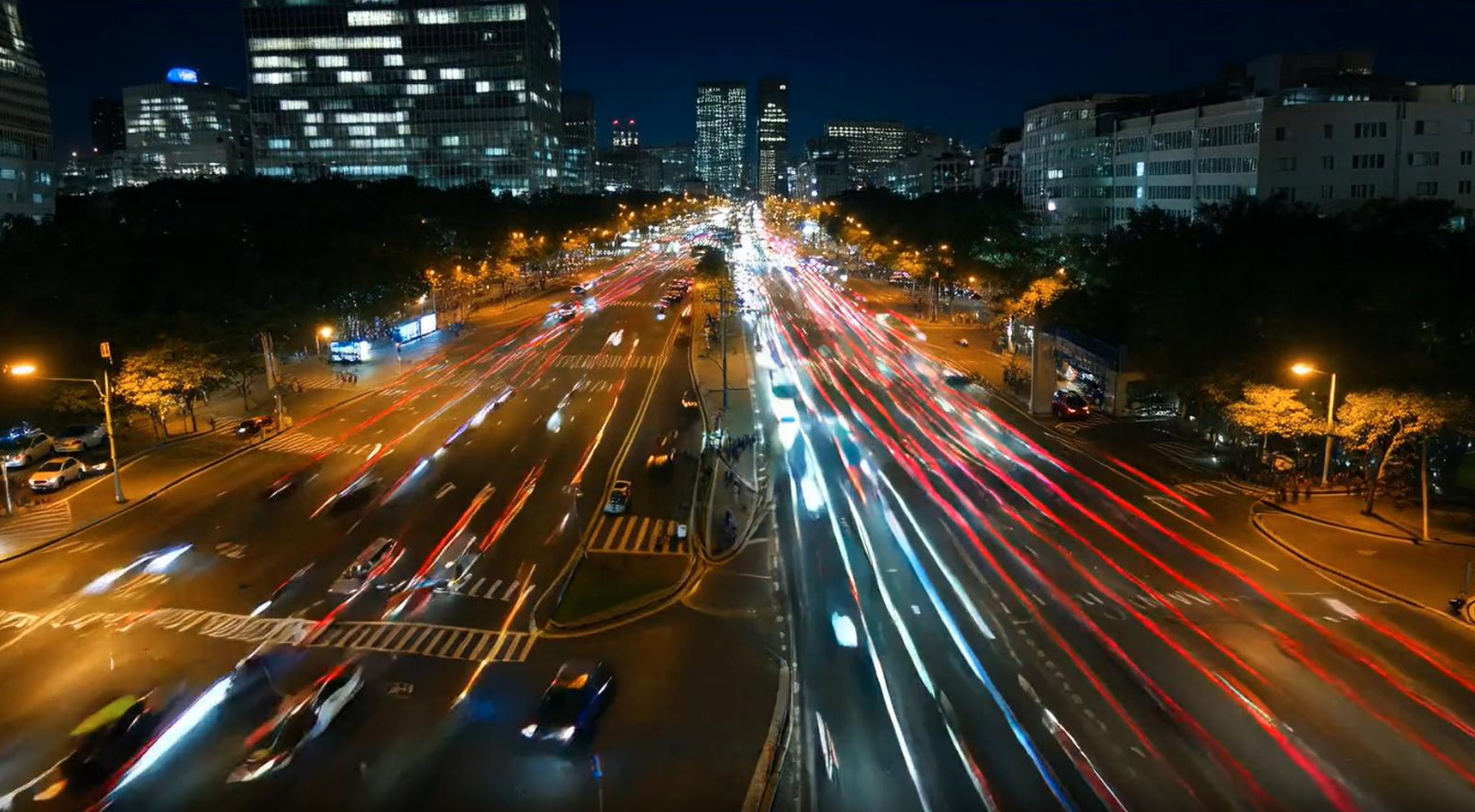} \hfill
            \includegraphics[width=0.195\textwidth]{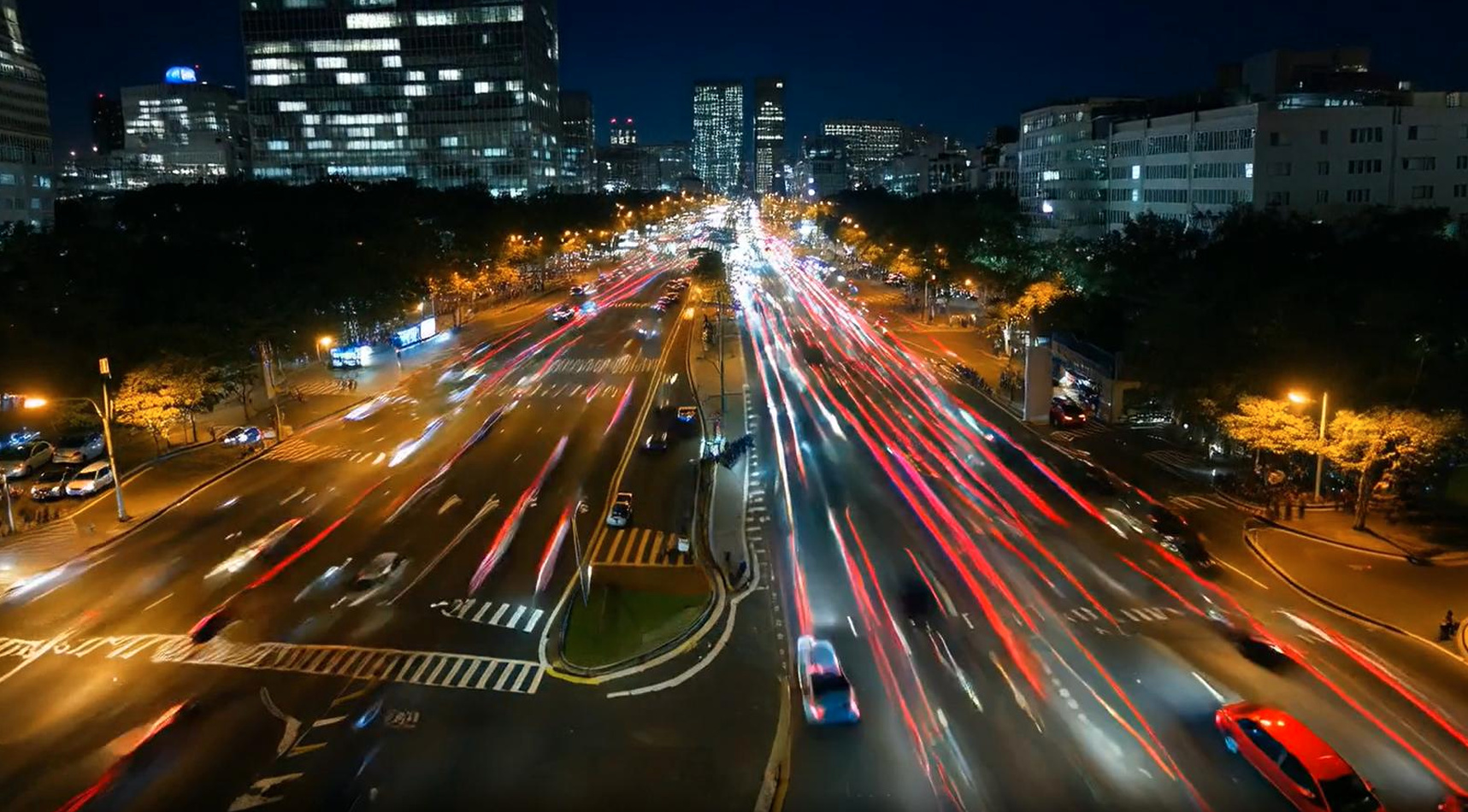}
        \end{subfigure}
        \vspace{1pt}
        \begin{subfigure}{\textwidth}
            \textbf{\scriptsize DynamicRad (Ours):} \\
            \includegraphics[width=0.195\textwidth]{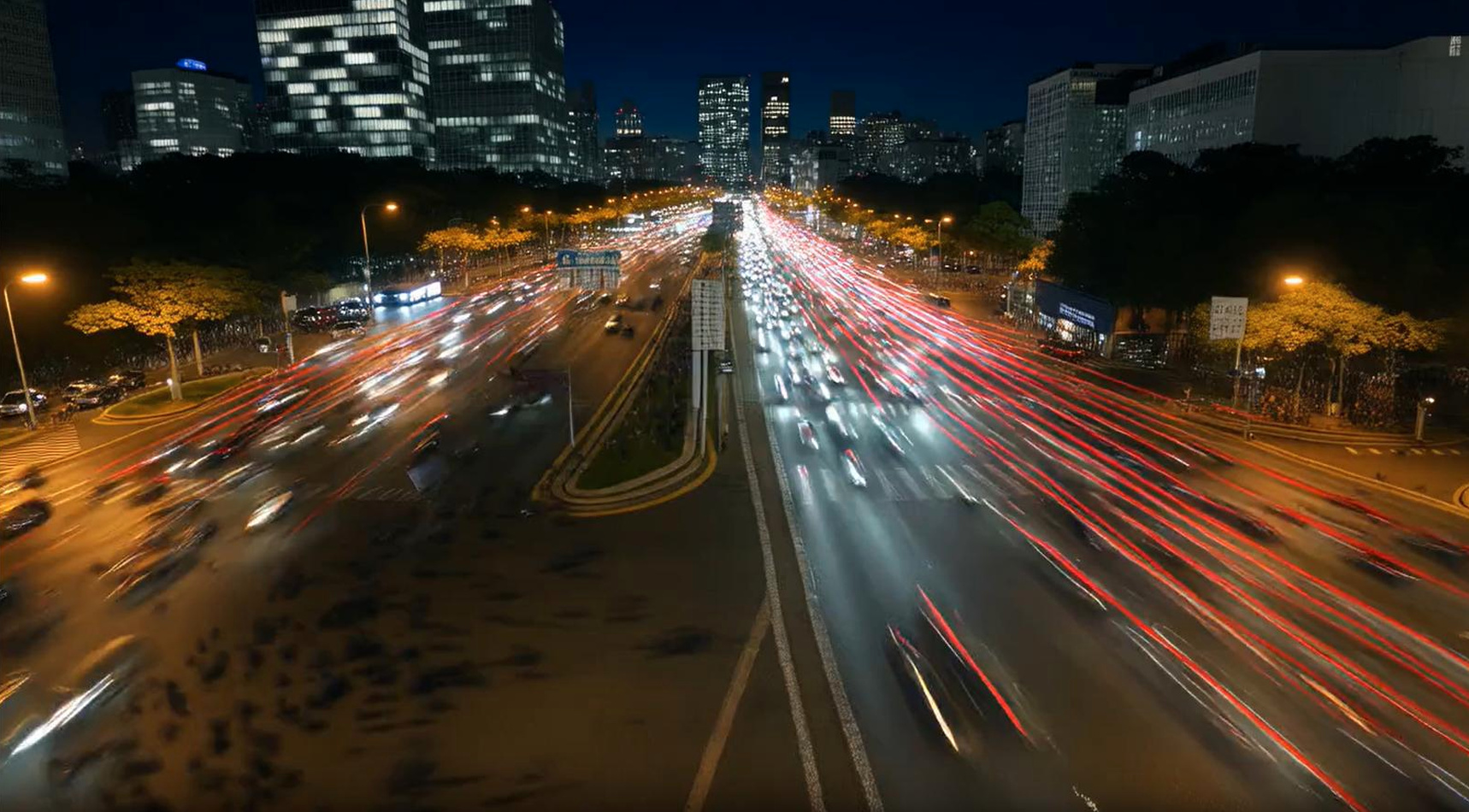} \hfill
            \includegraphics[width=0.195\textwidth]{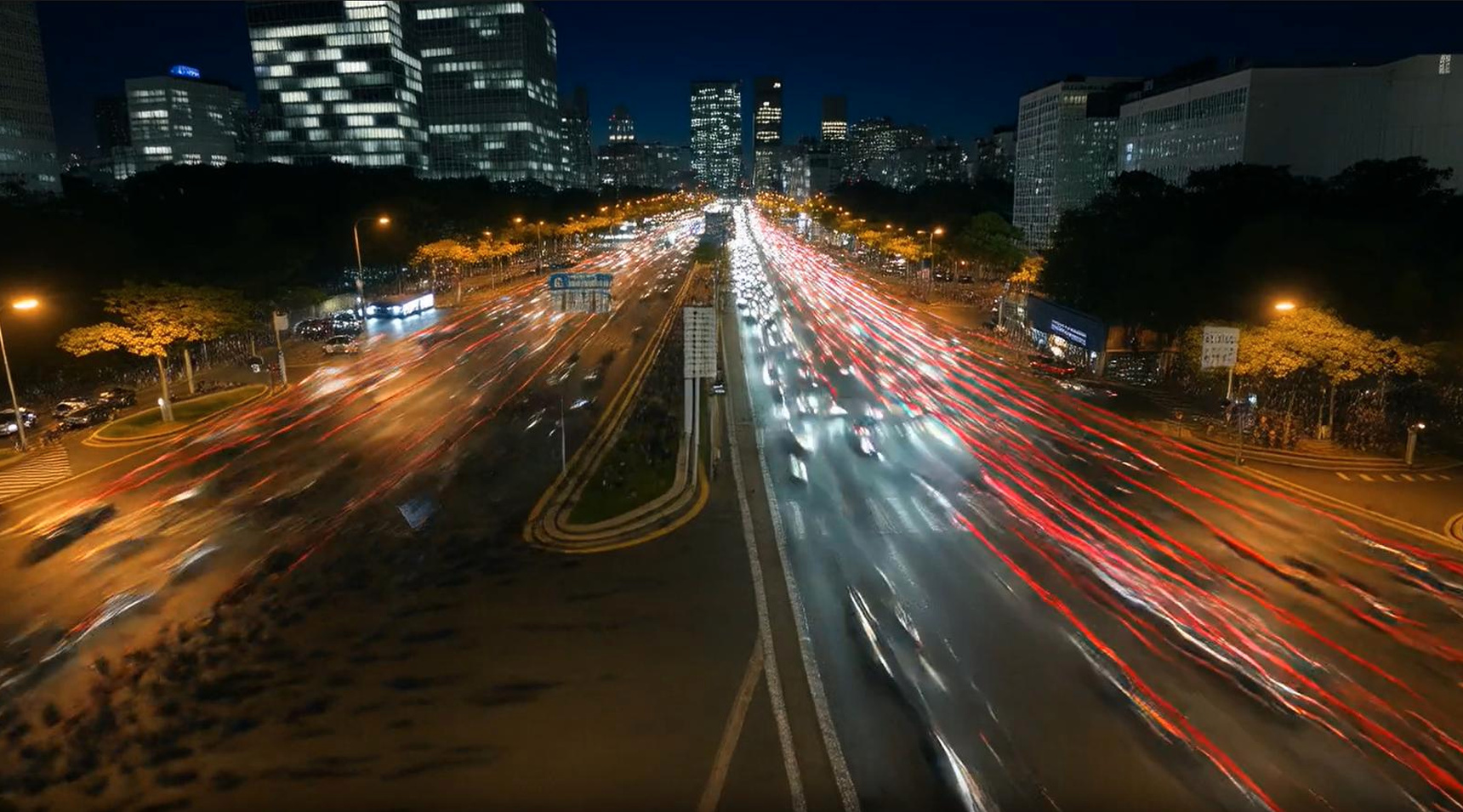} \hfill
            \includegraphics[width=0.195\textwidth]{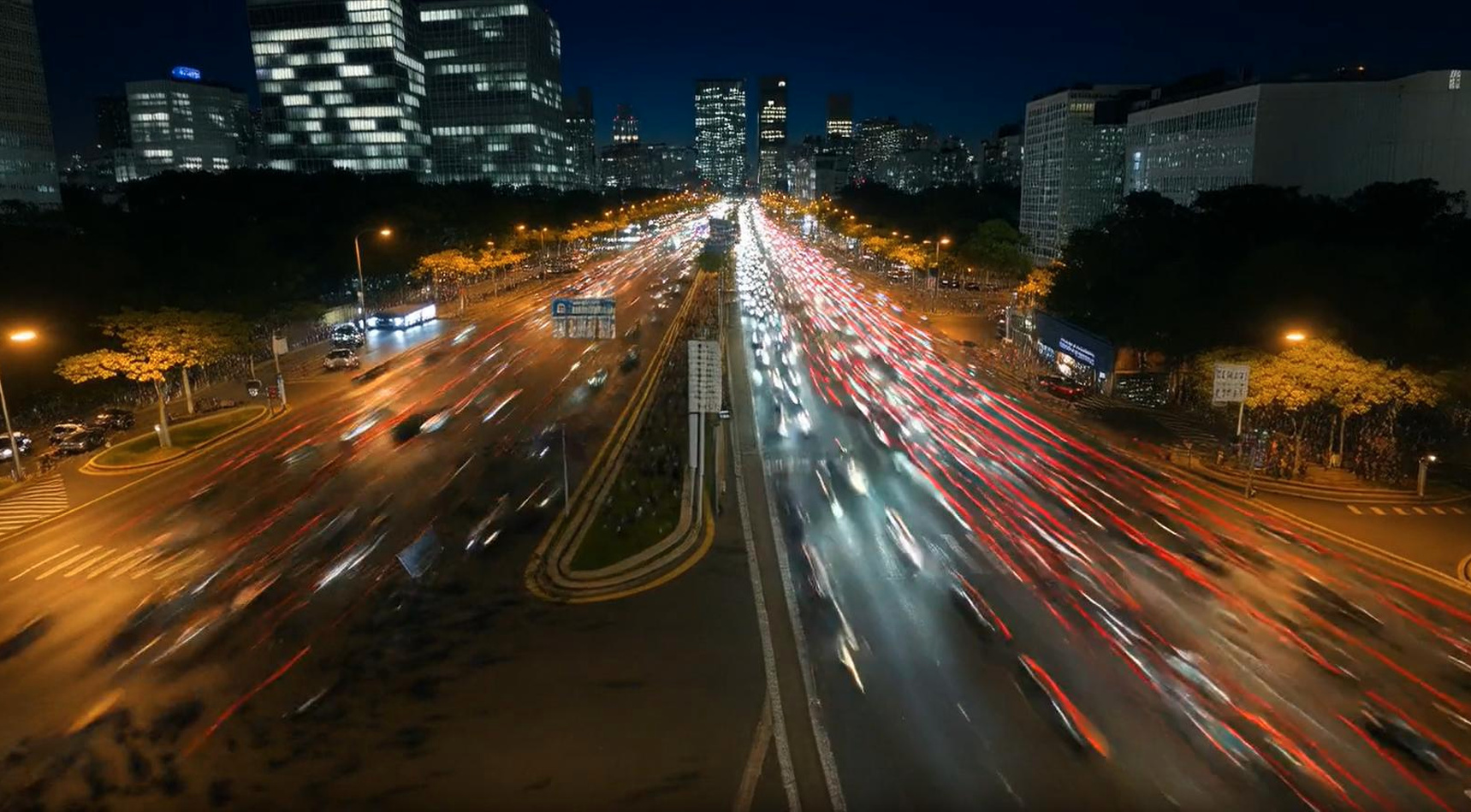} \hfill
            \includegraphics[width=0.195\textwidth]{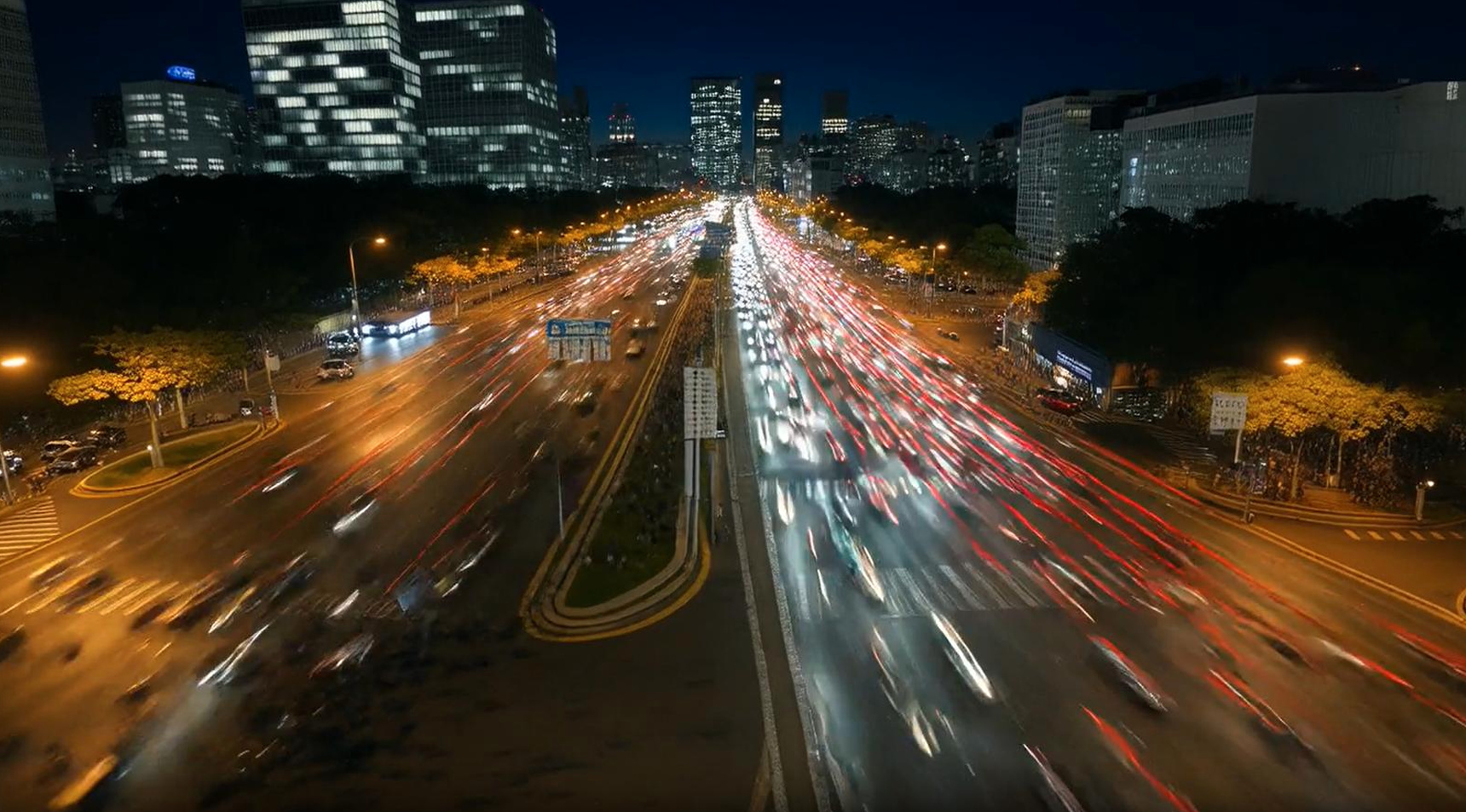} \hfill
            \includegraphics[width=0.195\textwidth]{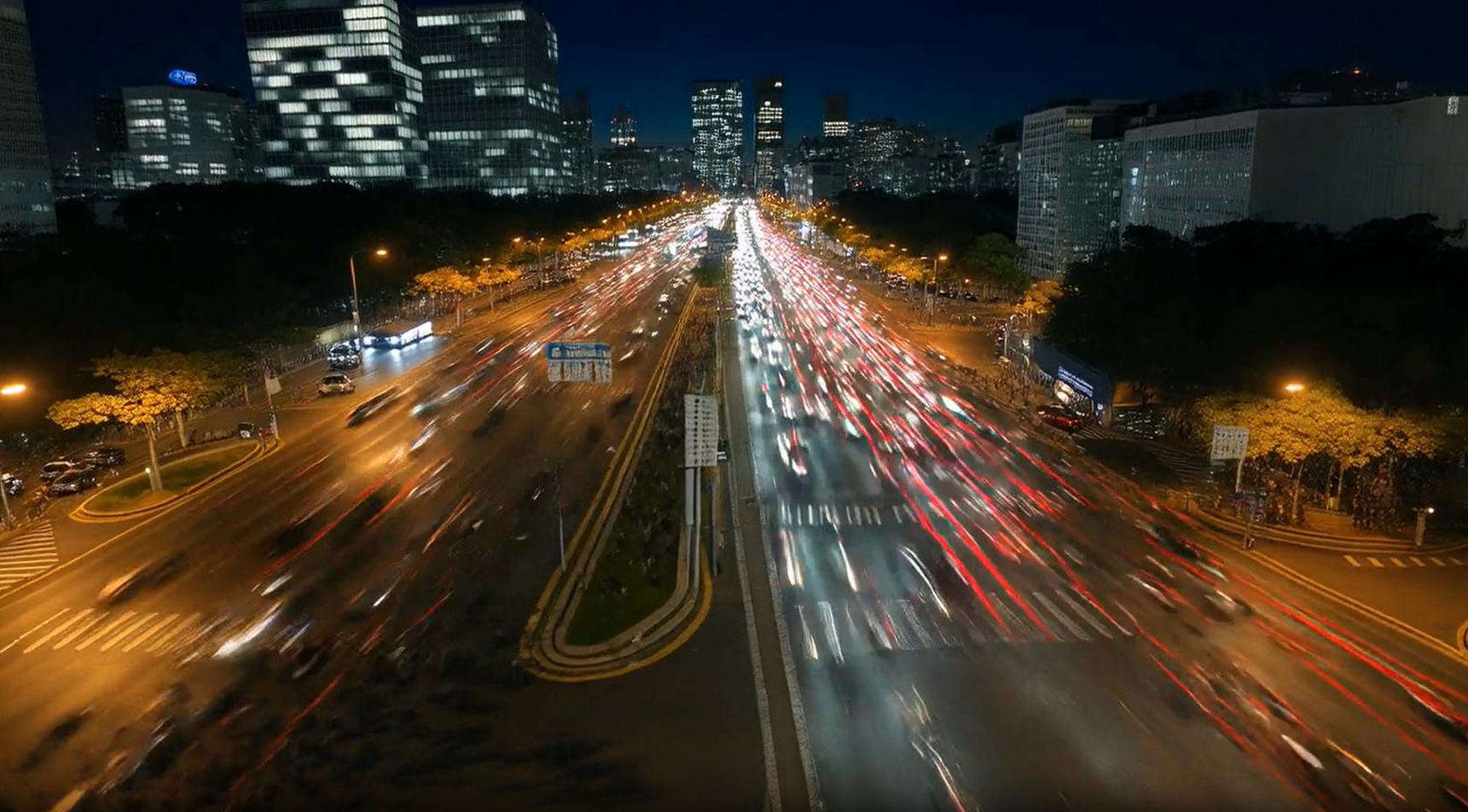}
        \end{subfigure}
    \end{minipage}

    \label{fig:hunyuan_gallery_2}
\end{figure}
\clearpage

\section{Broader Impacts and Existing Assets}
\label{app:ethics_assets}

\paragraph{Broader impacts.}
DynamicRad improves the inference efficiency of long-video diffusion models by reducing the attention-side computation and memory overhead. This can have positive impacts by lowering the computational cost of long-video generation research, enabling broader experimentation with resource-constrained hardware, and reducing the energy cost associated with repeated inference and ablation studies. At the same time, more efficient long-video generation may lower the barrier to producing synthetic videos, which could be misused for misleading media, impersonation, copyright-violating content, or disinformation. DynamicRad does not introduce a new generative backbone or release a new high-risk pretrained video generator; however, it can accelerate existing video generation systems. We therefore recommend that deployments using this technique follow the safeguards of the underlying generative model, including provenance tracking, watermarking where applicable, access control for high-risk use cases, and content-safety policies for generated media.

\paragraph{Existing assets.}
We use existing models, datasets, evaluation tools, and software packages
according to their respective licenses and terms of use. The licenses and usage
terms were checked from official repositories or model/dataset cards at the time
of submission. Table~\ref{tab:existing_assets} summarizes the main external
assets used in this work. We do not redistribute the original model weights,
datasets, or benchmark assets as part of this submission; users should obtain
them from the corresponding official sources and comply with their licenses and
usage terms.

\begin{table}[h]
\centering
\caption{Existing assets used in this work.}
\label{tab:existing_assets}
\scriptsize
\setlength{\tabcolsep}{3pt}
\renewcommand{\arraystretch}{1.12}
\begin{tabularx}{\linewidth}{@{}p{0.17\linewidth}p{0.16\linewidth}X X@{}}
\toprule
Asset & Type & Usage & License / Terms \\
\midrule
HunyuanVideo 
& Model/backbone 
& Long-video diffusion backbone for evaluation 
& Tencent Hunyuan Community License Agreement; includes territorial restrictions; third-party components follow their original licenses \\

Wan2.1-14B 
& Model/backbone 
& Video diffusion backbone for evaluation 
& Apache-2.0 \\

OpenVid-1M 
& Dataset 
& Router supervision and Mask-Aware LoRA fine-tuning subset 
& CC-BY-4.0 for dataset metadata; dataset card states intended research/non-commercial use and requires compliance with original source licenses \\

VBench 
& Evaluation benchmark 
& Subject consistency, aesthetic quality, and imaging quality evaluation 
& Apache-2.0 \\

VisionReward 
& Evaluation model/code 
& Human-aligned video quality scoring 
& Code: Apache-2.0; model weights follow the Hugging Face model card / upstream CogVLM2-based checkpoint terms \\

FlashInfer 
& Software/kernel library 
& Block-sparse attention kernel support and inference acceleration 
& Apache-2.0 \\

RAFT 
& Optical-flow model/code 
& Optical-flow pseudo-labeling for semantic motion router supervision 
& BSD-3-Clause \\

PEFT / LoRA 
& Software library 
& Parameter-efficient Mask-Aware LoRA adaptation 
& Apache-2.0 \\
\bottomrule
\end{tabularx}
\end{table}

\clearpage
\section*{NeurIPS Paper Checklist}

\begin{enumerate}

\item {\bf Claims}
    \item[] Question: Do the main claims made in the abstract and introduction accurately reflect the paper's contributions and scope?
    \item[] Answer: \answerYes{}
    \item[] Justification: The abstract and introduction state the main methodological contributions and the intended scope of the paper. These claims are supported by the experimental results in Secs.~4.1--4.3 and the stated limitations in Sec.~5.
    \item[] Guidelines:
    \begin{itemize}
        \item The answer \answerNA{} means that the abstract and introduction do not include the claims made in the paper.
        \item The abstract and/or introduction should clearly state the claims made, including the contributions made in the paper and important assumptions and limitations. A \answerNo{} or \answerNA{} answer to this question will not be perceived well by the reviewers. 
        \item The claims made should match theoretical and experimental results, and reflect how much the results can be expected to generalize to other settings. 
        \item It is fine to include aspirational goals as motivation as long as it is clear that these goals are not attained by the paper. 
    \end{itemize}

\item {\bf Limitations}
    \item[] Question: Does the paper discuss the limitations of the work performed by the authors?
    \item[] Answer: \answerYes{}
    \item[] Justification: The paper discusses limitations explicitly in Sec.~5 under ``Limitations and Future Work.'' These include extreme fast-motion cases, ambiguous prompts, and hardware generalization issues.
    \item[] Guidelines:
    \begin{itemize}
        \item The answer \answerNA{} means that the paper has no limitation while the answer \answerNo{} means that the paper has limitations, but those are not discussed in the paper. 
        \item The authors are encouraged to create a separate ``Limitations'' section in their paper.
        \item The paper should point out any strong assumptions and how robust the results are to violations of these assumptions (e.g., independence assumptions, noiseless settings, model well-specification, asymptotic approximations only holding locally). The authors should reflect on how these assumptions might be violated in practice and what the implications would be.
        \item The authors should reflect on the scope of the claims made, e.g., if the approach was only tested on a few datasets or with a few runs. In general, empirical results often depend on implicit assumptions, which should be articulated.
        \item The authors should reflect on the factors that influence the performance of the approach. For example, a facial recognition algorithm may perform poorly when image resolution is low or images are taken in low lighting. Or a speech-to-text system might not be used reliably to provide closed captions for online lectures because it fails to handle technical jargon.
        \item The authors should discuss the computational efficiency of the proposed algorithms and how they scale with dataset size.
        \item If applicable, the authors should discuss possible limitations of their approach to address problems of privacy and fairness.
        \item While the authors might fear that complete honesty about limitations might be used by reviewers as grounds for rejection, a worse outcome might be that reviewers discover limitations that aren't acknowledged in the paper. The authors should use their best judgment and recognize that individual actions in favor of transparency play an important role in developing norms that preserve the integrity of the community. Reviewers will be specifically instructed to not penalize honesty concerning limitations.
    \end{itemize}

\item {\bf Theory assumptions and proofs}
    \item[] Question: For each theoretical result, does the paper provide the full set of assumptions and a complete (and correct) proof?
    \item[] Answer: \answerNA{}
    \item[] Justification: The paper does not claim a formal complexity bound for the implemented method. Appendix A gives an explicitly idealized scaling illustration for the radial-decay prior, with its assumptions and short derivation stated there; no theorem/proof contribution is presented.
    \item[] Guidelines:
    \begin{itemize}
        \item The answer \answerNA{} means that the paper does not include theoretical results. 
        \item All the theorems, formulas, and proofs in the paper should be numbered and cross-referenced.
        \item All assumptions should be clearly stated or referenced in the statement of any theorems.
        \item The proofs can either appear in the main paper or the supplemental material, but if they appear in the supplemental material, the authors are encouraged to provide a short proof sketch to provide intuition. 
        \item Inversely, any informal proof provided in the core of the paper should be complemented by formal proofs provided in appendix or supplemental material.
        \item Theorems and Lemmas that the proof relies upon should be properly referenced. 
    \end{itemize}

    \item {\bf Experimental result reproducibility}
    \item[] Question: Does the paper fully disclose all the information needed to reproduce the main experimental results of the paper to the extent that it affects the main claims and/or conclusions of the paper (regardless of whether the code and data are provided or not)?
    \item[] Answer: \answerYes{}
    \item[] Justification: The paper provides the method description, experimental setup, model backbones, hardware setup, BO search description, and evaluation metrics in the main paper and appendix. A code link is also provided to support reproducibility.
    \item[] Guidelines:
    \begin{itemize}
        \item The answer \answerNA{} means that the paper does not include experiments.
        \item If the paper includes experiments, a \answerNo{} answer to this question will not be perceived well by the reviewers: Making the paper reproducible is important, regardless of whether the code and data are provided or not.
        \item If the contribution is a dataset and\slash or model, the authors should describe the steps taken to make their results reproducible or verifiable. 
        \item Depending on the contribution, reproducibility can be accomplished in various ways. For example, if the contribution is a novel architecture, describing the architecture fully might suffice, or if the contribution is a specific model and empirical evaluation, it may be necessary to either make it possible for others to replicate the model with the same dataset, or provide access to the model. In general. releasing code and data is often one good way to accomplish this, but reproducibility can also be provided via detailed instructions for how to replicate the results, access to a hosted model (e.g., in the case of a large language model), releasing of a model checkpoint, or other means that are appropriate to the research performed.
        \item While NeurIPS does not require releasing code, the conference does require all submissions to provide some reasonable avenue for reproducibility, which may depend on the nature of the contribution. For example
        \begin{enumerate}
            \item If the contribution is primarily a new algorithm, the paper should make it clear how to reproduce that algorithm.
            \item If the contribution is primarily a new model architecture, the paper should describe the architecture clearly and fully.
            \item If the contribution is a new model (e.g., a large language model), then there should either be a way to access this model for reproducing the results or a way to reproduce the model (e.g., with an open-source dataset or instructions for how to construct the dataset).
            \item We recognize that reproducibility may be tricky in some cases, in which case authors are welcome to describe the particular way they provide for reproducibility. In the case of closed-source models, it may be that access to the model is limited in some way (e.g., to registered users), but it should be possible for other researchers to have some path to reproducing or verifying the results.
        \end{enumerate}
    \end{itemize}

\item {\bf Open access to data and code}
    \item[] Question: Does the paper provide open access to the data and code, with sufficient instructions to faithfully reproduce the main experimental results, as described in supplemental material?
    \item[] Answer: \answerNo{}
    \item[] Justification: Code and usage instructions for the Wan implementation are publicly available. The current release does not include all materials needed to reproduce every main experimental result.
    \item[] Guidelines:
    \begin{itemize}
        \item The answer \answerNA{} means that paper does not include experiments requiring code.
        \item Please see the NeurIPS code and data submission guidelines (\url{https://neurips.cc/public/guides/CodeSubmissionPolicy}) for more details.
        \item While we encourage the release of code and data, we understand that this might not be possible, so \answerNo{} is an acceptable answer. Papers cannot be rejected simply for not including code, unless this is central to the contribution (e.g., for a new open-source benchmark).
        \item The instructions should contain the exact command and environment needed to run to reproduce the results. See the NeurIPS code and data submission guidelines (\url{https://neurips.cc/public/guides/CodeSubmissionPolicy}) for more details.
        \item The authors should provide instructions on data access and preparation, including how to access the raw data, preprocessed data, intermediate data, and generated data, etc.
        \item The authors should provide scripts to reproduce all experimental results for the new proposed method and baselines. If only a subset of experiments are reproducible, they should state which ones are omitted from the script and why.
        \item At submission time, to preserve anonymity, the authors should release anonymized versions (if applicable).
        \item Providing as much information as possible in supplemental material (appended to the paper) is recommended, but including URLs to data and code is permitted.
    \end{itemize}

\item {\bf Experimental setting/details}
    \item[] Question: Does the paper specify all the training and test details (e.g., data splits, hyperparameters, how they were chosen, type of optimizer) necessary to understand the results?
    \item[] Answer: \answerYes{}
    \item[] Justification: The experimental setting is described in Sec.~4.1, including the backbones, frame settings, hardware environment, BO-based configuration pipeline, and LoRA setup. Additional hyperparameter ranges and BO details are provided in the appendix.
    \item[] Guidelines:
    \begin{itemize}
        \item The answer \answerNA{} means that the paper does not include experiments.
        \item The experimental setting should be presented in the core of the paper to a level of detail that is necessary to appreciate the results and make sense of them.
        \item The full details can be provided either with the code, in appendix, or as supplemental material.
    \end{itemize}

\item {\bf Experiment statistical significance}
    \item[] Question: Does the paper report error bars suitably and correctly defined or other appropriate information about the statistical significance of the experiments?
    \item[] Answer: \answerNo{}
    \item[] Justification: The five-prompt end-to-end latency study reports standard deviations across prompt-level means after three timing runs per prompt. The video-quality comparisons do not include formal significance tests; conclusions are supported by matched protocols, two backbones, and ablations.
    \item[] Guidelines:
    \begin{itemize}
        \item The answer \answerNA{} means that the paper does not include experiments.
        \item The authors should answer \answerYes{} if the results are accompanied by error bars, confidence intervals, or statistical significance tests, at least for the experiments that support the main claims of the paper.
        \item The factors of variability that the error bars are capturing should be clearly stated (for example, train/test split, initialization, random drawing of some parameter, or overall run with given experimental conditions).
        \item The method for calculating the error bars should be explained (closed form formula, call to a library function, bootstrap, etc.)
        \item The assumptions made should be given (e.g., Normally distributed errors).
        \item It should be clear whether the error bar is the standard deviation or the standard error of the mean.
        \item It is OK to report 1-sigma error bars, but one should state it. The authors should preferably report a 2-sigma error bar than state that they have a 96\% CI, if the hypothesis of Normality of errors is not verified.
        \item For asymmetric distributions, the authors should be careful not to show in tables or figures symmetric error bars that would yield results that are out of range (e.g., negative error rates).
        \item If error bars are reported in tables or plots, the authors should explain in the text how they were calculated and reference the corresponding figures or tables in the text.
    \end{itemize}

\item {\bf Experiments compute resources}
    \item[] Question: For each experiment, does the paper provide sufficient information on the computer resources (type of compute workers, memory, time of execution) needed to reproduce the experiments?
    \item[] Answer: \answerYes{}
    \item[] Justification: Sec.~4.1 reports the main compute environment, including single-H100 and 8$\times$H100 settings, and the paper reports representative runtime and profiling information. Additional BO efficiency and routing overhead details are provided in the appendix.
    \item[] Guidelines:
    \begin{itemize}
        \item The answer \answerNA{} means that the paper does not include experiments.
        \item The paper should indicate the type of compute workers CPU or GPU, internal cluster, or cloud provider, including relevant memory and storage.
        \item The paper should provide the amount of compute required for each of the individual experimental runs as well as estimate the total compute. 
        \item The paper should disclose whether the full research project required more compute than the experiments reported in the paper (e.g., preliminary or failed experiments that didn't make it into the paper). 
    \end{itemize}
    
\item {\bf Code of ethics}
    \item[] Question: Does the research conducted in the paper conform, in every respect, with the NeurIPS Code of Ethics \url{https://neurips.cc/public/EthicsGuidelines}?
    \item[] Answer: \answerYes{}
    \item[] Justification: To the best of our knowledge, the research conforms to the NeurIPS Code of Ethics.
    \item[] Guidelines:
    \begin{itemize}
        \item The answer \answerNA{} means that the authors have not reviewed the NeurIPS Code of Ethics.
        \item If the authors answer \answerNo, they should explain the special circumstances that require a deviation from the Code of Ethics.
        \item The authors should make sure to preserve anonymity (e.g., if there is a special consideration due to laws or regulations in their jurisdiction).
    \end{itemize}

\item {\bf Broader impacts}
    \item[] Question: Does the paper discuss both potential positive societal impacts and negative societal impacts of the work performed?
    \item[] Answer: \answerYes{}
    \item[] Justification: We discuss both positive and negative societal impacts in Appendix~\ref{app:ethics_assets}. The positive impacts include reduced computational cost and broader accessibility for long-video generation research, while the negative impacts include potential misuse for synthetic media or disinformation.
    \item[] Guidelines:
    \begin{itemize}
        \item The answer \answerNA{} means that there is no societal impact of the work performed.
        \item If the authors answer \answerNA{} or \answerNo, they should explain why their work has no societal impact or why the paper does not address societal impact.
        \item Examples of negative societal impacts include potential malicious or unintended uses (e.g., disinformation, generating fake profiles, surveillance), fairness considerations (e.g., deployment of technologies that could make decisions that unfairly impact specific groups), privacy considerations, and security considerations.
        \item The conference expects that many papers will be foundational research and not tied to particular applications, let alone deployments. However, if there is a direct path to any negative applications, the authors should point it out. For example, it is legitimate to point out that an improvement in the quality of generative models could be used to generate Deepfakes for disinformation. On the other hand, it is not needed to point out that a generic algorithm for optimizing neural networks could enable people to train models that generate Deepfakes faster.
        \item The authors should consider possible harms that could arise when the technology is being used as intended and functioning correctly, harms that could arise when the technology is being used as intended but gives incorrect results, and harms following from (intentional or unintentional) misuse of the technology.
        \item If there are negative societal impacts, the authors could also discuss possible mitigation strategies (e.g., gated release of models, providing defenses in addition to attacks, mechanisms for monitoring misuse, mechanisms to monitor how a system learns from feedback over time, improving the efficiency and accessibility of ML).
    \end{itemize}
    
\item {\bf Safeguards}
    \item[] Question: Does the paper describe safeguards that have been put in place for responsible release of data or models that have a high risk for misuse (e.g., pre-trained language models, image generators, or scraped datasets)?
    \item[] Answer: \answerNA{}
    \item[] Justification: The paper does not release new pretrained generative model weights or new scraped datasets. The contribution provides an implementation of the method rather than a new high-risk generative model asset.
    \item[] Guidelines:
    \begin{itemize}
        \item The answer \answerNA{} means that the paper poses no such risks.
        \item Released models that have a high risk for misuse or dual-use should be released with necessary safeguards to allow for controlled use of the model, for example by requiring that users adhere to usage guidelines or restrictions to access the model or implementing safety filters. 
        \item Datasets that have been scraped from the Internet could pose safety risks. The authors should describe how they avoided releasing unsafe images.
        \item We recognize that providing effective safeguards is challenging, and many papers do not require this, but we encourage authors to take this into account and make a best faith effort.
    \end{itemize}

\item {\bf Licenses for existing assets}
    \item[] Question: Are the creators or original owners of assets (e.g., code, data, models), used in the paper, properly credited and are the license and terms of use explicitly mentioned and properly respected?
    \item[] Answer: \answerYes{}
    \item[] Justification: The external models, datasets, benchmarks, and software packages used in this work are cited in the main paper, and their licenses or usage terms are summarized in Appendix~\ref{app:ethics_assets}, Table~\ref{tab:existing_assets}. We do not redistribute the original external assets and use them according to their respective licenses and terms.
    \item[] Guidelines:
    \begin{itemize}
        \item The answer \answerNA{} means that the paper does not use existing assets.
        \item The authors should cite the original paper that produced the code package or dataset.
        \item The authors should state which version of the asset is used and, if possible, include a URL.
        \item The name of the license (e.g., CC-BY 4.0) should be included for each asset.
        \item For scraped data from a particular source (e.g., website), the copyright and terms of service of that source should be provided.
        \item If assets are released, the license, copyright information, and terms of use in the package should be provided. For popular datasets, \url{paperswithcode.com/datasets} has curated licenses for some datasets. Their licensing guide can help determine the license of a dataset.
        \item For existing datasets that are re-packaged, both the original license and the license of the derived asset (if it has changed) should be provided.
        \item If this information is not available online, the authors are encouraged to reach out to the asset's creators.
    \end{itemize}

\item {\bf New assets}
    \item[] Question: Are new assets introduced in the paper well documented and is the documentation provided alongside the assets?
    \item[] Answer: \answerYes{}
    \item[] Justification: The paper provides a new implementation of the proposed method together with documentation and usage instructions in the accompanying code repository and supplemental material.
    \item[] Guidelines:
    \begin{itemize}
        \item The answer \answerNA{} means that the paper does not release new assets.
        \item Researchers should communicate the details of the dataset\slash code\slash model as part of their submissions via structured templates. This includes details about training, license, limitations, etc. 
        \item The paper should discuss whether and how consent was obtained from people whose asset is used.
        \item At submission time, remember to anonymize your assets (if applicable). You can either create an anonymized URL or include an anonymized zip file.
    \end{itemize}

\item {\bf Crowdsourcing and research with human subjects}
    \item[] Question: For crowdsourcing experiments and research with human subjects, does the paper include the full text of instructions given to participants and screenshots, if applicable, as well as details about compensation (if any)? 
    \item[] Answer: \answerNA{}
    \item[] Justification: The paper reports no crowdsourcing or research involving human participants.
    \item[] Guidelines:
    \begin{itemize}
        \item The answer \answerNA{} means that the paper does not involve crowdsourcing nor research with human subjects.
        \item Including this information in the supplemental material is fine, but if the main contribution of the paper involves human subjects, then as much detail as possible should be included in the main paper. 
        \item According to the NeurIPS Code of Ethics, workers involved in data collection, curation, or other labor should be paid at least the minimum wage in the country of the data collector. 
    \end{itemize}

\item {\bf Institutional review board (IRB) approvals or equivalent for research with human subjects}
    \item[] Question: Does the paper describe potential risks incurred by study participants, whether such risks were disclosed to the subjects, and whether Institutional Review Board (IRB) approvals (or an equivalent approval/review based on the requirements of your country or institution) were obtained?
    \item[] Answer: \answerNA{}
    \item[] Justification: The paper reports no research involving human participants.
    \item[] Guidelines:
    \begin{itemize}
        \item The answer \answerNA{} means that the paper does not involve crowdsourcing nor research with human subjects.
        \item Depending on the country in which research is conducted, IRB approval (or equivalent) may be required for any human subjects research. If you obtained IRB approval, you should clearly state this in the paper. 
        \item We recognize that the procedures for this may vary significantly between institutions and locations, and we expect authors to adhere to the NeurIPS Code of Ethics and the guidelines for their institution. 
        \item For initial submissions, do not include any information that would break anonymity (if applicable), such as the institution conducting the review.
    \end{itemize}

\item {\bf Declaration of LLM usage}
    \item[] Question: Does the paper describe the usage of LLMs if it is an important, original, or non-standard component of the core methods in this research? Note that if the LLM is used only for writing, editing, or formatting purposes and does \emph{not} impact the core methodology, scientific rigor, or originality of the research, declaration is not required.
    \item[] Answer: \answerNA{}
    \item[] Justification: LLMs are not an important, original, or non-standard component of the proposed method. The core contribution is a sparse-attention and routing mechanism for long video diffusion.
    \item[] Guidelines:
    \begin{itemize}
        \item The answer \answerNA{} means that the core method development in this research does not involve LLMs as any important, original, or non-standard components.
        \item Please refer to our LLM policy in the NeurIPS handbook for what should or should not be described.
    \end{itemize}

\end{enumerate}

\end{document}